\documentclass[acmtog]{acmart}
\acmSubmissionID{241}

\usepackage{booktabs} %

\acmYear{2023} %
\copyrightyear{2023} %

\usepackage{hhline}
\usepackage{algorithm}
\usepackage{algpseudocode}
\usepackage{subcaption}
\usepackage{pifont}
\definecolor{limegreen}{HTML}{32CD32}
\usepackage{arydshln}
\usepackage{multirow}
\usepackage{multicol}
\usepackage{xcolor}
\usepackage[export]{adjustbox}
\usepackage[normalem]{ulem}
\usepackage{xspace}
\usepackage{rotating}
\usepackage{xhfill}

\usepackage{appendix}

\usepackage[capitalize]{cleveref}
\Crefname{section}{Section}{Sections}
\crefname{section}{Sec.}{Secs.}
\Crefname{table}{Table}{Tables}
\crefname{table}{Tab.}{Tabs.}
\crefname{figure}{Figure}{Figures}
\crefname{figure}{Figure}{Figures}
\crefname{algocf}{alg.}{algs.}
\Crefname{algocf}{Alg.}{Algs.}

\makeatletter
\DeclareRobustCommand\onedot{\futurelet\@let@token\@onedot}
\def\@onedot{\ifx\@let@token.\else.\null\fi\xspace}

\def\etal{\emph{et al}\onedot}
\def\eg{\emph{e.g}\onedot}

\def\ie{\emph{i.e}\onedot}

\def\etal{\emph{et al}\onedot}

\algrenewcommand\algorithmicrequire{\textbf{Input:}}
\algrenewcommand\algorithmicensure{\textbf{Output:}}

\definecolor{mydarkgreen}{rgb}{0.02,0.6,0.02}
\definecolor{jazzberryjam}{rgb}{0.65, 0.04, 0.37}

\setcopyright{acmlicensed}
\acmJournal{TOG}
\acmYear{2023} \acmVolume{42} \acmNumber{4} \acmArticle{1} \acmMonth{8} \acmPrice{15.00}\acmDOI{10.1145/3592116}

\title{Attend-and-Excite: Attention-Based Semantic Guidance for Text-to-Image Diffusion Models}

\author{Hila Chefer$^*$}
\affiliation{%
 \institution{Tel Aviv University}
 \country{Israel}
}
\email{hilach70@gmail.com}
\author{Yuval Alaluf$^*$}
\affiliation{%
 \institution{Tel Aviv University}
 \country{Israel}
}
\email{yuvalalaluf@gmail.com}
\author{Yael Vinker}
\affiliation{%
 \institution{Tel Aviv University}
 \country{Israel}
}
\email{yael.vinker@mail.huji.ac.il}
\author{Lior Wolf}
\affiliation{%
 \institution{Tel Aviv University}
 \country{Israel}
}
\email{liorwolf@gmail.com}

\author{Daniel Cohen-Or}
\affiliation{%
 \institution{Tel Aviv University}
 \country{Israel}
}
\email{cohenor@gmail.com}

\begin{document}

\begin{abstract}
Recent text-to-image generative models have demonstrated an unparalleled ability to generate diverse and creative imagery guided by a target text prompt.
While revolutionary, current state-of-the-art diffusion models may still fail in generating images that fully convey the semantics in the given text prompt. 
We analyze the publicly available Stable Diffusion model and assess the existence of \textit{catastrophic neglect}, where the model fails to generate one or more of the subjects from the input prompt. Moreover, we find that in some cases the model also fails to correctly bind attributes (\eg, colors) to their corresponding subjects. 
To help mitigate these failure cases, we introduce the concept of \textit{Generative Semantic Nursing (GSN)}, where we seek to intervene in the generative process on the fly during inference time to improve the faithfulness of the generated images.
Using an attention-based formulation of GSN, dubbed \textit{Attend-and-Excite}, we guide the model to refine the cross-attention units to \textit{attend} to all subject tokens in the text prompt and strengthen --- or \textit{excite} --- their activations, encouraging the model to generate all subjects described in the text prompt. 
We compare our approach to alternative approaches and demonstrate that it conveys the desired concepts more faithfully across a range of text prompts. 
Code is available at our project page:  \url{https://yuval-alaluf.github.io/Attend-and-Excite/}.

\end{abstract}

\ccsdesc[500]{Computing methodologies~Computer graphics}
\ccsdesc[500]{Computing methodologies~Image processing}

\keywords{Image Generation, Diffusion Models}

\begin{teaserfigure}
\centering
\includegraphics[width=0.975\textwidth]{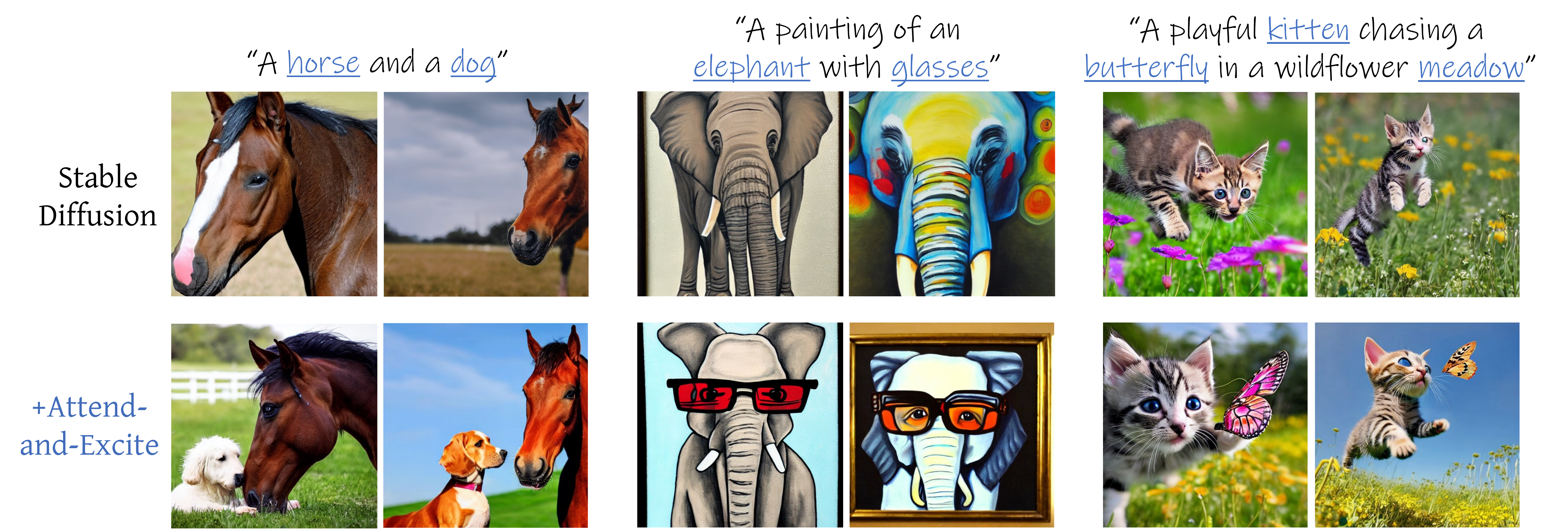}
    \\[-5pt]
    \caption{Given a pre-trained text-to-image diffusion model (\eg, Stable Diffusion~\cite{rombach2022high}) our method, Attend-and-Excite, guides the generative model to modify the cross-attention values during the image synthesis process to generate images that more faithfully depict the input text prompt. Stable Diffusion alone (top row) struggles to generate multiple objects (\eg, a horse and a dog). However, by incorporating Attend-and-Excite (bottom row) to strengthen the subject tokens (marked in blue), we achieve images that are more semantically faithful with respect to the input text prompts. 
    }
    \label{fig:teaser}
\end{teaserfigure}

\maketitle

\def\thefootnote{*}\footnotetext{Denotes equal contribution.}

\section{Introduction}

Recent advancements in text-based image generation~\cite{rombach2022high, Saharia2022PhotorealisticTD, gafni2022make, Balaji2022eDiffITD,Ramesh2022Hierarchical} have demonstrated an unprecedented ability to generate diverse and creative imagery provided a free-form text prompt. 
However, it has been shown~\cite{Feng2022Training,wang2022diffusiondb} that images produced by such models do not always faithfully reflect the semantic meaning of the target prompt.

\begin{figure}
    \centering
    \setlength{\tabcolsep}{0.5pt}
    \addtolength{\belowcaptionskip}{-12.5pt}
    {\small
    \begin{tabular}{c c c @{\hspace{0.1cm}} c c}

        &
        \multicolumn{2}{c}{\begin{tabular}{c}``A yellow \textcolor{blue}{bowl} and a blue \textcolor{blue}{cat}'' \end{tabular}} &
        \multicolumn{2}{c}{\begin{tabular}{c} ``A yellow \textcolor{blue}{bow}
        and a brown \textcolor{blue}{bench}'' \end{tabular}} \\

        \raisebox{0.115in}{\rotatebox{90}{\begin{tabular}{c} Stable \\ Diffusion \end{tabular}}} &
        \includegraphics[width=0.102\textwidth]{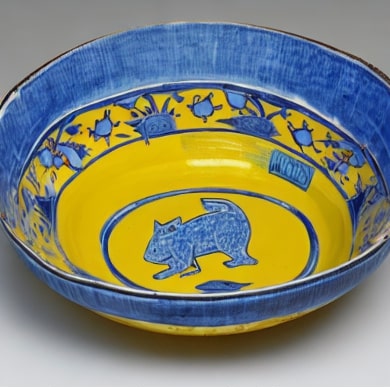} &
        \includegraphics[width=0.102\textwidth]{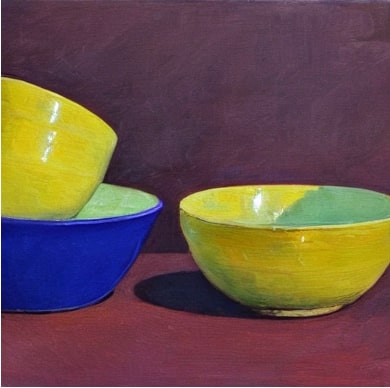} &
        \hspace{0.05cm}
        \includegraphics[width=0.102\textwidth]{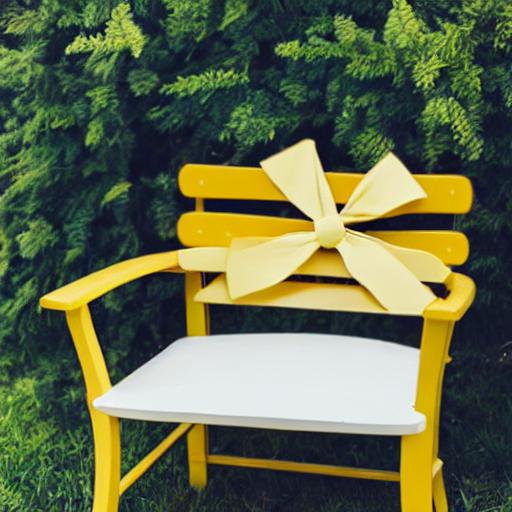} &
        \includegraphics[width=0.102\textwidth]{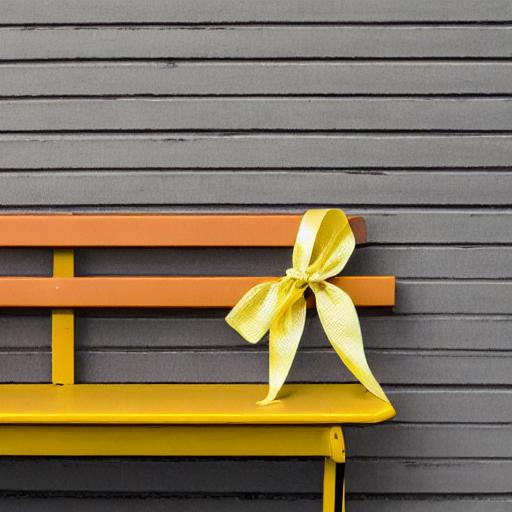} \\

        \raisebox{0.125in}{\rotatebox{90}{\begin{tabular}{c} SD with \\ \textcolor{blue}{A\&E} \end{tabular}}} &
        \includegraphics[width=0.102\textwidth]{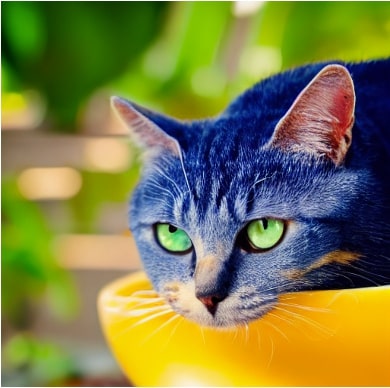} &
        \includegraphics[width=0.102\textwidth]{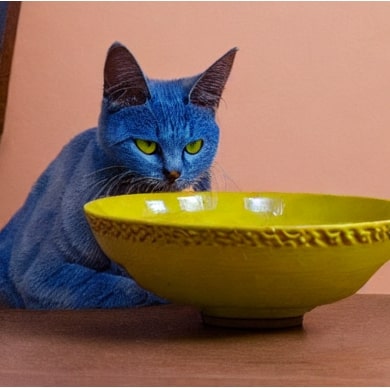} &
        \hspace{0.05cm}
        \includegraphics[width=0.102\textwidth]{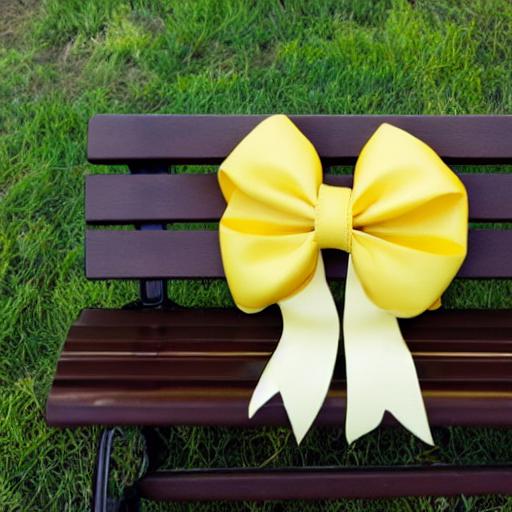} &
        \includegraphics[width=0.102\textwidth]{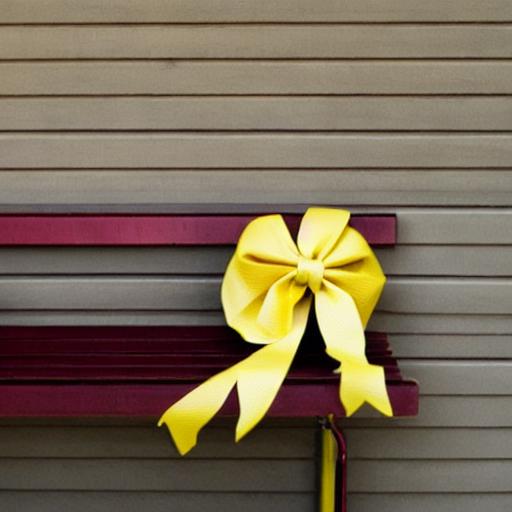} \\

        &
        \multicolumn{2}{c}{Catastrophic Neglect} &
        \multicolumn{2}{c}{Incorrect Attribute Binding}
    \\[-0.2cm]
    \end{tabular}

    }
    \caption{Failure cases of Stable Diffusion (SD)~\cite{rombach2022high}. In the top row, we show examples of two failure settings: catastrophic neglect (left) and incorrect attribute binding (right). In the bottom row, we show images obtained when applying Attend-and-Excite over SD using the same seeds.
    }
    \label{fig:sd_failure_cases}
\end{figure}

We observe two key semantic issues in state-of-the-art text-based image generation models: (i) ``catastrophic neglect'', where one or more of the subjects of the prompt are not generated; and (ii) incorrect ``attribute binding'', where the model binds attributes to the wrong subjects or fails to bind them entirely. 
Examples of cases where the aforementioned issues arise can be found in the top row of~\Cref{fig:sd_failure_cases}, which depicts images generated by the state-of-the-art Stable Diffusion model~\cite{rombach2022high}.
In the left column, we provide an example of catastrophic neglect where the model fails to generate the blue cat, choosing to focus solely on generating the bowl. 
In the right column, we demonstrate incorrect attribute binding where the color ``yellow'' is incorrectly binded to the bench.

To mitigate these semantic issues, we introduce the concept of \textit{``Generative Semantic Nursing'' (GSN)}. In the GSN process, one slightly shifts the latent code at each timestep of the denoising process such that the latent is encouraged to better consider the semantic information passed from the input text prompt. 

We propose a form of GSN dubbed \textit{Attend-and-Excite}, which leverages the powerful cross-attention maps of a pre-trained diffusion model. 
The attention maps define a probability distribution over the text tokens for each image patch, which determines the dominant tokens in the patch.
We observe that this text-image interaction is susceptible to neglect. Although each patch can attend freely to all text tokens, there is no mechanism to ensure that \textit{all} tokens are attended to by some patch in the image.
In cases where a subject token is not attended to, the corresponding subject will not be manifested in the output image. 

Thus, intuitively, in order for a subject to be present in the generated image, the model should assign at least one image patch to the subject's token.
Attend-and-Excite embodies this intuition by demanding that each subject token is dominant in some patch in the image. 
We carefully guide the latent at each denoising timestep and encourage the model to \textit{attend} to all subject tokens and strengthen --- or \textit{excite} --- their activations. 
Importantly, our approach is applied on the fly during inference time and requires no additional training or fine-tuning. We instead choose to preserve the strong semantics already learned by the pre-trained generative model and text encoder. 
Example generations with our approach applied over Stable Diffusion are shown in the bottom row of~\Cref{fig:teaser}.

As shall be demonstrated, although Attend-and-Excite explicitly tackles only the issue of catastrophic neglect, our solution implicitly encourages correct bindings between attributes and their subjects. This can be attributed to the connection between the two issues of catastrophic neglect and attribute binding. The embedding of the text, obtained by a pre-trained text encoder, links information between each subject and its corresponding attributes. For example, in the prompt ``a yellow bowl and a blue cat'', the token ``cat'' receives information from the token ``blue'' during the text encoding process. Therefore, mitigating catastrophic neglect over the cat should ideally result in enhancing the color attribute (\ie, allowing for correct binding between ``cat'' and ``blue'').

We demonstrate Attend-and-Excite's superiority in generating semantically-faithful images over Stable Diffusion and alternative methods that explore similar semantic issues. 
We additionally analyze the cross-attention maps realized with and without Attend-and-Excite and demonstrate the importance of applying our method to mitigate catastrophic neglect, while enabling the use of cross-attention as a form of explanation for the generated content.

\section{Related Work}
Early works studied text-guided image synthesis in the context of GANs \cite{tao2022df,zhu2019dm,xu2018attngan,zhang2021cross,ye2021improving}. More recently, impressive results were achieved with large-scale auto-regressive models~\cite{ramesh2021zero,yu2022scaling} and diffusion models~\cite{Ramesh2022Hierarchical,nichol2021glide,rombach2022high,Saharia2022PhotorealisticTD}.
Yet, generating images that faithfully align with the input prompt is often difficult. To enforce heavier reliance on the text, classifier-free guidance~\cite{Saharia2022PhotorealisticTD,nichol2021glide,ho2022classifier} allows extrapolating text-driven gradients to better guide the generation by strengthening the reliance on the text. However, even when employing this technique, extensive prompt engineering is often required to achieve the expected result~\cite{witteveen2022investigating,wang2022diffusiondb,liu2022design,marcus2022very}.

To provide users with more control over the synthesis process, several works employ a segmentation map or spatial conditioning~\cite{gafni2022make,avrahami2022spatext,zhaobo2019layout2im}. In the context of image editing, while most methods are generally limited to global edits~\cite{crowson2022vqgan,kwon2022clipstyler,gal2022stylegan,chefer2022targetclip}, several works introduce a user-provided mask to specify the region that should be altered~\cite{avrahami2022blended,bau2021paint,nichol2021glide,couairon2022diffedit}. 

Another related line of work aims to introduce specific concepts to a pre-trained text-to-image model by learning to map a set of images to a ``word'' in the embedding space of the model~\cite{gal2022image, ruiz2022dreambooth,kumari2022multi}.
Several works have also explored providing users with more control over the synthesis process solely through the use of the input text prompt~\cite{hertz2022prompt,kawar2022imagic,brooks2022instructpix2pix,valevski2022unitune}. 

Recently, two works have explored the semantic flaws of text-to-image models. 
First, Liu~\etal~\shortcite{liu2022compositional} propose Composable Diffusion models where an image is generated by composing multiple outputs of a pre-trained diffusion model. Each output is tasked with capturing different image components which are then joined using compositional operators to attain a unified image. Yet, we observe that this method often struggles in achieving realistic compositions of multiple objects (see~\Cref{sec:results}). Moreover, the approach is limited to operating over conjunctions and negations of subjects.

Feng~\etal~\shortcite{Feng2022Training} propose StructureDiffusion which employs consistency trees or scene graphs to split the prompt into several noun phrases. 
An attention map is computed for each noun phrase and the output of the cross-attention unit is the average of all attention operations.
In contrast, our Attend-and-Excite technique  directly optimizes the noised latent, allowing us to synthesize images that vary significantly from those produced by Stable Diffusion.
We find that results obtained by StructureDiffusion often resemble those produced by Stable Diffusion, falling short of achieving meaningful modifications that amend the semantic faults (see~\Cref{sec:results}).

It should be noted that there are additional semantic issues in text-based image synthesis, \eg, object relations and compositions. Addressing such issues may require additional models to determine the object relations~\cite{johnson2018image,ashual2019specifying}. However, this deviates from the scope of this work where we focus on inference-time guidance of a pre-trained generative model.

\section{Preliminaries}

\paragraph{\textbf{Latent Diffusion Models}}
We apply our method over the state-of-the-art Stable Diffusion model (SD)~\cite{rombach2022high}. 
Instead of operating in the image space, SD operates in the latent space of an autoencoder. First, an encoder $\mathcal{E}$ is trained to map a given image $x \in \mathcal{X}$ into a spatial latent code $z = \mathcal{E}(x)$. A decoder $\mathcal{D}$ is then tasked with reconstructing the input image such that $\mathcal{D}(\mathcal{E}(x)) \approx x$.

Given the trained autoencoder, a denoising diffusion probabilistic model (DDPM)~\cite{sohl2015deep,ho2020denoising} operates over the learned latent space to produce a denoised version of an input latent $z_t$ at each timestep $t$. During the denoising process, the diffusion model can be conditioned on an additional input vector. In Stable Diffusion, this additional input is typically a text encoding produced by a pre-trained CLIP text encoder~\cite{radford2021learning}. Given a conditioning prompt $y$, we denote the conditioning vector by $c(y)$. The DDPM model $\varepsilon_\theta$ is trained to minimize the loss,
\begin{equation}
    \mathcal{L} = \mathbb{E}_{z\sim\mathcal{E}(x),y,\varepsilon\sim\mathcal{N}(0,1),t} \left [ || \varepsilon - \varepsilon_\theta(z_t, t, c(y)) ||_2^2 \right ].
\end{equation}
In words, at each timestep $t$, the denoising network $\varepsilon_\theta$ is tasked with correctly removing the noise $\varepsilon$ added to the latent code $z$, given the noised latent $z_t$, timestep $t$, and conditioning encoding $c(y)$. Here, $\varepsilon_\theta$ is a UNet network~\cite{ronneberger2015u} consisting of self-attention and cross-attention layers, discussed below. 

At inference, a latent $z_T$ is sampled from $\mathcal{N}(0, 1)$ and is iteratively denoised to produce a latent $z_0$ using the DDPM. The denoised latent is then passed to the decoder to obtain the image $x' = \mathcal{D}(z_0)$.

\begin{figure}
    \centering
    \includegraphics[width=0.475\textwidth]{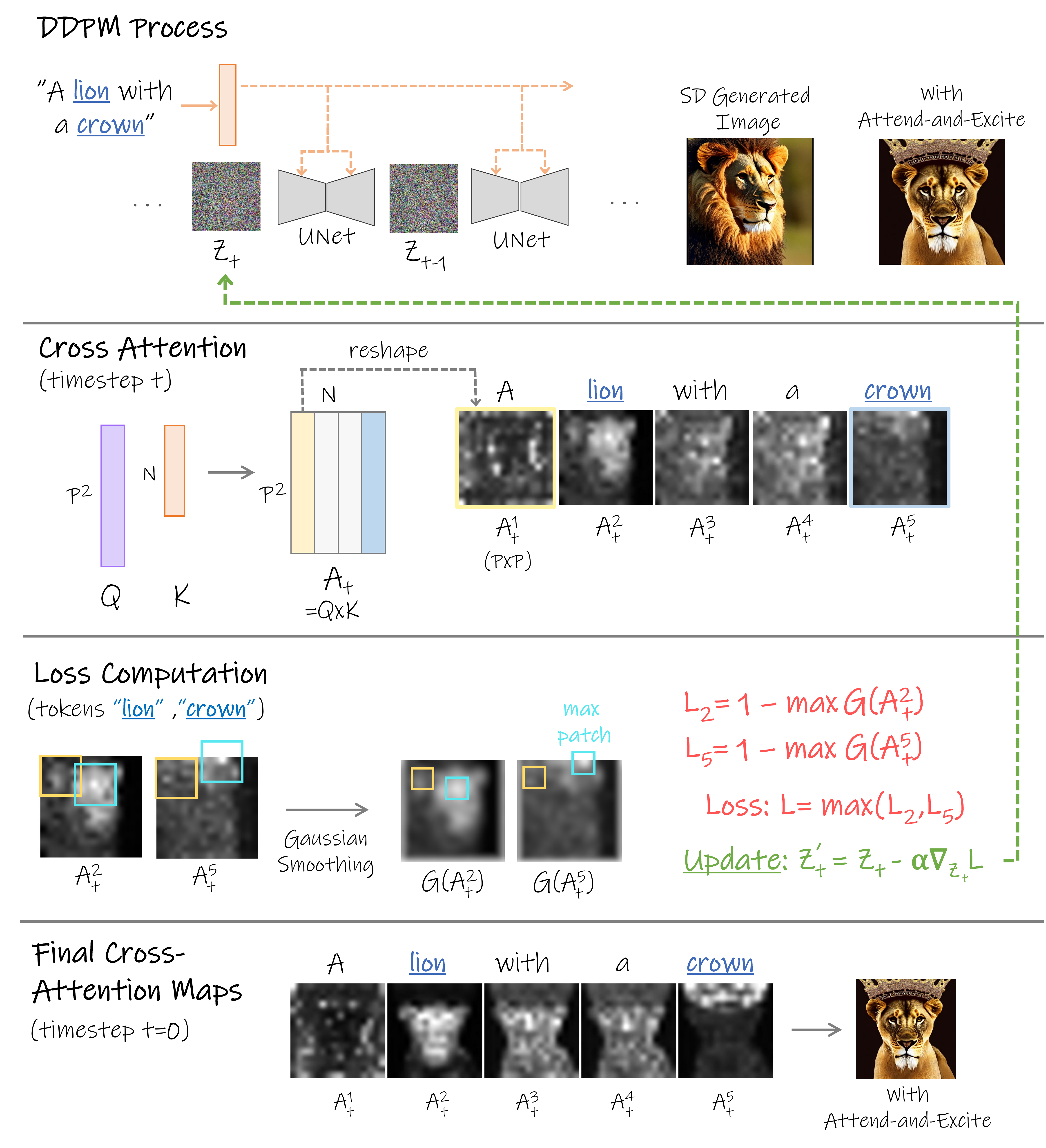}
    \\[-0.35cm]
    \caption{Overview of Attend-and-Excite. Given a prompt (\eg ``A lion with a crown''), we extract the subject tokens (lion, crown), and their corresponding attention maps ($A_t^2, A_t^5$). We apply a Gaussian kernel on each attention map to obtain smoothed attention maps that consider the neighboring patches. Our optimization enhances the maximal activation for the most neglected token at timestep $t$ and updates the latent code $z_t$ accordingly. The final cross-attention maps at $t=0$ are illustrated in the final row.
    }
    \vspace{-0.3cm}
    \label{fig:overview}
\end{figure}

\paragraph{\textbf{Text-Conditioning Via Cross-Attention}}
Text guidance in Stable Diffusion is performed using the cross-attention mechanism. The denoising UNet network consists of self-attention layers followed by cross-attention layers at resolutions of $64, 32, 16$, and $8$.

Denote by $P$ the spatial dimension of the intermediate feature map (\ie, $P\in\{64,32,16,8\}$), and by $N$ the number of text tokens in the prompt. An attention map $A_t \in \mathbb{R}^{P\times P \times N}$ is calculated over linear projections of the intermediate features ($Q$) and text embedding ($K$), as illustrated in the second row of~\Cref{fig:overview}. 
$A_t$ defines a distribution over the text tokens for each spatial patch $(i,j)$. Specifically, $A_t[i,j,n]$ denotes the probability assigned to token $n$ for the $(i,j)$-th spatial patch of the intermediate feature map. Intuitively, this probability indicates the amount of information that will be passed from token $n$ to patch $(i,j)$. Note that the maximum value of each of the $P\times P$ cells is $1$. 

We operate over the $16\times16$ attention units since they have been shown to contain the most semantic information~\cite{hertz2022prompt}.

\section{Attend-and-Excite}

\algnewcommand{\algorithmicgoto}{\textbf{Go to}}%
\algnewcommand{\Goto}[1]{\algorithmicgoto~\ref{#1}}%

\begin{algorithm}[t!]
\caption{A Single Denoising Step using Attend-and-Excite}

\begin{flushleft}
\textbf{Input:} A text prompt $\mathcal{P}$, a set of subject token indices $\mathcal{S}$, a timestep $t$, a set of iterations for refinement $\{t_1,\dots,t_k\}$, a set of thresholds $\{T_1,\dots,T_k\}$, 
and a trained Stable Diffusion model $SD$.\\
\textbf{Output:} A noised latent $z_{t-1}$ for the next timestep
\end{flushleft}

\begin{algorithmic}[1]

\State $\_, A_t \gets SD(z_t, \mathcal{P}, t)$
\State $A_t \gets \text{Softmax}(A_t - \langle sot \rangle)$
\For{$s \in \mathcal{S}$}
    \State $A_t^s \gets A_t[:, :, s]$
    \State $A_t^s \gets \text{Gaussian}(A_t^s)$
    \State $\mathcal{L}_s \gets 1 - \max(A_t^s)$
\EndFor
\State $\mathcal{L} \gets \max_s(\mathcal{L}_s)$
\State $z_t' \gets z_t - \alpha_t \cdot \nabla_{z_t} \mathcal{L}$

\If{$t \in \{t_1, \dots, t_k\}$}  \Comment{\small{If performing iterative refinement at $t$}}
    \If{$\mathcal{L} > 1 - T_{t}$}
        \State $z_t \gets z_t'$
        \State \textbf{Go to} Step 1
    \EndIf
\EndIf

\State $z_{t-1}, \_ \gets SD(z_t', \mathcal{P}, t)$
\State \textbf{Return} $z_{t-1}$

\end{algorithmic}
\label{alg:attend-and-excite}
\end{algorithm}

At the core of our method is the idea of \textit{generative semantic nursing}, where we gradually shift the noised latent code at each timestep $t$ toward a more semantically-faithful generation. 
At each denoising step $t$, we consider the attention maps of the subject tokens in the prompt $\mathcal{P}$. Intuitively, for a subject to be present in the synthesized image, it should have a high influence on some patch in the image. As such, we define a loss objective that attempts to maximize the attention values for each subject token. We then update the noised latent at time $t$ according to the gradient of the computed loss. This encourages the latent at the next timestep to better incorporate all subject tokens in its representation. This manipulation occurs on the fly during inference (\ie, no additional training is performed).

In the next sections, we discuss each of the steps presented in~\Cref{alg:attend-and-excite} for a single denoising timestep $t$ as illustrated in~\Cref{fig:overview}.

\paragraph{\textbf{Extracting the Cross-Attention Maps}}
Given the input prompt $\mathcal{P}$, we consider the set of all subject tokens (\eg, nouns) $\mathcal{S} = \{s_1, ..., s_k\}$ present in $\mathcal{P}$. 
Our objective is to extract a spatial attention map for each token $s \in \mathcal{S}$, indicating the influence of $s$ on each image patch.

Given the noised latent $z_t$ at the current timestep, we perform a forward pass through the pre-trained UNet network using $z_t$ and $\mathcal{P}$ (Step 1 in~\Cref{alg:attend-and-excite}).
We then consider the resulting cross-attention map obtained after averaging all $16\times16$ attention layers and  heads. The resulting aggregated map $A_t$ contains $N$ spatial attention maps, one for each of the tokens of $\mathcal{P}$, \ie $A_t \in \mathbb{R}^{16\times 16 \times N}$. 

The pre-trained CLIP text encoder prepends a specialized token $\langle sot \rangle$ to $\mathcal{P}$ indicating the start of the text. 
We note that Stable Diffusion learns to consistently assign a high attention value to the $\langle sot \rangle$ token in the token distribution defined in $A_t$.
Since we are interested in enhancing the actual prompt tokens, we re-weigh the attention values by ignoring the attention of $\langle sot \rangle$ and performing a Softmax operation on the remaining tokens (Step 2 in~\Cref{alg:attend-and-excite}).
After the Softmax operation, the $(i,j)$-th entry of the resulting matrix $A_t$ indicates the probability of each of the textual tokens being present in the corresponding image patch.
We then extract the $16\times 16$ normalized attention map for each subject token $s$ (Step 4).

\paragraph{\textbf{Obtaining Smooth Attention Maps}}
Observe that the attention values $A_t^s$ calculated above may not fully reflect whether an object is generated in the resulting image. 
Specifically, a single patch with a high attention value could stem from partial information being passed from the token $s$.
This may occur when the model does not generate the full subject, but rather a patch that resembles some part of the subject, \eg, a silhouette that resembles an animal's body part. See~\Cref{sec:ablations} for such failure cases. 

To avoid such adversarial solutions, we apply a Gaussian filter over $A_t^s$ in Step 5 of~\Cref{alg:attend-and-excite}. After doing so, the attention value of the maximally-activated patch is dependent on its neighboring patches since, after this operation, each patch is a linear combination of its neighboring patches in the original map.

\begin{figure}[t]
    \addtolength{\belowcaptionskip}{-7.5pt}
    \begin{center}
    \includegraphics[width=0.9\linewidth]{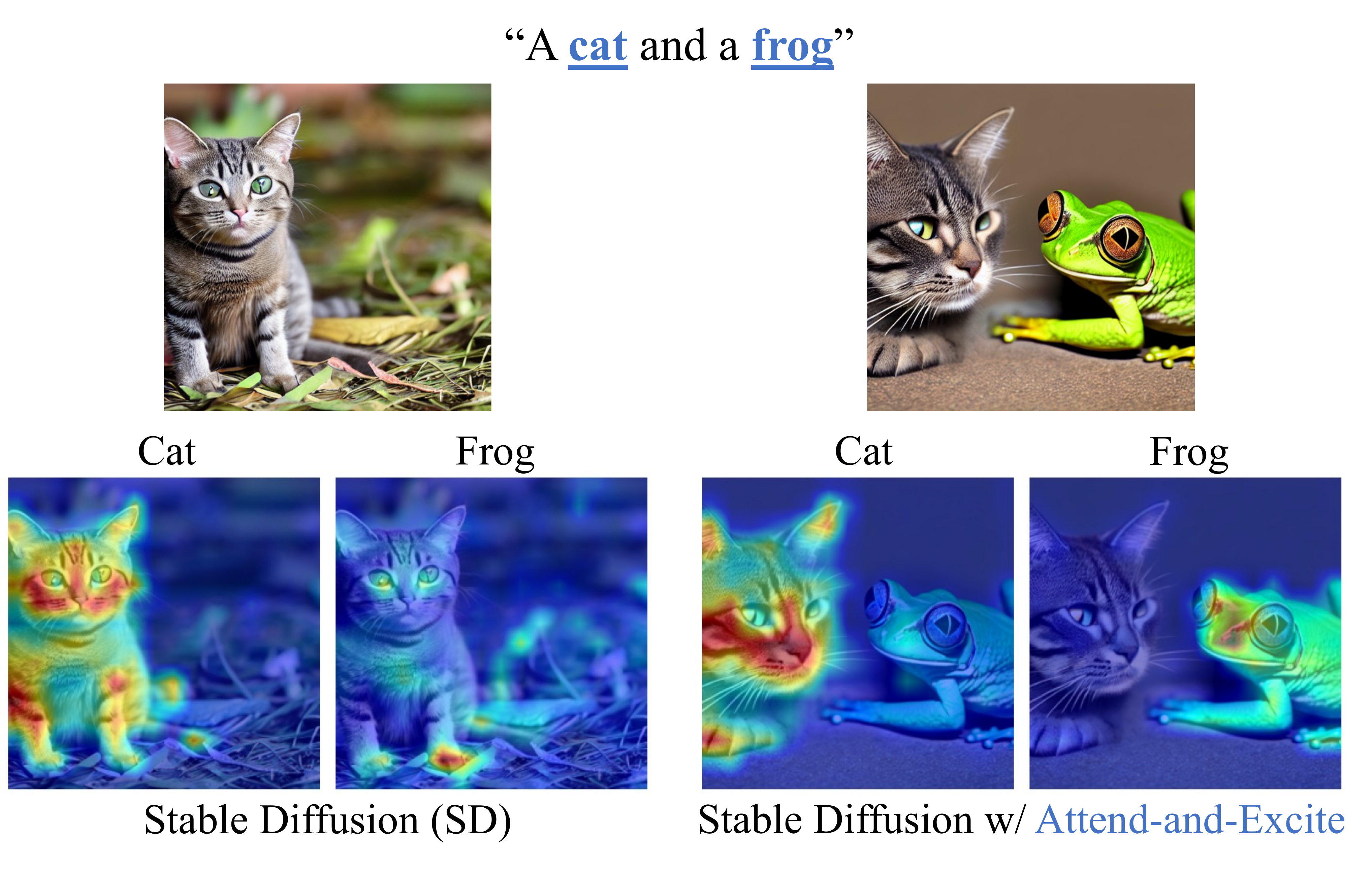}
    \\[-0.4cm]
    \caption{Visualization of the cross-attention maps for each subject token with and without Attend-and-Excite  over Stable Diffusion.}
    \label{fig:cls_spes}
    \end{center}
\end{figure}

\setlength{\abovedisplayskip}{7.5pt}
\setlength{\belowdisplayskip}{7.5pt}

\paragraph{\textbf{Performing On the Fly Optimization.}}
Intuitively, successfully generated subjects should have an image patch that significantly attends to their corresponding token. Our optimization objective embodies this intuition directly.

For each subject token in $S$, our optimization encourages the existence of at least one patch of $A_t^s$ with a high activation value. Therefore, we define the loss quantifying this desired behavior as
\begin{align}
      \mathcal{L} &= \max_{s \in S} \mathcal{L}_s  &  &\text{where} & \mathcal{L}_s &= 1 - \max(A_t^s).
\end{align}
That is, the loss attempts to strengthen the activations of the most neglected subject token at the current timestep $t$. It should be noted that different timesteps may strengthen different tokens, encouraging all neglected subject tokens to be strengthened at some timestep. 

Having computed our loss $\mathcal{L}$, we shift the current latent $z_t$ by
\begin{equation}~\label{eq:latent_update}
    z'_t \gets z_t - \alpha_t \cdot  \nabla_{z_t} \mathcal{L},
\end{equation}
where $\alpha_t$ is a scalar defining the step size of the gradient update.
Finally, we perform another forward pass through $SD$ using $z'_t$ to calculate $z_{t-1}$ for the next denoising step (Step 16 of~\Cref{alg:attend-and-excite}).  
The above update process is repeated for a subset of the timesteps $t = T,T-1,\dots,t_{end}$ where we set $T=50$, following Stable Diffusion, and $t_{end} = 25$. This is based on the observation that the final timesteps do not alter the spatial locations of objects in the generated image. 

\begin{figure*}
    \centering
    \setlength{\tabcolsep}{0.5pt}
    \renewcommand{\arraystretch}{0.3}
    {\small
    \begin{tabular}{c c c @{\hspace{0.2cm}} c c @{\hspace{0.2cm}} c c @{\hspace{0.2cm}} c c }

        &
        \multicolumn{2}{c}{``A \textcolor{blue}{cat} and a \textcolor{blue}{dog}''} &
        \multicolumn{2}{c}{``A \textcolor{blue}{turtle} and a yellow \textcolor{blue}{bowl}''} &
        \multicolumn{2}{c}{``A \textcolor{blue}{frog} and a pink \textcolor{blue}{bench}''}
        &
        \multicolumn{2}{c}{\begin{tabular}{c} ``A red \textcolor{blue}{bench} and a yellow \textcolor{blue}{clock}''
        \end{tabular}} \\

        {\raisebox{0.275in}{
        \multirow{2}{*}{\rotatebox{90}{Stable Diffusion}}}} &
        \includegraphics[width=0.1015\textwidth]{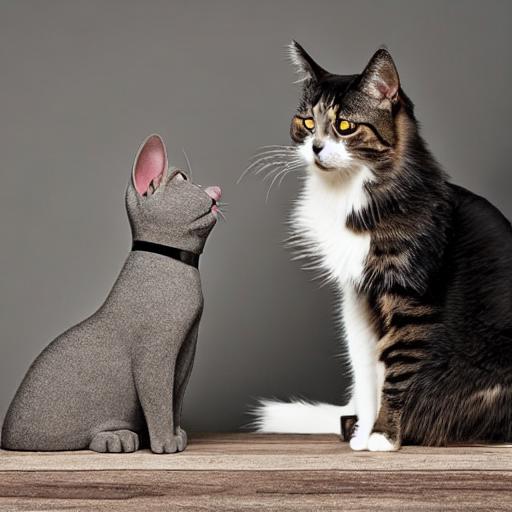} &
        \includegraphics[width=0.1015\textwidth]{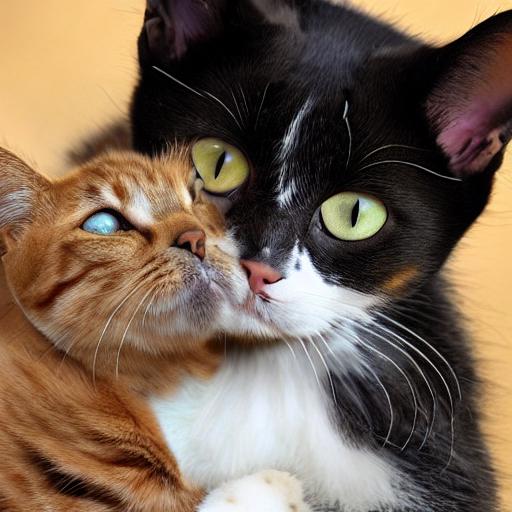} &
        \hspace{0.05cm}
        \includegraphics[width=0.1015\textwidth]{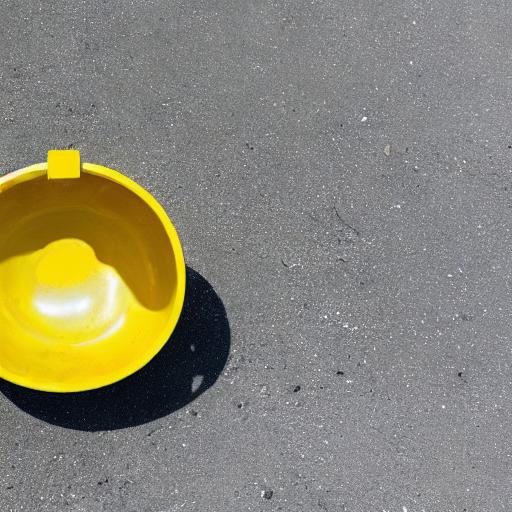} &
        \includegraphics[width=0.1015\textwidth]{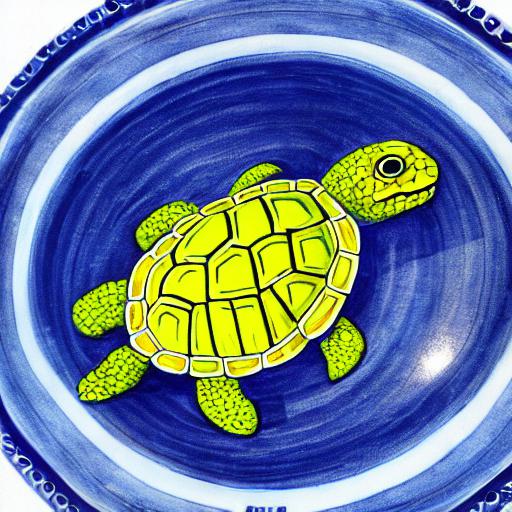} &
        \hspace{0.05cm}
        \includegraphics[width=0.1015\textwidth]{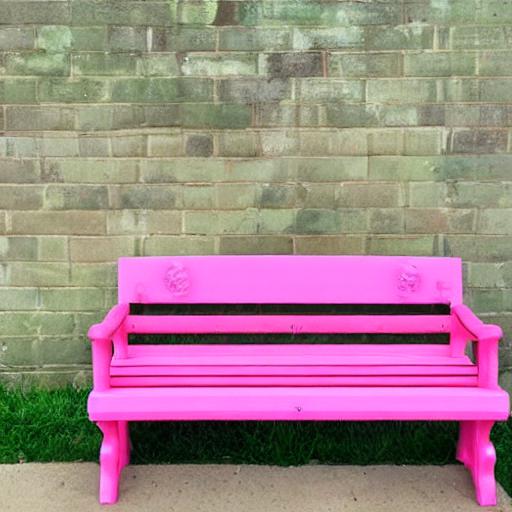} &
        \includegraphics[width=0.1015\textwidth]{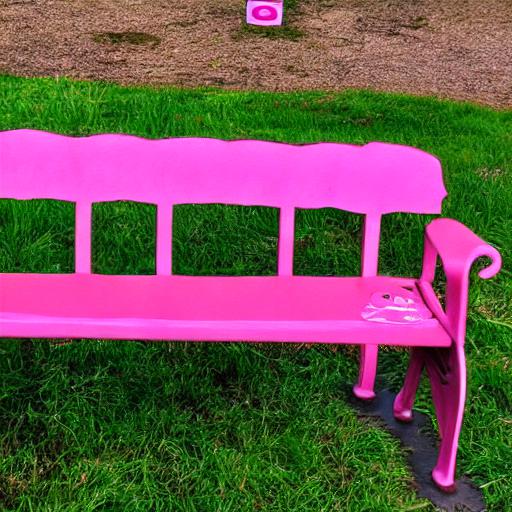} &
        \hspace{0.05cm}
        \includegraphics[width=0.1015\textwidth]{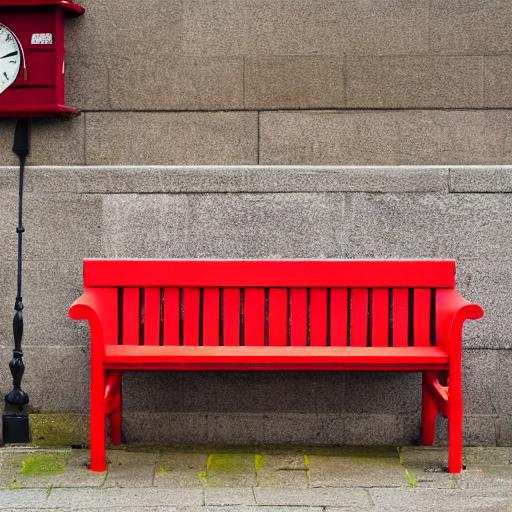} &
        \includegraphics[width=0.1015\textwidth]{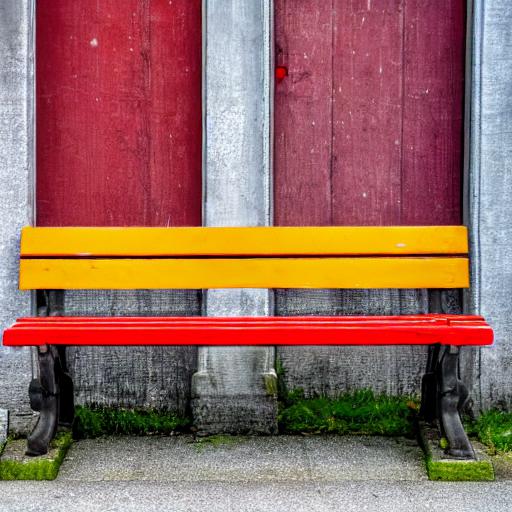} \\

        &
        \includegraphics[width=0.1015\textwidth]{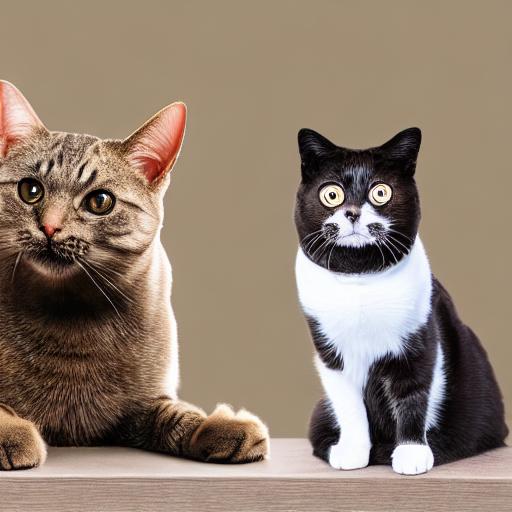} &
        \includegraphics[width=0.1015\textwidth]{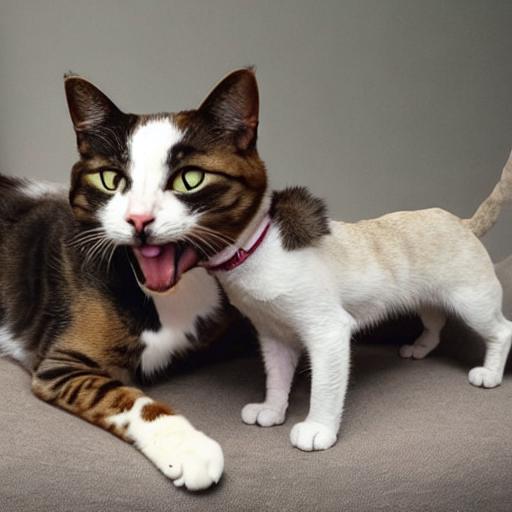} &
        \hspace{0.05cm}
        \includegraphics[width=0.1015\textwidth]{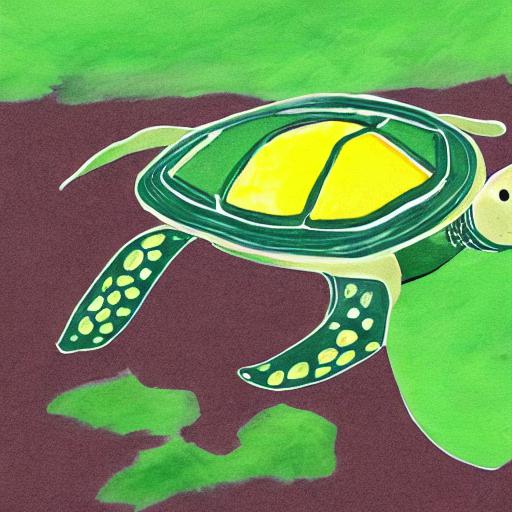} &
        \includegraphics[width=0.1015\textwidth]{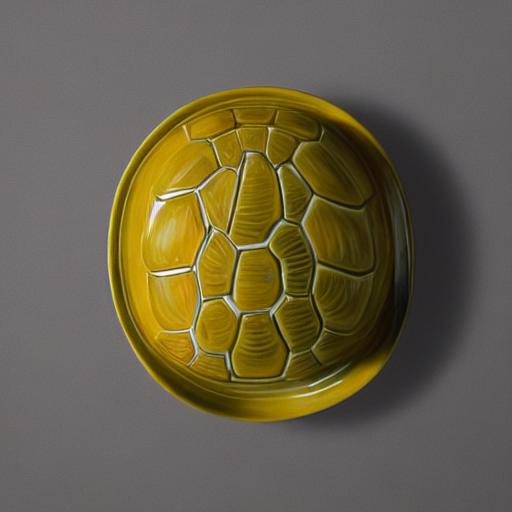} &
        \hspace{0.05cm}
        \includegraphics[width=0.1015\textwidth]{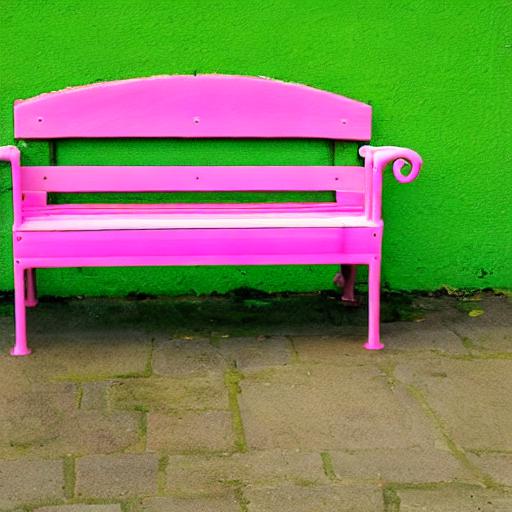} &
        \includegraphics[width=0.1015\textwidth]{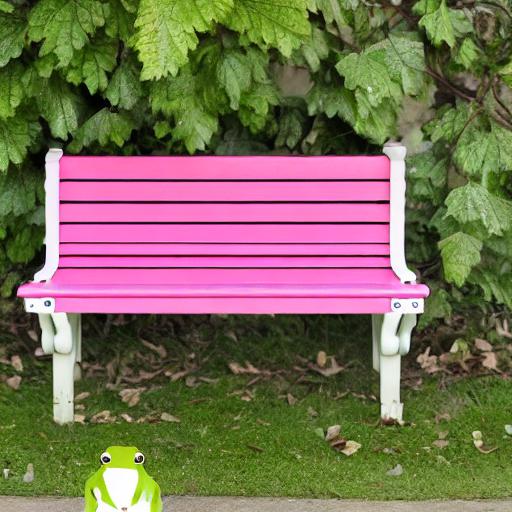} &
        \hspace{0.05cm}
        \includegraphics[width=0.1015\textwidth]{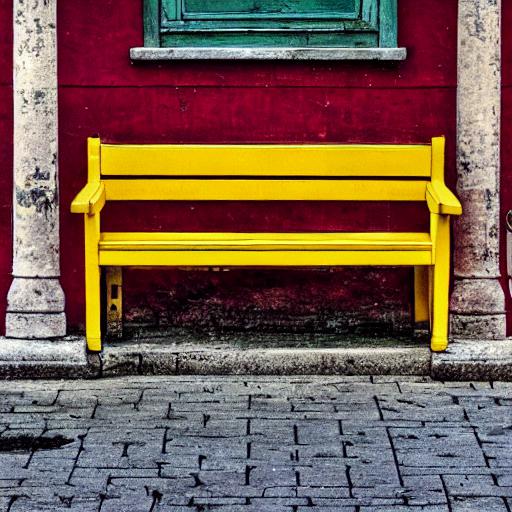} &
        \includegraphics[width=0.1015\textwidth]{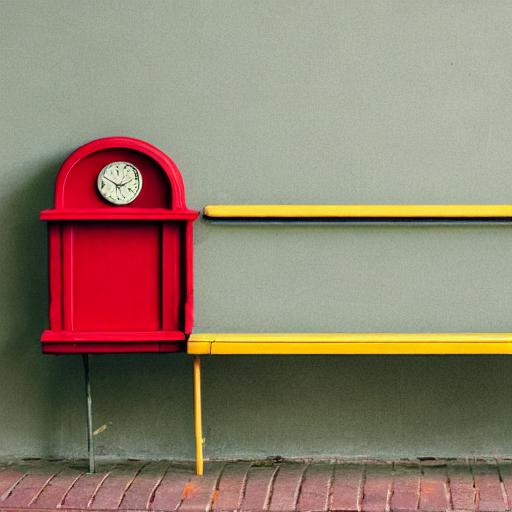}  \\ \\

        {\raisebox{0.425in}{
        \multirow{2}{*}{\rotatebox{90}{Composable Diffusion}}}} &
        \includegraphics[width=0.1015\textwidth]{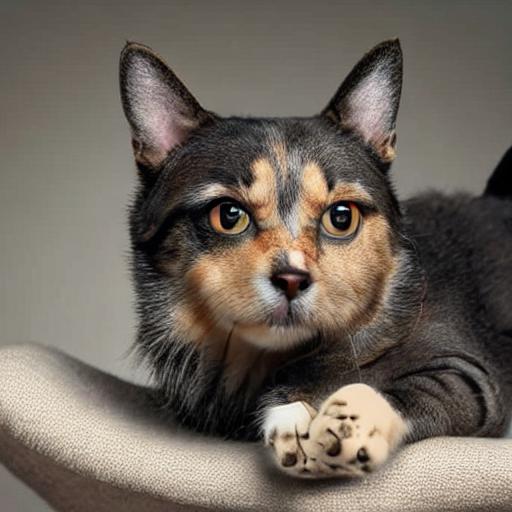} &
        \includegraphics[width=0.1015\textwidth]{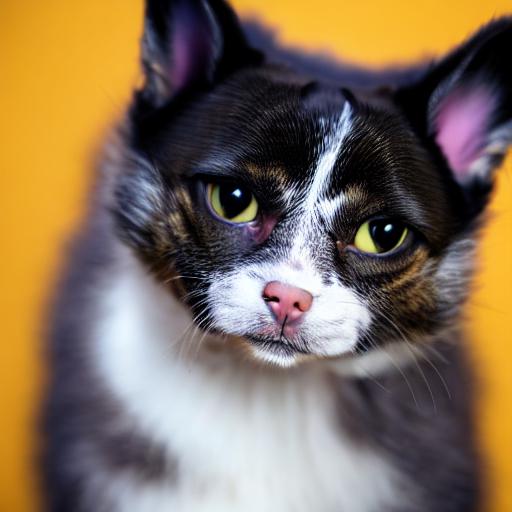} &
        \hspace{0.05cm}
        \includegraphics[width=0.1015\textwidth]{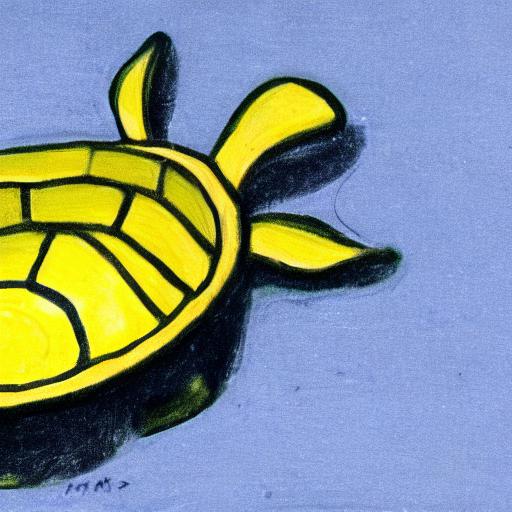} &
        \includegraphics[width=0.1015\textwidth]{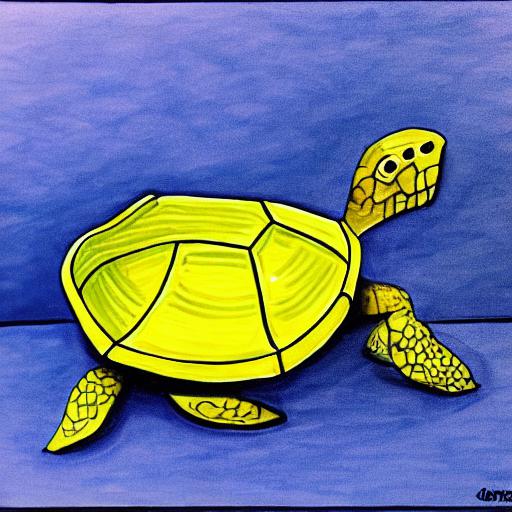} &
        \hspace{0.05cm}
        \includegraphics[width=0.1015\textwidth]{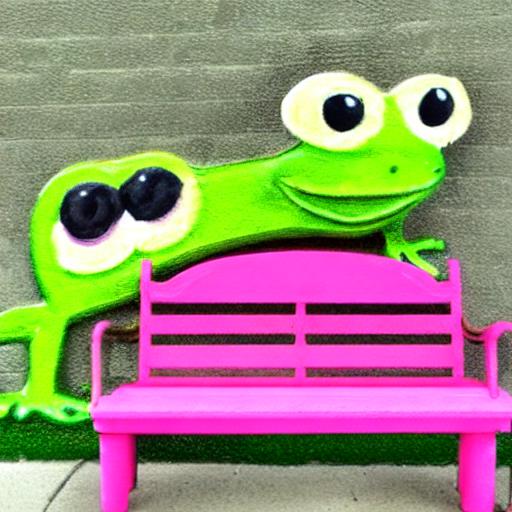} &
        \includegraphics[width=0.1015\textwidth]{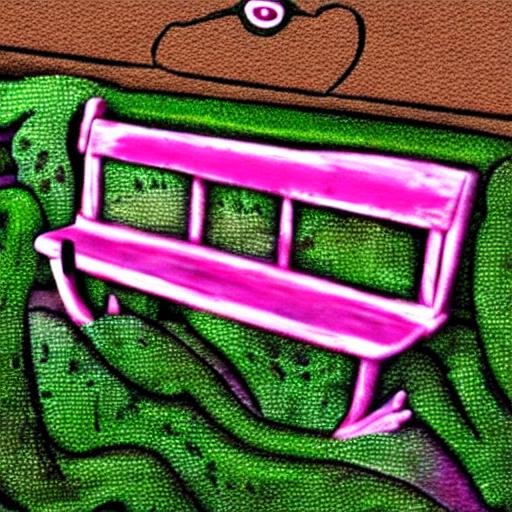} &
        \hspace{0.05cm}
        \includegraphics[width=0.1015\textwidth]{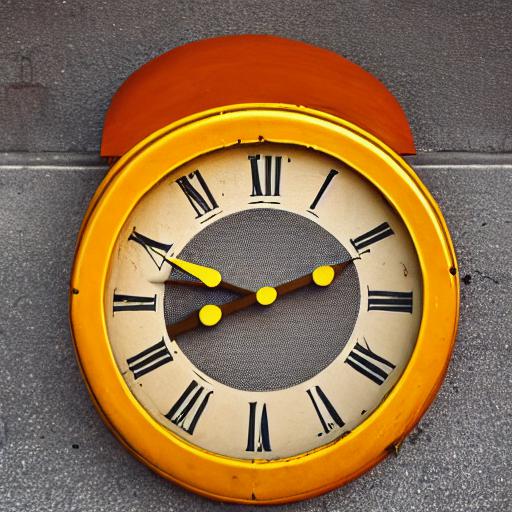} &
        \includegraphics[width=0.1015\textwidth]{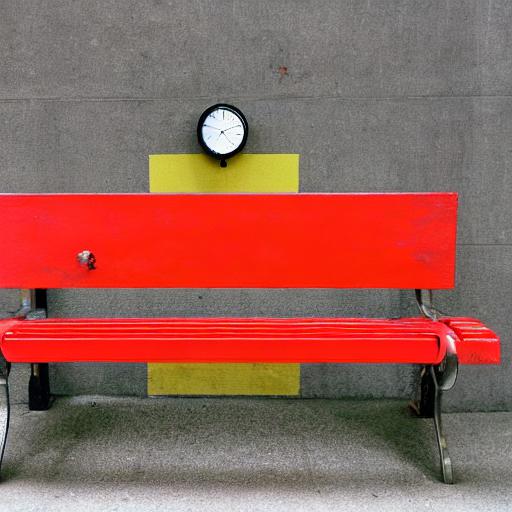} \\

        &
        \includegraphics[width=0.1015\textwidth]{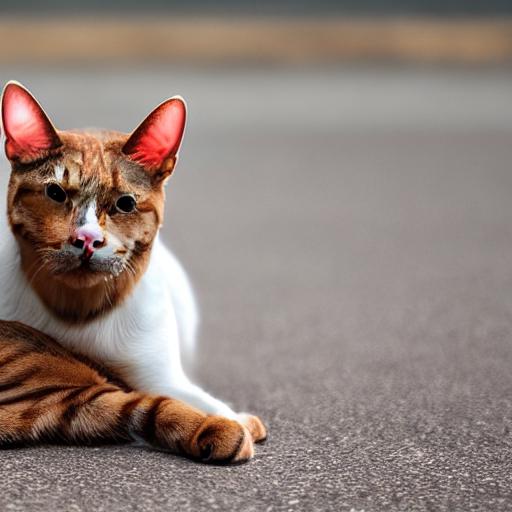} &
        \includegraphics[width=0.1015\textwidth]{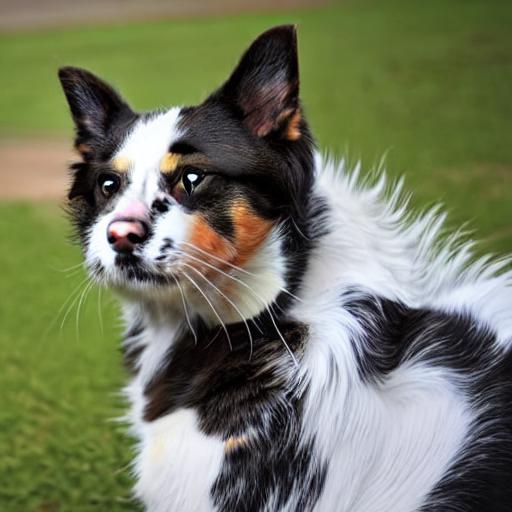} &
        \hspace{0.05cm}
        \includegraphics[width=0.1015\textwidth]{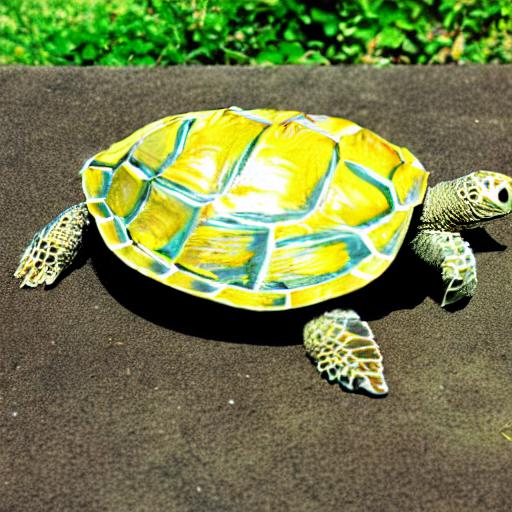} &
        \includegraphics[width=0.1015\textwidth]{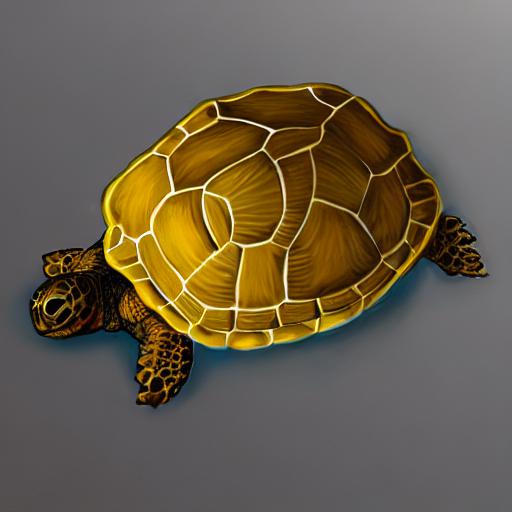} &
        \hspace{0.05cm}
        \includegraphics[width=0.1015\textwidth]{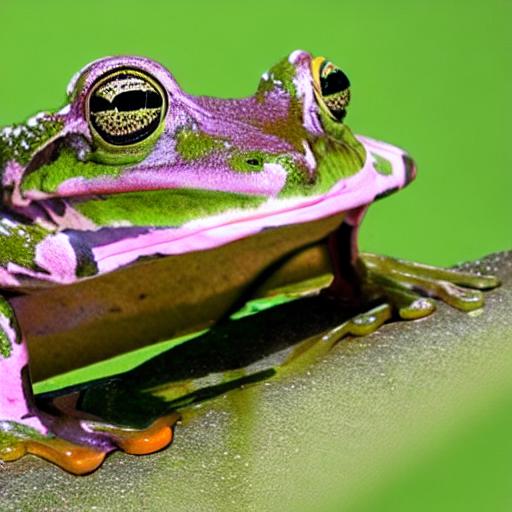} &
        \includegraphics[width=0.1015\textwidth]{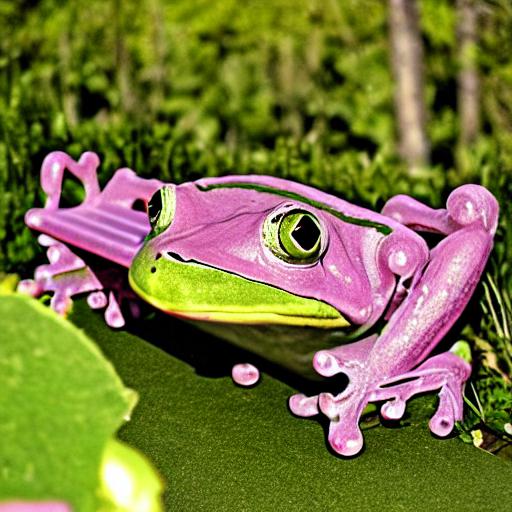} &
        \hspace{0.05cm}
        \includegraphics[width=0.1015\textwidth]{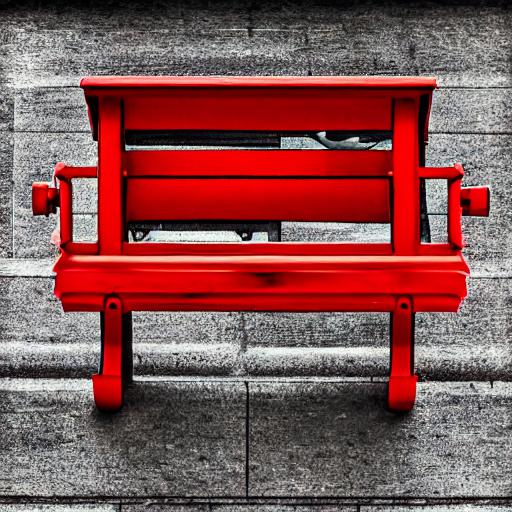} &
        \includegraphics[width=0.1015\textwidth]{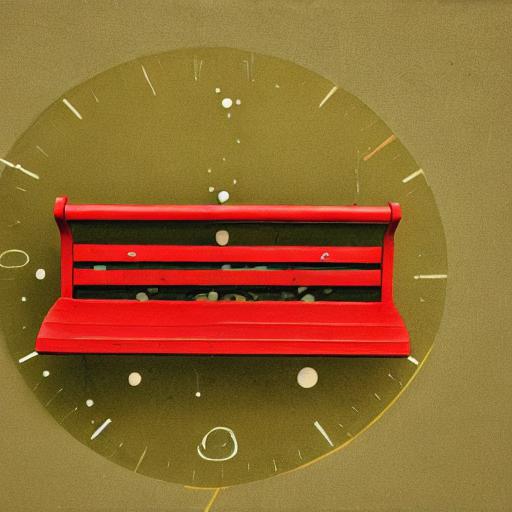}  \\ \\

        {\raisebox{0.325in}{
        \multirow{2}{*}{\rotatebox{90}{StructureDiffusion}}}} &
        \includegraphics[width=0.1015\textwidth]{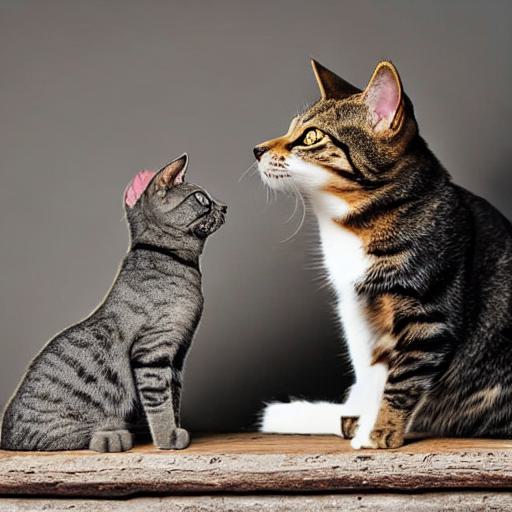} &
        \includegraphics[width=0.1015\textwidth]{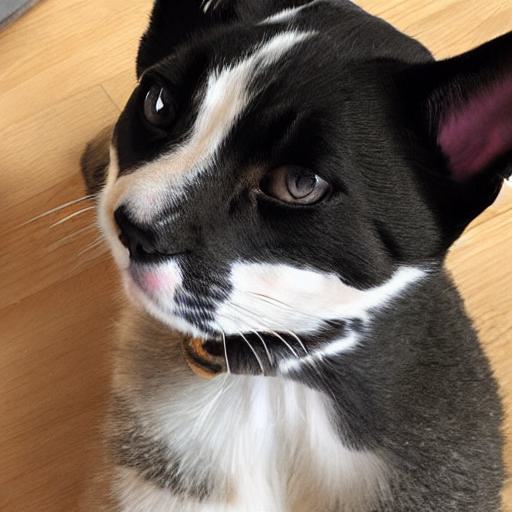} &
        \hspace{0.05cm}
        \includegraphics[width=0.1015\textwidth]{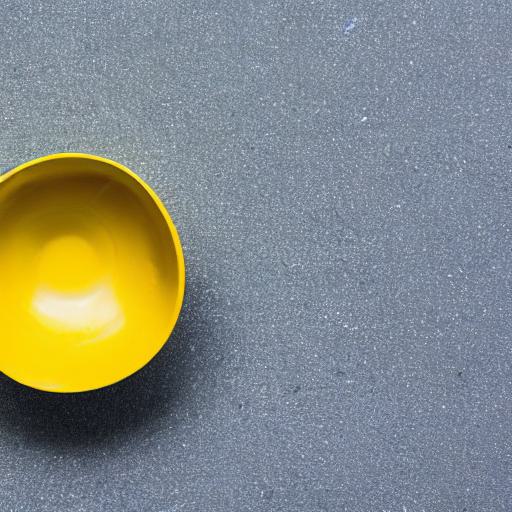} &
        \includegraphics[width=0.1015\textwidth]{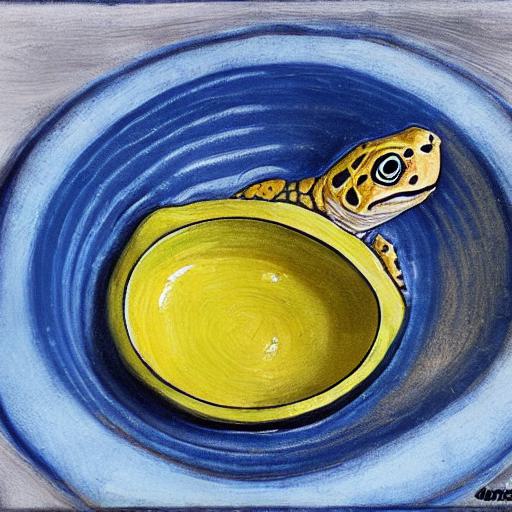} &
        \hspace{0.05cm}
        \includegraphics[width=0.1015\textwidth]{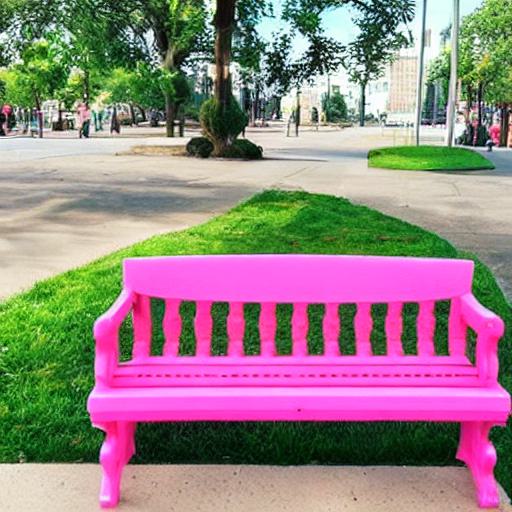} &
        \includegraphics[width=0.1015\textwidth]{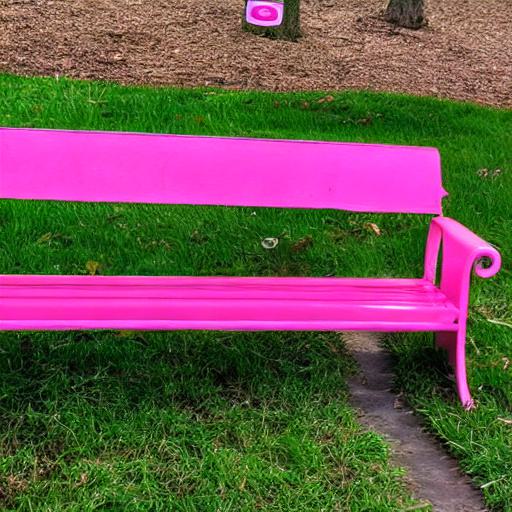} &
        \hspace{0.05cm}
        \includegraphics[width=0.1015\textwidth]{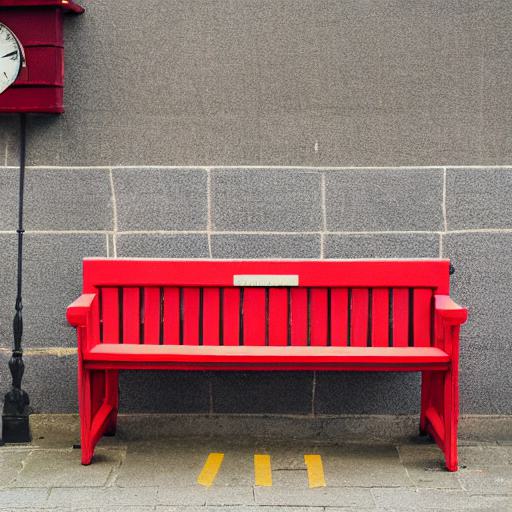} &
        \includegraphics[width=0.1015\textwidth]{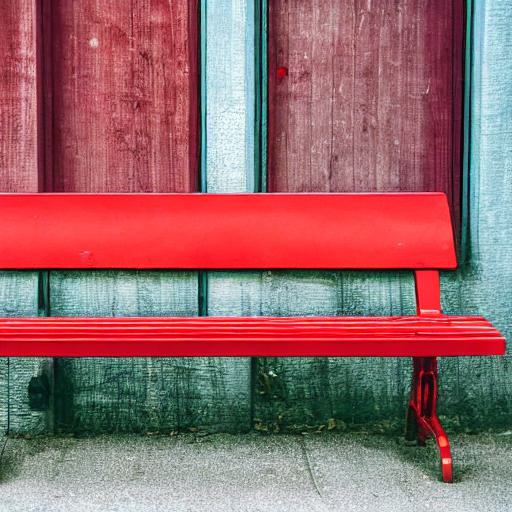} \\

        &
        \includegraphics[width=0.1015\textwidth]{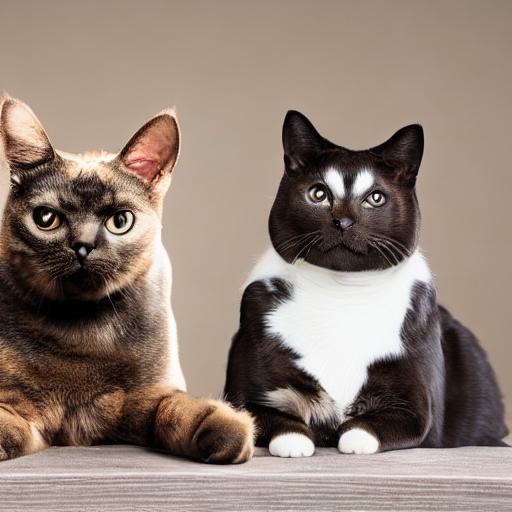} &
        \includegraphics[width=0.1015\textwidth]{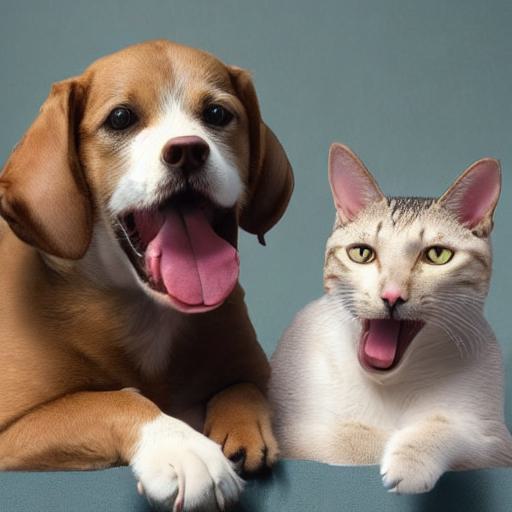} &
        \hspace{0.05cm}
        \includegraphics[width=0.1015\textwidth]{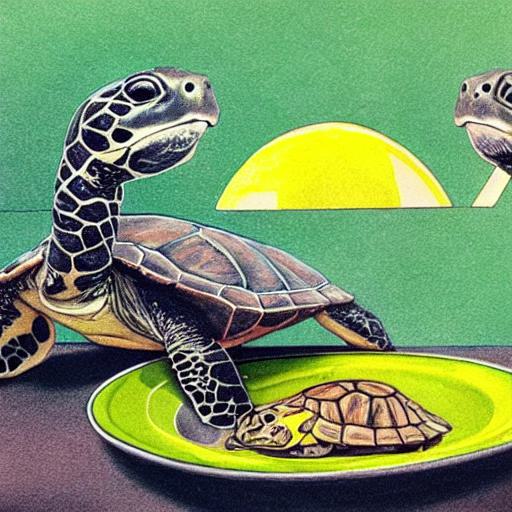} &
        \includegraphics[width=0.1015\textwidth]{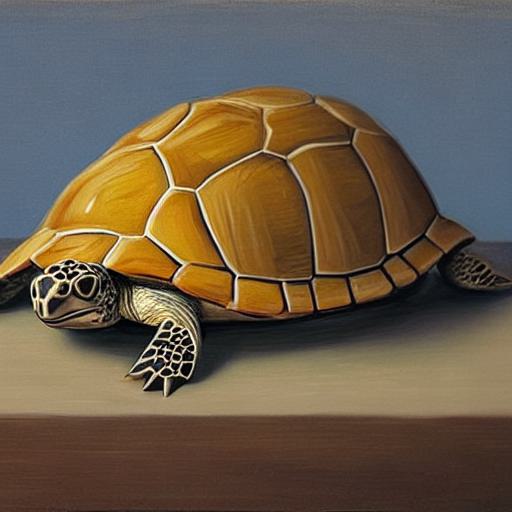} &
        \hspace{0.05cm}
        \includegraphics[width=0.1015\textwidth]{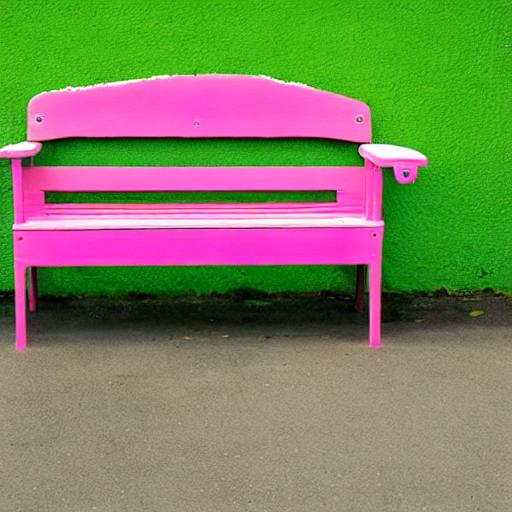} &
        \includegraphics[width=0.1015\textwidth]{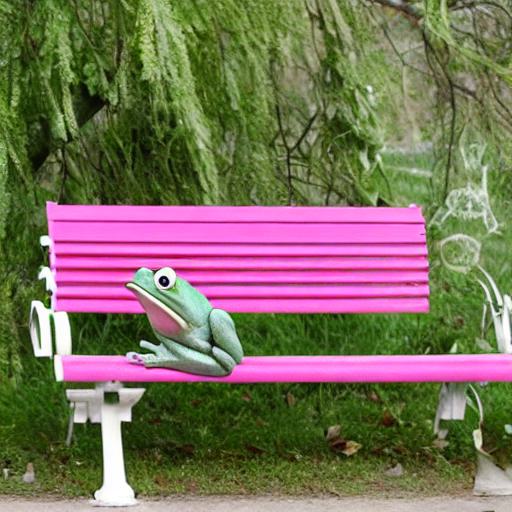} &
        \hspace{0.05cm}
        \includegraphics[width=0.1015\textwidth]{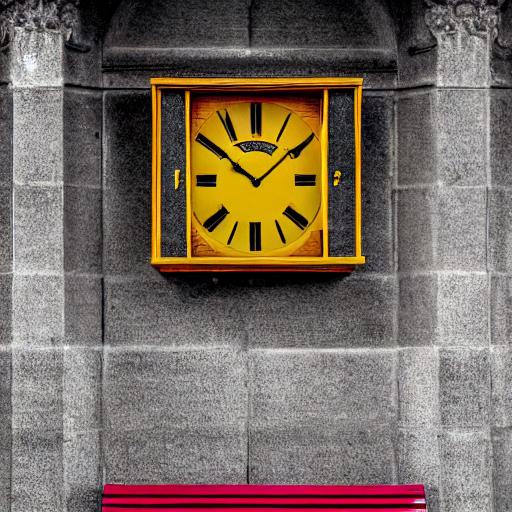} &
        \includegraphics[width=0.1015\textwidth]{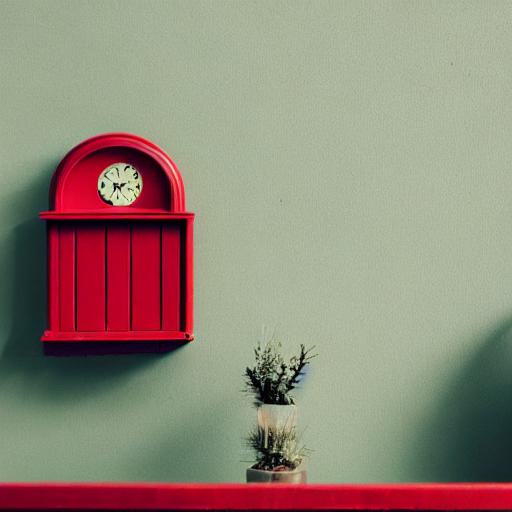}  \\ \\

        {\raisebox{0.425in}{
        \multirow{2}{*}{\rotatebox{90}{\begin{tabular}{c} Stable Diffusion with \\ \textcolor{blue}{Attend-and-Excite} \\ \\ \end{tabular}}}}} &
        \includegraphics[width=0.1015\textwidth]{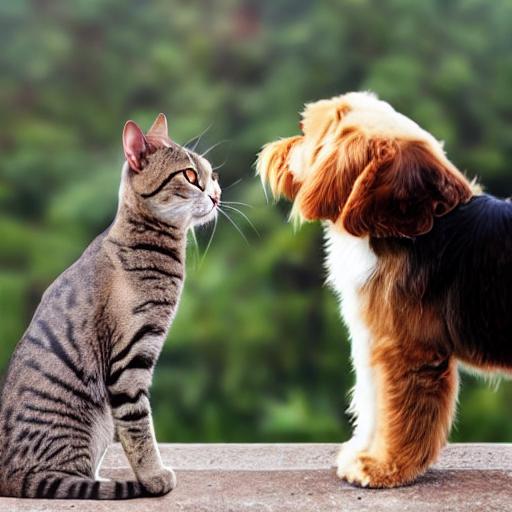} &
        \includegraphics[width=0.1015\textwidth]{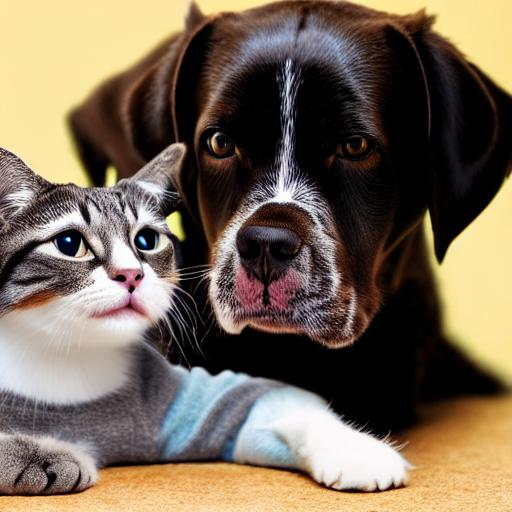} &
        \hspace{0.05cm}
         \includegraphics[width=0.1015\textwidth]{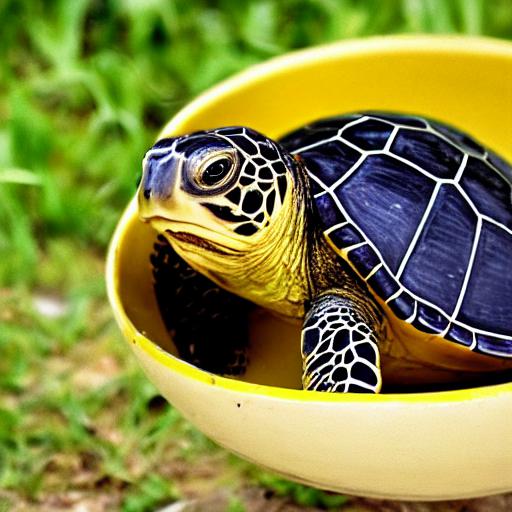} &
        \includegraphics[width=0.1015\textwidth]{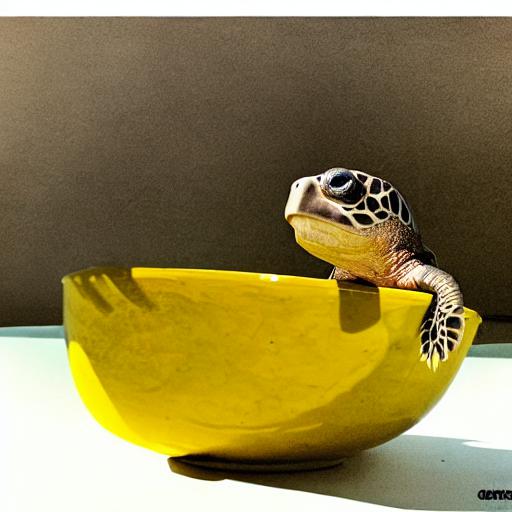} &
        \hspace{0.05cm}
        \includegraphics[width=0.1015\textwidth]{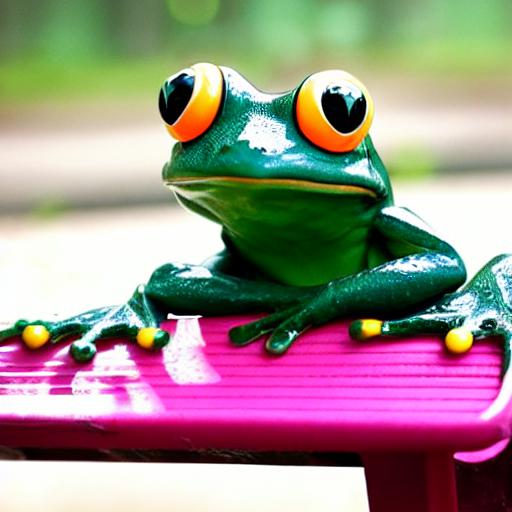} &
        \includegraphics[width=0.1015\textwidth]{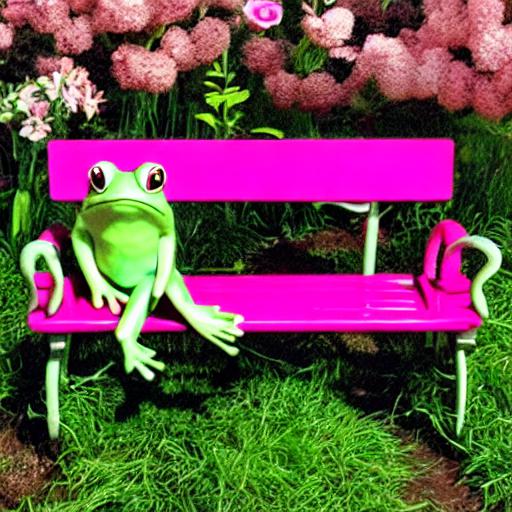} &
        \hspace{0.05cm}
        \includegraphics[width=0.1015\textwidth]{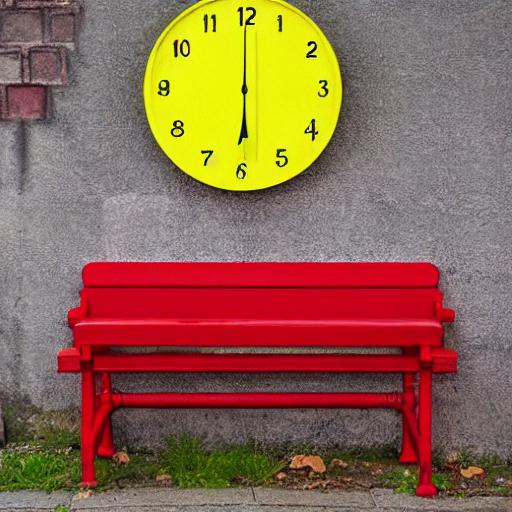} &
        \includegraphics[width=0.1015\textwidth]{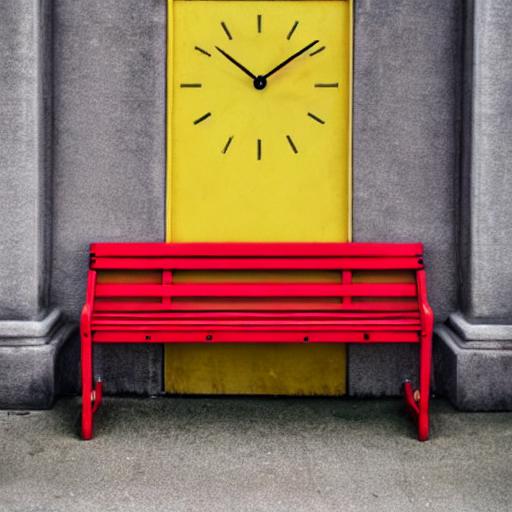}  \\

        &
        \includegraphics[width=0.1015\textwidth]{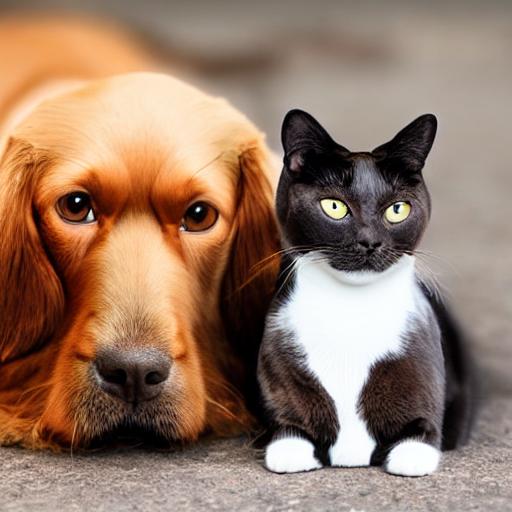} &
        \includegraphics[width=0.1015\textwidth]{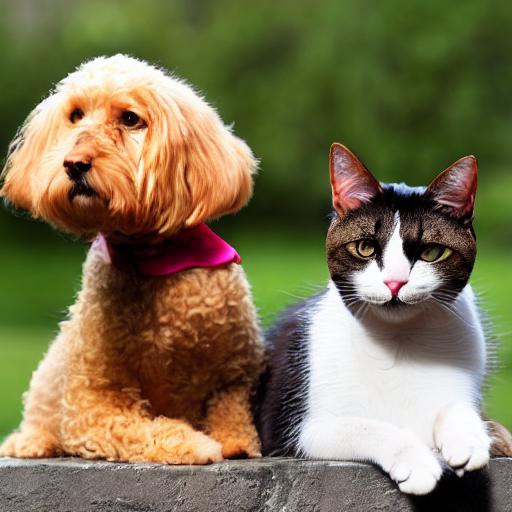} &
        \hspace{0.05cm}
        \includegraphics[width=0.1015\textwidth]{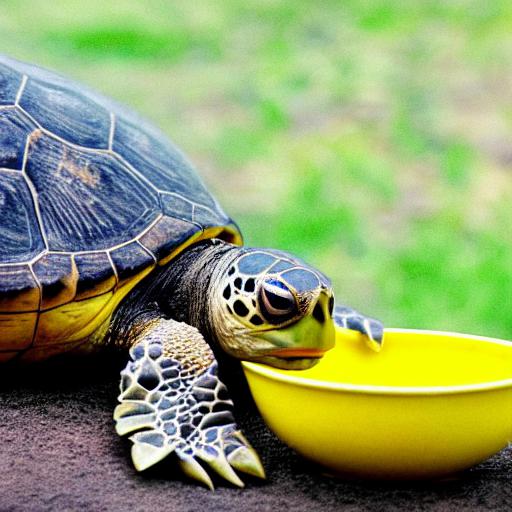} &
        \includegraphics[width=0.1015\textwidth]{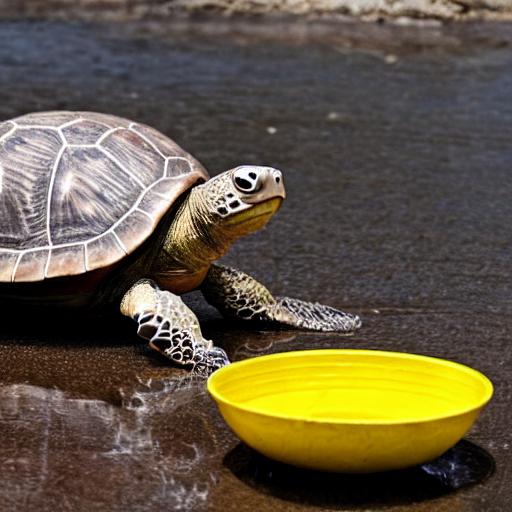} &
        \hspace{0.05cm}
        \includegraphics[width=0.1015\textwidth]{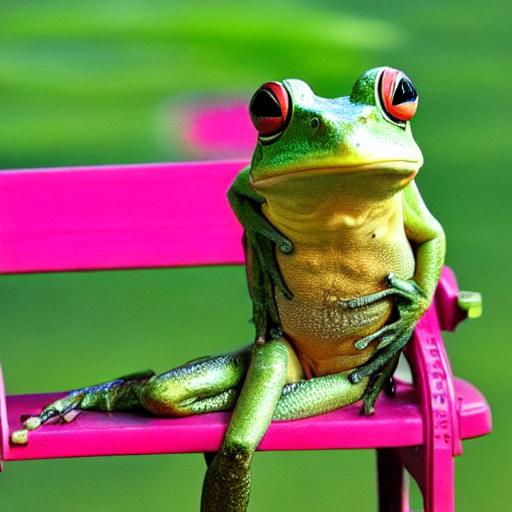} &
        \includegraphics[width=0.1015\textwidth]{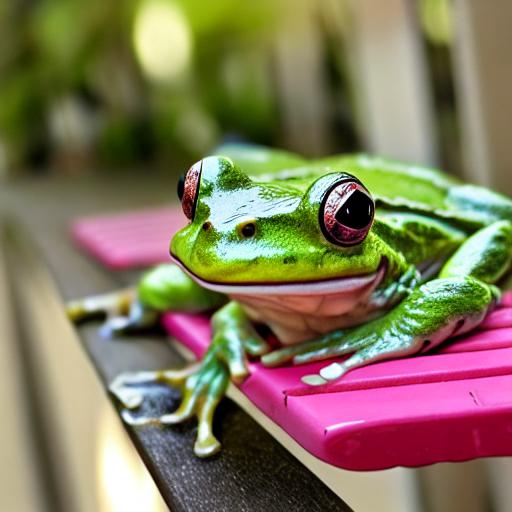} &
        \hspace{0.05cm}
        \includegraphics[width=0.1015\textwidth]{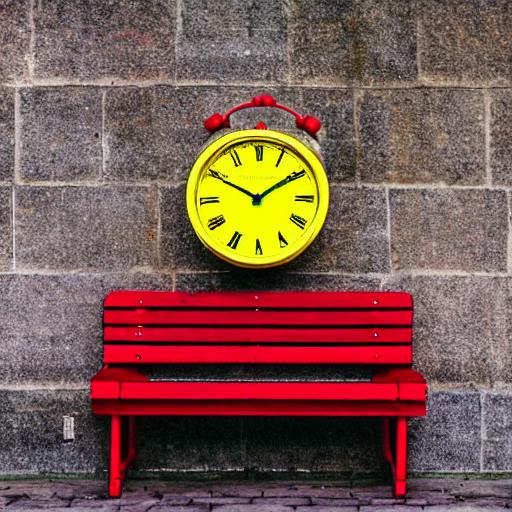} &
        \includegraphics[width=0.1015\textwidth]{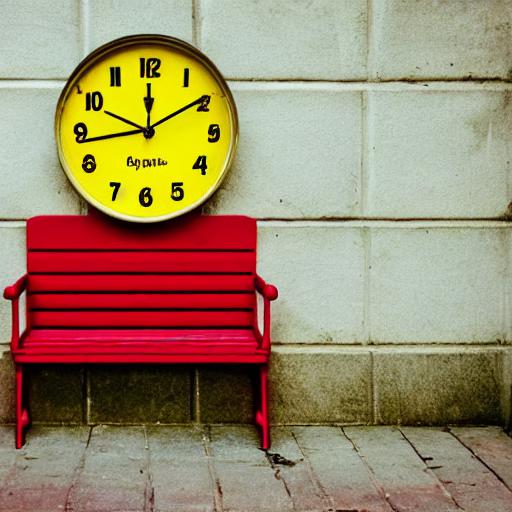}

    \end{tabular}
    \\[-0.3cm]
    }
    \caption{{Qualitative Comparison using prompts from our dataset. For each prompt, we show four images generated by each of the four considered methods where we use the same set of seeds across all approaches.
    The subject tokens optimized by Attend-and-Excite are highlighted in \textcolor{blue}{blue}.}
    }
    \label{fig:our results}
\end{figure*}

\paragraph{Iterative Latent Refinement}
So far, we have made a single latent update at each denoising timestep. However, if the attention values of a token do not reach a certain value in the early denoising stages, the corresponding object will not be generated. Therefore, we iteratively update $z_t$ until a pre-defined minimum attention value is achieved for \textit{all} subject tokens. Yet, many updates of $z_t$ may lead to the latent becoming out-of-distribution, resulting in incoherent images. As such, this refinement is performed \textit{gradually} across a \textit{small} subset of timesteps.

Specifically, we demand that each subject token reaches a maximum attention value of at least $0.8$. To do so gradually, we perform the iterative updates at various denoising steps (Steps $10$-$15$ in~\Cref{alg:attend-and-excite}). We set the iterations to $t_1=0, t_2=10$, and $t_3=20$ with minimum required attention values of $T_1=0.05, T_2=0.5$, and $T_3=0.8$. 
This gradual refinement prevents $z_t$ from becoming out-of-distribution while encouraging more faithful generations.

\paragraph{\textbf{Obtaining Explainable Image Generators}}
The extent to which attention can be used as an explanation has been widely explored~\cite{Abnar2020QuantifyingAF,Chefer_2021_ICCV,chefer2022optimizing}. In the context of text-based image generation, the cross-attention maps have been considered a natural explanation for the model~\cite{hertz2022prompt}. 

\begin{figure*}
    \centering
    \setlength{\tabcolsep}{0.5pt}
    \renewcommand{\arraystretch}{0.3}
    \addtolength{\belowcaptionskip}{-5pt}
    {\small
    \begin{tabular}{c c c @{\hspace{0.1cm}} c c @{\hspace{0.1cm}} c c @{\hspace{0.1cm}} c c }

        &
        \multicolumn{2}{c}{\begin{tabular}{c} ``A grizzly \textcolor{blue}{bear} catching \\ a \textcolor{blue}{salmon} in a crystal clear \\ \textcolor{blue}{river} surrounded by a \textcolor{blue}{forest}''\end{tabular}} &
        \multicolumn{2}{c}{\begin{tabular}{c}``A pod of \textcolor{blue}{dolphins} leaping \\ out of the \textcolor{blue}{water} in an \textcolor{blue}{ocean} \\ with a \textcolor{blue}{ship} on the background'' \end{tabular}} &
        \multicolumn{2}{c}{\begin{tabular}{c} ``A Picasso  \\ \textcolor{blue}{painting} in a \textcolor{blue}{garden}'' \\\\
        \end{tabular}} &
         \multicolumn{2}{c}{\begin{tabular}{c} ``A \textcolor{blue}{cat} and a \textcolor{blue}{dog} \\  reading in the \textcolor{blue}{library}'' \\\\
        \end{tabular}} \\

        {\raisebox{0.3in}{
        \multirow{2}{*}{\rotatebox{90}{Stable Diffusion}}}} &
        \includegraphics[width=0.105\textwidth]{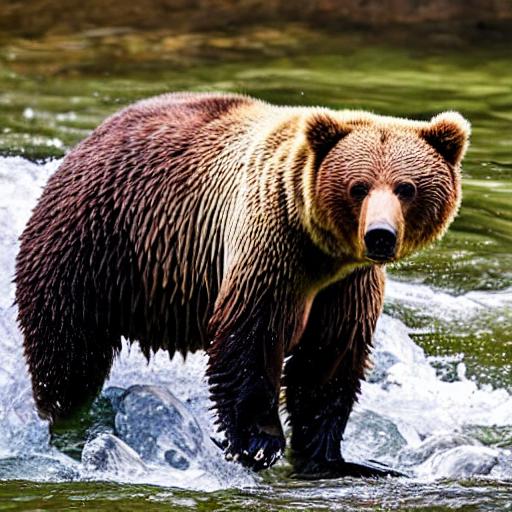} &
        \includegraphics[width=0.105\textwidth]{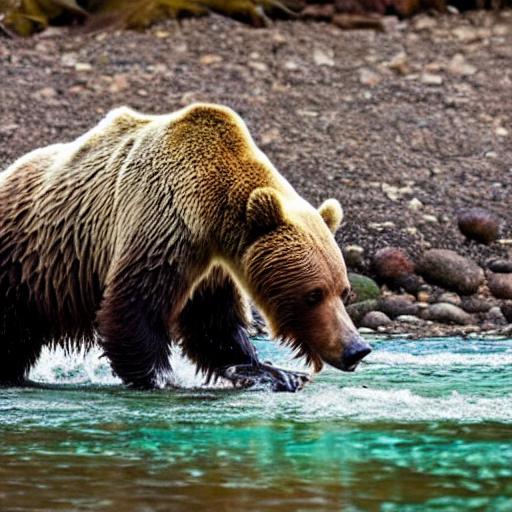} &
        \hspace{0.05cm}
        \includegraphics[width=0.105\textwidth]{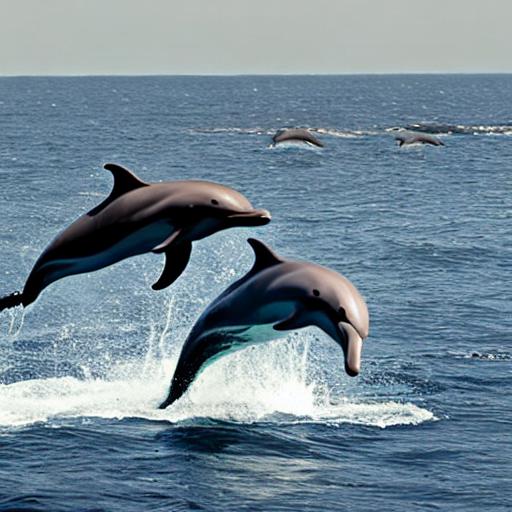} &
        \includegraphics[width=0.105\textwidth]{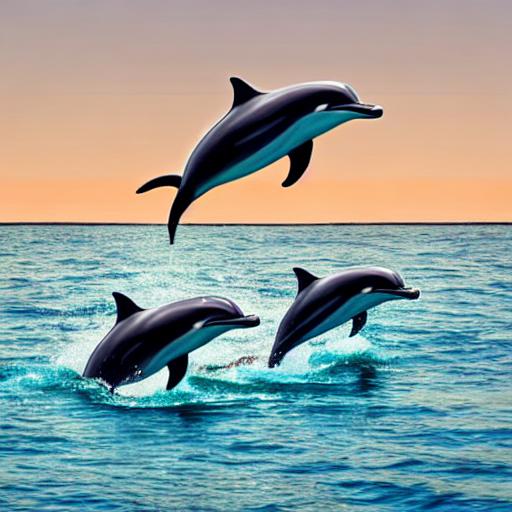} &
        \hspace{0.05cm}
        \includegraphics[width=0.105\textwidth]{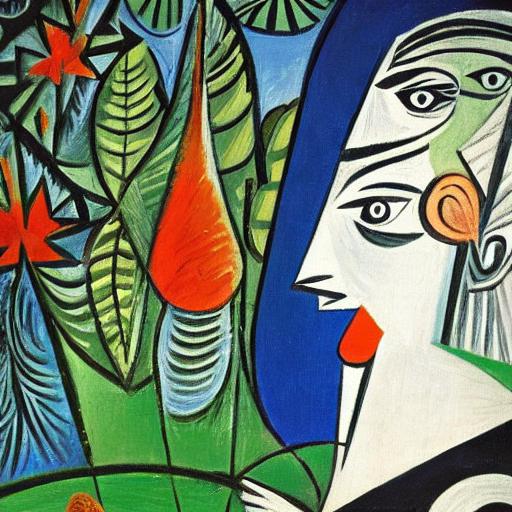} &
        \includegraphics[width=0.105\textwidth]{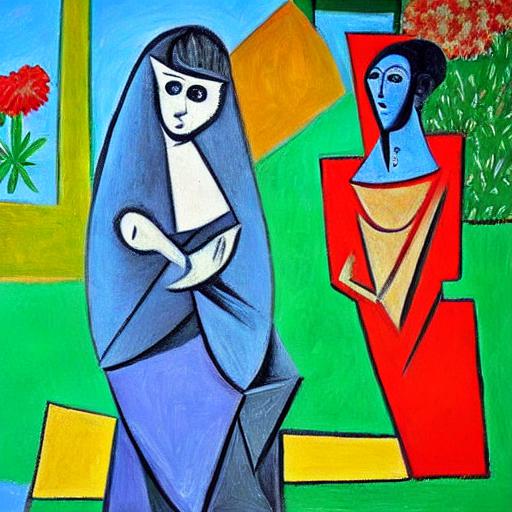} &
        \hspace{0.05cm}
        \includegraphics[width=0.105\textwidth]{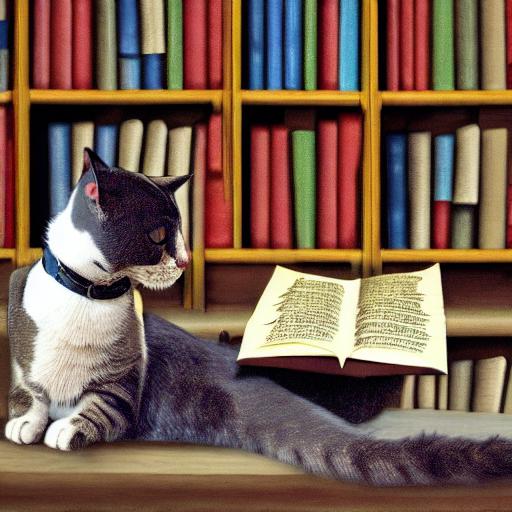} &
        \includegraphics[width=0.105\textwidth]{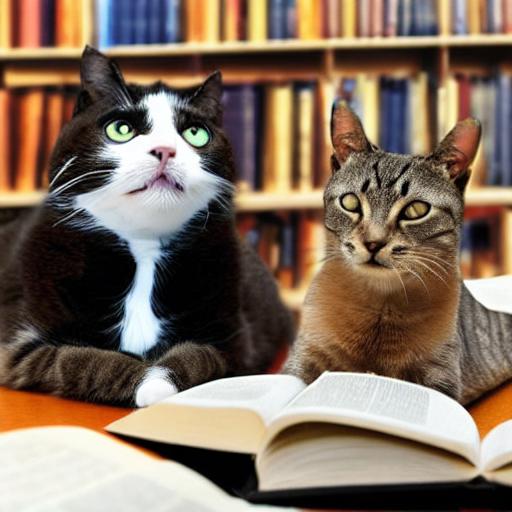} \\

        &
        \includegraphics[width=0.105\textwidth]{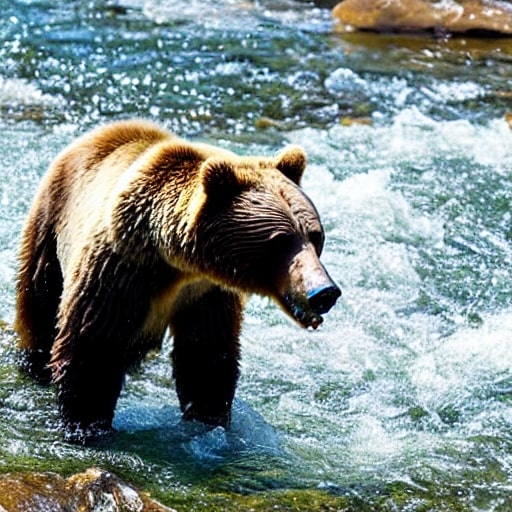} &
        \includegraphics[width=0.105\textwidth]{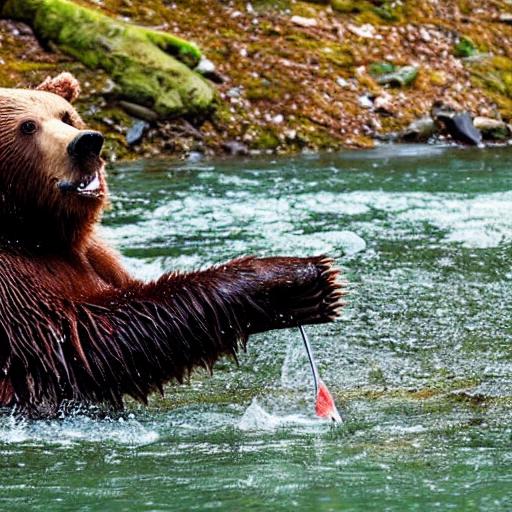} &
        \hspace{0.05cm}
        \includegraphics[width=0.105\textwidth]{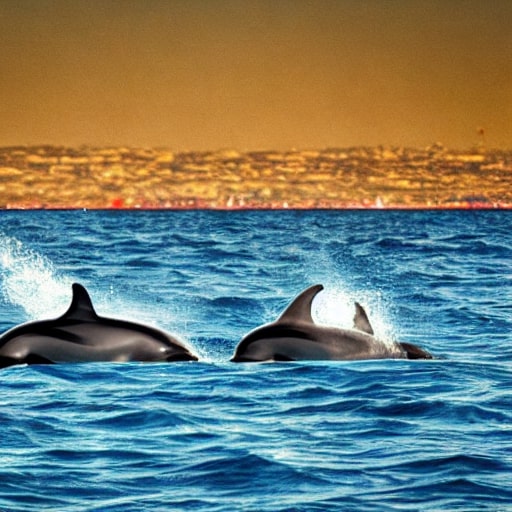} &
        \includegraphics[width=0.105\textwidth]{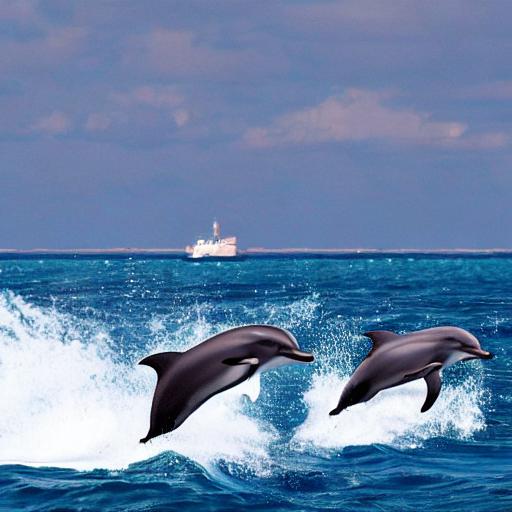} &
        \hspace{0.05cm}
        \includegraphics[width=0.105\textwidth]{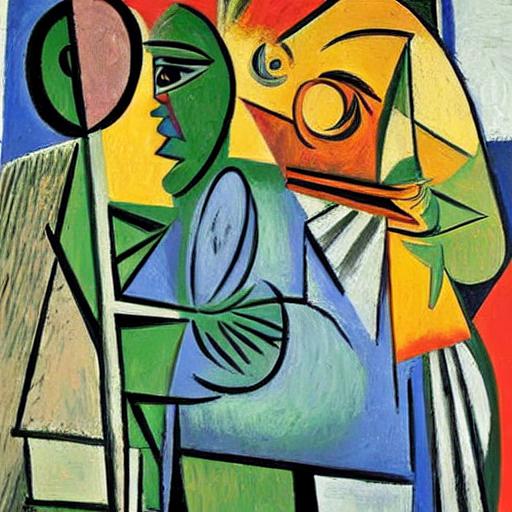} &
        \includegraphics[width=0.105\textwidth]{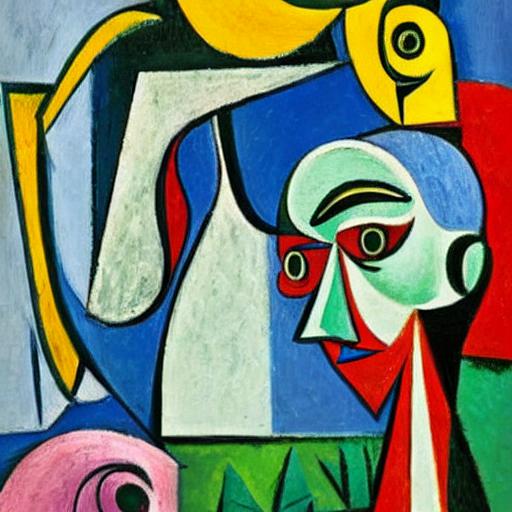} &
        \hspace{0.05cm}
        \includegraphics[width=0.105\textwidth]{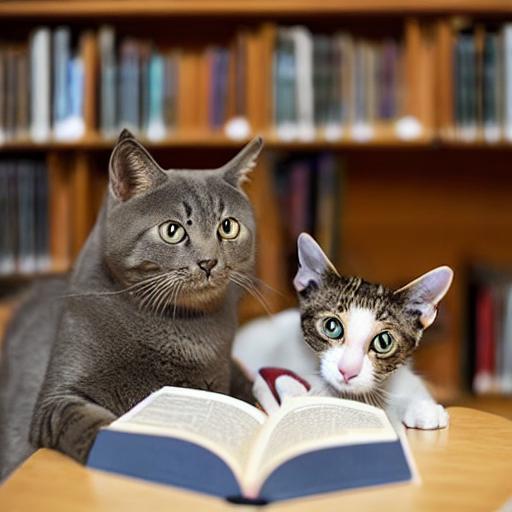} &
        \includegraphics[width=0.105\textwidth]{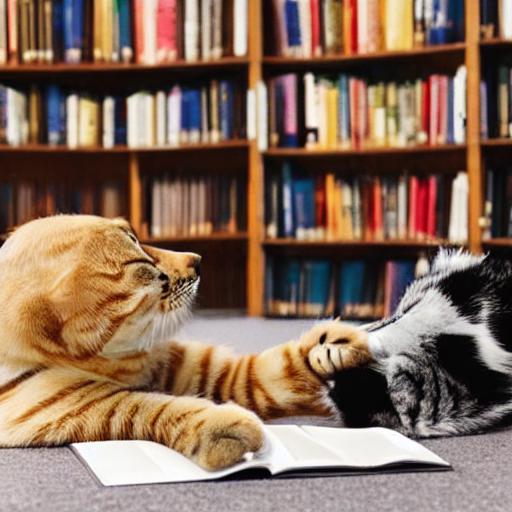} \\ \\ \\

         {\raisebox{0.425in}{
        \multirow{2}{*}{\rotatebox{90}{\begin{tabular}{c} Stable Diffusion with \\ \textcolor{blue}{Attend-and-Excite} \\ \\ \end{tabular}}}}} &
        \includegraphics[width=0.105\textwidth]{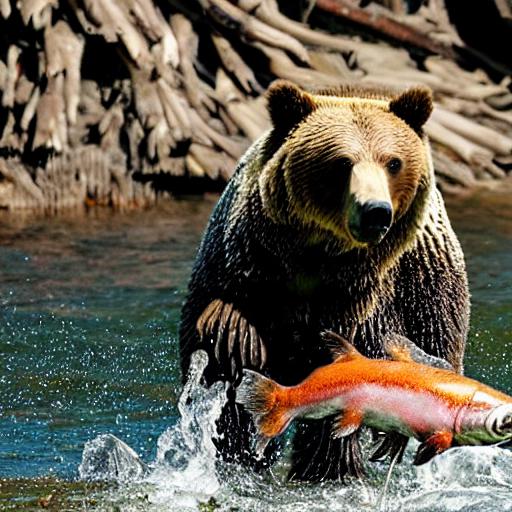} &
        \includegraphics[width=0.105\textwidth]{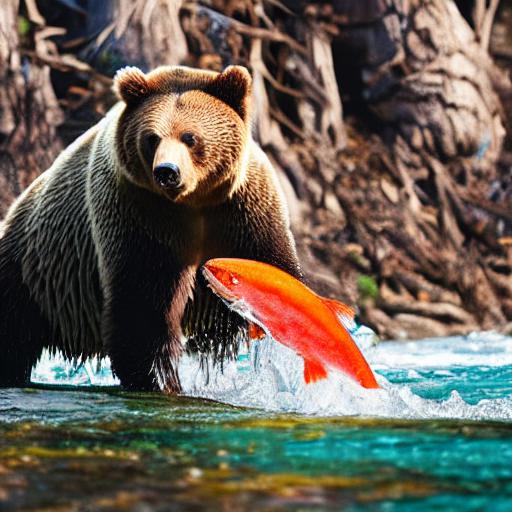} &
        \hspace{0.05cm}
        \includegraphics[width=0.105\textwidth]{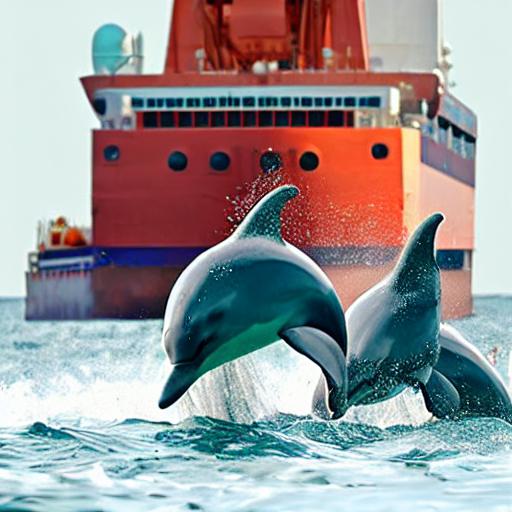} &
        \includegraphics[width=0.105\textwidth]{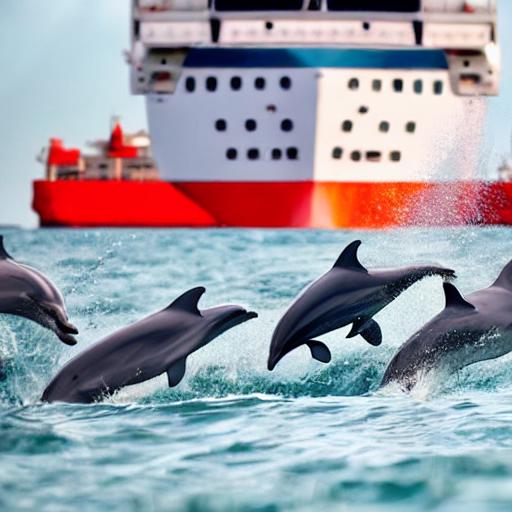} &
        \hspace{0.05cm}
        \includegraphics[width=0.105\textwidth]{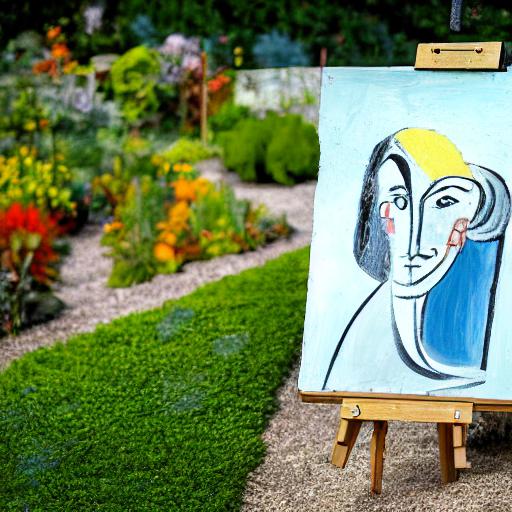} &
        \includegraphics[width=0.105\textwidth]{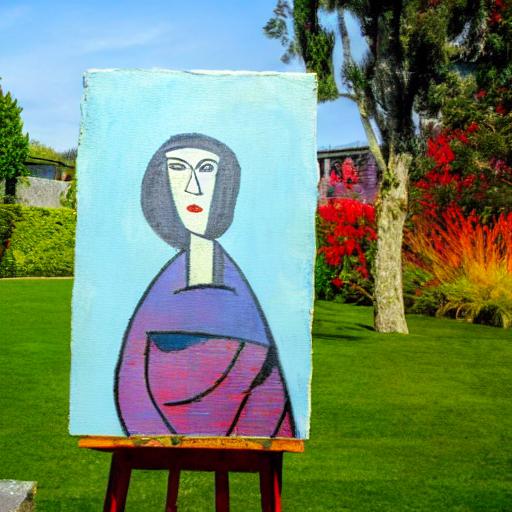} &
        \hspace{0.05cm}
        \includegraphics[width=0.105\textwidth]{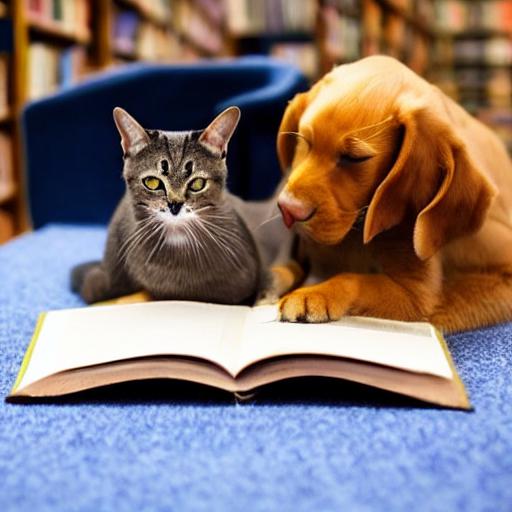} &
        \includegraphics[width=0.105\textwidth]{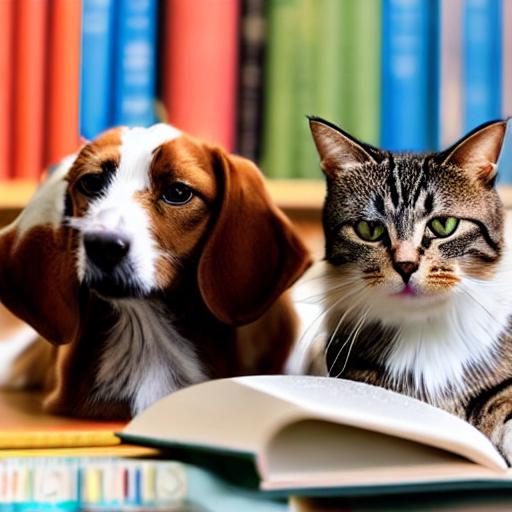} \\

        &
        \includegraphics[width=0.105\textwidth]{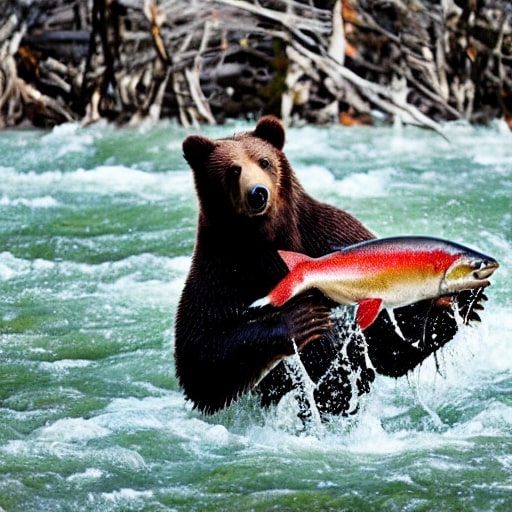} &
        \includegraphics[width=0.105\textwidth]{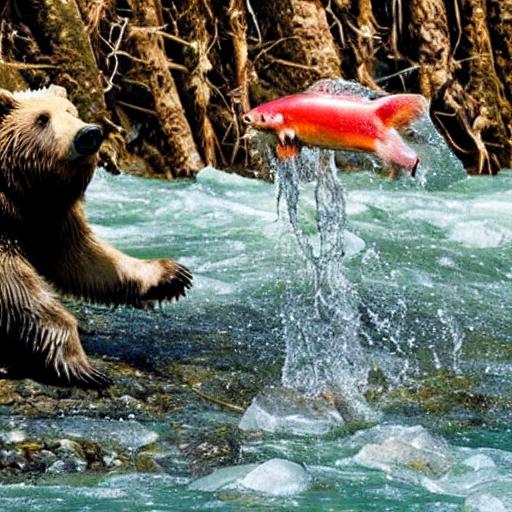} &
        \hspace{0.05cm}
        \includegraphics[width=0.105\textwidth]{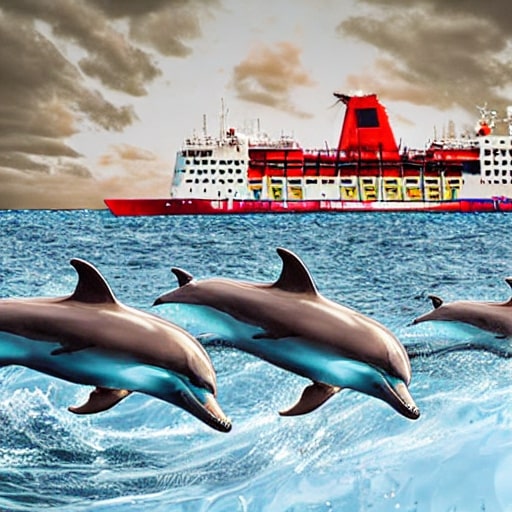} &
        \includegraphics[width=0.105\textwidth]{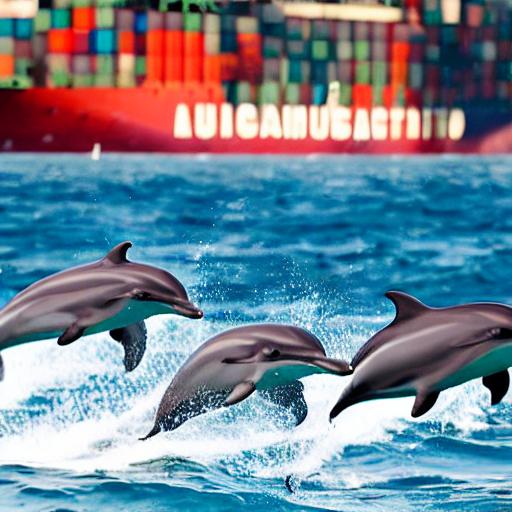} &
        \hspace{0.05cm}
        \includegraphics[width=0.105\textwidth]{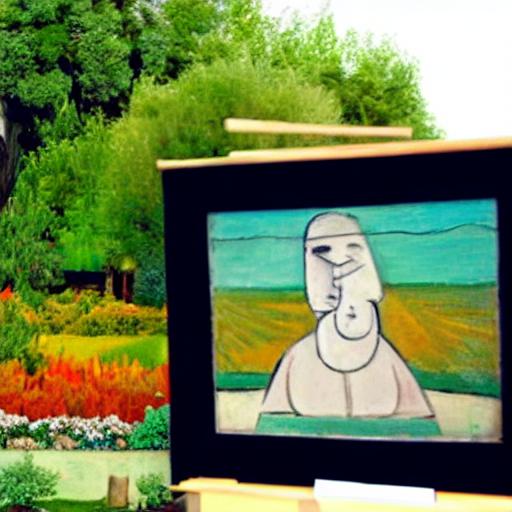} &
        \includegraphics[width=0.105\textwidth]{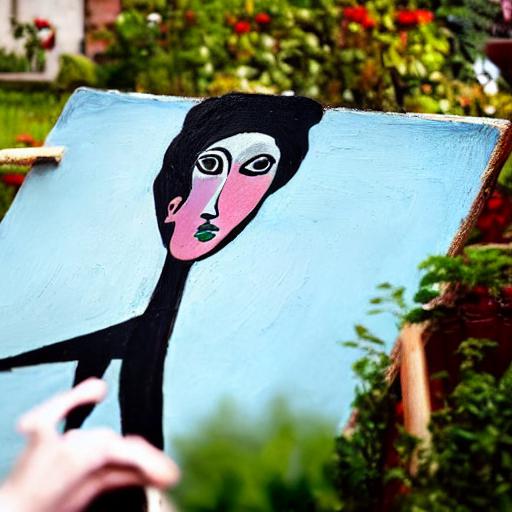} &
        \hspace{0.05cm}
        \includegraphics[width=0.105\textwidth]{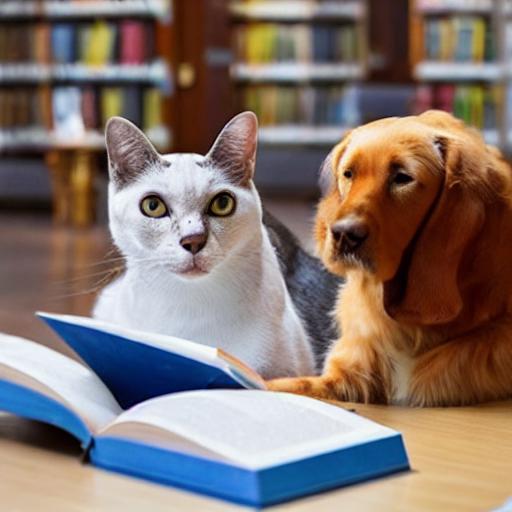} &
        \includegraphics[width=0.105\textwidth]{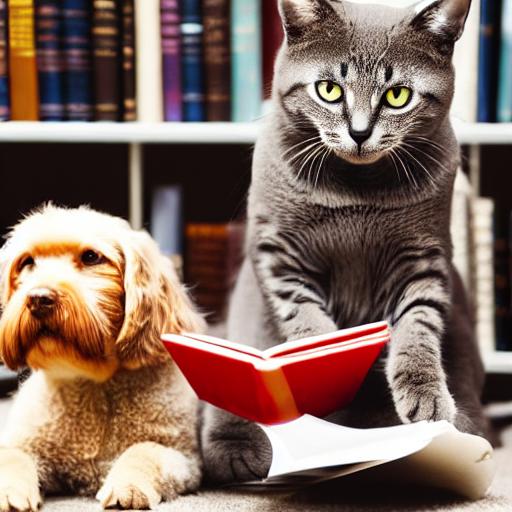} \\[-0.3cm]
    \end{tabular}
    }
    \caption{Additional comparisons with Stable Diffusion using prompts describing complex scenes and multiple subject tokens. For each prompt, we show four generated images where we use the same set of seeds for both approaches.
    The subject tokens optimized by Attend-and-Excite are highlighted in \textcolor{blue}{blue}.
    }
    \label{fig:creative}
\end{figure*}

However, a direct result of catastrophic neglect is that the attention map corresponding to the neglected subject no longer faithfully represents the subject's localization in the generated image, as can be seen in the left column of~\Cref{fig:cls_spes}. 
While the cross-attention map for the cat is correctly localized, the map corresponding to the frog highlights irrelevant regions since a frog is not present. Thus, the cross-attention maps do not constitute viable explanations, as they are misleading and inaccurate. 
Conversely, as can be seen on the right of~\Cref{fig:cls_spes}, by  mitigating neglect using Attend-and-Excite, both the cat and the frog are accurately localized in the attention maps, and the maps can now be considered a faithful explanation.

\section{Results}~\label{sec:results}

\vspace*{-0.4cm}
\paragraph{\textbf{Evaluation Setup.}} As there are currently no openly-available datasets that analyze semantic issues in text-based image generation, we construct a new benchmark to evaluate all methods. 
To analyze the existence of catastrophic neglect, we construct prompts containing two subjects. Additionally, to test correct attribute binding, the prompts should contain a variety of attributes matched to the subject tokens. 
Specifically, we consider three types of text prompts: (i) ``a [\textcolor{red}{\textit{animalA}}] and a [\textcolor{red}{\textit{animalB}}]'', (ii) ``a [\textcolor{red}{\textit{animal}}] and a [\textcolor{orange}{\textit{color}}][\textcolor{mydarkgreen}{\textit{object}}]'', and (iii) ``a [\textcolor{orange}{\textit{colorA}}][\textcolor{mydarkgreen}{\textit{objectA}}] and a [\textcolor{orange}{\textit{colorB
}}][\textcolor{mydarkgreen}{\textit{objectB}}]''. 
To compose the prompts, we consider $12$ animals and $12$ object items with $11$ colors, detailed in~\Cref{sec:additional_details}. For each prompt containing a subject-color pair, we randomly select a color for the subject.
This results in $66$ Animal-Animal and Object-Object pairs and $144$ Animal-Object pairs. For each prompt, we then generate $64$ images using $64$ random seeds applied across all methods.

\begin{figure}
    \centering
    \setlength{\tabcolsep}{0.7pt}
    \renewcommand{\arraystretch}{0.55}
    \addtolength{\belowcaptionskip}{-10pt}
    {\small
    \begin{tabular}{c c c @{\hspace{0.15cm}} c c c }

        \multicolumn{3}{c}{StructureDiffusion} &
        \multicolumn{3}{c}{Attend-and-Excite} \\

        \includegraphics[width=0.0725\textwidth]{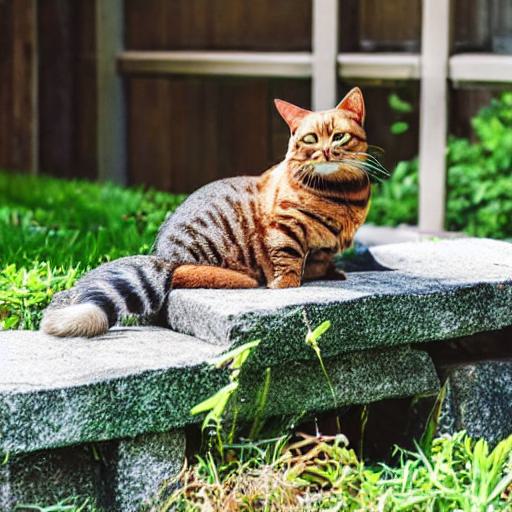} &
        \includegraphics[width=0.0725\textwidth]{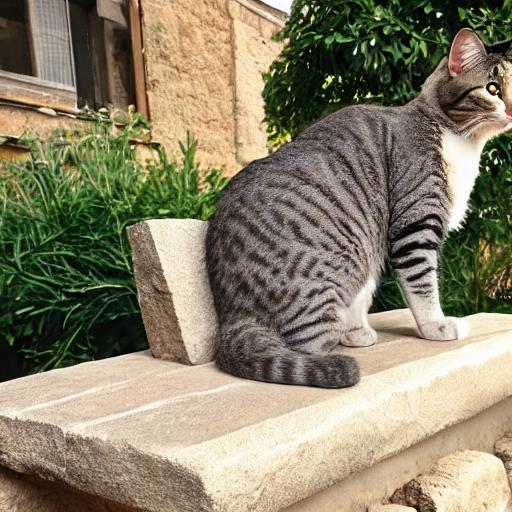} &
        \includegraphics[width=0.0725\textwidth]{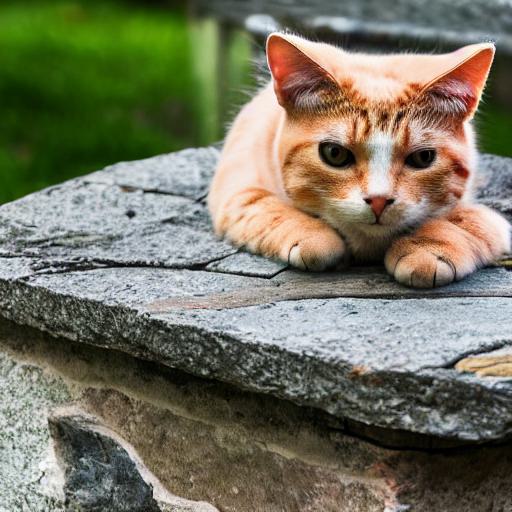} &
        \hspace{0.05cm}
        \includegraphics[width=0.0725\textwidth]{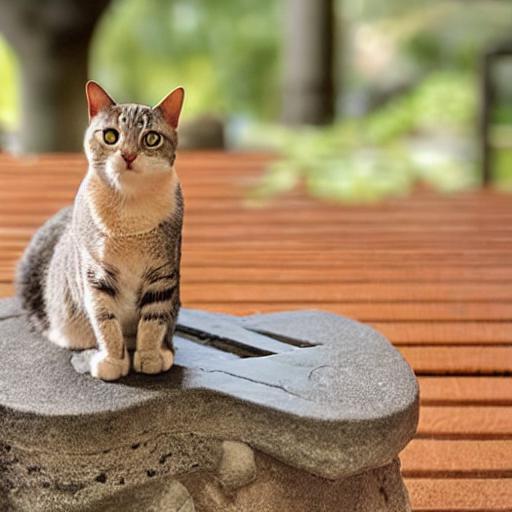} &
        \includegraphics[width=0.0725\textwidth]{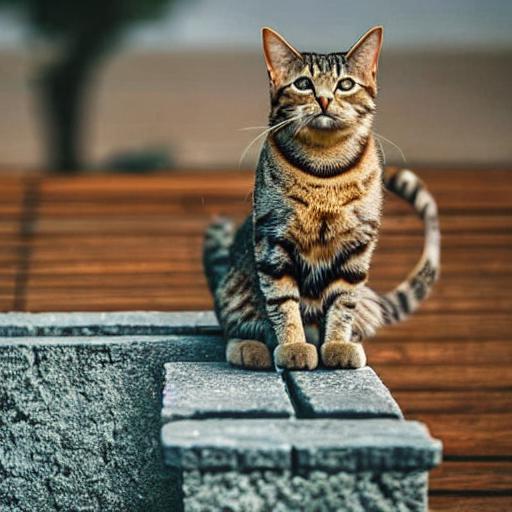} &
        \includegraphics[width=0.0725\textwidth]{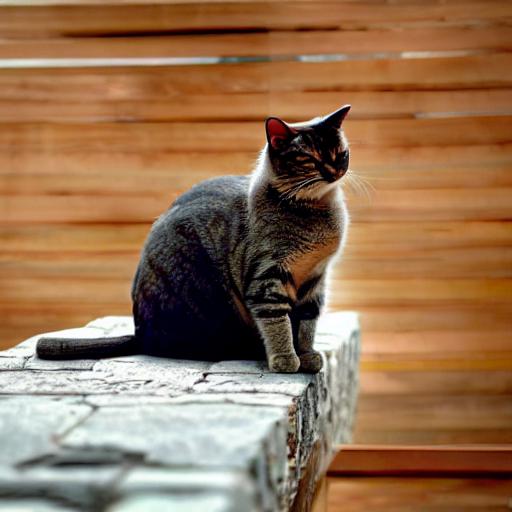} \\

        \multicolumn{6}{c}{\begin{tabular}{c} ``A \textcolor{blue}{cat} is perched upon a stone \textcolor{blue}{bench} that sits on a wooden \textcolor{blue}{patio}'' \end{tabular}} \\

        \includegraphics[width=0.0725\textwidth]{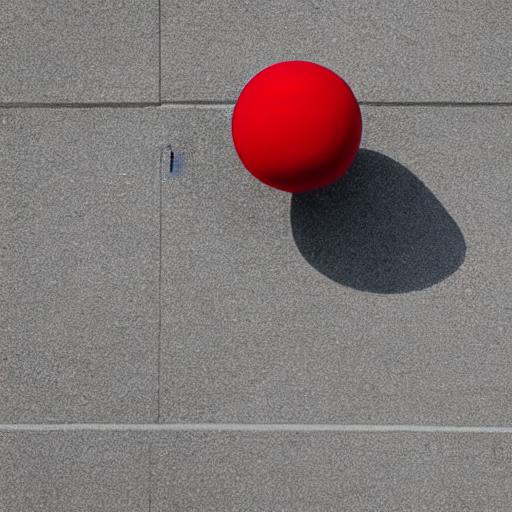} &
        \includegraphics[width=0.0725\textwidth]{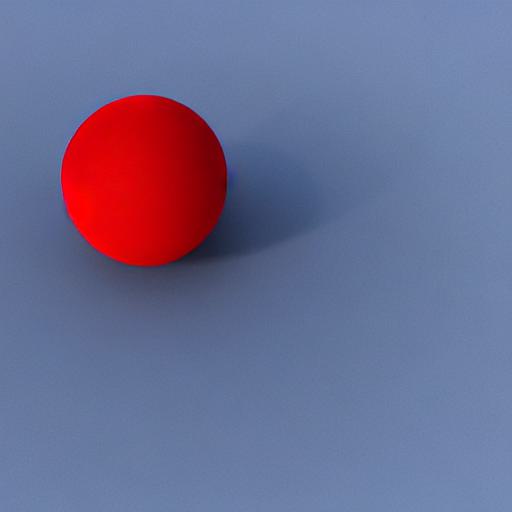} &
        \includegraphics[width=0.0725\textwidth]{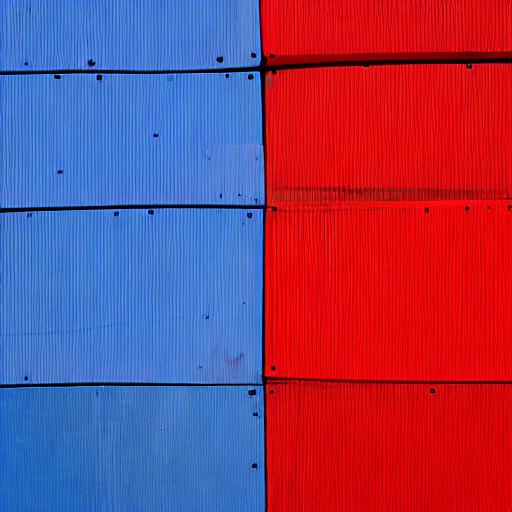} &
        \hspace{0.075cm}
        \includegraphics[width=0.0725\textwidth]{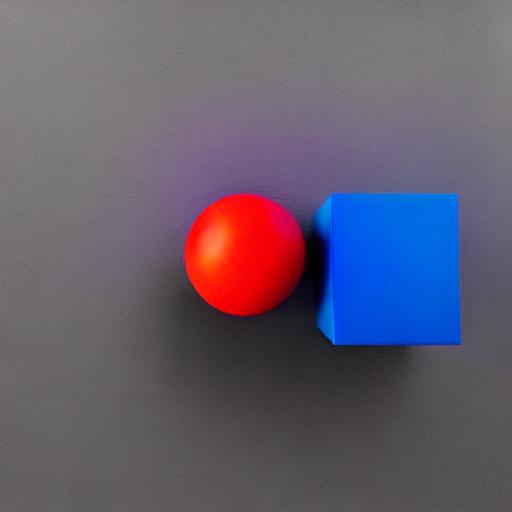} &
        \includegraphics[width=0.0725\textwidth]{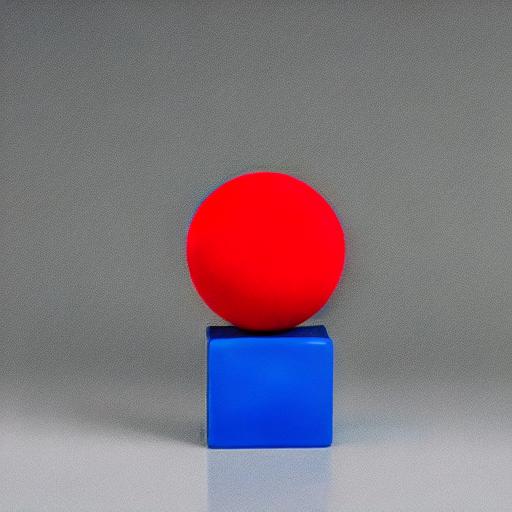} &
        \includegraphics[width=0.0725\textwidth]{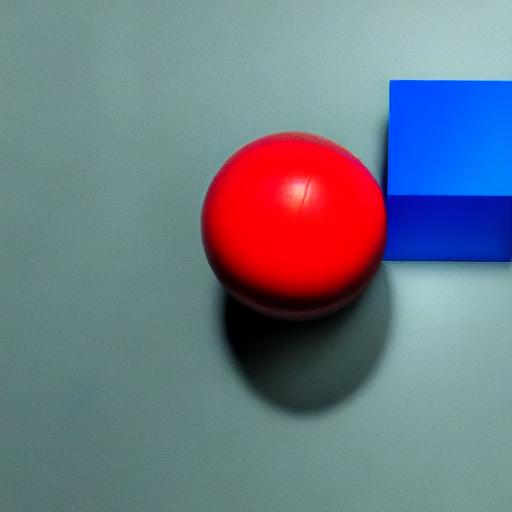} \\

        \multicolumn{6}{c}{\begin{tabular}{c} ``A red \textcolor{blue}{sphere} and a blue \textcolor{blue}{cube}'' \end{tabular}} \\

        \includegraphics[width=0.0725\textwidth]{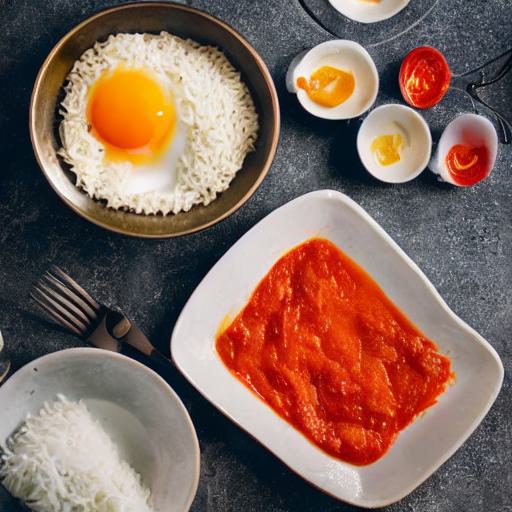} &
        \includegraphics[width=0.0725\textwidth]{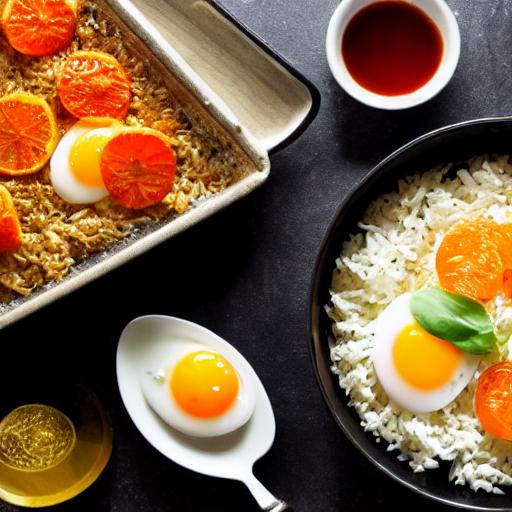} &
        \includegraphics[width=0.0725\textwidth]{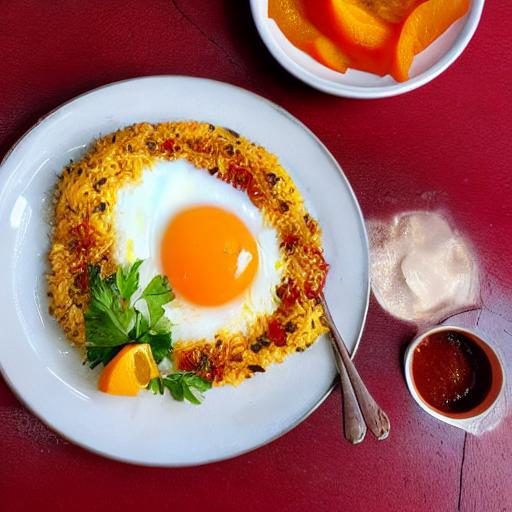} &
        \hspace{0.075cm}
        \includegraphics[width=0.0725\textwidth]{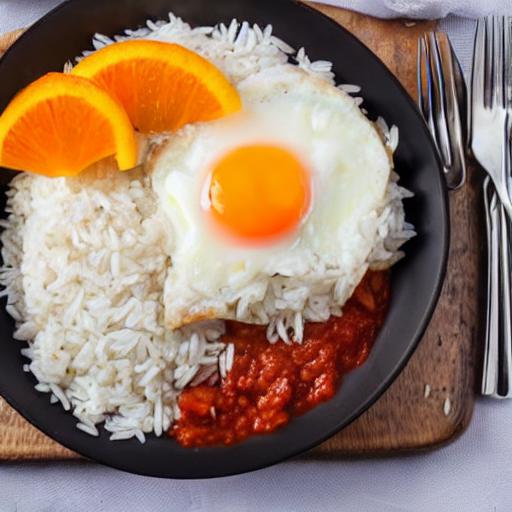} &
        \includegraphics[width=0.0725\textwidth]{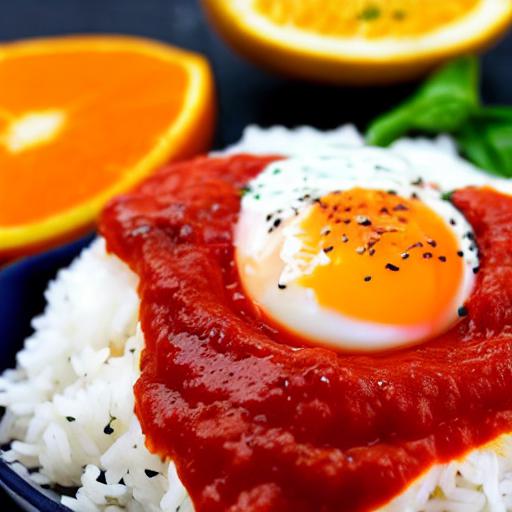} &
        \includegraphics[width=0.0725\textwidth]{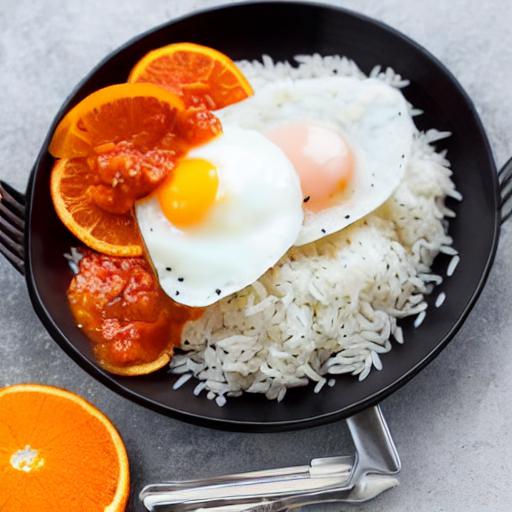} \\

        \multicolumn{6}{c}{\begin{tabular}{c} ``\textcolor{blue}{Rice} with red sauce with \textcolor{blue}{eggs} over the top and \textcolor{blue}{orange} slices on the side'' \end{tabular}} \\

        \includegraphics[width=0.0725\textwidth]{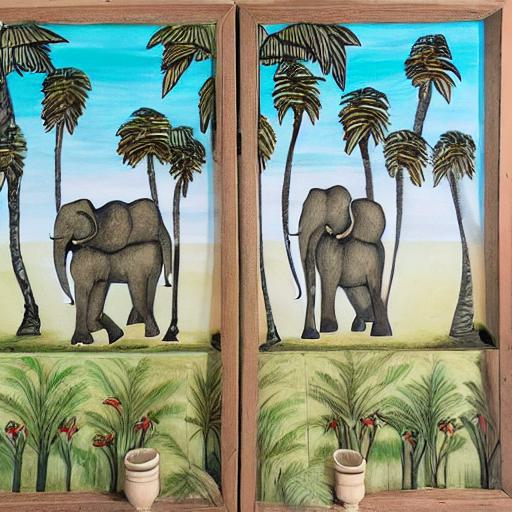} &
        \includegraphics[width=0.0725\textwidth]{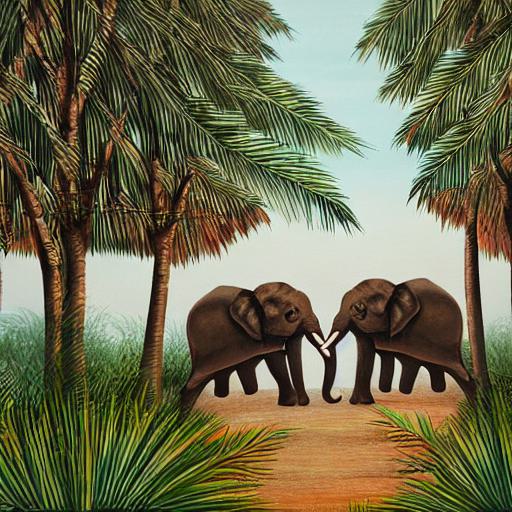} &
        \includegraphics[width=0.0725\textwidth]{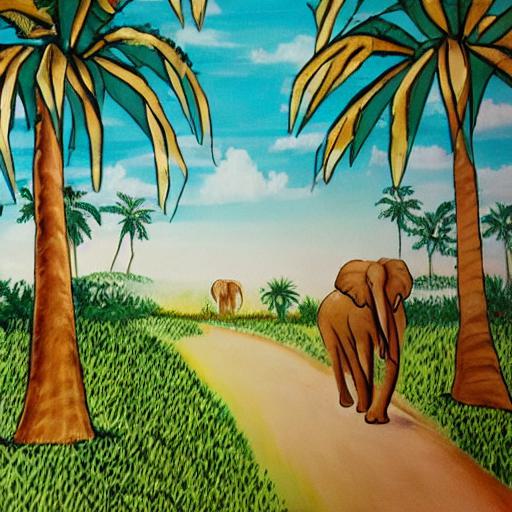} &
        \hspace{0.05cm}
        \includegraphics[width=0.0725\textwidth]{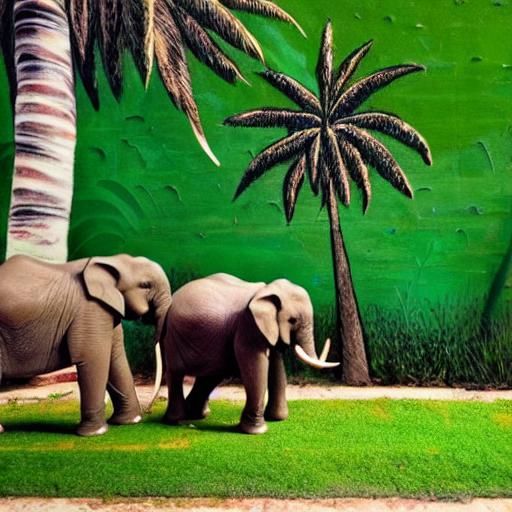} &
        \includegraphics[width=0.0725\textwidth]{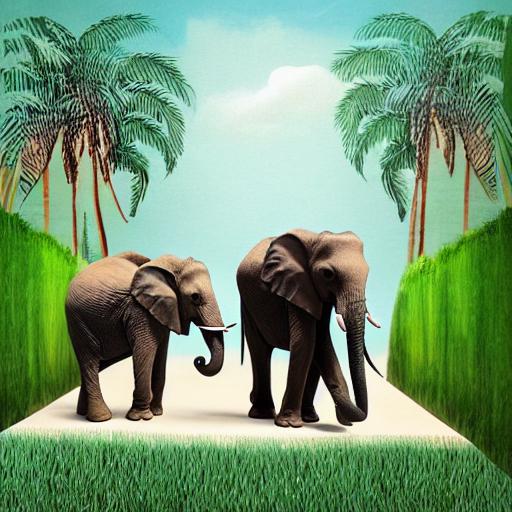} &
        \includegraphics[width=0.0725\textwidth]{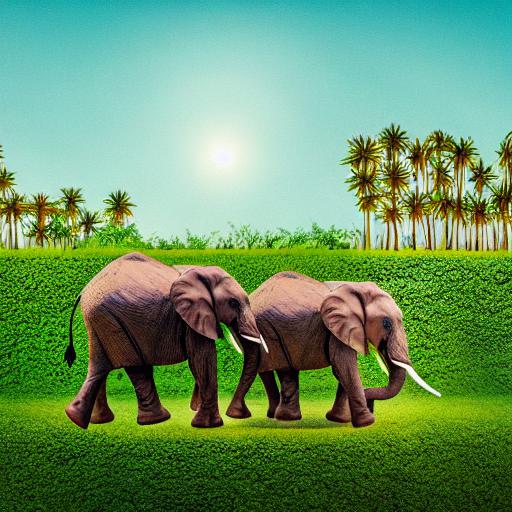}  \\

        \multicolumn{6}{c}{\begin{tabular}{c} ``Two \textcolor{blue}{elephants} walking by a green \textcolor{blue}{wall} with tan palm \textcolor{blue}{trees} painted on it'' \end{tabular}} \\ \\[-0.4cm]

    \end{tabular}

    }
    \caption{Comparison with prompts appearing in Feng~\etal~\shortcite{Feng2022Training}.
    For each prompt, we apply the same set of random seeds across the two methods.
    }
\label{fig:structured_prompts}
\end{figure}

For ease of evaluation, our prompts are constructed of conjunctions and color attributes. Yet, our method is not limited to such cases and can be applied to prompts with any number or type of subjects and attributes (see~\Cref{fig:creative,fig:structured_prompts} and~\Cref{sec:additional_results}).

\begin{figure*}
    \centering
    \begin{tabular}{c c c}
    \includegraphics[height=3.4cm]{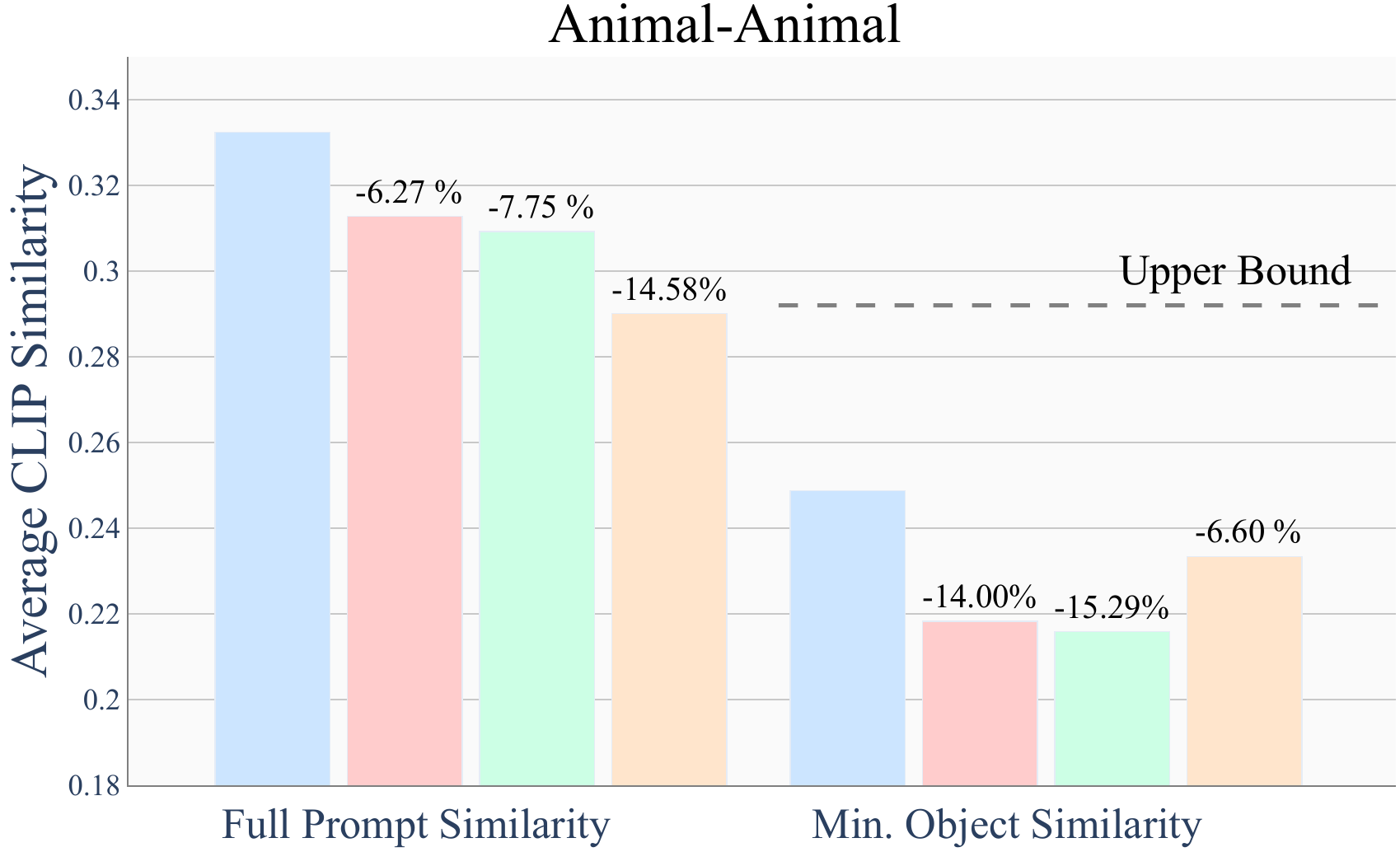} &
    \includegraphics[height=3.4cm]{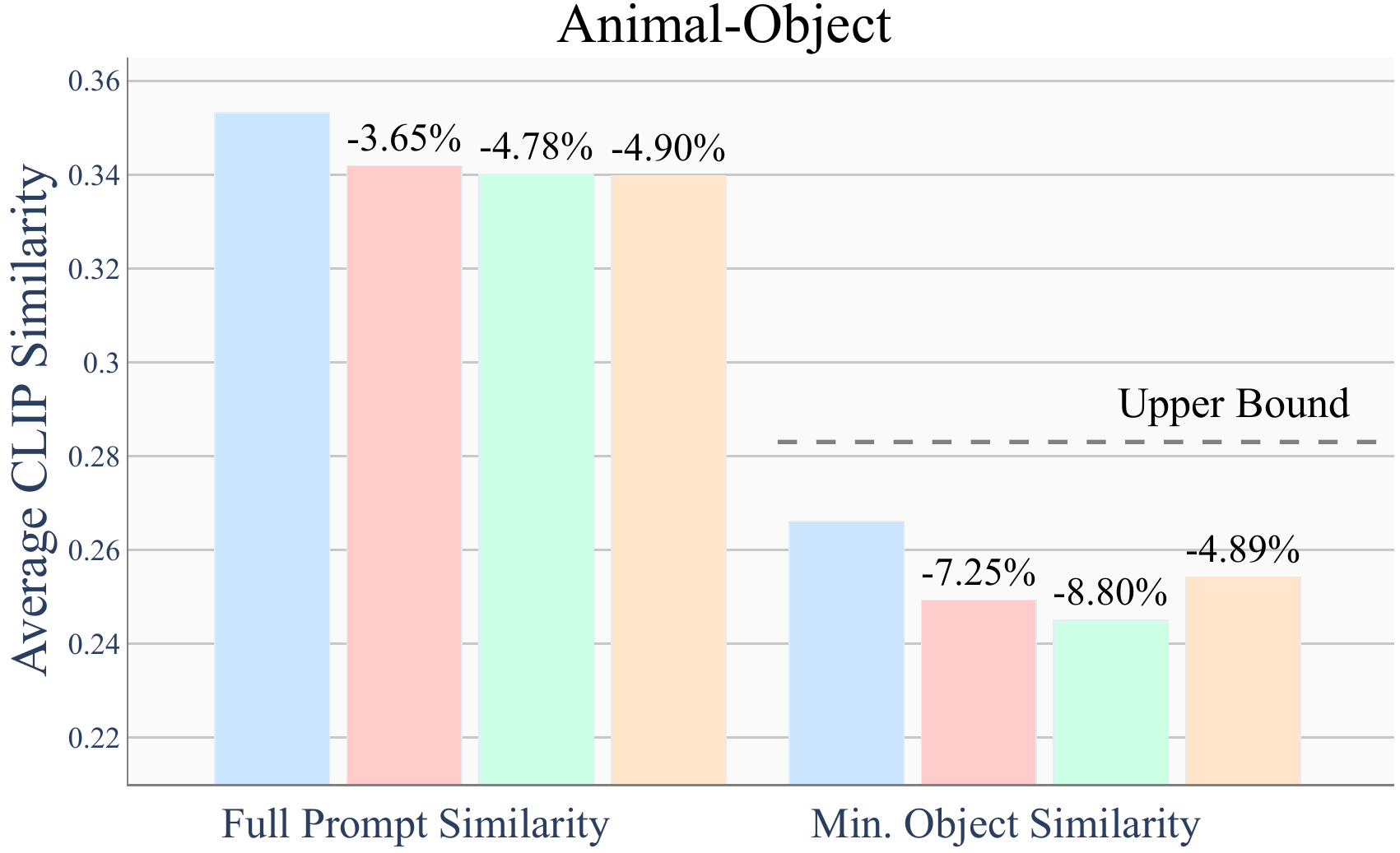}  &
    \includegraphics[height=3.4cm]{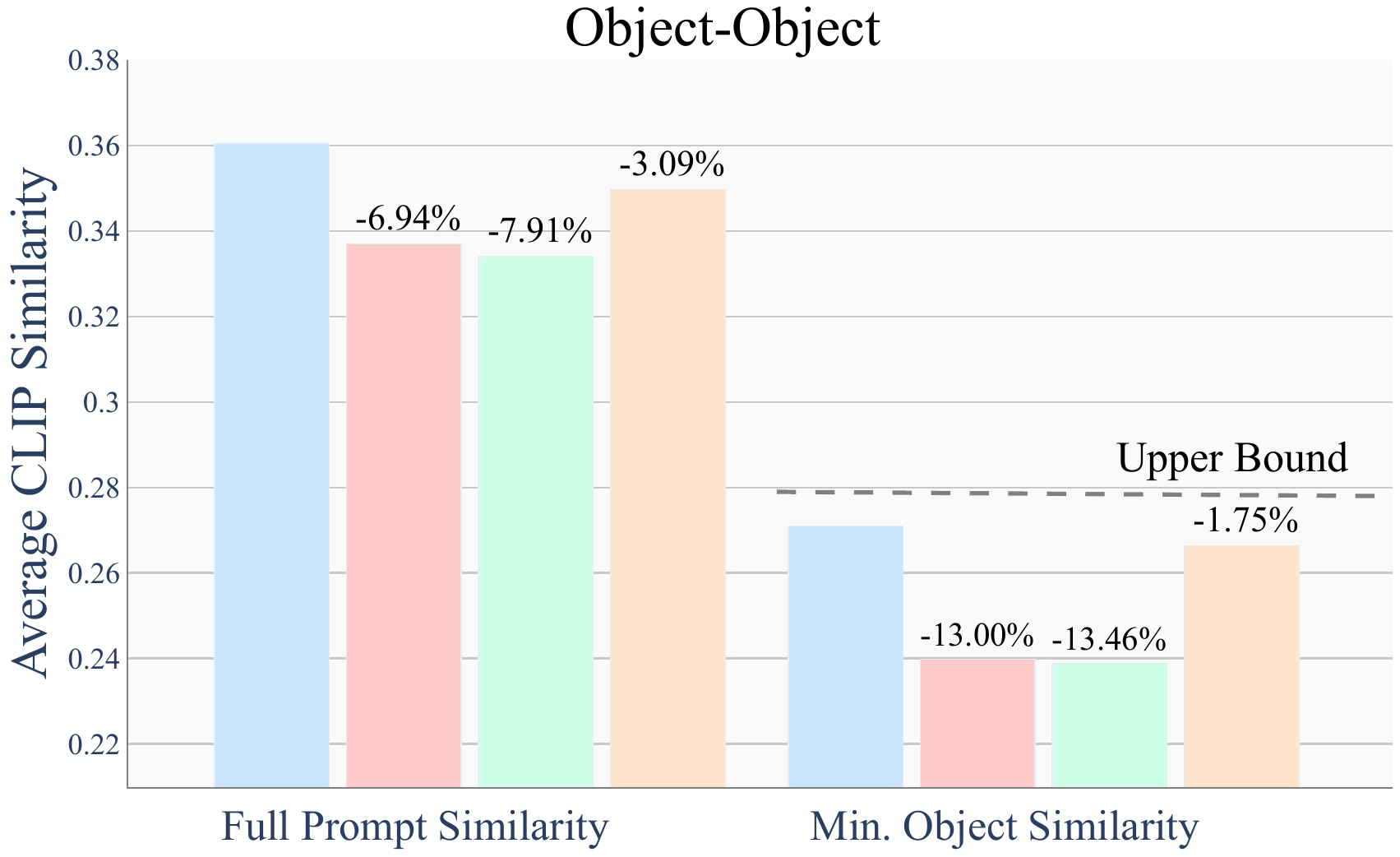}
    \vspace{-0.05cm}
    \end{tabular}
    \begin{tabular}{c c c}
    & \hspace{-0.15cm} \includegraphics[width=0.4\textwidth]{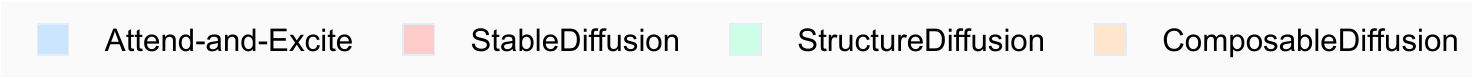} & \\[-0.4cm]
    \end{tabular}
    \caption{Average CLIP image-text similarities between the text prompts and the images generated by each method, split by subset. The \textit{Full Prompt Similarity} indicates the image-text similarity when considering the full text prompt while \textit{Minimum Object Similarity} represents the average CLIP similarity for the most neglected subject. Note, the \textit{Upper Bound} (the maximal-expected similarity) is applicable only to the \textit{Minimum Object Similarity}.}
    \label{fig:clip_image_similarities}
    \vspace{-0.25cm}
\end{figure*}

\subsection{Qualitative Comparisons}
In Figure~\ref{fig:our results}, we present results using prompts from our dataset. As can be seen, Composable Diffusion~\cite{liu2022compositional} tends to generate images containing a mixture of the subjects. For example, for ``A cat and a dog'', the images tend to mix the cat's body with the dog's face and vice versa. This can similarly be seen in the prompt ``A frog and a pink bench'' where the images may contain a frog in the shape of a bench. For StructureDiffusion~\cite{Feng2022Training}, the generated images tend to be very similar to those of Stable Diffusion, indicating that the approach fails to adequately address the semantic issues since it heavily relies on the inaccurate semantics captured by Stable Diffusion. 
Further, in the second and last column, the alternative methods either fail to generate all subjects or fail to correctly bind colors to each subject (\eg, a blue bowl instead of a yellow bowl and a red clock instead of a yellow clock).
In contrast, Attend-and-Excite is able to synthesize images that more faithfully contain all subjects with correctly binded colors. Although we explicitly tackle only the issue of neglect, we are able to implicitly improve attribute bindings between colors and subjects (\eg, the red bench and yellow clock).

Additionally, we provide examples of complex prompts in~\Cref{fig:creative} and in~\Cref{sec:additional_results}, including prompts with three or more subjects, complex attributes, and interactions between subjects. As can be seen, Attend-and-Excite is able to mitigate neglect while generating images that correspond to the input prompt, and the interactions between the subjects. For example, for the prompt ``A grizzly bear catching a salmon in a crystal clear river surrounded by a forest'' Attend-and-Excite mitigates the neglect over the salmon while generating images in which the bear catches the salmon, as specified in the prompt. Finally, Attend-and-Excite can also be used to correct global properties such as a background subject as shown with the ``garden'' in the third column.

In~\Cref{fig:structured_prompts} we consider prompts from the StructureDiffusion paper with more than two subjects or complex attributes (\eg, ``stone bench'', ``wooden patio''). As can be observed,  StructureDiffusion fails to mitigate both semantic issues. For example, in the second row, StructureDiffusion generates a sphere or cube-like object but fails to generate both. In the third row, it fails to correctly bind attributes such as the red sauce to the rice. Conversely, Attend-and-Excite generates semantically accurate images in both cases.

In~\Cref{sec:additional_results}, we provide additional qualitative and quantitative results, as well as an ablation study and additional comparisons to image editing techniques.

\vspace*{-0.1cm}
\subsection{Quantitative Analysis}
We quantify the performance of each method using CLIP-space distances along two fronts. First, we evaluate image-text similarities between the generated images and each text prompt. Second, several works~\cite{liang2022mind,Ashual2022KNNDiffusionIG} have analyzed the existence of a modality gap between CLIP's image and text embeddings. To overcome this gap, we consider an additional text-only metric.

\vspace*{-0.1cm}
\paragraph{\textbf{Text-Image Similarities.}}
For each prompt, we compute the average CLIP cosine similarity between the text prompt and the corresponding set of $64$ generated images. We denote this as the \textit{Full Prompt Similarity}. Yet, considering the full text may not accurately reflect the existence of neglect. It has been observed~\cite{paiss2022no} that CLIP's similarities resemble a bag-of-words behavior where a high score can be achieved even if the image does not fully correspond to the semantic meaning of the prompt. For example, an image of a cat may obtain a high similarity to ``a cat and a dog'' even though a dog is not present. 
In such cases, considering only the full-text similarity will not capture the existence of neglect.

As such, we evaluate the CLIP similarity for the \textit{most} neglected subject independently of the full text. To this end, we split the prompt into two sub-prompts, each containing a single subject (\eg, ``a cat'', ``a dog''). 
We then compute the CLIP similarity between each sub-prompt and each generated image. Given the two scores for each image, we are interested in maximizing the smaller of the two as this would correspond to minimizing neglect. We average the smaller of the two scores across all seeds and prompts and denote this as the \textit{Minimum Object Similarity}.
To provide intuition for the scale of the best-achievable Minimum Similarity, we compute an \textit{Upper Bound}. For each subject, we collect $50$ images from classification and detection datasets~\cite{animal_dataset,lin2014microsoft} and the internet.
We then compute the average CLIP similarity between the collected images and the subject prompt (\eg, ``a cat''). 
To obtain the bound for each subset, we average the scores of all subjects in the set.

\Cref{fig:clip_image_similarities} presents the results of the CLIP text-image metrics for all three subsets (Animal-Animal, Animal-Object, Object-Object). Observe that Attend-and-Excite outperforms all baselines across all subsets and for both metrics. Additionally, we provide the relative decrease in similarity (in percentage) compared to Attend-and-Excite. 

\begin{table}
\small
\centering
\setlength{\tabcolsep}{2pt}
\caption{Average CLIP text-text similarities between the text prompts and captions generated by BLIP over the generated images. \\[-0.65cm]} 
\begin{tabular}{l c c c} 
\toprule
Method & \begin{tabular}{c} Animal-Animal \end{tabular} & \begin{tabular}{c} Animal-Object \end{tabular} & \begin{tabular}{c} Object-Object \end{tabular} \\
\midrule
Stable Diffusion    & 0.767 (\textcolor{mydarkgreen}{-5.08\%})   & 0.793 (\textcolor{mydarkgreen}{-4.74\%})  
                     & 0.765 (\textcolor{mydarkgreen}{-5.89\%})          \\
Composable Diffusion & 0.692 (\textcolor{mydarkgreen}{-16.47\%})  & 0.769     (\textcolor{mydarkgreen}{-7.94\%}) 
                     & 0.759 (\textcolor{mydarkgreen}{-6.85\%})              \\
StructureDiffusion  & 0.761 (\textcolor{mydarkgreen}{-5.91\%})   & 0.781 (\textcolor{mydarkgreen}{-6.31\%})  
                     & 0.762 (\textcolor{mydarkgreen}{-6.49\%})          \\
\textbf{Attend-and-Excite}   & \textbf{0.806} & \textbf{0.830} & \textbf{0.811}  \\
\bottomrule \\[-0.4cm]
\end{tabular}
\label{tb:blip_captioning_similarity}
\end{table}

Notice that StructureDiffusion obtains scores similar to those of Stable Diffusion (albeit slightly lower). Attend-and-Excite significantly improves the Minimum Object Similarity in comparison to both by a gap of at least $7\%$ across all test cases, indicating that our method substantially improves the issue of neglect.
For some subsets, Composable Diffusion achieves results closest to those obtained by Attend-and-Excite. This can be attributed to a deficiency in the image-based metric where a high score can be achieved even when only a portion of a subject is present. As mentioned, Composable Diffusion often generates an object that is a mixture of the subjects in the input text. In such cases, the similarity to both subjects could be high, even though they are not generated separately. 

For example, an image featuring a car shaped like a bird may obtain a high similarity for both ``a bird'' and ``a car'' since the shape corresponds to ``a bird'' while the object itself is a car. We refer the reader to~\Cref{sec:additional_results} for examples of such behavior.
To overcome this limitation, we explore a text-based metric below. 

\begin{table}
\small
\centering
\setlength{\tabcolsep}{2pt}
\caption{User study conducted with $65$ respondents. We randomly select $10$ prompts from each subset and apply the same $4$ randomly-selected seeds to all methods. Users are asked to select the set of images that best corresponds to the input prompt. Results are averaged across all prompts in the subset.\\[-0.65cm]} 
\begin{tabular}{l c c c} 
\toprule
Method & \begin{tabular}{c} Animal-Animal \end{tabular} & \begin{tabular}{c} Animal-Object \end{tabular} & \begin{tabular}{c} Object-Object \end{tabular} \\
\midrule
Stable Diffusion    & 2.32\%    & 13.92\%   & 5.71\%           \\
Composable Diffusion & 0\%       & 1.69\%    & 9.82\%              \\
StructureDiffusion  & 6.98\%    & 6.75\%    & 7.31\%          \\
\textbf{Attend-and-Excite}   & \textbf{90.70\%} & \textbf{77.64\%} & \textbf{77.16\%}  \\
\bottomrule
\end{tabular}
\vspace{-0.2cm}
\label{tb:user_study}
\end{table}

\paragraph{\textbf{Text-Text Similarities.}}
Given the $64$ generated images for a given input prompt, we generate matching image captions using a pre-trained BLIP image-captioning model~\cite{li2022blip}. We then compute the average CLIP similarity between the prompt and all captions. This process is repeated for each subset and the results are averaged across the prompts in the subset. 
The choice of CLIP to compute the text-text similarity arises from the strong semantic prior of CLIP. We are less concerned with the exact phrasing and order of subjects in the captions. Instead, our focus is on capturing all subjects and attributes in the original prompt.

We present the text-text similarity results in~\Cref{tb:blip_captioning_similarity}. As shown, Attend-and-Excite outperforms all alternative methods across each of our three subsets by at least $4.7\%$. 
Additionally, observe that ComposableDiffusion is the lowest-performing approach when considering the text-text similarity metrics, indicating that the text-text metric captures the subject-mixing behavior discussed above.

\paragraph{\textbf{User Study.}}
Finally, we perform a user study to analyze the fidelity of the generated images. For each of the three evaluation subsets, we randomly sample $10$ prompts and generate images with each approach using the same $4$ randomly-selected seeds. For each prompt, we ask the respondents to select which set of images best reflects the prompt. The final score for each approach is calculated as the number of times respondents selected the approach averaged across all the prompts in the set (\eg, a score of $90\%$ indicates that $90\%$ of responses preferred the approach over all others). 

The study results are shown in~\Cref{tb:user_study}. Attend-and-Excite received the highest percentage of votes across \textit{all} subsets, with $90.70\%$ of responses preferring our method in the Animal-Animal subset, $77.64\%$ for the Animal-Object category, and $77.16\%$ for Object-Object.  
When evaluating each prompt individually, Attend-and-Excite is always preferred over the baselines by a majority of respondents. Even our lowest performing prompt received $59.09\%$ of votes (with SD and StructureDiffusion tied for second, each receiving $16\%$ of votes). 
This substantiates the effectiveness of Attend-and-Excite in alleviating semantic issues in text-based image generation.

\section{Limitations}

While our method offers increased fidelity with respect to the given prompt, there are several limitations to consider. 
First, our method is limited by the expressive power of the generative model since we do not apply additional training. In cases where the prompt resides outside the distribution of the textual descriptions the model learned, our method could lead to latents that are out of distribution, resulting in images that do not correspond to the text prompt.

\begin{figure}
    \centering
    \setlength{\tabcolsep}{0.5pt}
    \renewcommand{\arraystretch}{0.3}
    \addtolength{\belowcaptionskip}{-9pt}
    {\small
    \begin{tabular}{c c c @{\hspace{0.1cm}} c c c}

        \multicolumn{3}{c}{``A  \textcolor{blue}{dog} and a \textcolor{blue}{bear}''} &
        \multicolumn{3}{c}{``An \textcolor{blue}{elephant} with a som\textcolor{blue}{brero}''} \\ \\

        \includegraphics[width=0.075\textwidth]{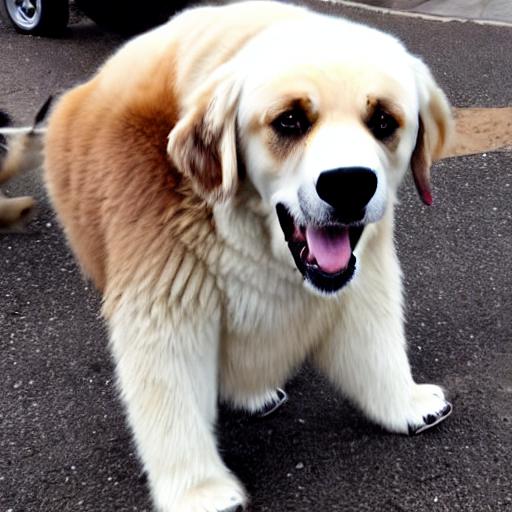} &
        \includegraphics[width=0.075\textwidth]{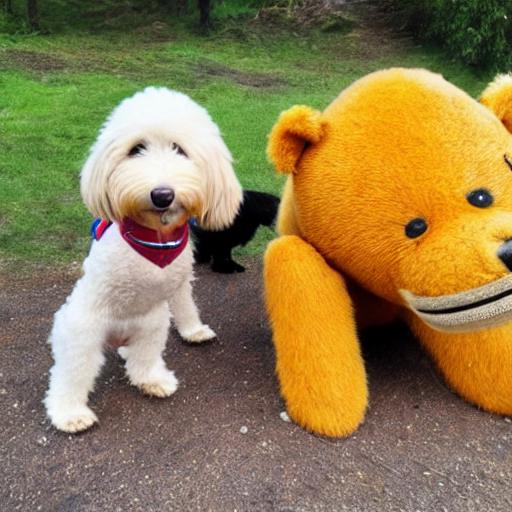} &
        \includegraphics[width=0.075\textwidth]{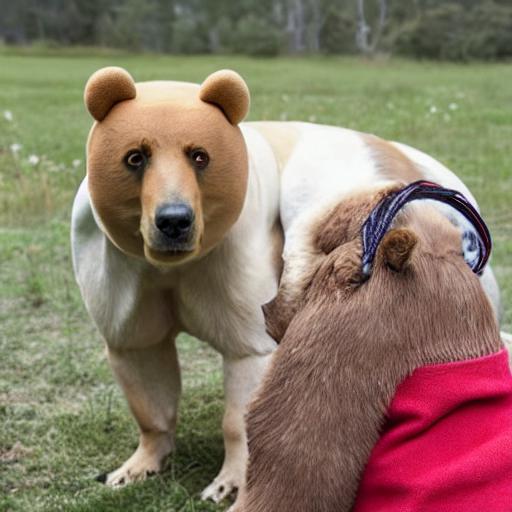} &
        \hspace{0.05cm}
        \includegraphics[width=0.075\textwidth]{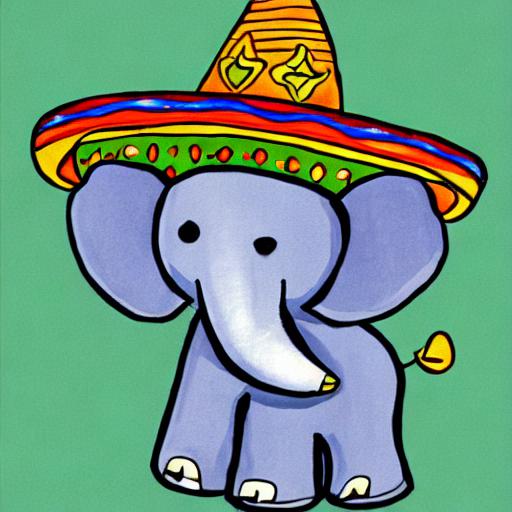} &
        \includegraphics[width=0.075\textwidth]{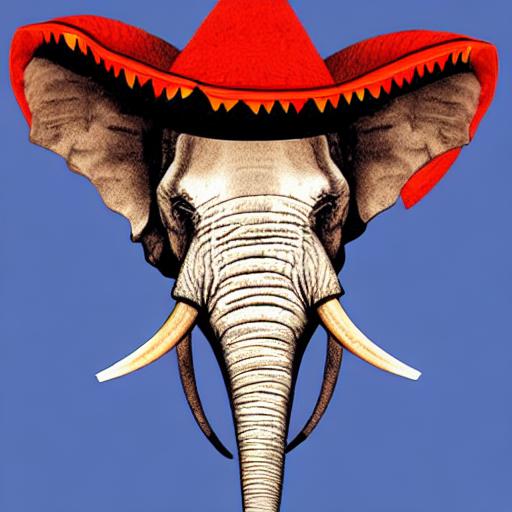} &
        \includegraphics[width=0.075\textwidth]{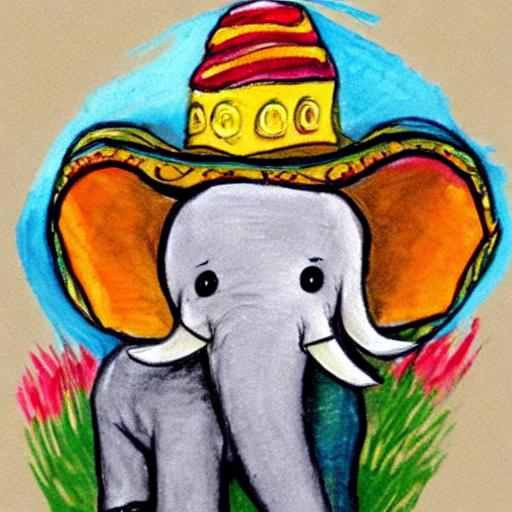}  \\ \\ \\[-0.4cm]

    \end{tabular}

    }
    \caption{Limitations. Left: Out-of-distribution results due to the limited expressive power of Stable Diffusion. Right: When the subject combination is not natural (``elephant'', ``sombrero''), the results may be less realistic.}
    \label{fig:limitations}
\end{figure}

Second, when synthesizing subjects that naturally do not appear together, the generated images may be less realistic (\eg, paintings). We attribute this to the fact that such combinations tend to reside outside the distribution that Stable Diffusion has learned for real images. Examples of these limitations are shown in~\Cref{fig:limitations}.

Finally, while we tackle two core semantic issues, the path to achieving semantically-accurate generation is still long, and there exist additional challenges to be addressed such as complex object compositions (\eg, ``riding on'', ``in front of'', ``beneath''). Additionally, while we have not explored applying Attend-and-Excite over a negation (\ie, ``not''), this could potentially be achieved by demanding a \textit{low} attention value for the subject.

\section{Conclusions}
Can a diffusion process be corrected once it takes a wrong turn? In this work, we introduce the concept of \textit{Generative Semantic Nursing} (GSN), which refers to a careful manipulation of latents during the denoising process of a pre-trained text-to-image diffusion model. 
We then present \textit{Attend-and-Excite}, a specific form of GSN that encourages all subject tokens in the text to be attended to by some image patch. We demonstrate that by applying this intuitive optimization, we are able to alleviate two core semantic issues on the fly, thus correcting the generator after it has taken a wrong turn. 

Similar to extrapolating text-driven gradients in classifier-free guidance, our approach aims to strengthen the text conditioning along the image generation process. 
While we explore the notion of GSN for mitigating semantic issues of text-conditioned generation, we believe GSN can potentially be applied to any image editing and generation task by defining an appropriate loss objective. 
Moreover, this guidance need not be through text and does not require conditioning at all, but is defined only by the task itself.

\begin{acks}
    We would like to thank Daniel Livshen, Elad Richardson, Matan Cohen, Or Patashnik, Rinon Gal, and Yotam Nitzan for their early feedback and insightful discussions. 
    The first author is supported by the Council for Higher Education in Israel.
    This work was supported in part by the 
    Israel Science Foundation under Grant No.~\grantnum{}{2366/16} and Grant No.~\grantnum{}{2492/20}.
\end{acks}

\bibliographystyle{ACM-Reference-Format}
\bibliography{main}

\appendix
\appendixpage

\section{Additional Details}~\label{sec:additional_details}

\vspace{-0.3cm}
\subsection{Implementation Details}

We use the official Stable Diffusion v1.4 text-to-image model employing the pre-trained text encoder from the CLIP ViT-L/14 model~\cite{radford2021learning}. We use a fixed guidance scale $s$ of $7.5$. For performing our gradient update defined in Equation (3) in the main paper, we set the scale factor $\alpha_t$ using a linear scheduling rate that starts from $20$ and decays linearly to a minimum value of $10$. 
The Gaussian filter used to smooth the cross-attention maps $A_t^s$ has a kernel size of $3$ and a standard deviation of $\sigma = 0.5$ unless specified otherwise. 

Finally, when performing the iterative latent refinement, we do so for at most $20$ latent updates or until the specified threshold value is attained, whichever comes first. Note that this is done to encourage the latent to remain in-distribution with respect to the latent distribution that Stable Diffusion has learned.

\subsection{Selecting the Subject Tokens} 
Our three data subsets are constructed such that each prompt is a conjunction of two subject tokens for ease of evaluation (\eg, for splitting the prompt into two subject prompts to calculate the Minimum Object Similarity), and to enable a comparison to all baseline methods (\eg Composable Diffusion~\cite{liu2022compositional} only operates over conjunctions and negations). 

However, in the more general setting, our approach allows users to determine the set of tokens to strengthen. A natural way of obtaining the set of subject tokens is employing a part-of-speech tagger and extracting the nouns from the prompt. However, the user is also free to define any set of tokens to strengthen, which could also depend on the results obtained by the original model. This includes tokens describing background settings (\eg, a kitchen or a library). 
Our method also allows flexibility in setting the target attention value to control the strengthening intensity. This can be done by setting the target attention to be lower than $1$ in Step 6 of Algorithm 1.

Additionally, in cases where the subject requires more than one token (\eg, ``sombrero'' is split by the tokenizer into ``som'', and ``brero''), we observe that the attention scores for the subject are typically dominated by a single token. To apply Attend-and-Excite in such cases, one must identify the dominant token  (in this case ``brero'') by examining the attention maps for all tokens using the subject prompt (\eg, ``an image of a sombrero''). Figure 8 of the main paper presents an example of such a case with the prompt ``an elephant with a sombrero''.

\subsection{Runtime}
Evaluated on an A100 GPU, Stable Diffusion generates a single image in approximately $5.6$ seconds. In cases where the iterative refinement is not activated, the runtime for Attend-and-Excite is approximately $9.7$ seconds per image. For more challenging prompts, where the iterative refinement is used, this increases to $\sim15.4$ seconds per image. By using float16 precision, the runtime can be reduced to $\sim6.6$ and $\sim11.8$ seconds per image, respectively.

\subsection{Quantitative Evaluations}

\begin{table}
\small
\centering
\caption{Datasets. We list the animals, objects, and colors used to define each of the three data subsets used in our quantitative evaluations.} 
\vspace{-0.2cm}
\begin{tabular}{l | l} 
\toprule
Category & \\
\midrule
Animals & \begin{tabular}{l} cat, dog, bird, bear, lion, horse, elephant, monkey, frog, \\ turtle, rabbit, mouse \end{tabular} \\
\midrule
Objects & \begin{tabular}{l} backpack, glasses, crown, suitcase, chair, balloon, bow,  \\ car, bowl, bench, clock, apple \end{tabular} \\
\midrule
Colors & \begin{tabular}{l} red, orange, yellow, green, blue, purple, pink, brown, \\ gray, black, white \end{tabular} \\
\bottomrule
\end{tabular}
\label{tb:datasets}
\end{table}

\paragraph{Datasets.} As discussed in Section 5 of the main paper, we define three different data subsets for evaluating Attend-and-Excite and the alternative text-to-image approaches. In~\Cref{tb:datasets}, we specify the animals, objects, and colors used to define our three subsets. We will release the list of prompts used for each subset to help facilitate future comparisons and evaluations.

\paragraph{CLIP-based Metrics.}
In our quantitative experiments based on CLIP-space distances, we employ the official CLIP ViT-B/16 model \cite{radford2021learning,dosovitskiy2020image}. For computing the image captions of all generated images, we use the official BLIP image captioning model~\cite{li2022blip} fine-tuned on the MS-COCO Captions dataset~\cite{chen2015microsoft}. The official implementation was taken from the LAVIS library~\cite{li2022lavis}.

When computing the text embeddings of an input prompt or sub-prompt (\eg, for computing the \textit{Minimum Object Similarity}), we follow the evaluation setup used in CLIP and apply $80$ different prompt templates with our input prompt. These include templates of the form ``a photo a \{\}'', ``a cartoon of \{\}'', and ``a a drawing of a \{\}''. The text embedding of the original prompt is computed as the average CLIP embedding across all $80$ constructed prompts. This aggregated embedding is then used to compute the cosine similarity between either the generated images or generated image captions. Note that we do not use these templates for defining the text embeddings of the BLIP-generated captions.

Finally, when computing the aggregated metrics for either the text-to-image similarities or text-to-text similarity, we omit seeds where the method returned a black image (meaning NSFW content was discovered and removed).

\begin{figure}
    \setlength{\tabcolsep}{0.5pt}
    \renewcommand{\arraystretch}{0.3}
    {\small
    \begin{tabular}{c c @{\hspace{0.1cm}} c c}

    \multicolumn{2}{c}{\begin{tabular}{c} Attend-and-Excite \\ w/o Gaussian \end{tabular}} &
    \multicolumn{2}{c}{Attend-and-Excite} \\

    \includegraphics[width=0.11\textwidth]{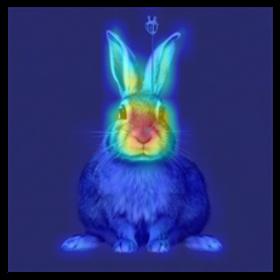} &
    \includegraphics[width=0.11\textwidth]{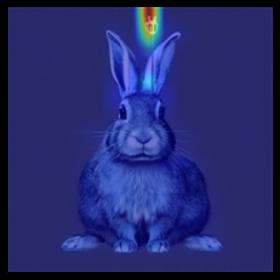} &
    \hspace{0.05cm}
    \includegraphics[width=0.11\textwidth]{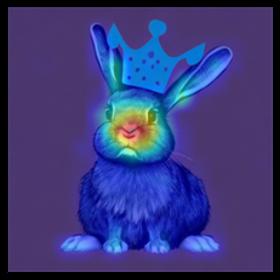} &
    \includegraphics[width=0.11\textwidth]{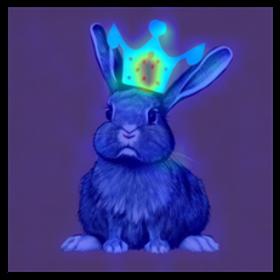} \\

    Rabbit & Crown & \hspace{0.05cm} Rabbit & Crown \\
    
    \includegraphics[width=0.11\textwidth]{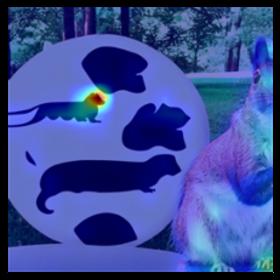} &
    \includegraphics[width=0.11\textwidth]{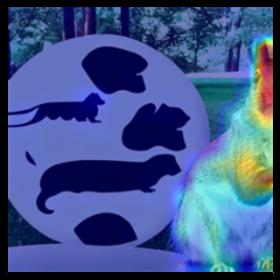} &
    \hspace{0.05cm}
    \includegraphics[width=0.11\textwidth]{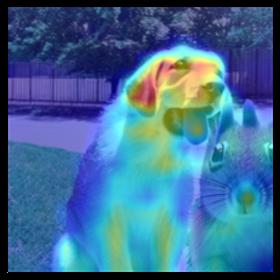} &
    \includegraphics[width=0.11\textwidth]{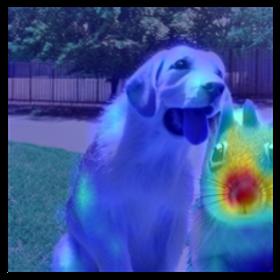} \\

    Dog & Squirrel & \hspace{0.05cm} Dog & Squirrel \\
        
    \end{tabular}
    }
    \vspace{-0.2cm}
    \caption{Visualization of the cross-attention maps per subject using Attend-and-Excite on Stable Diffusion with and without Gaussian Smoothing. The prompts used are: ``A \textcolor{blue}{rabbit} with a \textcolor{blue}{crown}'' and ``A \textcolor{blue}{dog} and \textcolor{blue}{squirrel}''.}
    \label{fig:gaussian_expl}
\end{figure}

\section{Ablation Study}~\label{sec:ablations}
In this section, we explore three key design choices of Attend-and-Excite. First, we validate the use of the Gaussian kernel for smoothing the cross-attention maps before 
computing our loss objective. As discussed, this encourages each patch in the smooth cross-attention map to consider its neighboring patches in the original map. By doing so, we find that after the denoising process, the final cross-attention map for each subject token attains high activation values in multiple image patches. This is illustrated in~\Cref{fig:gaussian_expl}, where we visualize the cross-attention maps for the same input prompt and the same seed with and without the use of the Gaussian kernel. As can be seen, when we do not apply Gaussian smoothing, the subject may collapse into a single or small number of patches that only generate partial information. As a result, our objective of maintaining a patch with a high attention value may be satisfied even though the actual semantic issue of neglect was not solved. For example, for the prompt ``a rabbit with a crown'' (top row in~\Cref{fig:gaussian_expl}), notice how the cross-attention map for ``crown'' attains a very high activation on a small image region on the rabbit's head even though a full crown is not generated. For the example on the bottom row, a silhouette containing a dog-like head achieves a very high activation, although a full dog is not generated. In contrast, for the results of our full method with the Gaussian smoothing, both subjects take a significant part of the image. Accordingly, the cross-attention maps act as a good medium for explainability as the correct subject is highlighted in each case.

Additionally, we validate the use of iterative latent refinement, which ensures that each subject token achieves a certain maximum activation value at specific timesteps along the denoising process. We find that doing so helps encourage the presence of \textit{all} subject tokens in the generated image.

\begin{figure*}
    \centering
    \setlength{\tabcolsep}{0.5pt}
    \renewcommand{\arraystretch}{0.3}
    {\small
    \begin{tabular}{c c c @{\hspace{0.1cm}} c c c @{\hspace{0.1cm}} c c c }

        \multicolumn{3}{c}{No Iterative Refinement} &
        \multicolumn{3}{c}{No Gaussian Smoothing} &
        \multicolumn{3}{c}{\begin{tabular}{c} Attend-and-Excite \end{tabular}} \\ \\

        \includegraphics[width=0.1075\textwidth]{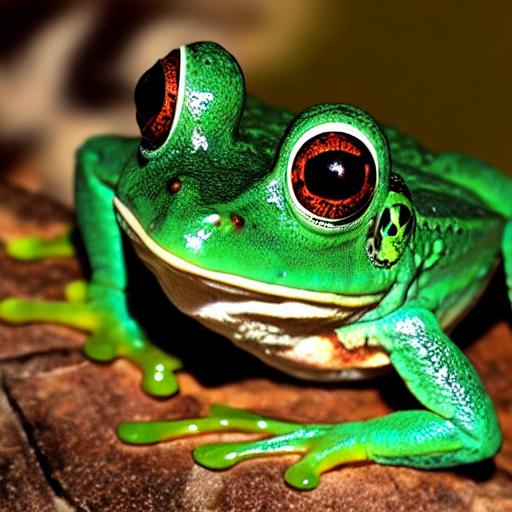} &
        \includegraphics[width=0.1075\textwidth]{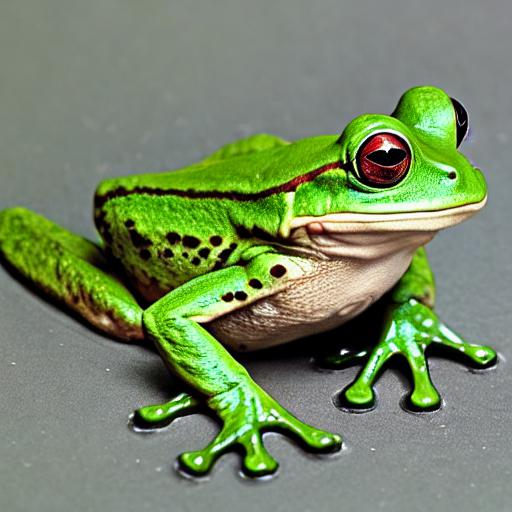} &
        \includegraphics[width=0.1075\textwidth]{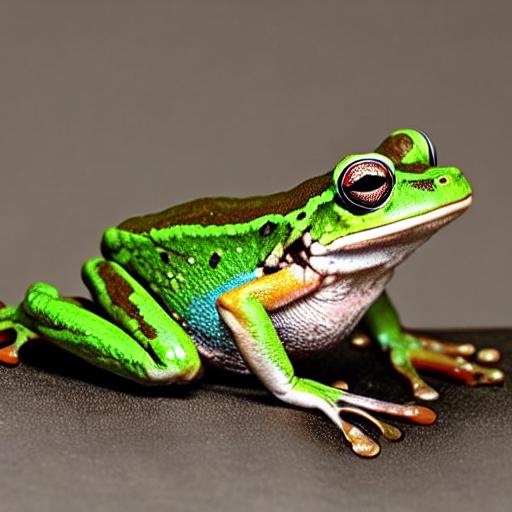} &
        \hspace{0.05cm}
        \includegraphics[width=0.1075\textwidth]{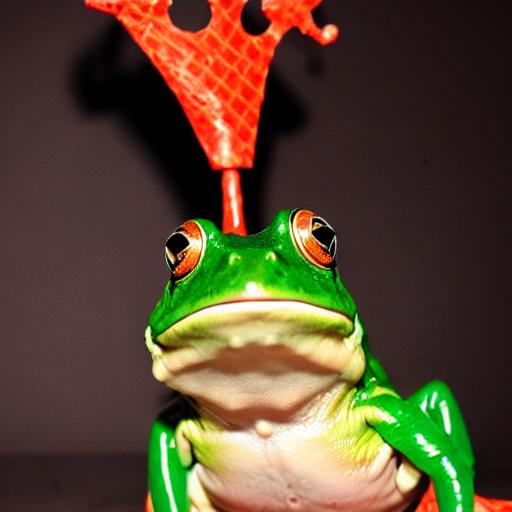} &
        \includegraphics[width=0.1075\textwidth]{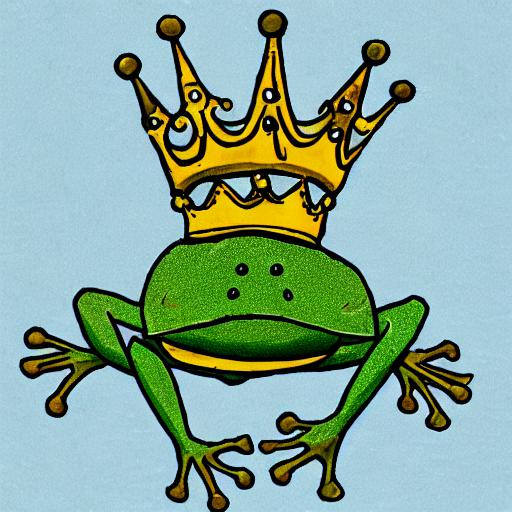} &
        \includegraphics[width=0.1075\textwidth]{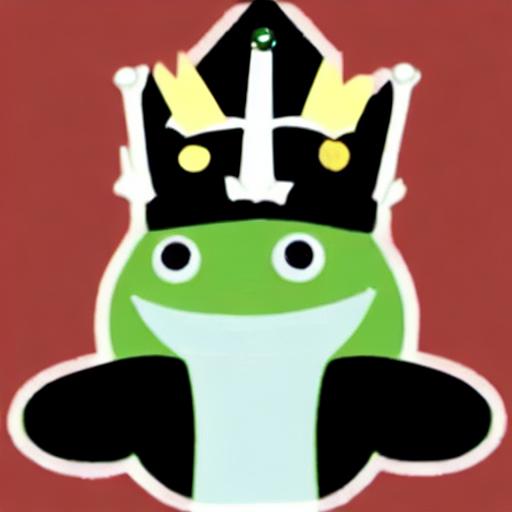} &
        \hspace{0.05cm}
        \includegraphics[width=0.1075\textwidth]{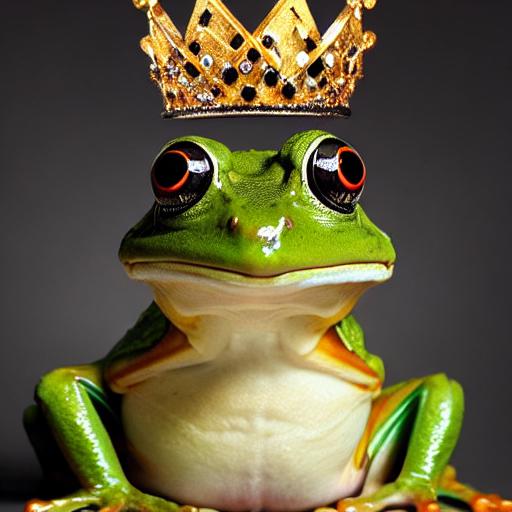} &
        \includegraphics[width=0.1075\textwidth]{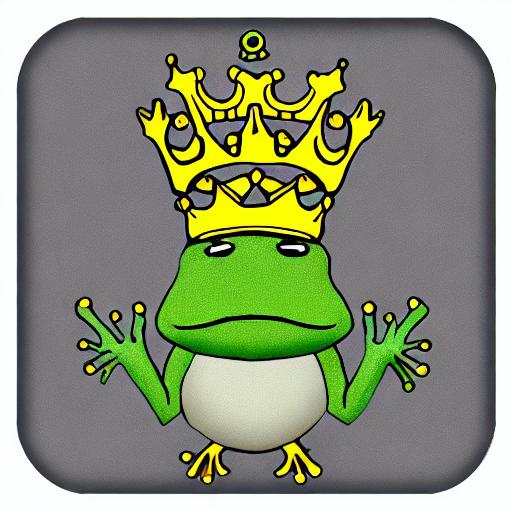} & 
        \includegraphics[width=0.1075\textwidth]{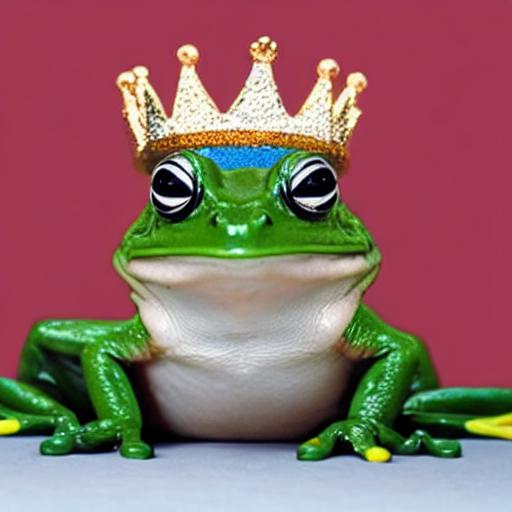} \\\\

        \multicolumn{9}{c}{``A \textcolor{blue}{frog} with a \textcolor{blue}{crown}''} \\ \\ \\

        \includegraphics[width=0.1075\textwidth]{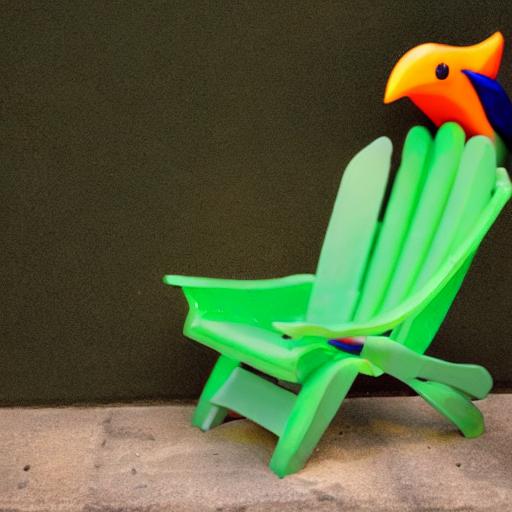} &
        \includegraphics[width=0.1075\textwidth]{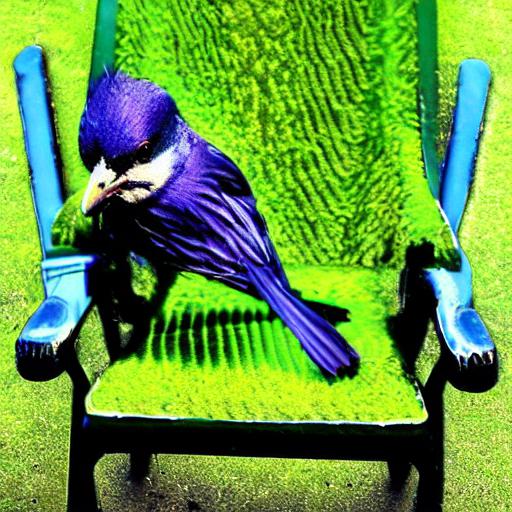} &
        \includegraphics[width=0.1075\textwidth]{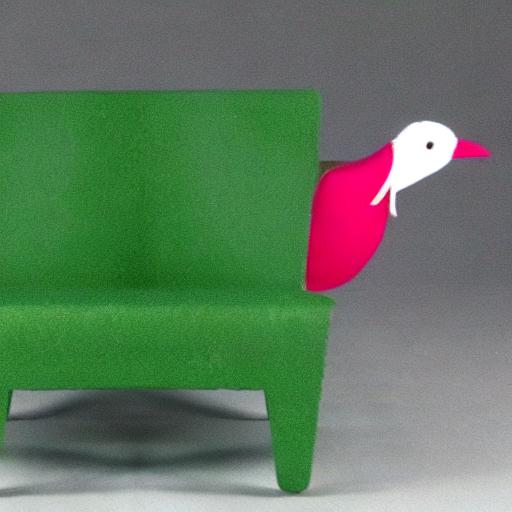} &
        \hspace{0.05cm}
        \includegraphics[width=0.1075\textwidth]{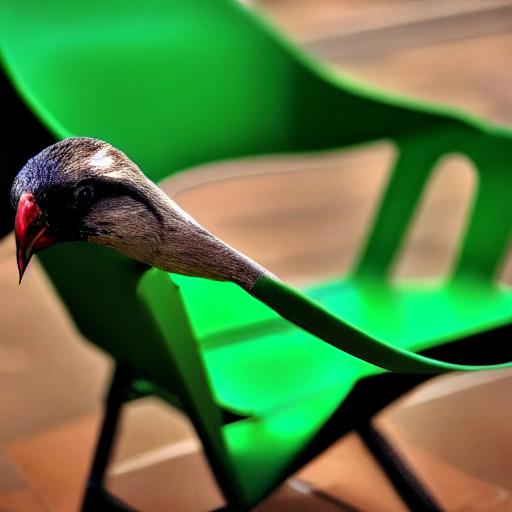} &
        \includegraphics[width=0.1075\textwidth]{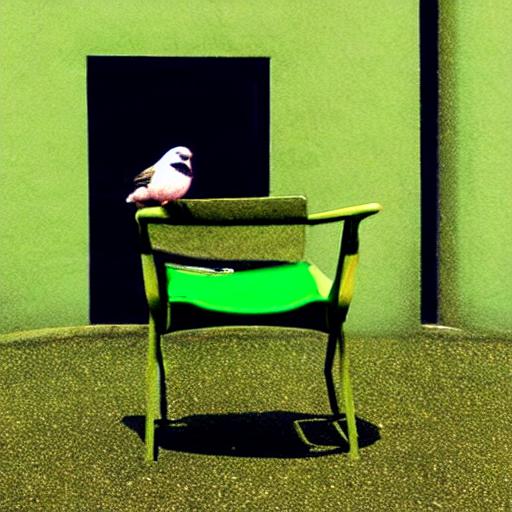} &
        \includegraphics[width=0.1075\textwidth]{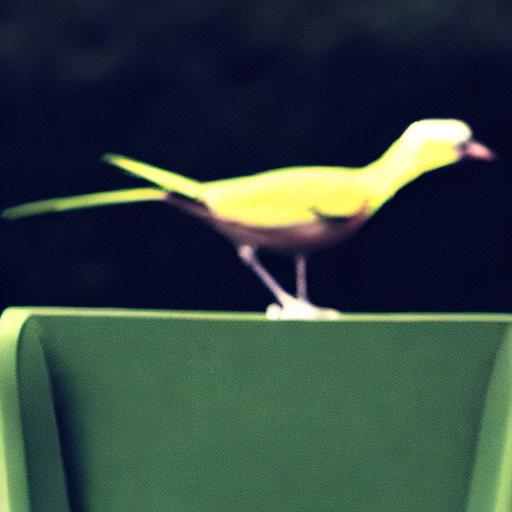} &
        \hspace{0.05cm}
        \includegraphics[width=0.1075\textwidth]{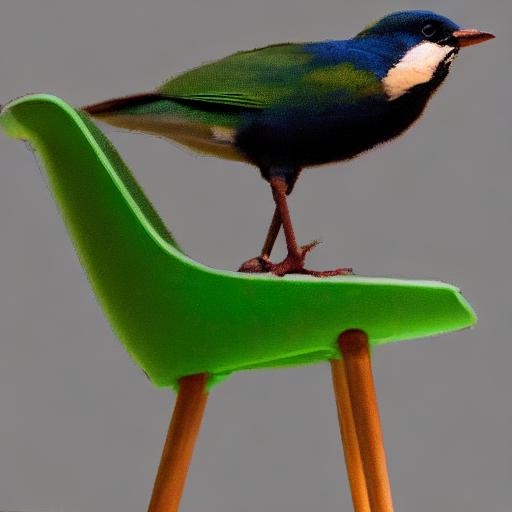} &
        \includegraphics[width=0.1075\textwidth]{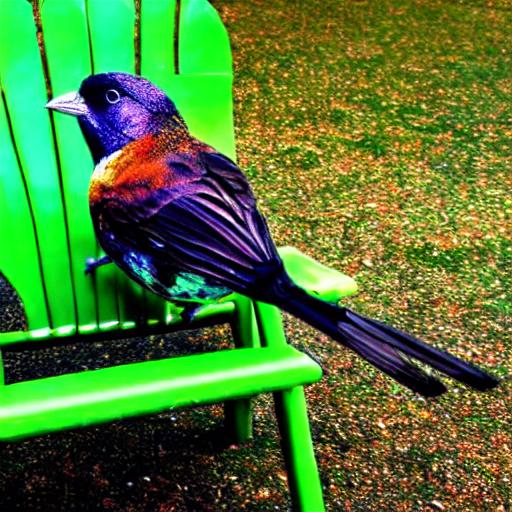} &
        \includegraphics[width=0.1075\textwidth]{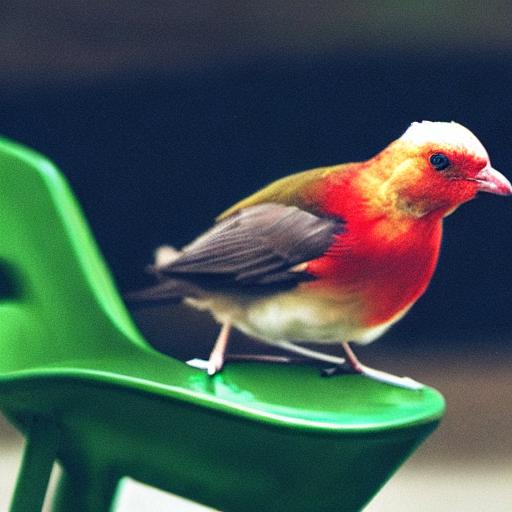}  \\\\
        
        \multicolumn{9}{c}{``A \textcolor{blue}{bird} and a green \textcolor{blue}{chair}''} \\ \\ \\

        \includegraphics[width=0.1075\textwidth]{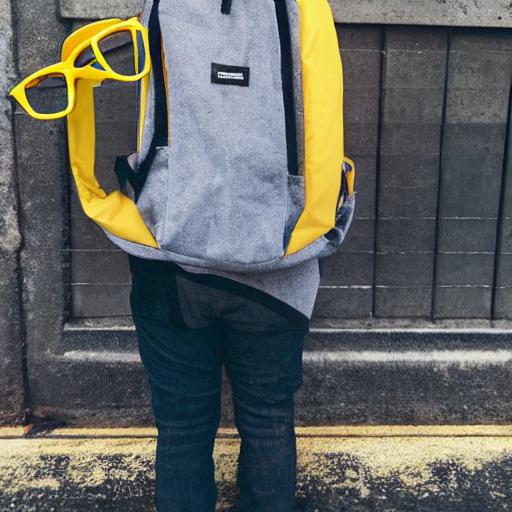} &
        \includegraphics[width=0.1075\textwidth]{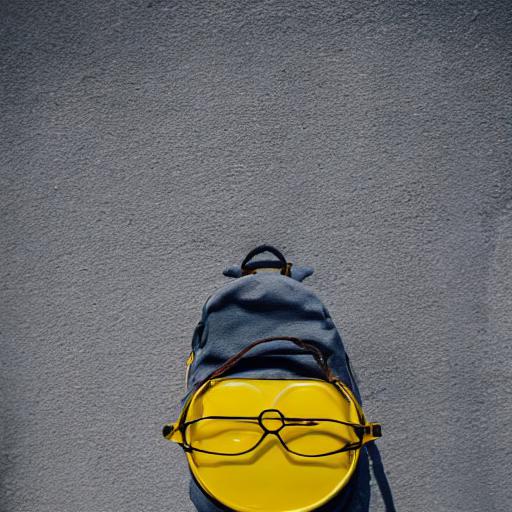} &
        \includegraphics[width=0.1075\textwidth]{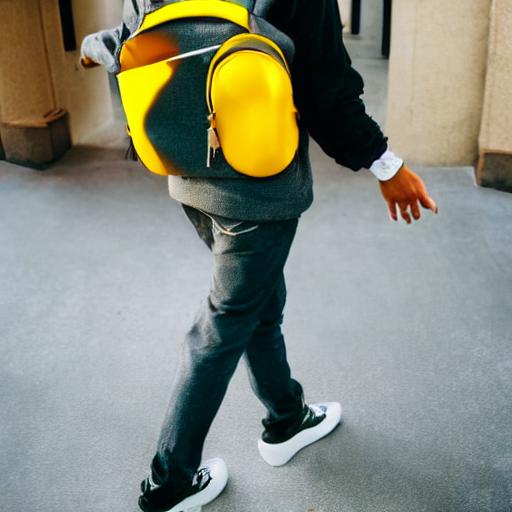} &
        \hspace{0.05cm}
        \includegraphics[width=0.1075\textwidth]{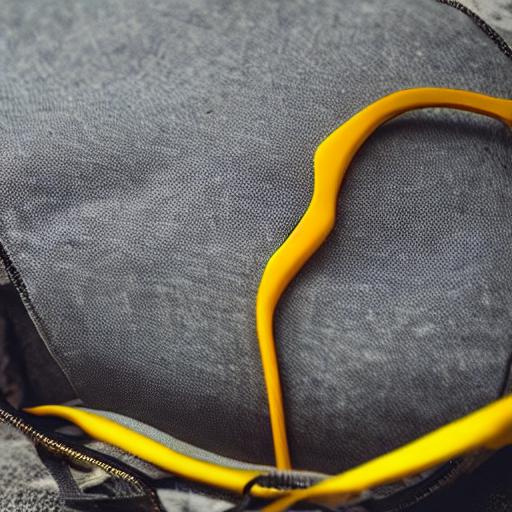} &
        \includegraphics[width=0.1075\textwidth]{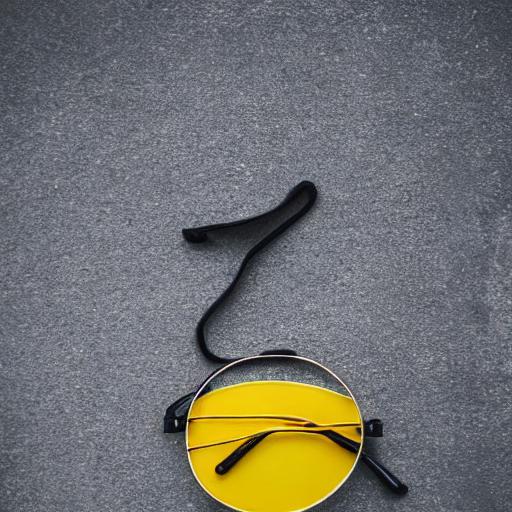} &
        \includegraphics[width=0.1075\textwidth]{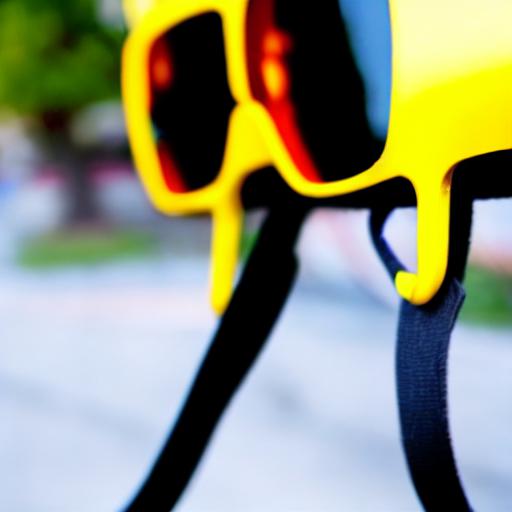} &
        \hspace{0.05cm}
        \includegraphics[width=0.1075\textwidth]{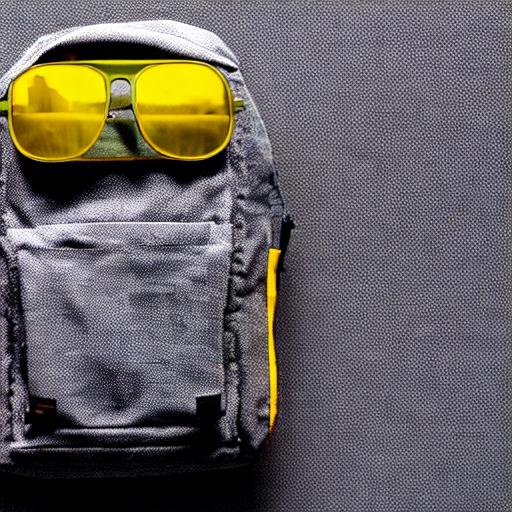} &
        \includegraphics[width=0.1075\textwidth]{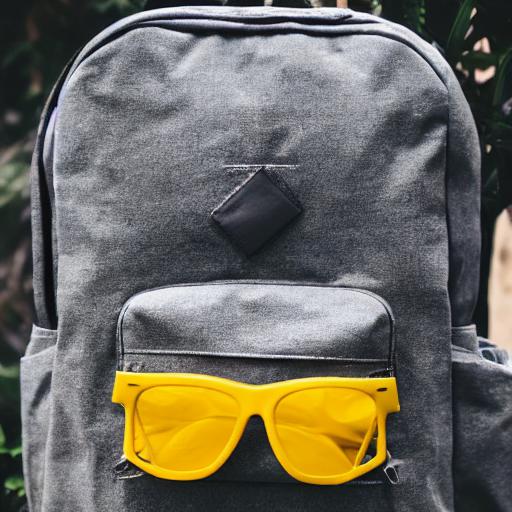} &
        \includegraphics[width=0.1075\textwidth]{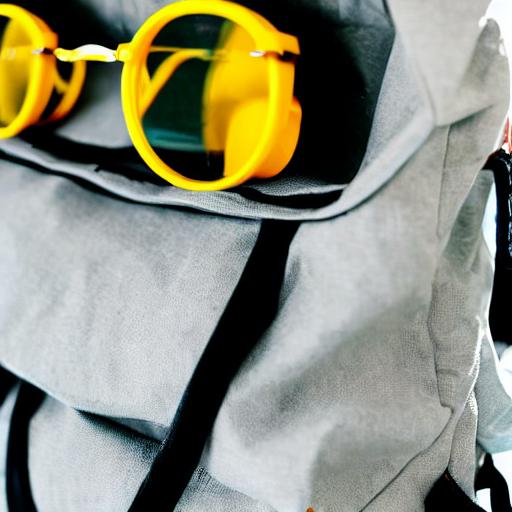} \\\\

        \multicolumn{9}{c}{``A gray \textcolor{blue}{backpack} and yellow \textcolor{blue}{glasses}''} \\ \\ \\

        \includegraphics[width=0.1075\textwidth]{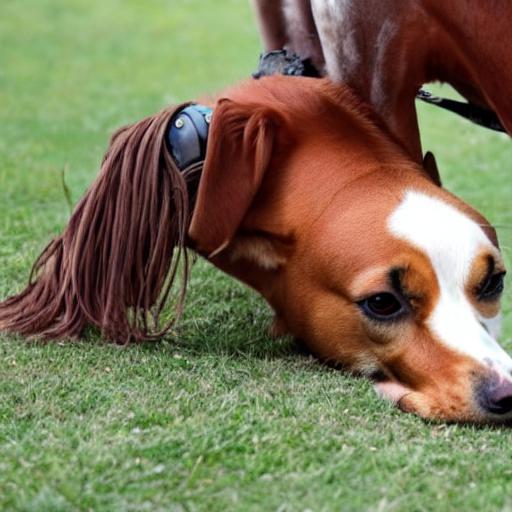} &
        \includegraphics[width=0.1075\textwidth]{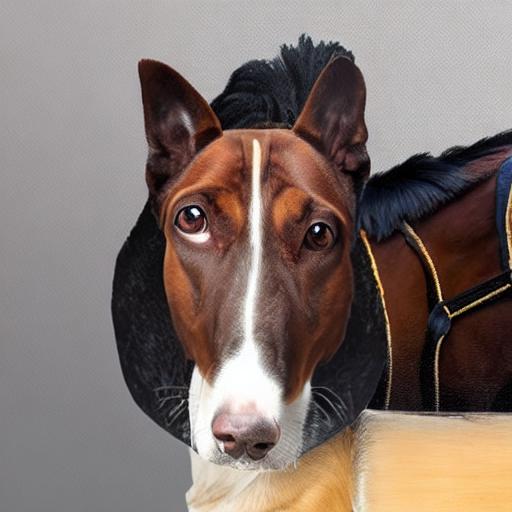} &
        \includegraphics[width=0.1075\textwidth]{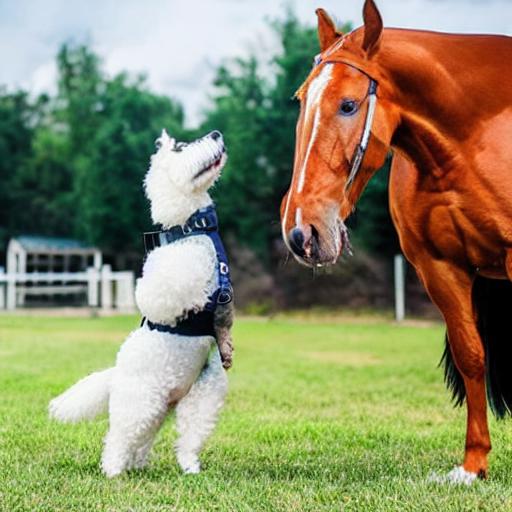} &
        \hspace{0.05cm}
        \includegraphics[width=0.1075\textwidth]{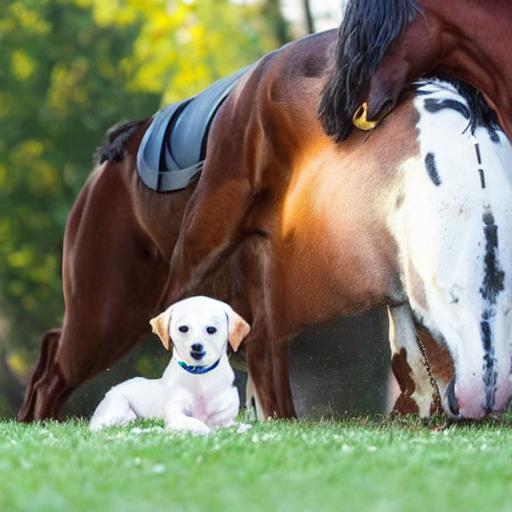} &
        \includegraphics[width=0.1075\textwidth]{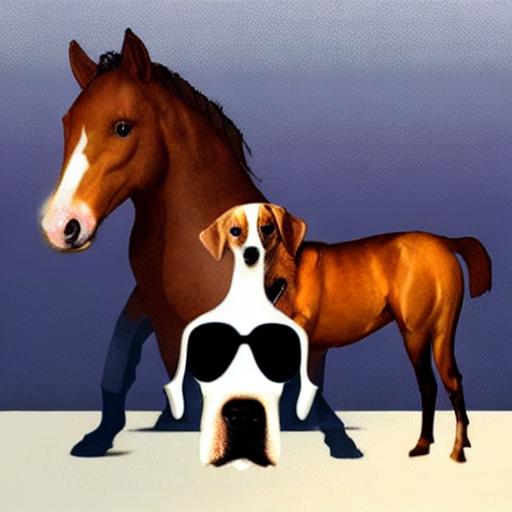} &
        \includegraphics[width=0.1075\textwidth]{images/ablations/no_iterative_58000.jpg} &
        \hspace{0.05cm}
        \includegraphics[width=0.1075\textwidth]{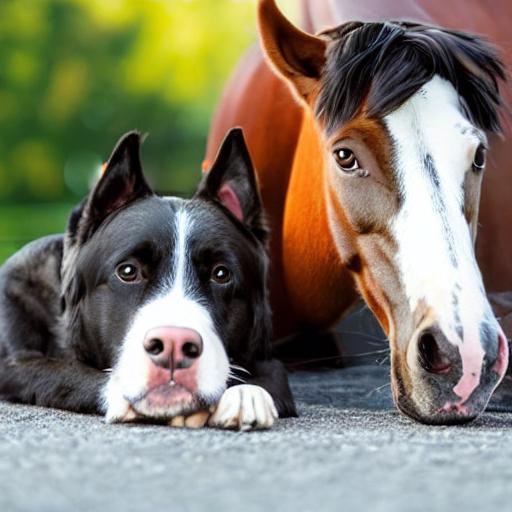} &
        \includegraphics[width=0.1075\textwidth]{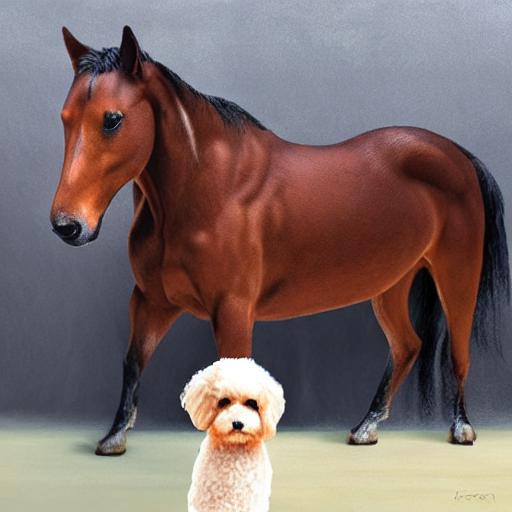} &
        \includegraphics[width=0.1075\textwidth]{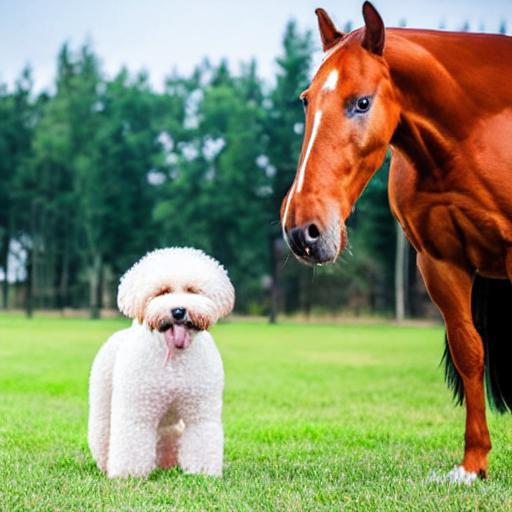} \\\\

        \multicolumn{9}{c}{``A \textcolor{blue}{dog} and a \textcolor{blue}{horse}''} \\ \\ \\

        \includegraphics[width=0.1075\textwidth]{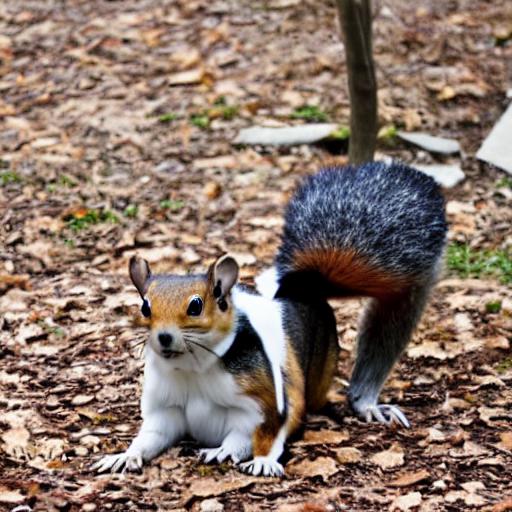} &
        \includegraphics[width=0.1075\textwidth]{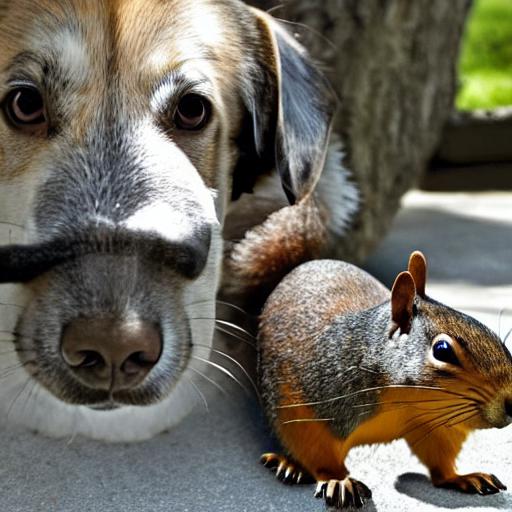} &
        \includegraphics[width=0.1075\textwidth]{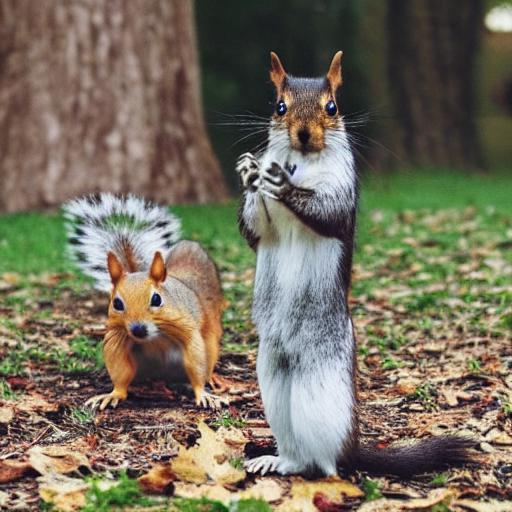} &
        \hspace{0.05cm}
        \includegraphics[width=0.1075\textwidth]{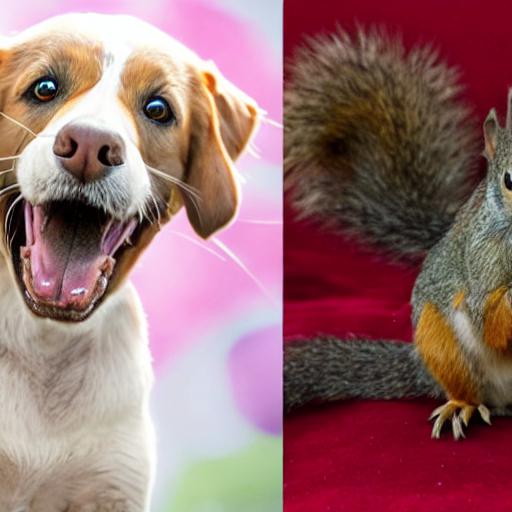} &
        \includegraphics[width=0.1075\textwidth]{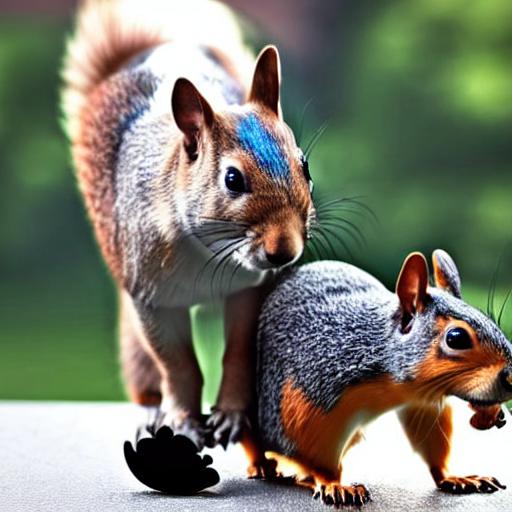} &
        \includegraphics[width=0.1075\textwidth]{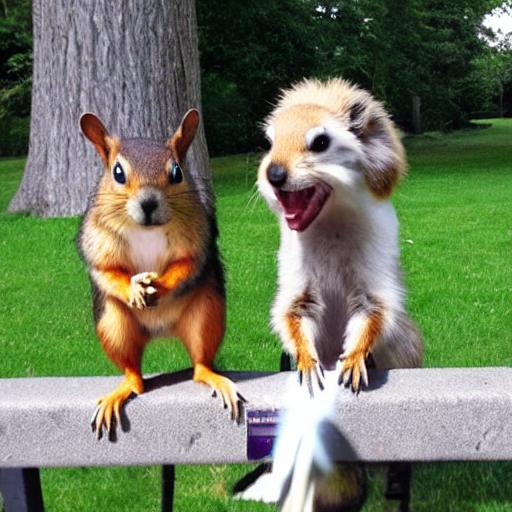} &
        \hspace{0.05cm}
        \includegraphics[width=0.1075\textwidth]{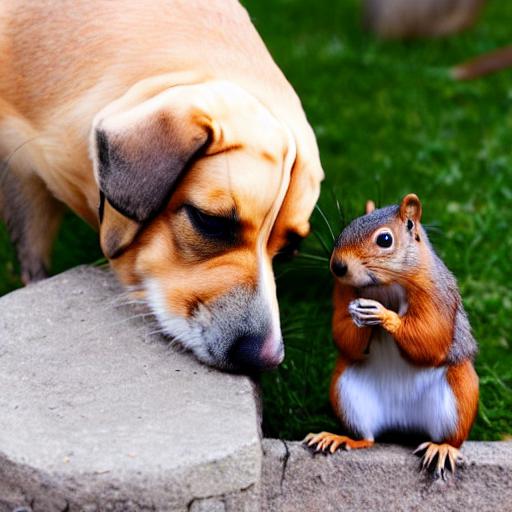} &
        \includegraphics[width=0.1075\textwidth]{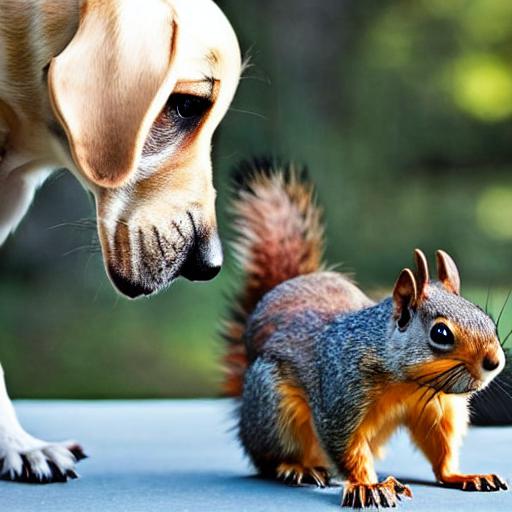} &
        \includegraphics[width=0.1075\textwidth]{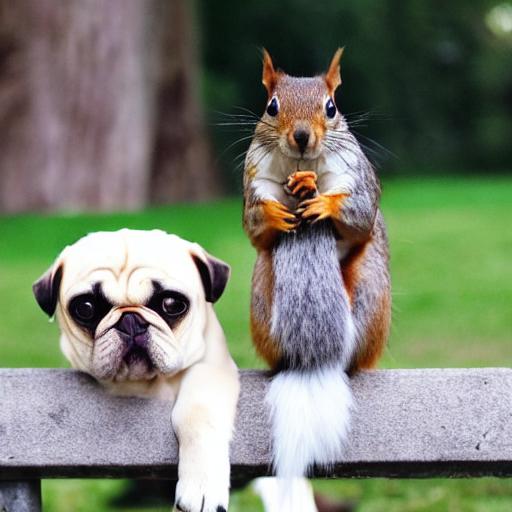} \\\\ 

        \multicolumn{9}{c}{``A \textcolor{blue}{dog} and a \textcolor{blue}{squirrel}''}
        
    \end{tabular}
    
    }
    \vspace{-0.25cm}
    \caption{Ablation Study. We examine the impact of each component of our method by removing it, and comparing the results against those of our full method. Each comparison is ran with the same $3$ seeds.}
    \label{fig:ablations}
\end{figure*}

In~\Cref{fig:ablations}, we provide a qualitative comparison of Attend-and-Excite without either the iterative latent refinement or without the Gaussian smoothing. For each text prompt, we show results obtained over three seeds shared across the three variants. As can be seen, applying each of the two components assists with mitigating catastrophic neglect. For example, in the first row, when omitting the iterative refinement, the crown is not generated in any of the shown images. In addition, in the third row, the gray backpack is not generated when omitting the Gaussian smoothing. Overall, we found that applying both components leads to more semantically-faithful, higher-quality generations across the three considered subsets.

We note that the iterative refinement process is not always applied (\eg, when the threshold is already met for all subject tokens). In cases where Stable Diffusion is able to successfully generate the two subjects, applying the iterative refinement will not have a strong influence on the generated image. 

\begin{figure*}
    \centering
    \setlength{\tabcolsep}{0.5pt}
    \renewcommand{\arraystretch}{0.3}
    {\small
    \begin{tabular}{c c c @{\hspace{0.1cm}} c c c @{\hspace{0.1cm}} c c c }

        & 
        \multicolumn{2}{c}{``A \textcolor{blue}{turtle} and a yellow \textcolor{blue}{bowl}''} &
        \multicolumn{2}{c}{``A \textcolor{blue}{cat} and a \textcolor{blue}{dog}''} &
        \multicolumn{2}{c}{``Yellow \textcolor{blue}{glasses} and a gray \textcolor{blue}{bowl}''} &
        \multicolumn{2}{c}{\begin{tabular}{c} ``A purple \textcolor{blue}{chair} and a red \textcolor{blue}{bow}'' \\\\
        \end{tabular}} \\

        {\raisebox{0.15in}{
        \rotatebox{90}{\begin{tabular}{c} Modify \\ All Steps \\ \\ \end{tabular}}}} &
        \includegraphics[width=0.115\textwidth]{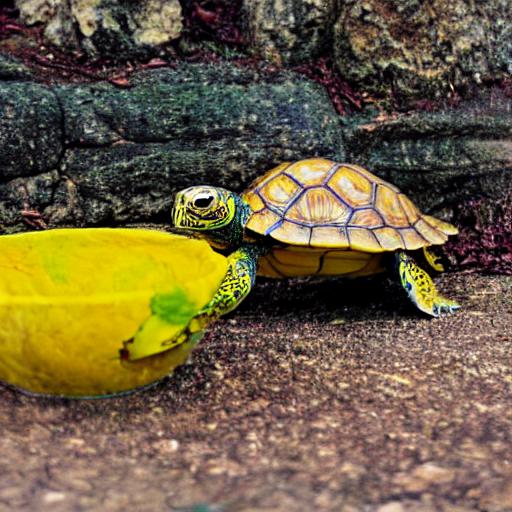} &
        \includegraphics[width=0.115\textwidth]{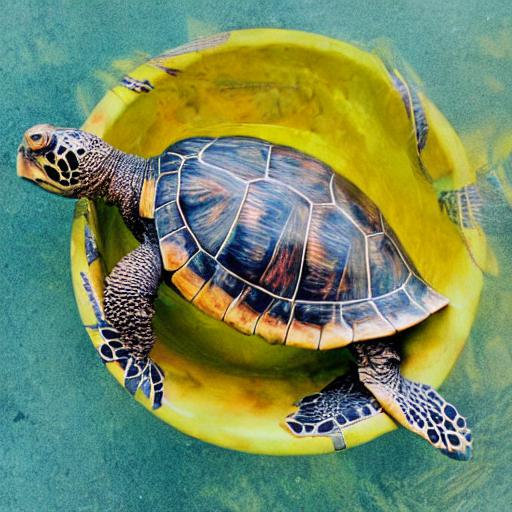} &
        \hspace{0.05cm}
        \includegraphics[width=0.115\textwidth]{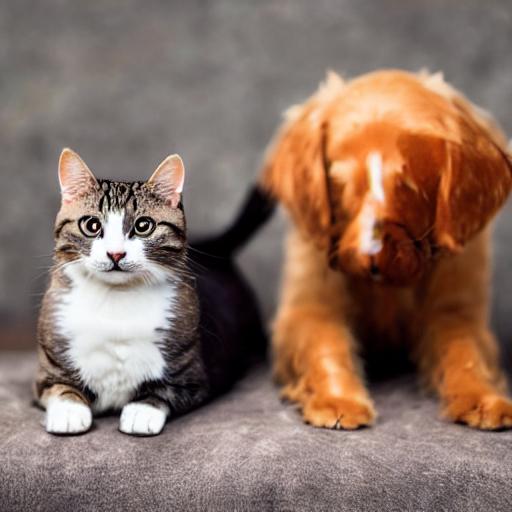} &
        \includegraphics[width=0.115\textwidth]{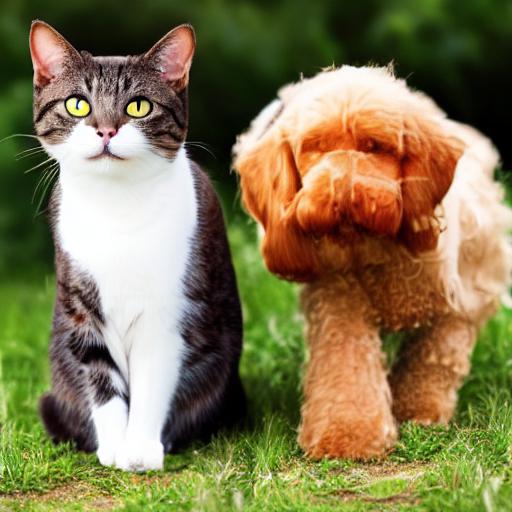} &
        \hspace{0.05cm}
        \includegraphics[width=0.115\textwidth]{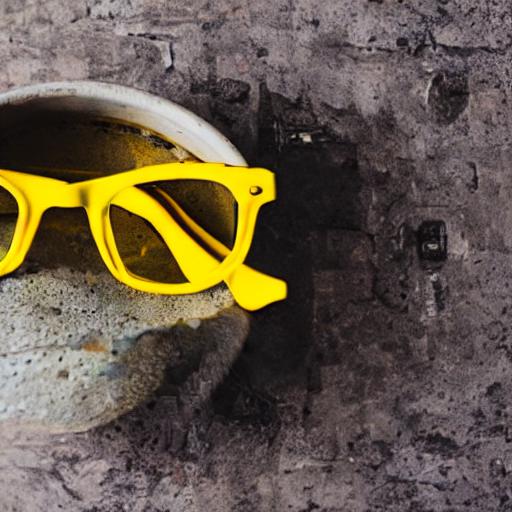} &
        \includegraphics[width=0.115\textwidth]{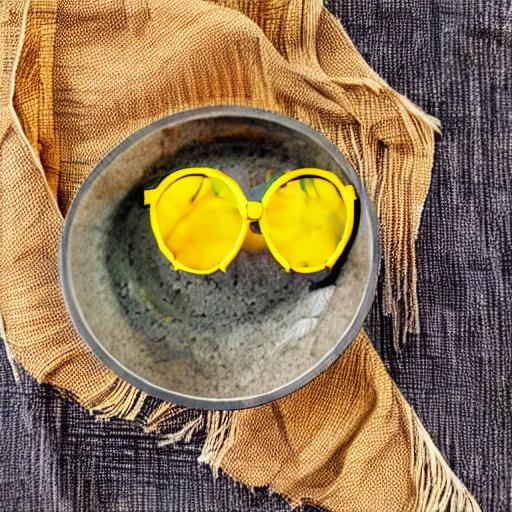} &
        \hspace{0.05cm}
        \includegraphics[width=0.115\textwidth]{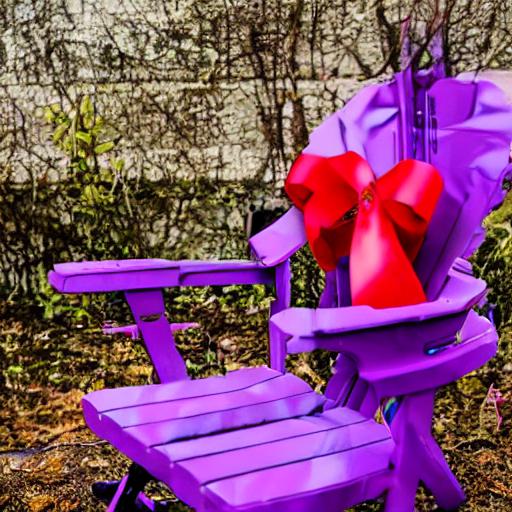} &
        \includegraphics[width=0.115\textwidth]{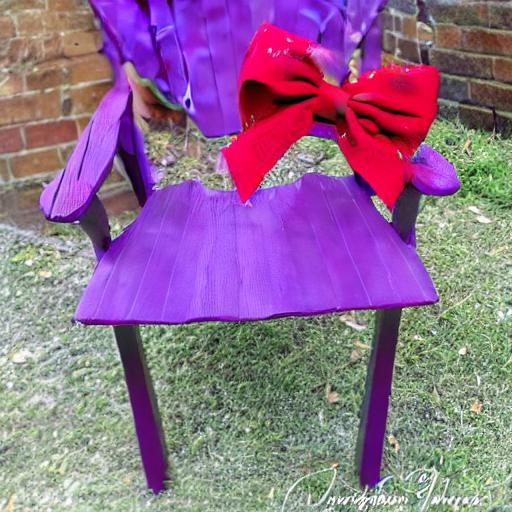} \\ \\ \\

        {\raisebox{0.025in}{\rotatebox{90}{Attend-\&-Excite}}} &
        \includegraphics[width=0.115\textwidth]{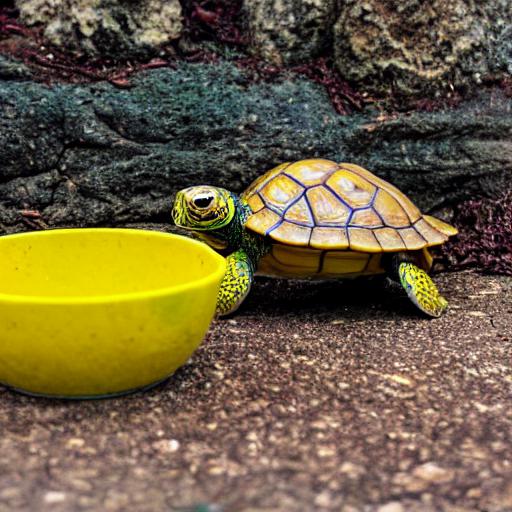} &
        \includegraphics[width=0.115\textwidth]{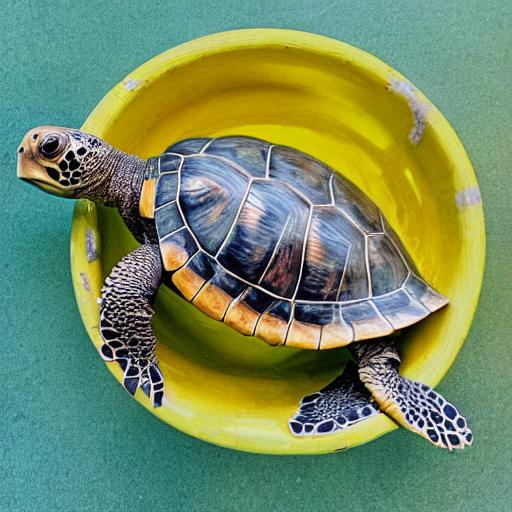} &
        \hspace{0.05cm}
        \includegraphics[width=0.115\textwidth]{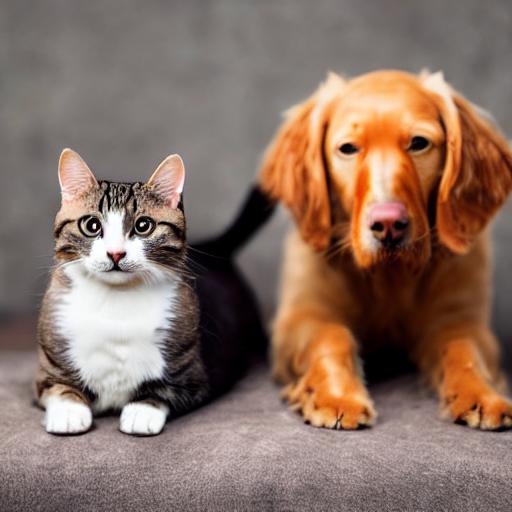} &
        \includegraphics[width=0.115\textwidth]{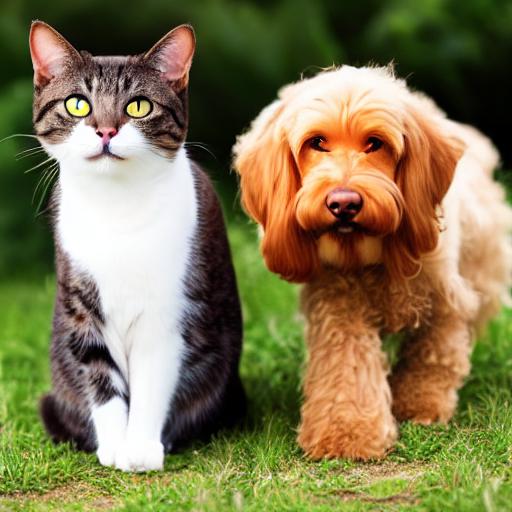} &
        \hspace{0.05cm}
        \includegraphics[width=0.115\textwidth]{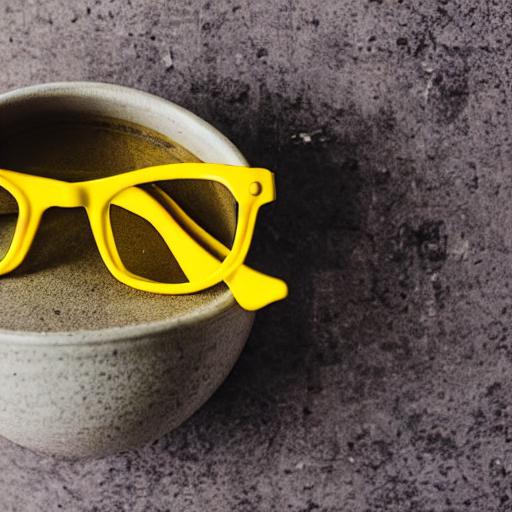} &
        \includegraphics[width=0.115\textwidth]{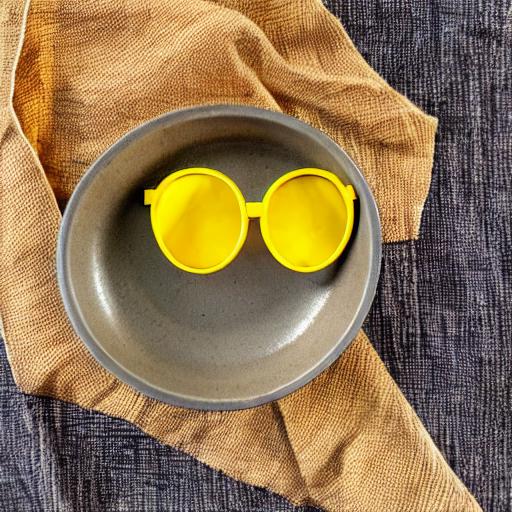} &
        \hspace{0.05cm}
        \includegraphics[width=0.115\textwidth]{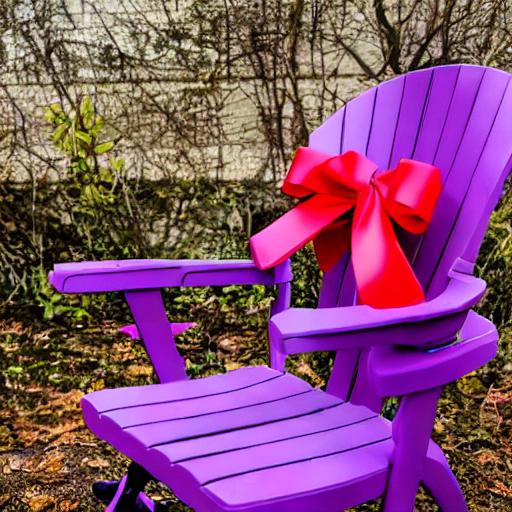} &
        \includegraphics[width=0.115\textwidth]{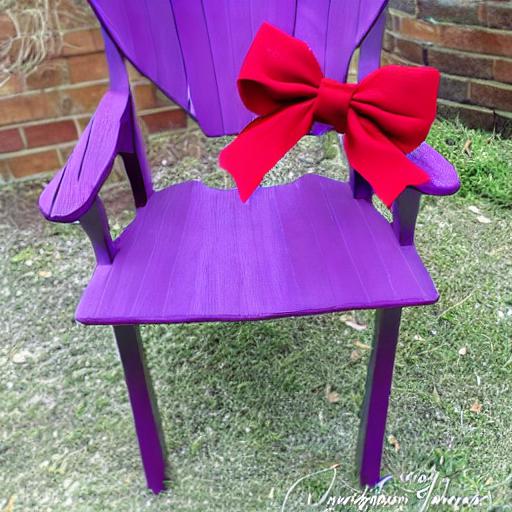} \\ \\ \\

    \end{tabular}
    
    }
    \vspace{-0.25cm}
    \caption{Ablation Study. We demonstrate the results obtained by our method with and without early stopping after $25$ steps. Modifying all steps does not result in a significant semantic change, and adds artifacts to the resulting image.}
    \label{fig:ablation_all_steps}
\end{figure*}

Next, we validate the decision to stop the latent modification after $25$ denoising steps (\ie, after half of all denoising steps). This follows from the observation that the spatial location of each subject is determined in the early denoising steps~\cite{hertz2022prompt,voynov2022sketch}. As such, applying our latent update toward the end of the denoising process will most likely have a negligible effect on the spatial layout of the resulting image. Moreover, we found that applying the latent updates after $25$ iterations leads to unwanted artifacts in the resulting images. This is illustrated in~\Cref{fig:ablation_all_steps} where we compare our Attend-and-Excite technique where we either modify the latent at all timesteps (top) or modify only during the first $25$ timesteps (bottom). As can be seen, stopping the modification early leads to sharper, higher-quality images. For example, our complete Attend-and-Excite method is able to better capture the shape of the yellow bowl on the left or the purple chair on the right while generating finer details such as the face of the dog in the second column.

\begin{figure*}
    \centering
    \setlength{\tabcolsep}{0.5pt}
    \renewcommand{\arraystretch}{0.3}
    {\small
    \begin{tabular}{c c c @{\hspace{0.2cm}} c c @{\hspace{0.2cm}} c c @{\hspace{0.2cm}} c c }

        &
        \multicolumn{2}{c}{``A \textcolor{blue}{cat} and a \textcolor{blue}{dog}''} &
        \multicolumn{2}{c}{``A \textcolor{blue}{turtle} and a yellow \textcolor{blue}{bowl}''} &
        \multicolumn{2}{c}{``A \textcolor{blue}{frog} and a pink \textcolor{blue}{bench}''}
        &
        \multicolumn{2}{c}{\begin{tabular}{c} ``A red \textcolor{blue}{bench} \\ \\ and a yellow \textcolor{blue}{clock}''
        \end{tabular}} \\

        {\raisebox{0.375in}{
        \multirow{2}{*}{\rotatebox{90}{Prompt-to-Prompt}}}} &
        \includegraphics[width=0.11\textwidth]{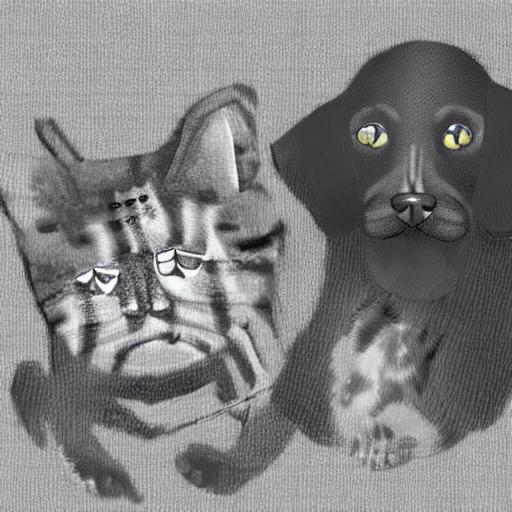} &
        \includegraphics[width=0.11\textwidth]{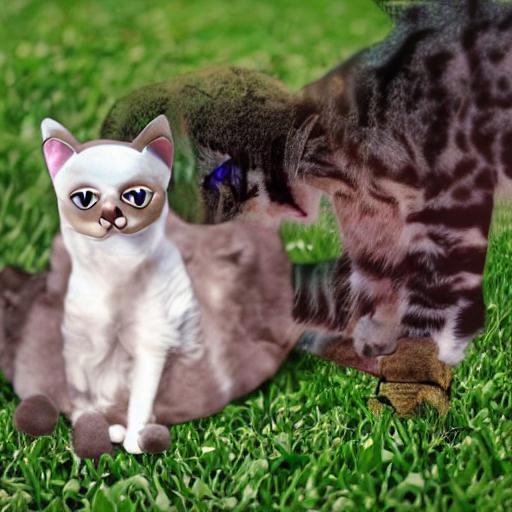} &
        \hspace{0.05cm}
        \includegraphics[width=0.11\textwidth]{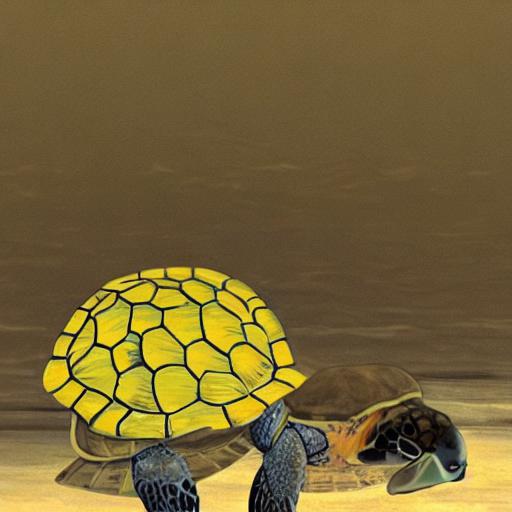} &
        \includegraphics[width=0.11\textwidth]{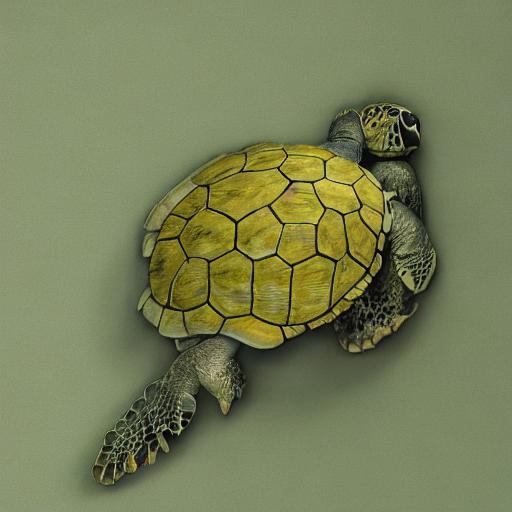} &
        \hspace{0.05cm}
        \includegraphics[width=0.11\textwidth]{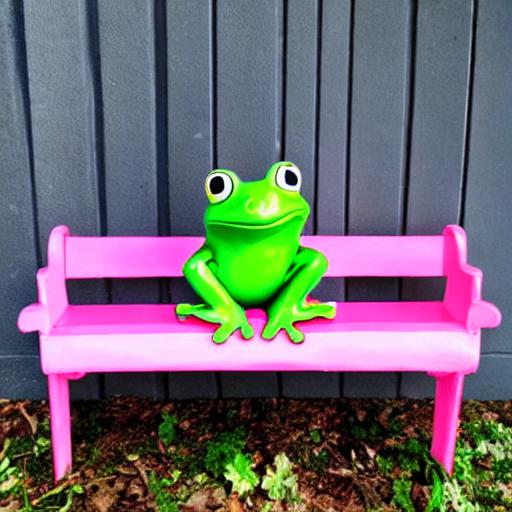} &
        \includegraphics[width=0.11\textwidth]{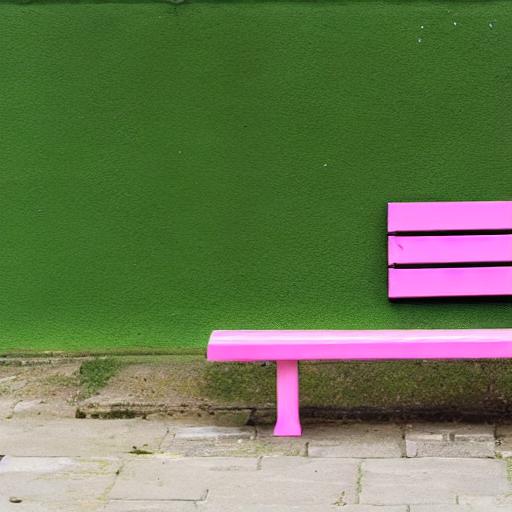} &
        \hspace{0.05cm}
        \includegraphics[width=0.11\textwidth]{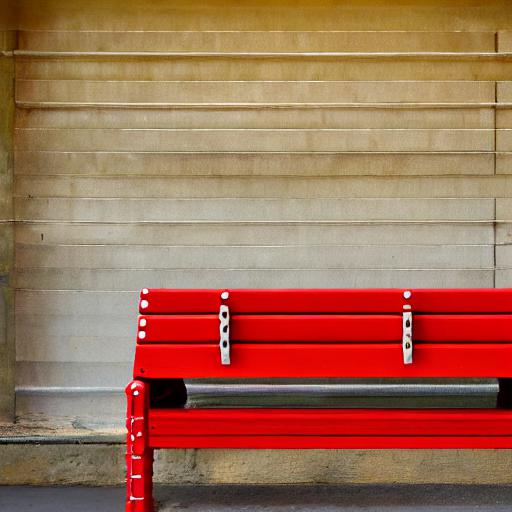} &
        \includegraphics[width=0.11\textwidth]{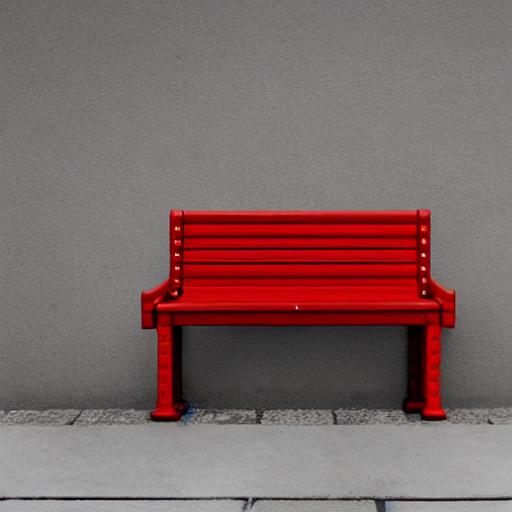} \\

        &
        \includegraphics[width=0.11\textwidth]{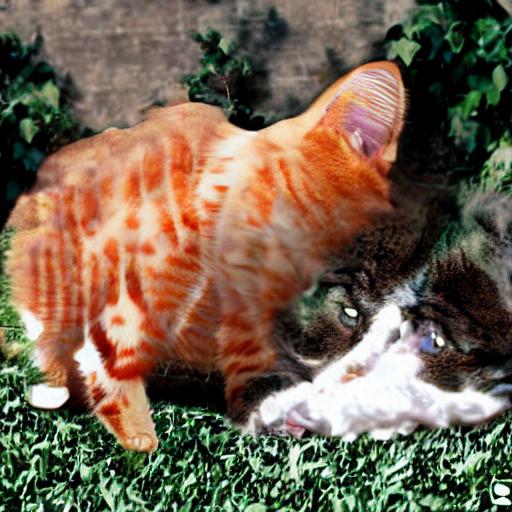} &
        \includegraphics[width=0.11\textwidth]{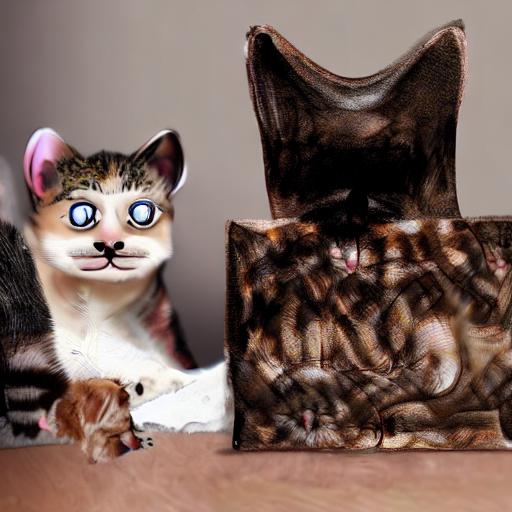} &
        \hspace{0.05cm}
        \includegraphics[width=0.11\textwidth]{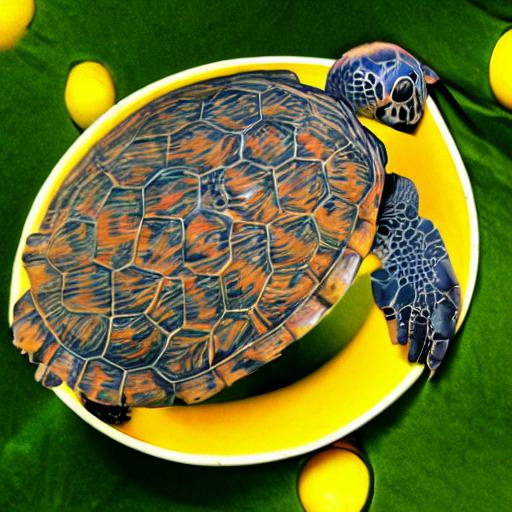} &
        \includegraphics[width=0.11\textwidth]{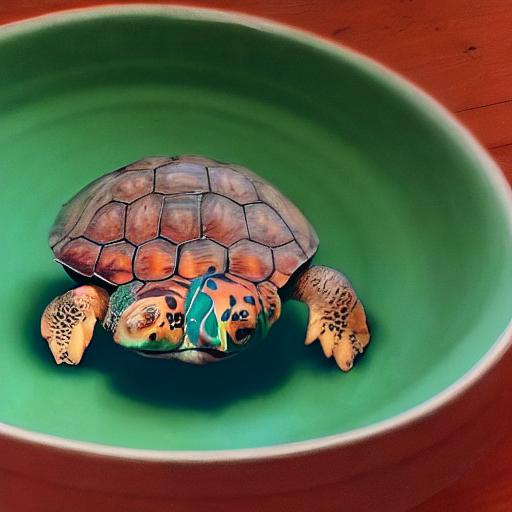} &
        \hspace{0.05cm}
        \includegraphics[width=0.11\textwidth]{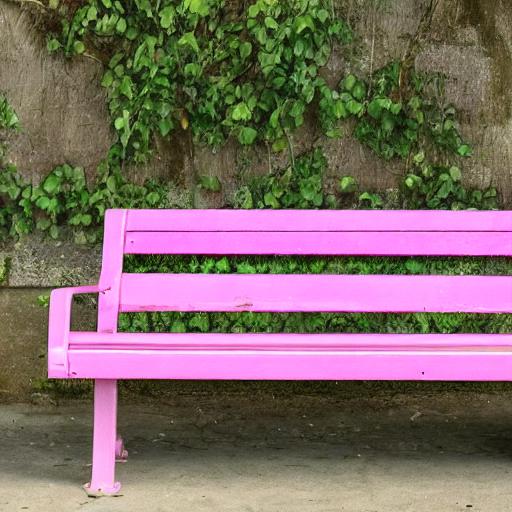} &
        \includegraphics[width=0.11\textwidth]{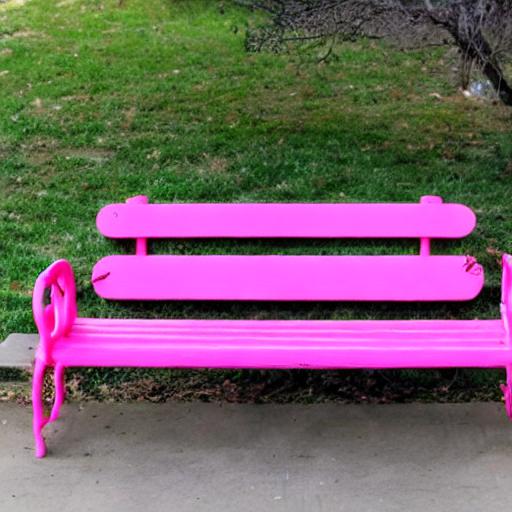} &
        \hspace{0.05cm}
        \includegraphics[width=0.11\textwidth]{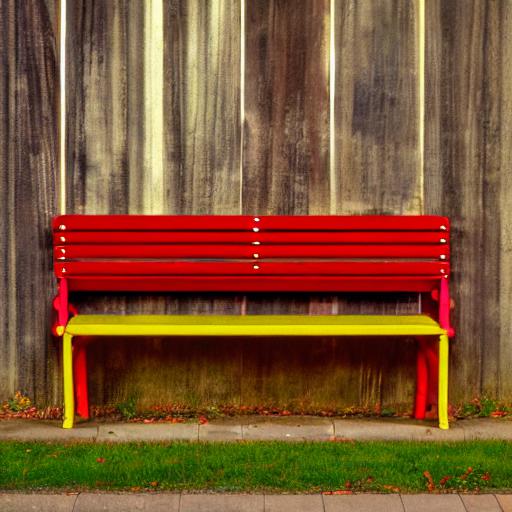} &
        \includegraphics[width=0.11\textwidth]{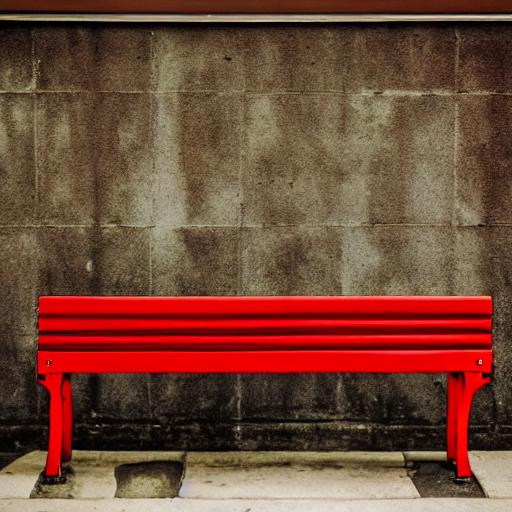}  \\ \\

        {\raisebox{0.425in}{
        \multirow{2}{*}{\rotatebox{90}{\begin{tabular}{c} Stable Diffusion with \\ \textcolor{blue}{Attend-and-Excite} \\ \\ \end{tabular}}}}} &
        \includegraphics[width=0.11\textwidth]{images/outputs/a_cat_and_a_dog/6119.jpg} &
        \includegraphics[width=0.11\textwidth]{images/outputs/a_cat_and_a_dog/ours_10936.jpg} &
        \hspace{0.05cm}
         \includegraphics[width=0.11\textwidth]{images/outputs/a_turtle_and_a_yellow_bowl/44471.jpg} &
        \includegraphics[width=0.11\textwidth]{images/outputs/a_turtle_and_a_yellow_bowl/32636.jpg} &
        \hspace{0.05cm}
        \includegraphics[width=0.11\textwidth]{images/outputs/a_frog_and_a_pink_bench/33.jpg} &
        \includegraphics[width=0.11\textwidth]{images/outputs/a_frog_and_a_pink_bench/6714.jpg} &
        \hspace{0.05cm}
        \includegraphics[width=0.11\textwidth]{images/outputs/a_red_bench_and_a_yellow_clock/46.jpg} &
        \includegraphics[width=0.11\textwidth]{images/outputs/a_red_bench_and_a_yellow_clock/1077.jpg}  \\

        &
        \includegraphics[width=0.11\textwidth]{images/outputs/a_cat_and_a_dog/ours_637.jpg} &
        \includegraphics[width=0.11\textwidth]{images/outputs/a_cat_and_a_dog/25.jpg} &
        \hspace{0.05cm}
        \includegraphics[width=0.11\textwidth]{images/outputs/a_turtle_and_a_yellow_bowl/ours_3399.jpg} &
        \includegraphics[width=0.11\textwidth]{images/outputs/a_turtle_and_a_yellow_bowl/32134.jpg} &
        \hspace{0.05cm}
        \includegraphics[width=0.11\textwidth]{images/outputs/a_frog_and_a_pink_bench/6177.jpg} &
        \includegraphics[width=0.11\textwidth]{images/outputs/a_frog_and_a_pink_bench/392.jpg} &
        \hspace{0.05cm}
        \includegraphics[width=0.11\textwidth]{images/outputs/a_red_bench_and_a_yellow_clock/3798.jpg} &
        \includegraphics[width=0.11\textwidth]{images/outputs/a_red_bench_and_a_yellow_clock/8909.jpg}

    \end{tabular}

    }
    \vspace{-0.3cm}
    \caption{{Qualitative comparison. For each prompt, we show four images generated by Prompt-to-Prompt and Attend-and-Excite where we use the same set of seeds as in the main paper.
    The subject tokens optimized by Attend-and-Excite are highlighted in \textcolor{blue}{blue}.}
    }
    \vspace{-0.1cm}
    \label{fig:ptp_comp}
\end{figure*}

\null\newpage

\section{Additional Results and Comparisons}~\label{sec:additional_results}

\subsection{Comparison to Image Editing Methods}
In the context of image editing, Hertz~\etal~\shortcite{hertz2022prompt} propose Prompt-to-Prompt (ptp) which manipulates the cross-attention units to edit images generated by text-to-image diffusion models. Three variants are presented; (i) word swapping, where a single word in the prompt is replaced with a target word (\ie, object replacement), (ii) prompt extension, where additional text is added to the input prompt, and (iii) attention re-weighting, where the attention map of a word is amplified or reduced to increase or decrease the presence of the word in the generated image. This variant is introduced mainly to allow the user to control the magnitude of a property of the image (\eg, control the amount of snow in the image). Note that the first two variants are irrelevant to the task of obtaining semantically faithful images, as they refer to cases where the input prompt is modified such that the semantic meaning of the text is changed. The third variant is closest to our task at hand. In this section, we explore the attention re-weighting variant as an additional baseline.

Attention re-weighting is performed such that given a set $X$ of tokens to amplify or reduce, ptp scales the spatial attention maps corresponding to the tokens $x\in X$ by a parameter $c_x\in [-2, 2]$, as follows,
\begin{equation}
    A_t^x \gets c_x\cdot A_t^x.
\end{equation}
In other words, the existing spatial map for each token $x\in X$ is scaled such that if $c_x > 1$ the presence of $x$ is amplified, and otherwise reduced (or ``reversed'' if $c_x < 0$). 

In order to mitigate the issue of catastrophic neglect, one could attempt to use attention re-weighting for the subject tokens in the prompt ($S$) and apply $c_x >1$ in order to encourage the presence of all subject tokens in the generated image. We note that while the basic idea of manipulating the attention values is also employed by our method, ptp's attention re-weighting performs \textit{local editing}, \ie, the spatial location of the token in the image is \textit{not} modified in the editing process, but rather the objects are amplified \textit{within the same spatial location}. When a subject is not present in the generated image, it is not allocated a spatial location. As such, amplifying its presence in its existing spatial location will have no effect on the generated image. Therefore, intuitively, attention re-weighting should not assist in mitigating catastrophic neglect.

However, for completeness of evaluation, we present additional comparisons to the attention re-weighting variant of ptp in this section. We employ the official implementation of ptp over Stable Diffusion and use the same configuration as presented in the code to perform the attention re-weighting over the three subsets of our constructed dataset (Animal-Animal, Animal-Object, and Object-Object) where we amplify the subject tokens in each prompt. Due to computational limitations, we perform the comparisons on $20$ randomly-selected prompts from each subset.

\begin{table}
    \small
    \centering
    \setlength{\tabcolsep}{2pt}
    \caption{\textit{CLIP-based Text-Text Similarity Comparison with Prompt-to-Prompt~\cite{hertz2022prompt}}. We show the average CLIP text-text similarities between the text prompts and captions generated by BLIP~\cite{li2022blip} over the generated images. 
    } 
    \vspace{-0.2cm}
    \begin{tabular}{l c c c} 
    \toprule
    Method & \begin{tabular}{c} Animal-Animal \end{tabular} & \begin{tabular}{c} Animal-Object \end{tabular} & \begin{tabular}{c} Object-Object \end{tabular} \\
    \midrule
    Prompt-to-Prompt    & 0.771 (\textcolor{mydarkgreen}{-4.29\%})   & 0.809 (\textcolor{mydarkgreen}{-2.99\%})  
                         & 0.782 (\textcolor{mydarkgreen}{-3.61\%})          \\
    \textbf{Attend-and-Excite}   & \textbf{0.803} & \textbf{0.833} & \textbf{0.810}  \\
    \bottomrule
    \end{tabular}
    \label{tb:blip_captioning_similarity_ptp}
\end{table}

\begin{figure*}
    \centering
    \begin{tabular}{c c c}
    \includegraphics[height=3.45cm]{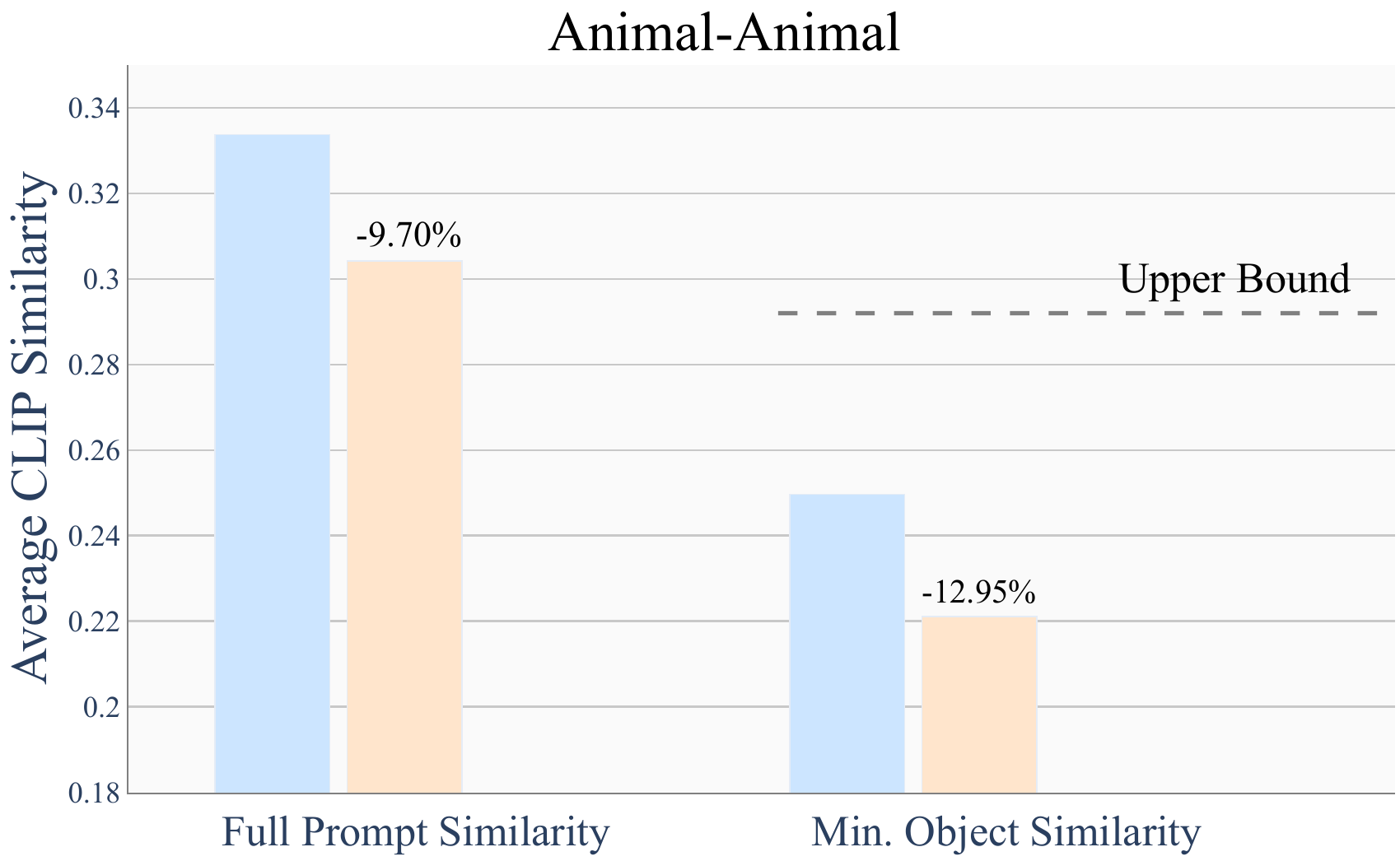} &
    \includegraphics[height=3.45cm]{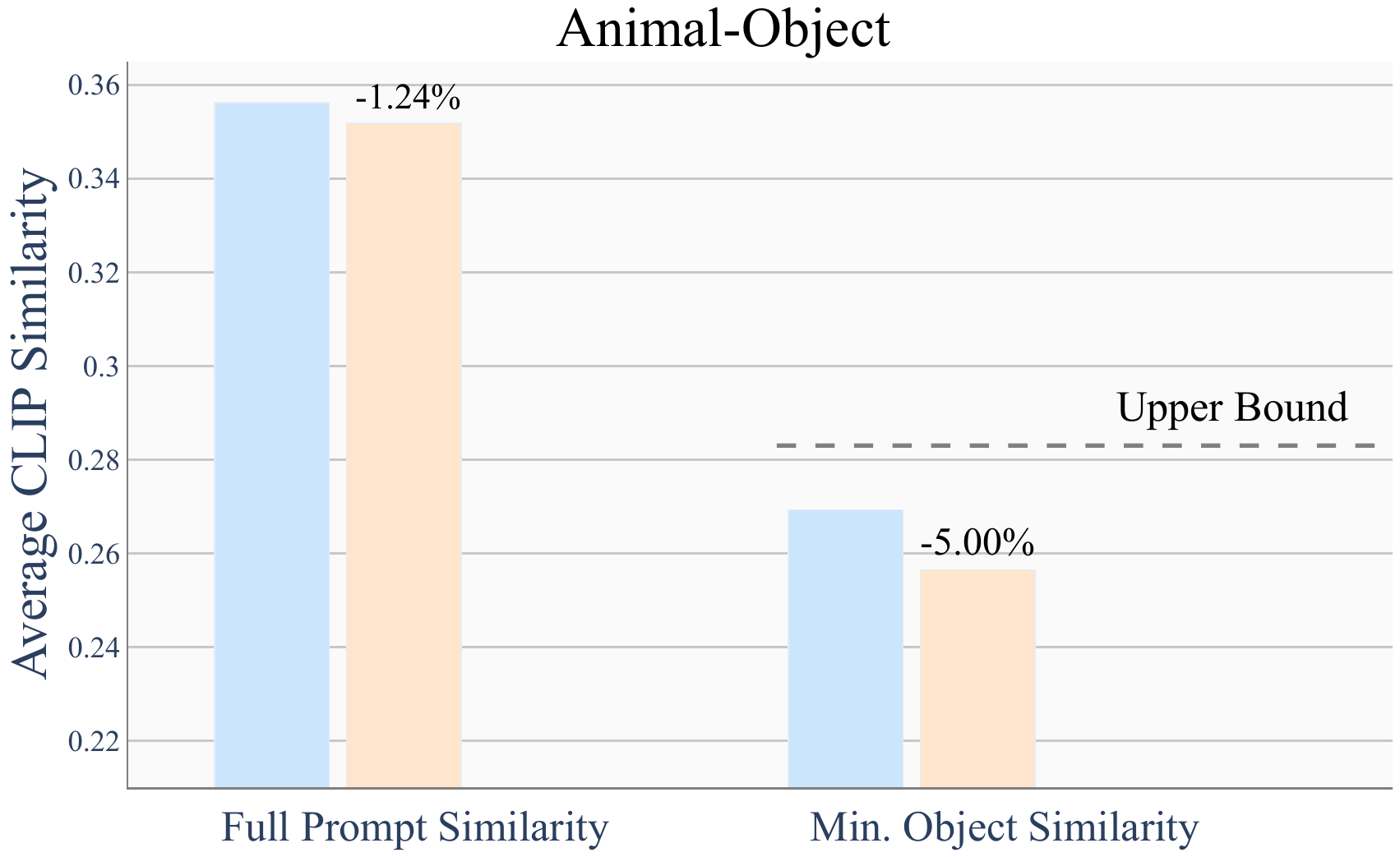}  &
    \includegraphics[height=3.45cm]{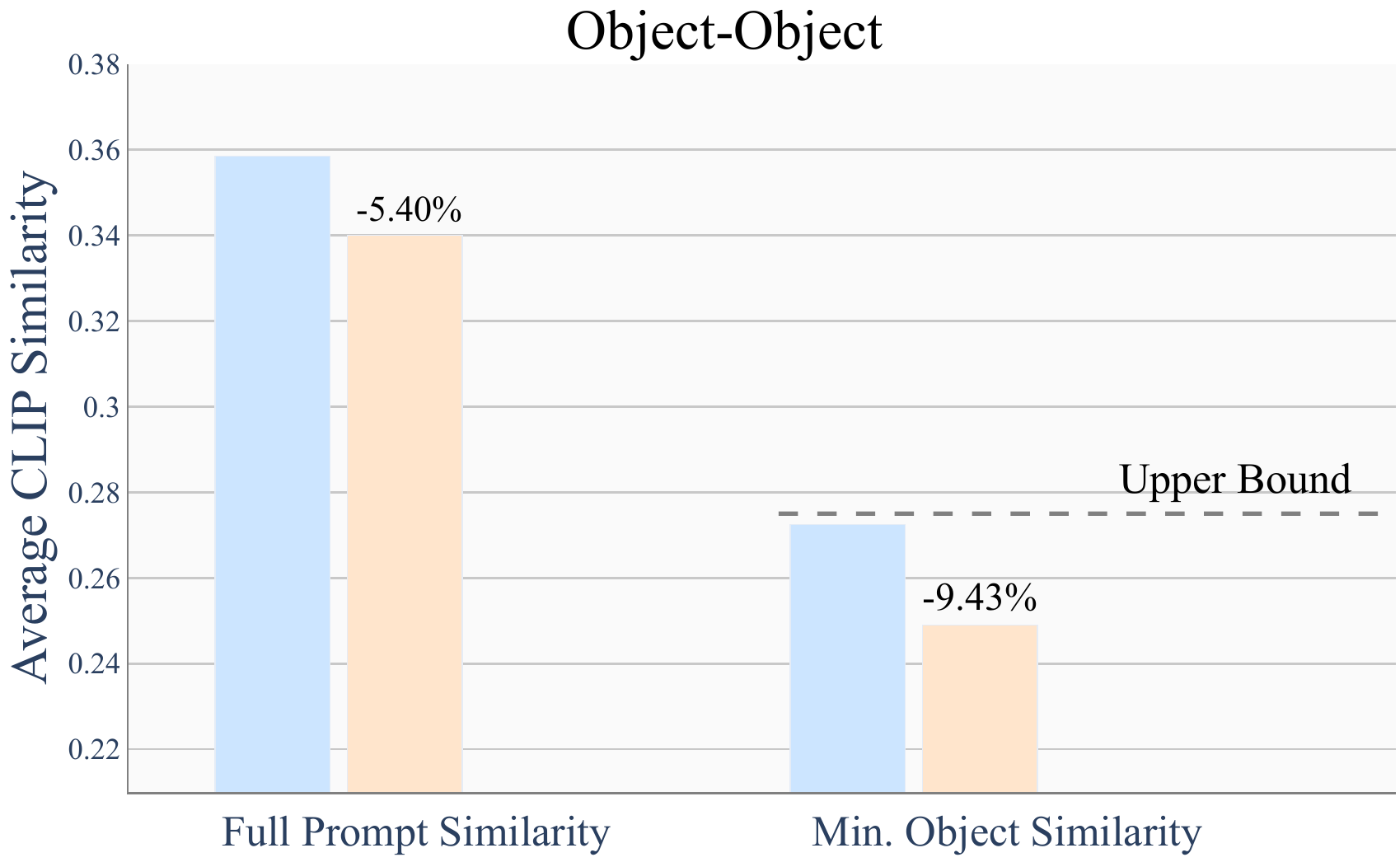}
    \vspace{-0.05cm}
    \end{tabular}
    \begin{tabular}{c c c}
    & \hspace{-0.15cm} \includegraphics[width=0.25\textwidth]{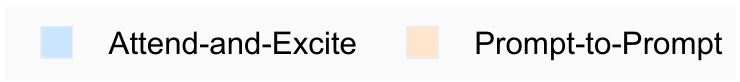} & \\
    \end{tabular}
    \vspace{-0.5cm}
    \caption{Average CLIP image-text similarities between the original text prompts and the images generated by Prompt-to-Prompt~\cite{hertz2022prompt} and our Attend-and-Excite method, split by subset. The \textit{Full Prompt Similarity} indicates the image-text similarity when considering the full-text prompt while \textit{Minimum Object Similarity} represents the average CLIP similarity for the most neglected object. Note, the \textit{Upper Bound} (the maximal-expected similarity) is applicable only to the \textit{Minimum Object Similarity}.}
    \label{fig:clip_image_similarities_ptp}
    \vspace{-0.25cm}
\end{figure*}

\Cref{fig:ptp_comp} presents a qualitative comparison between results obtained by ptp and Attend-and-Excite using the same seeds and prompts as the figure presented in the main paper. As can be seen, ptp often fails to mitigate catastrophic neglect (\eg, no clock is synthesized in the last column), and does not resolve incorrect attribute binding (\eg, in the second column, a green bowl is generated instead of a yellow one). In contrast, Attend-and-Excite produces semantically accurate results that correspond to the input prompts.

Additionally, in~\Cref{fig:clip_image_similarities_ptp} we present the results of the CLIP-based image-text metrics for ptp and Attend-and-Excite across the three subsets (Animal-Animal, Object-Objcet, Animal-Object) for the $20$ randomly selected prompts from each subset.
As can be seen, Attend-and-Excite outperforms ptp across all experiments and does so by at least $5\%$ when considering the \textit{Minimum Object Similarity Metric}, indicating that ptp falls short of Attend-and-Excite in mitigation of the presented semantic issues. 
Finally,~\Cref{tb:blip_captioning_similarity_ptp} contains the results of the BLIP-based text-text similarities across all three subsets. Observe that here too, Attend-and-Excite outperforms ptp by a margin of at least $3\%$ in all three subsets.

\begin{table}
    \small
    \centering
    \setlength{\tabcolsep}{1pt}
    \caption{\textit{CLIP-based Image-Text Similarity Comparison over complex prompts}. We show the average CLIP image-text similarities between the text prompts and generated images. Results are computed over $40$ input prompts with $64$ generated images for each text prompt.}
    \vspace{-0.2cm}
    \begin{tabular}{l c c c} 
    \toprule
    & \begin{tabular}{c} Stable Diffusion \end{tabular} & \begin{tabular}{c} StructureDiffusion \end{tabular} & \begin{tabular}{c} \textbf{Attend-and-Excite} \end{tabular} \\
    \midrule
    CLIP Similarity & 0.338 (\textcolor{mydarkgreen}{-3.85\%}) & 0.336 (\textcolor{mydarkgreen}{-4.40\%}) & \textbf{0.351} \\
    \bottomrule
    \end{tabular}
    \label{tb:complex_prompts}
    \vspace{-0.3cm}
\end{table}

\subsection{Quantitative Evaluation on Complex Prompts}
In the main text, we present an extensive quantitative evaluation over prompts that are constructed as conjunctions of two subjects. This design serves two goals. First, as mentioned, since Composable Diffusion~\cite{liu2022compositional} only operates over conjunctions and negations, we consider prompts that allow us to compare against all relevant baselines. 
Second, we purposefully designed the prompt templates to allow us to analyze neglect \emph{for each subject separately} by splitting each prompt into per-subject sub-prompts, allowing us to compute the \textit{Minimum Object Similarity} score. This analysis is critical since full prompt metrics may not reliably capture neglect~\cite{paiss2022no}.

With that, for completeness of evaluation, in this section, we conduct a quantitative analysis on complex prompts with three or more subjects, challenging attribute bindings, and background subjects. We collected $40$ 
prompts extracted from the qualitative examples in the StructureDiffusion paper~\cite{Feng2022Training}, and the Conceptual Captions dataset~\cite{sharma2018conceptual}. We compare our method against both StructureDiffusion and Stable Diffusion and use the same $64$ random seeds for all methods.
Since the prompts are no longer separable into per-subject prompts, we present the \textit{Full Prompt Similarity} metric, which estimates the average CLIP similarity between the generated images and their corresponding prompts. As can be seen in~\Cref{tb:complex_prompts}, Attend-and-Excite out-performs all baselines, with an improvement of $3.8\%$ over Stable Diffusion, and a $4.4\%$ improvement over StructureDiffusion. This further validates our method's ability to tackle challenging prompts and mitigate semantic issues in more complex cases.

\subsection{Evaluation of Generation Quality}
Since Attend-and-Excite shifts the input latent to the UNet network, it is useful to evaluate the quality of the generated images compared to the results produced by Stable Diffusion. We evaluate two aspects of the generation quality with Attend-and-Excite. First, the perceived image quality (\ie, evaluation of artifacts that may be introduced by our method), and second, in accordance with our \textit{Minimum Object Similarity} metric, we evaluate the presence and quality of the maximal subject in the input prompt. 

\begin{table}
    \small
    \centering
    \setlength{\tabcolsep}{2pt}
    \caption{\textit{Average Maximum Object CLIP-based Image-Text Similarity metric}. We show the Max Object Similarity metric obtained for images generated by Stable Diffusion and images generated by Attend-and-Excite.  
    }
    \vspace{-0.2cm}
    \begin{tabular}{l c c c} 
    \toprule
    Method & \begin{tabular}{c} Animal-Animal \end{tabular} & \begin{tabular}{c} Animal-Object \end{tabular} & \begin{tabular}{c} Object-Object \end{tabular} \\
    \midrule
    Stable Diffusion    & \textbf{0.287} (\textcolor{red}{+2.93\%}) & 0.305 (\textcolor{red}{+2.66\%}) & 0.319 (\textcolor{red}{+2.41\%})          \\
    \textbf{Attend-and-Excite}   & 0.279 & 0.297 & 0.312  \\
    \bottomrule
    \end{tabular}
    \label{tb:max_similarity}
    \vspace{-0.4cm}
\end{table}

The perceived image quality is evaluated in~\Cref{fig:ablation_all_steps}, where we test the choice to employ early stopping at step $25$ of the denoising process. 
As can be seen, this design choice assists in maintaining high-quality images since we only perform manipulation in the early denoising steps to encourage the generation of all subjects, while the final timesteps are performed without intervention. 

Second, we evaluate the generation quality of the non-neglected subjects by evaluating the \textit{Maximum Object Similarity}. Computing the maximal similarity score is used to validate that Attend-and-Excite does not harm the generated subjects when strengthening the neglected subjects(s).
In accordance with the \textit{Minimum Object Similarity} metric, we evaluate the CLIP similarity for the \textit{least} neglected subject independently of the full text. To this end, we split the prompt into two sub-prompts, each containing a single subject (\eg, ``a cat'', ``a dog''). 
We then compute the CLIP similarity between each sub-prompt and each generated image. Given the two scores for each image, we are interested in the higher score, as it corresponds to the least neglected subject in the image. We then average the  scores across all seeds and prompts. As shown in~\Cref{tb:max_similarity}, Attend-and-Excite obtains results that are on par with Stable Diffusion, albeit slightly lower. 

We note that a slightly lower score is to be expected since Stable Diffusion suffers from neglect. For example, consider the prompt ``A cat and a dog'' and assume that only a cat is generated by Stable Diffusion, while Attend-and-Excite generates both subjects. When considering the \textit{Maximum Object Similarity} for Stable Diffusion, we would compute the similarity between an image of a cat and the prompt ``a cat''. This will naturally be slightly higher than the similarity of the text ``A cat'' to an image of a cat and a dog, which is produced by our method. While a slightly lower score is to be expected, the gap is small, especially in comparison to the improvement attained in the \textit{Minimum Object Similarity} metric that is presented in the main text.

\subsection{Additional Qualitative Results}~\label{sec:additional_qualitative}
\begin{figure*}
    \centering
    \setlength{\tabcolsep}{0.5pt}
    \renewcommand{\arraystretch}{0.3}
    {\small
    \begin{tabular}{c c @{\hspace{0.4cm}} c}
        &
        {\raisebox{0.1in}{{Stable Diffusion}}} &
        {\raisebox{0.1in}{
        {{\begin{tabular}{c} Stable Diffusion with \\ \textcolor{blue}{Attend-and-Excite} 
        \end{tabular}}}}}
        \\
         \raisebox{0.3in}{\rotatebox{90}{``A \textcolor{blue}{parrot} and a \textcolor{blue}{bear}''}}
         &
        \includegraphics[width=0.435\textwidth]{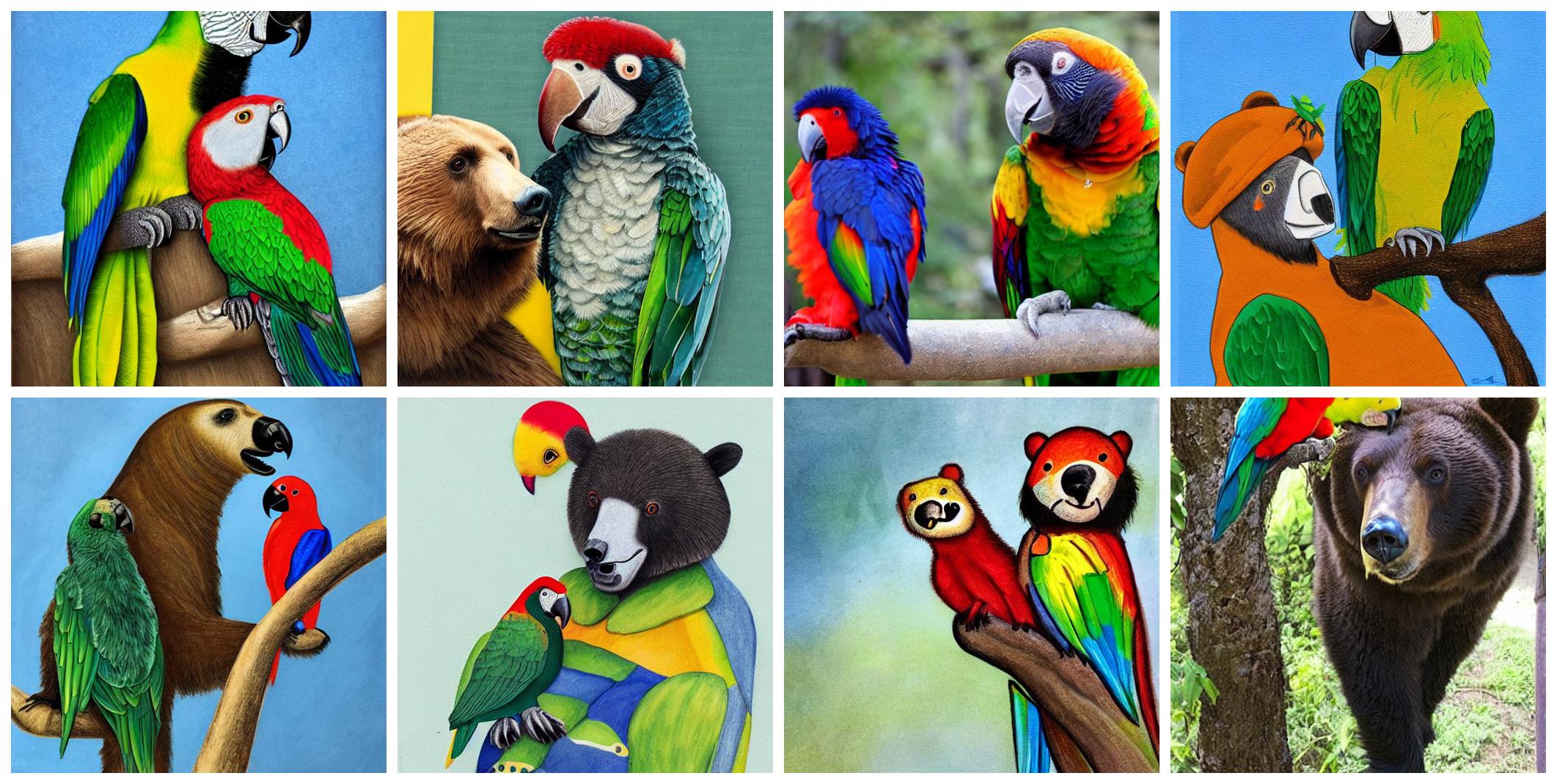} &
        \includegraphics[width=0.435\textwidth]{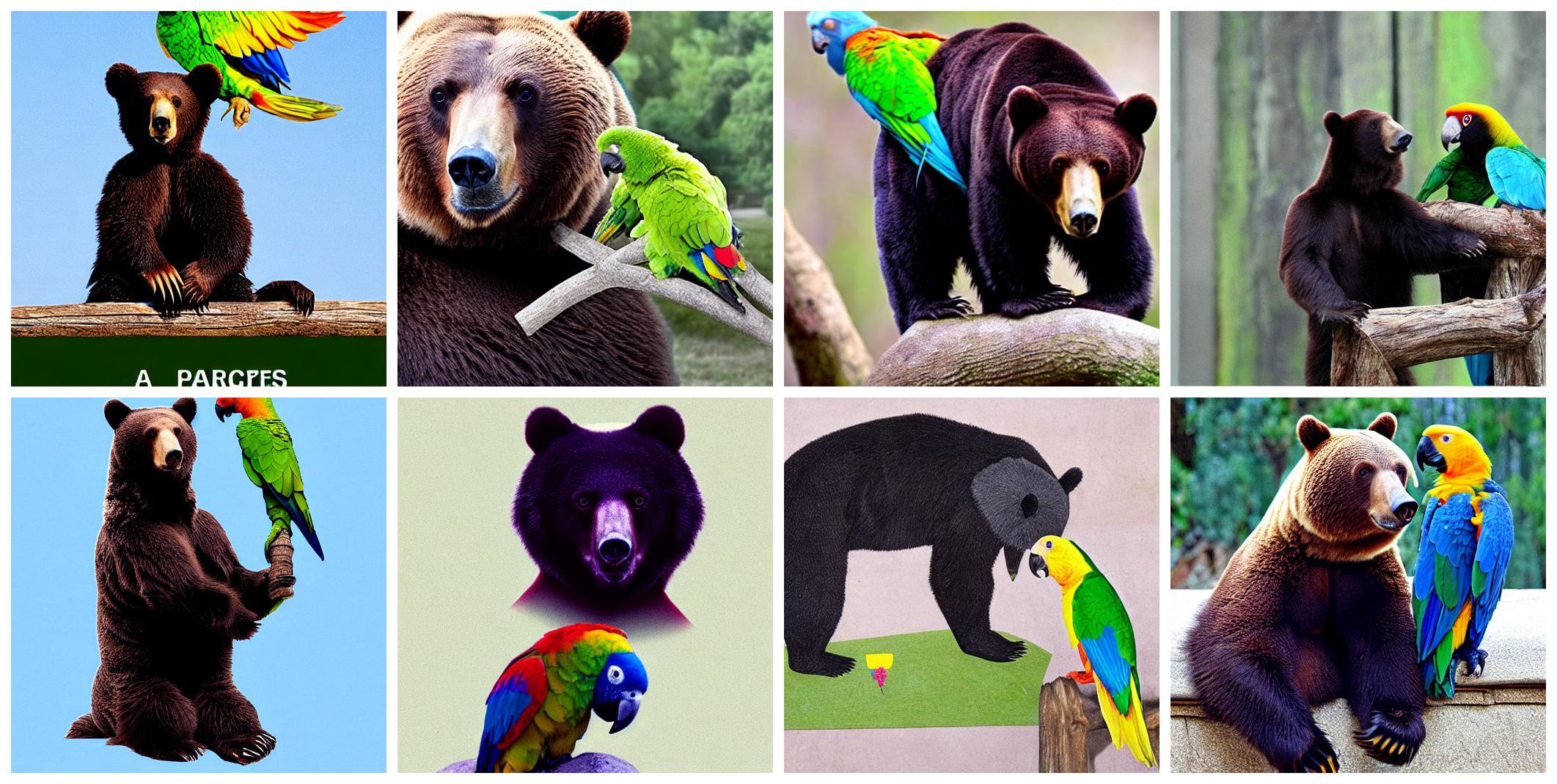} 
        \\\\

         \raisebox{0.3in}{\rotatebox{90}{``A \textcolor{blue}{lion} with a \textcolor{blue}{crown}''}}
         &
        \includegraphics[width=0.435\textwidth]{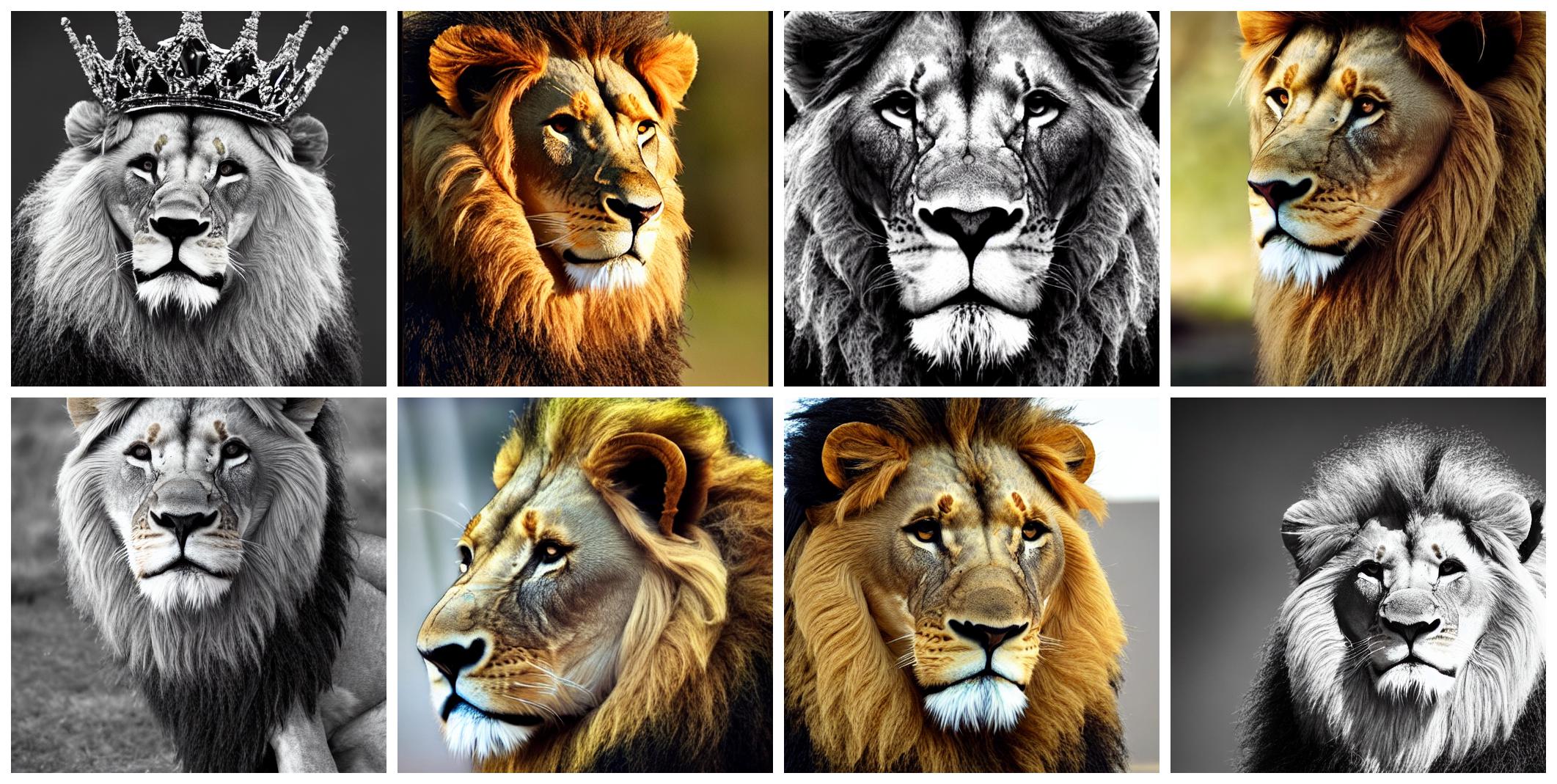} &
        \includegraphics[width=0.435\textwidth]{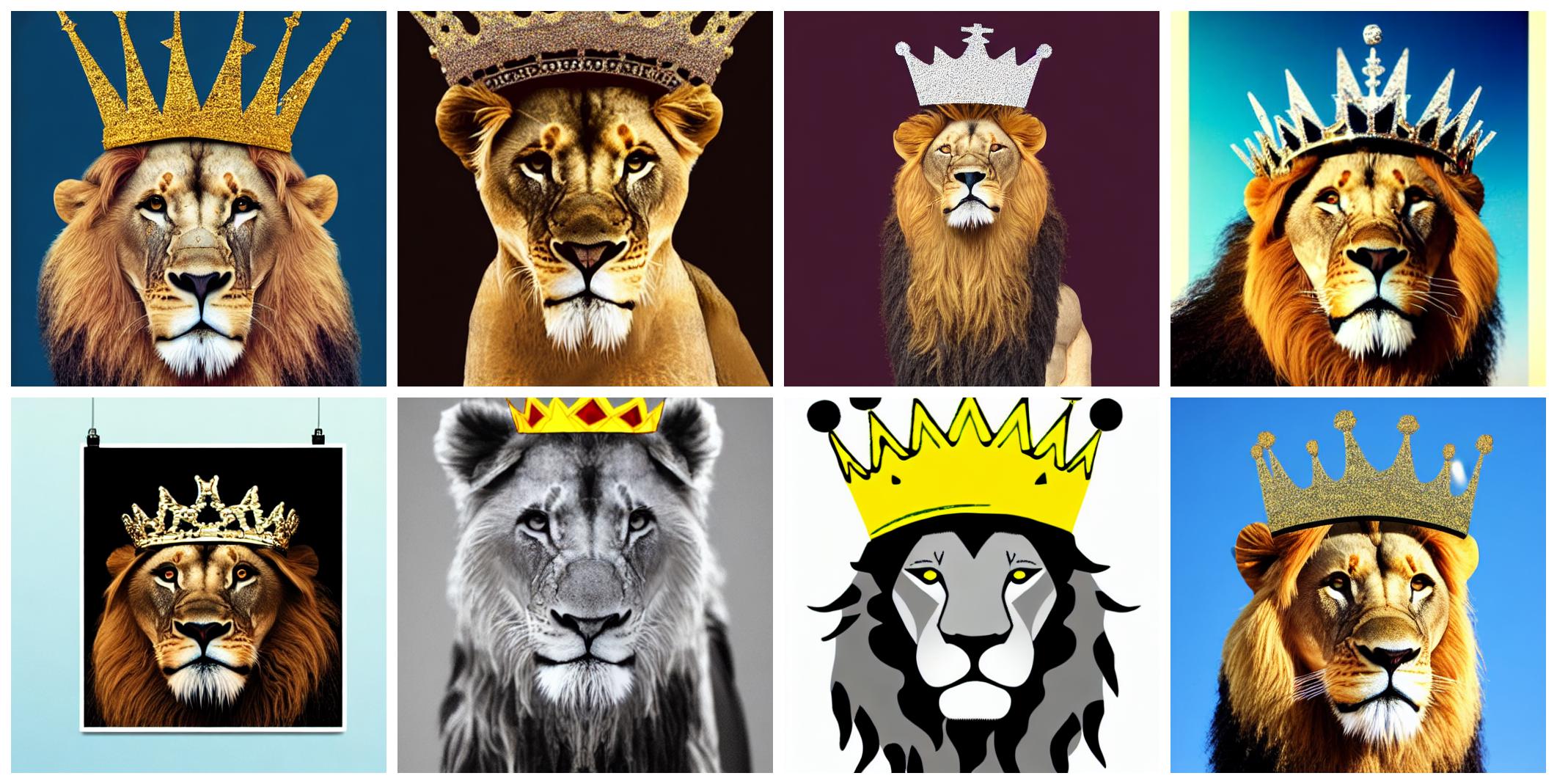} 
        \\\\

        \raisebox{0.2in}{\rotatebox{90}{``A \textcolor{blue}{bird} and a yellow \textcolor{blue}{chair}''}}
         &
        \includegraphics[width=0.435\textwidth]{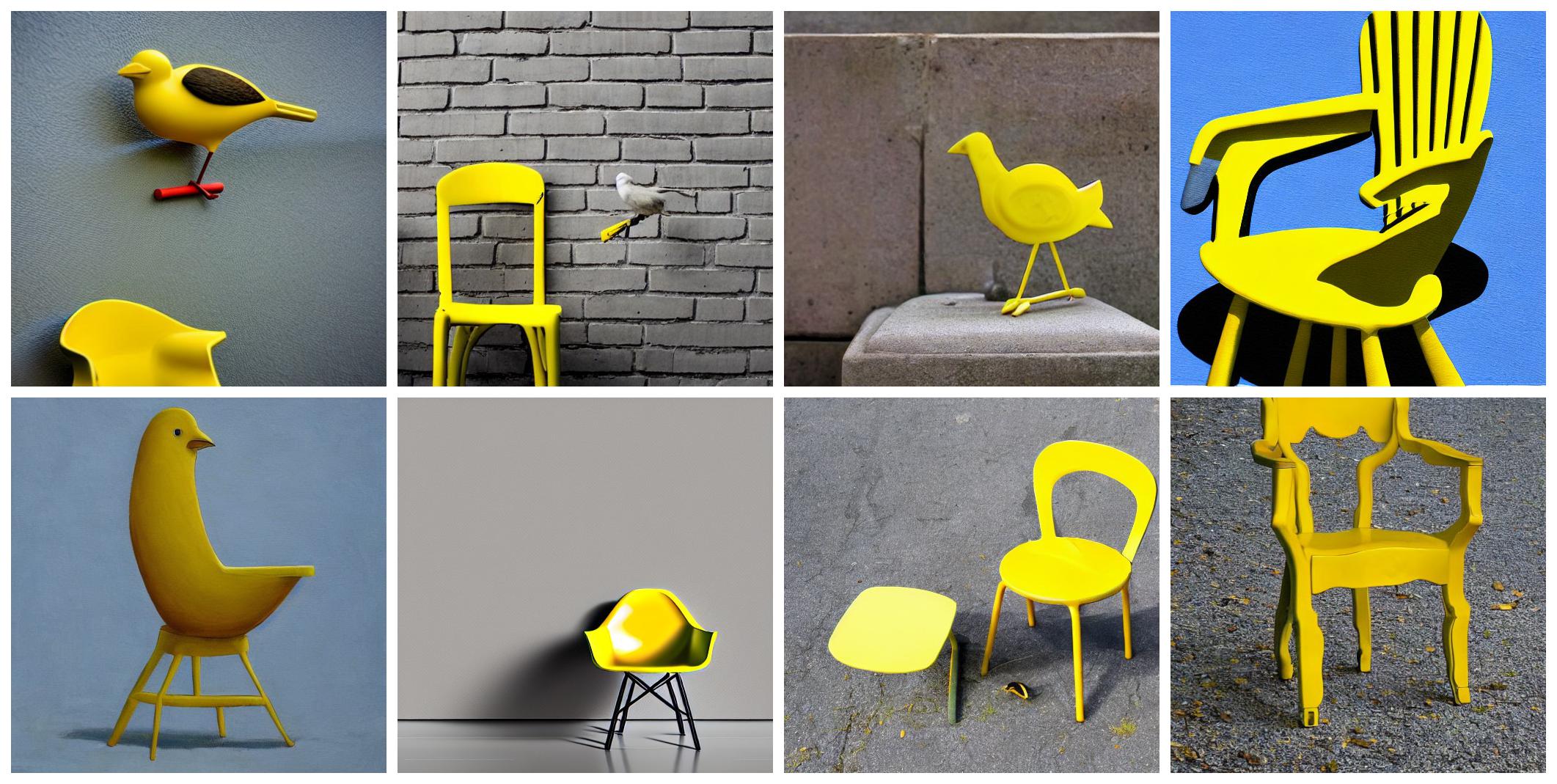} &
        \includegraphics[width=0.435\textwidth]{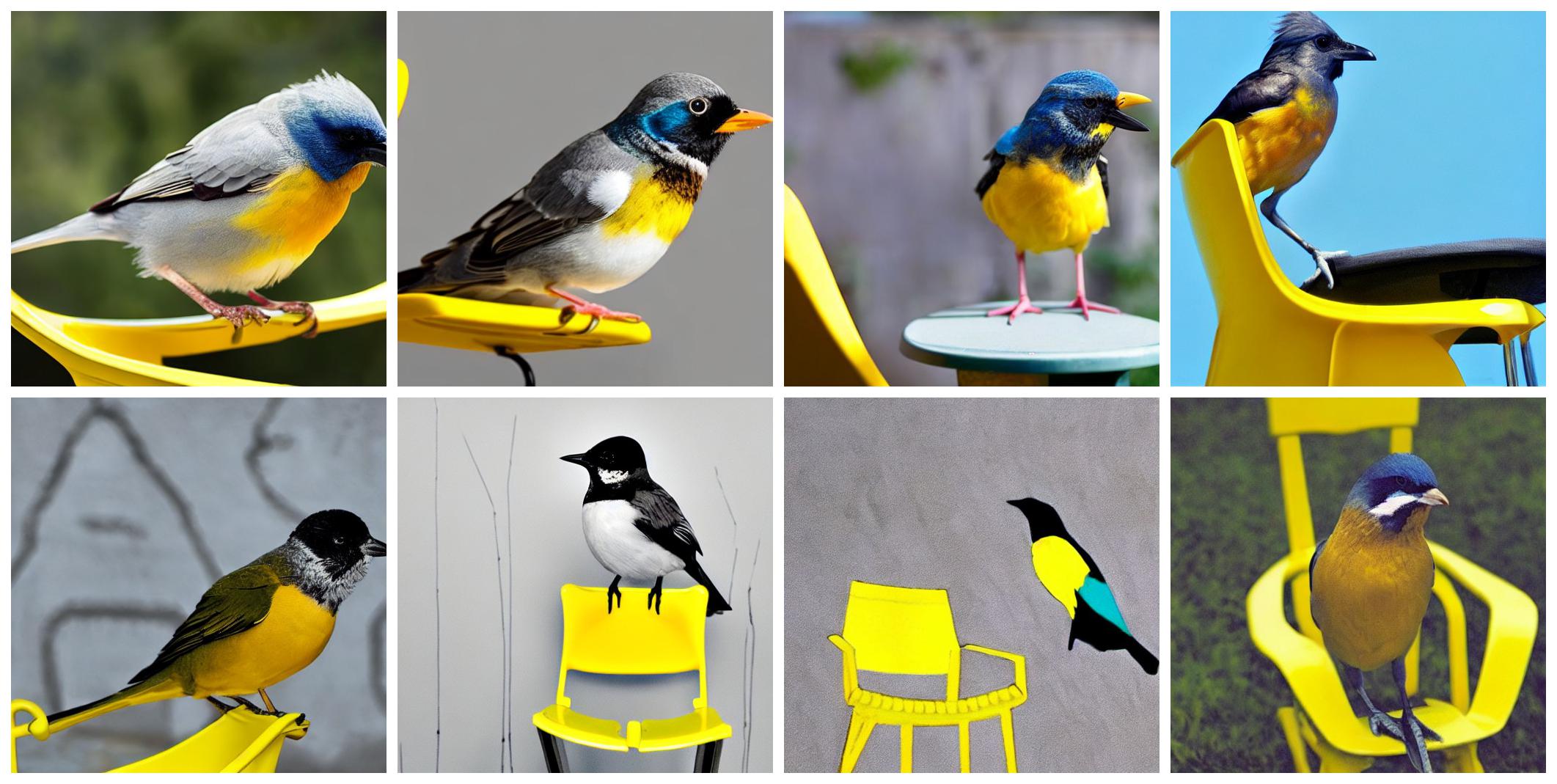} 
        \\\\

        \raisebox{0.2in}{\rotatebox{90}{``A \textcolor{blue}{rabbit} and a blue \textcolor{blue}{bowl}''}}
         &
        \includegraphics[width=0.435\textwidth]{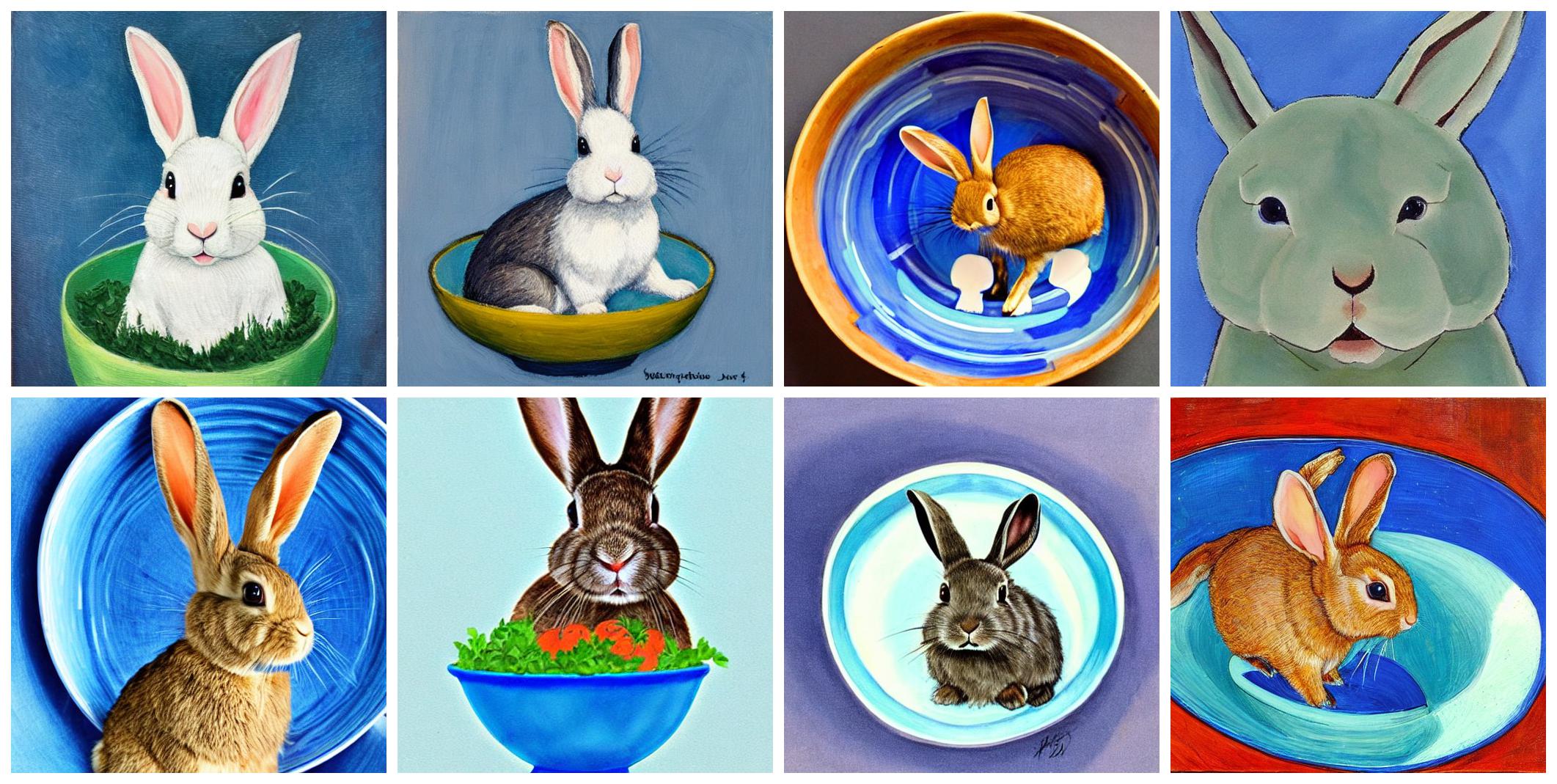} &
        \includegraphics[width=0.435\textwidth]{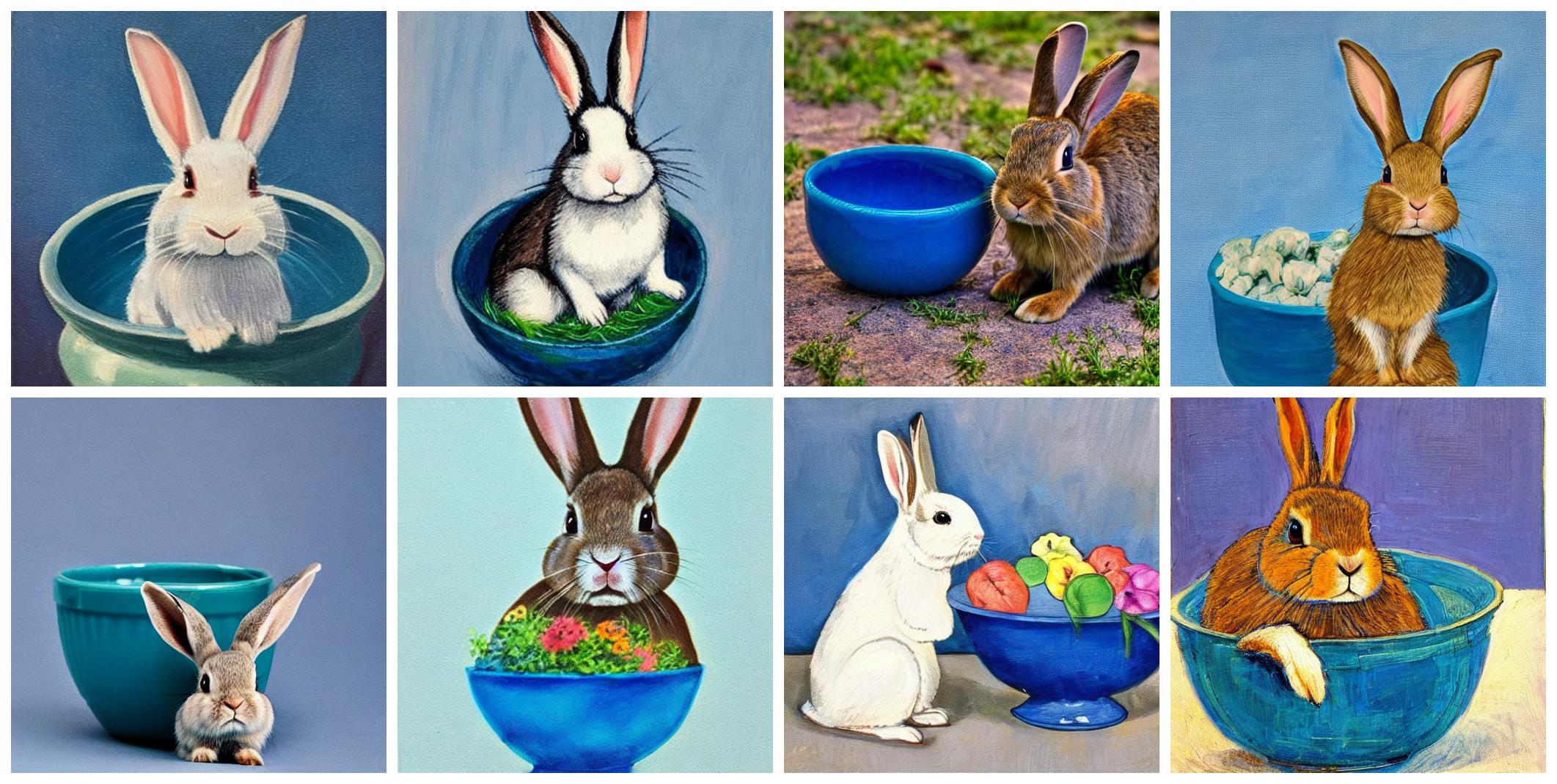} 
        \\\\

        \raisebox{0.1in}{\rotatebox{90}{``A red \textcolor{blue}{bow} and a green \textcolor{blue}{bowl}''}}
         &
        \includegraphics[width=0.435\textwidth]{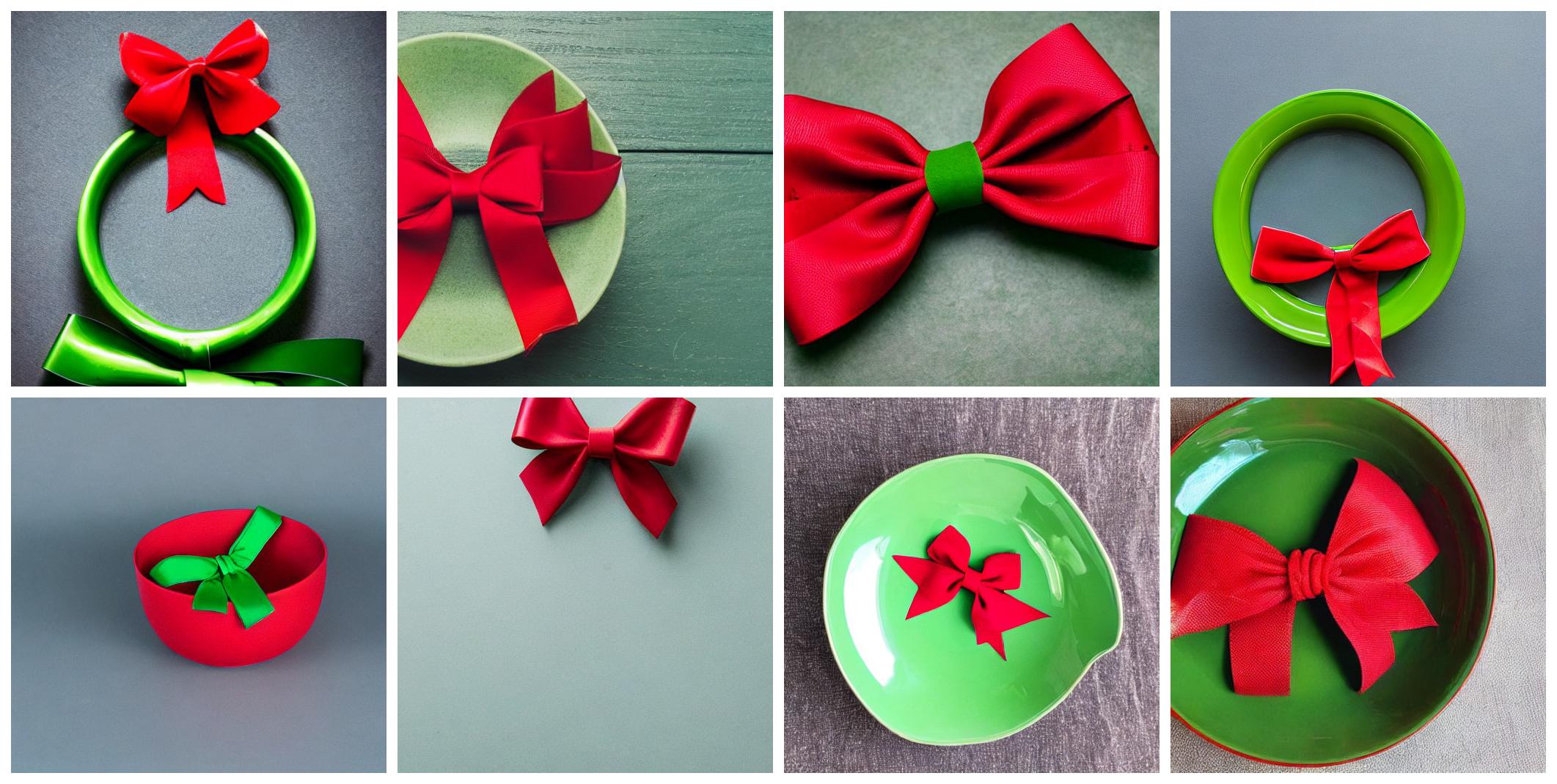} &
        \includegraphics[width=0.435\textwidth]{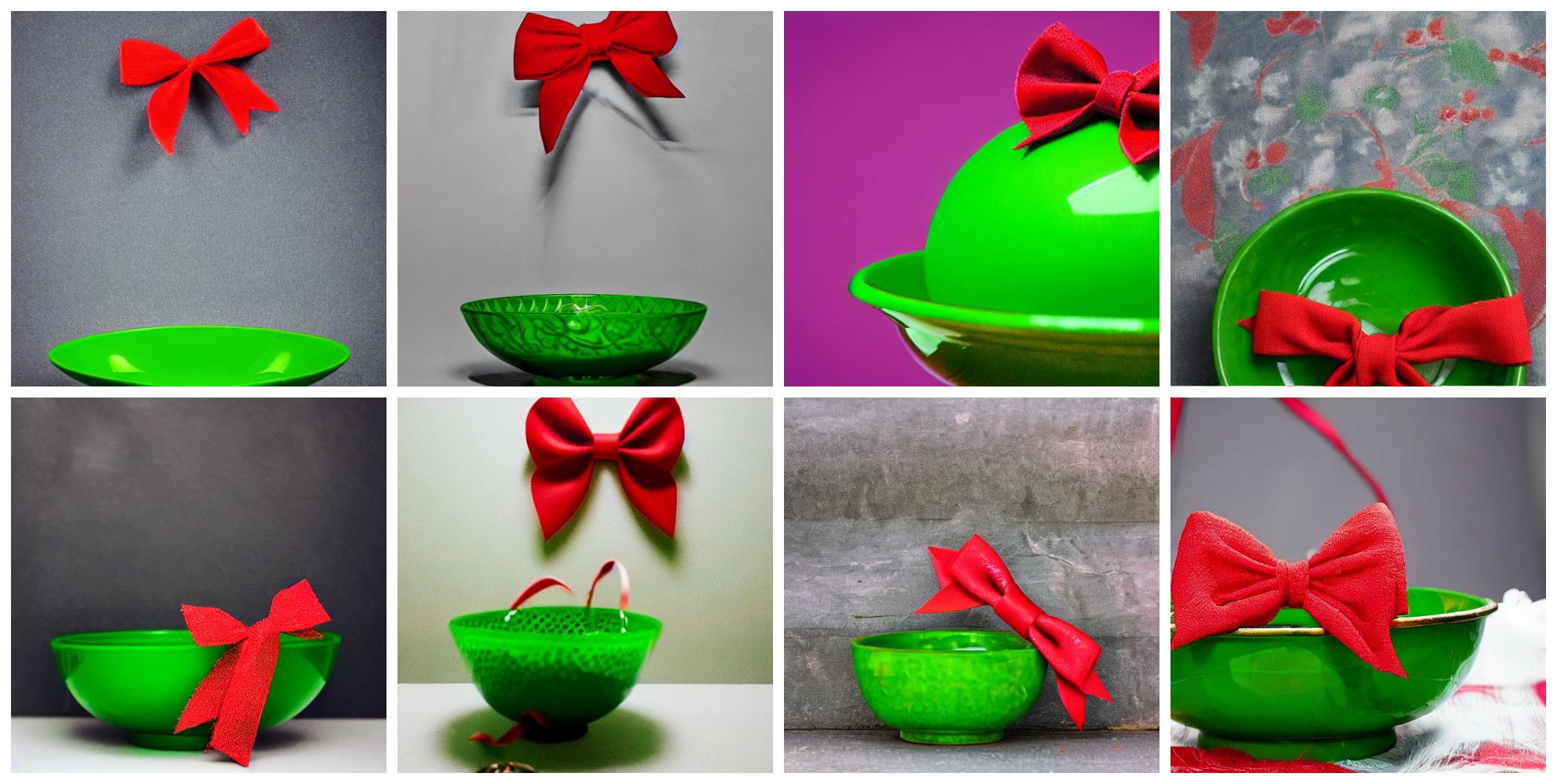} 
    \end{tabular}
    
    }
    \vspace{-0.25cm}
    \caption{Sample uncurated results achieved with Stable Diffusion~\cite{rombach2022high} with and without our Attend-and-Excite approach. For each prompt, we show eight images synthesized when optimizing over the subject tokens highlighted in \textcolor{blue}{blue}. When displaying results with and without Attend-and-Excite we use the same set of random seeds without cherry-picking.}
    \label{fig:uncurated}
\end{figure*}

\begin{figure*}
    \centering
    \setlength{\tabcolsep}{0.5pt}
    \renewcommand{\arraystretch}{0.3}
    {\small
     \begin{tabular}{c c @{\hspace{0.4cm}} c}
        &
        {\raisebox{0.1in}{{Stable Diffusion}}} &
        {\raisebox{0.1in}{
        {{\begin{tabular}{c} Stable Diffusion with \\ \textcolor{blue}{Attend-and-Excite} 
        \end{tabular}}}}}
        \\
         {\rotatebox{90}{``A blue \textcolor{blue}{parrot} and a golden \textcolor{blue}{cat}''}}
         &
        \includegraphics[width=0.42\textwidth]{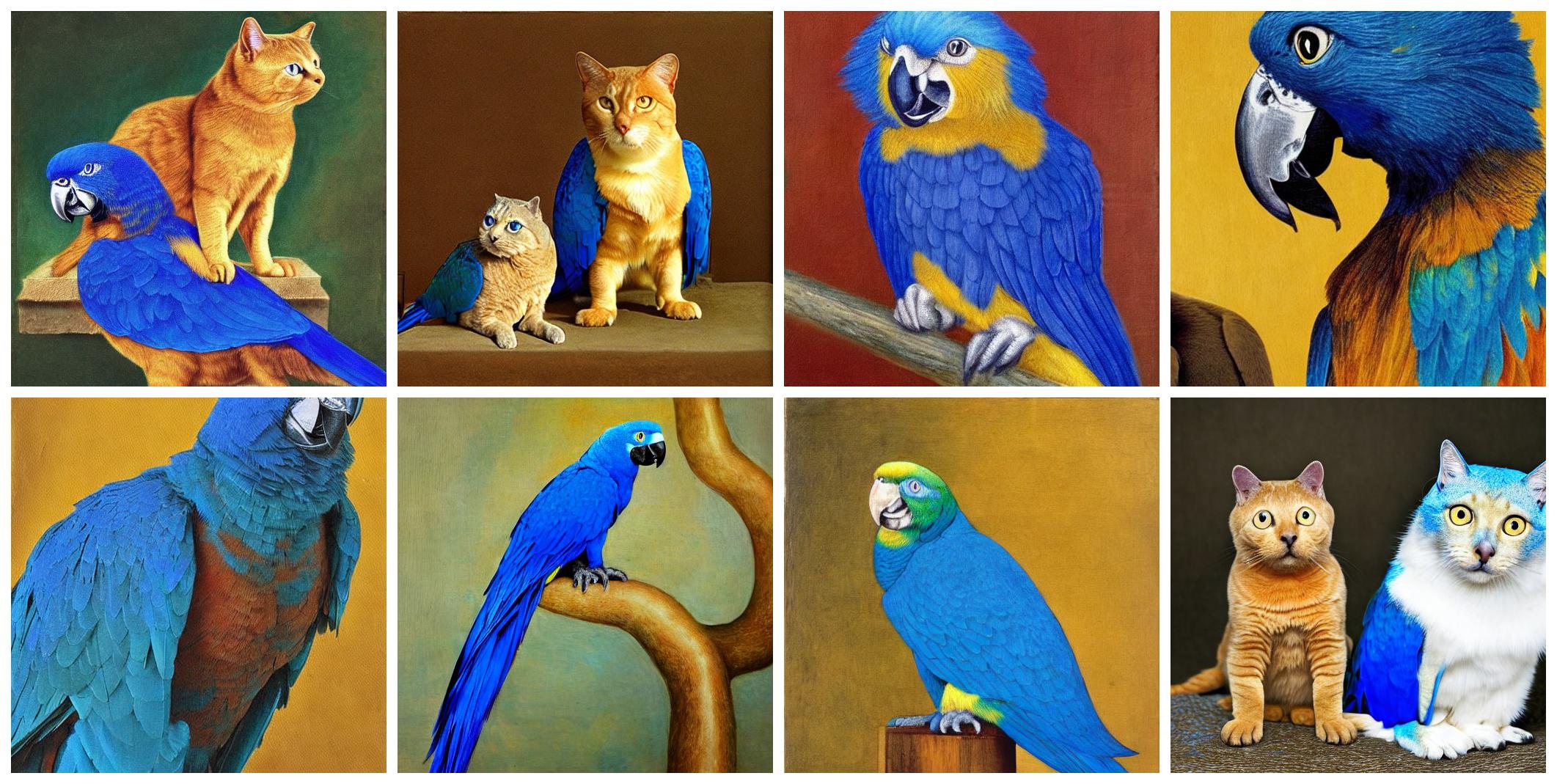} &
        \includegraphics[width=0.42\textwidth]{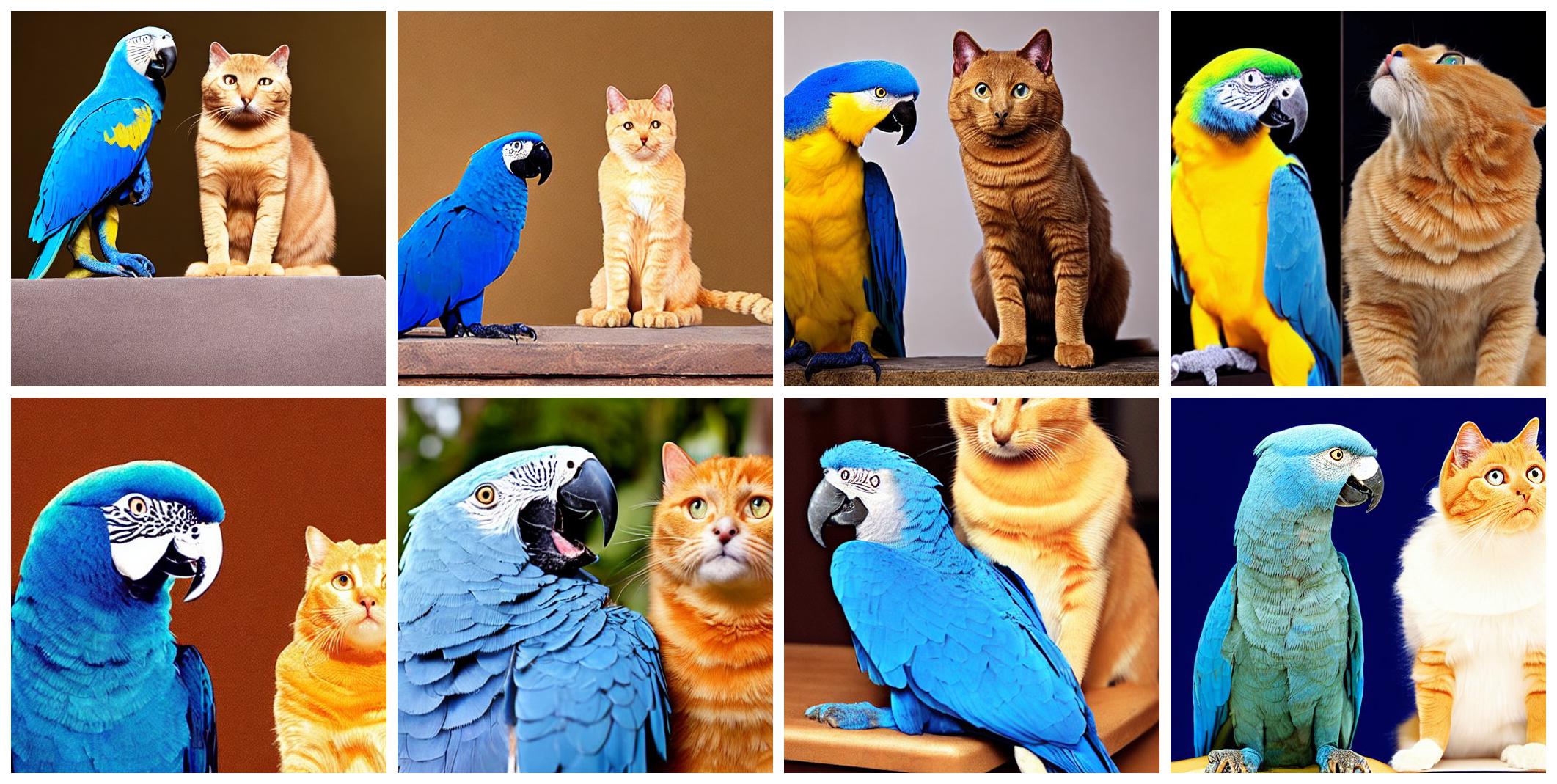} 
        \\\\
        
        \raisebox{0.3in}{\rotatebox{90}{``A \textcolor{blue}{cow} and a \textcolor{blue}{dog}''}}
         &
        \includegraphics[width=0.42\textwidth]{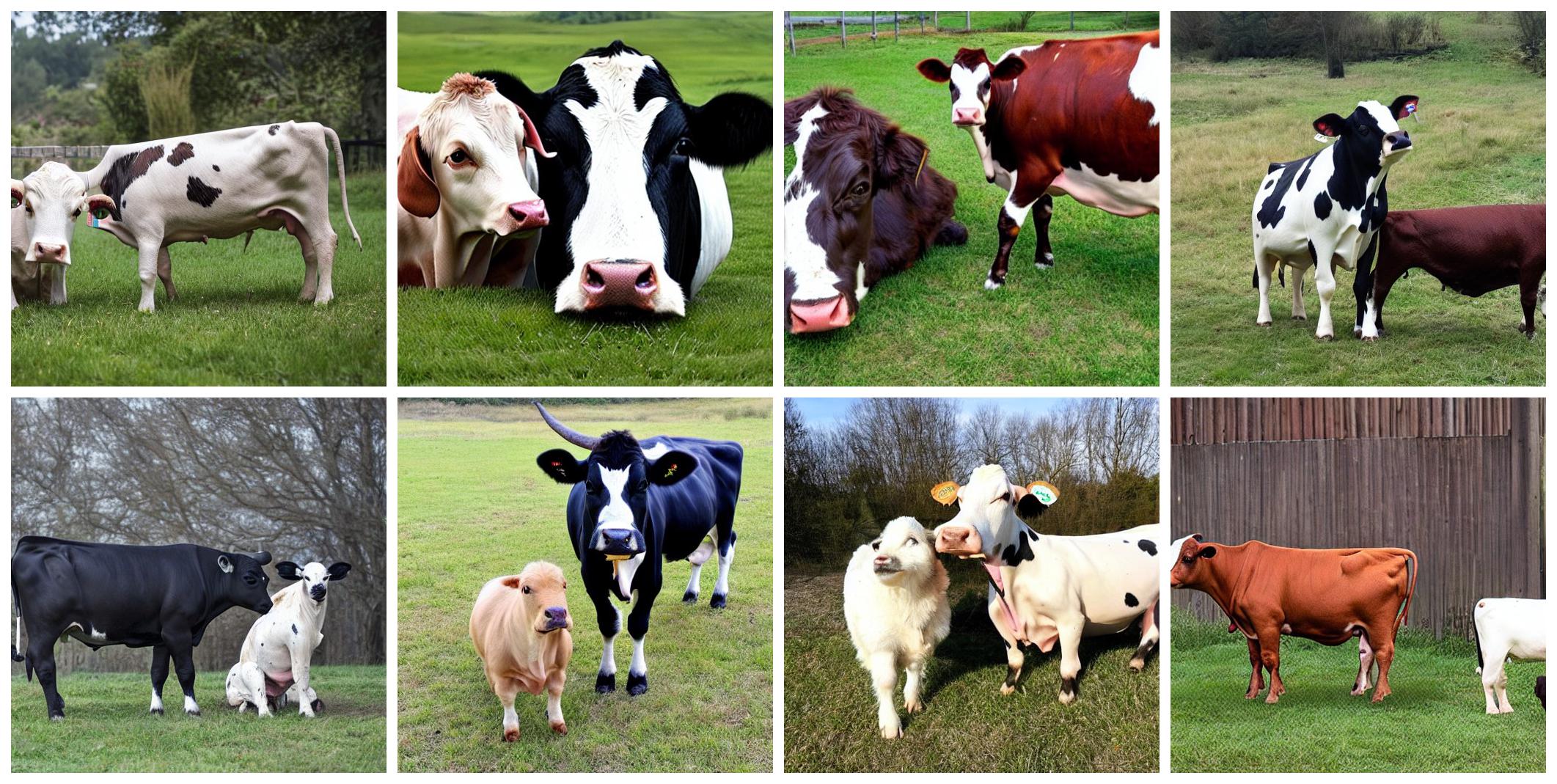} &
        \includegraphics[width=0.42\textwidth]{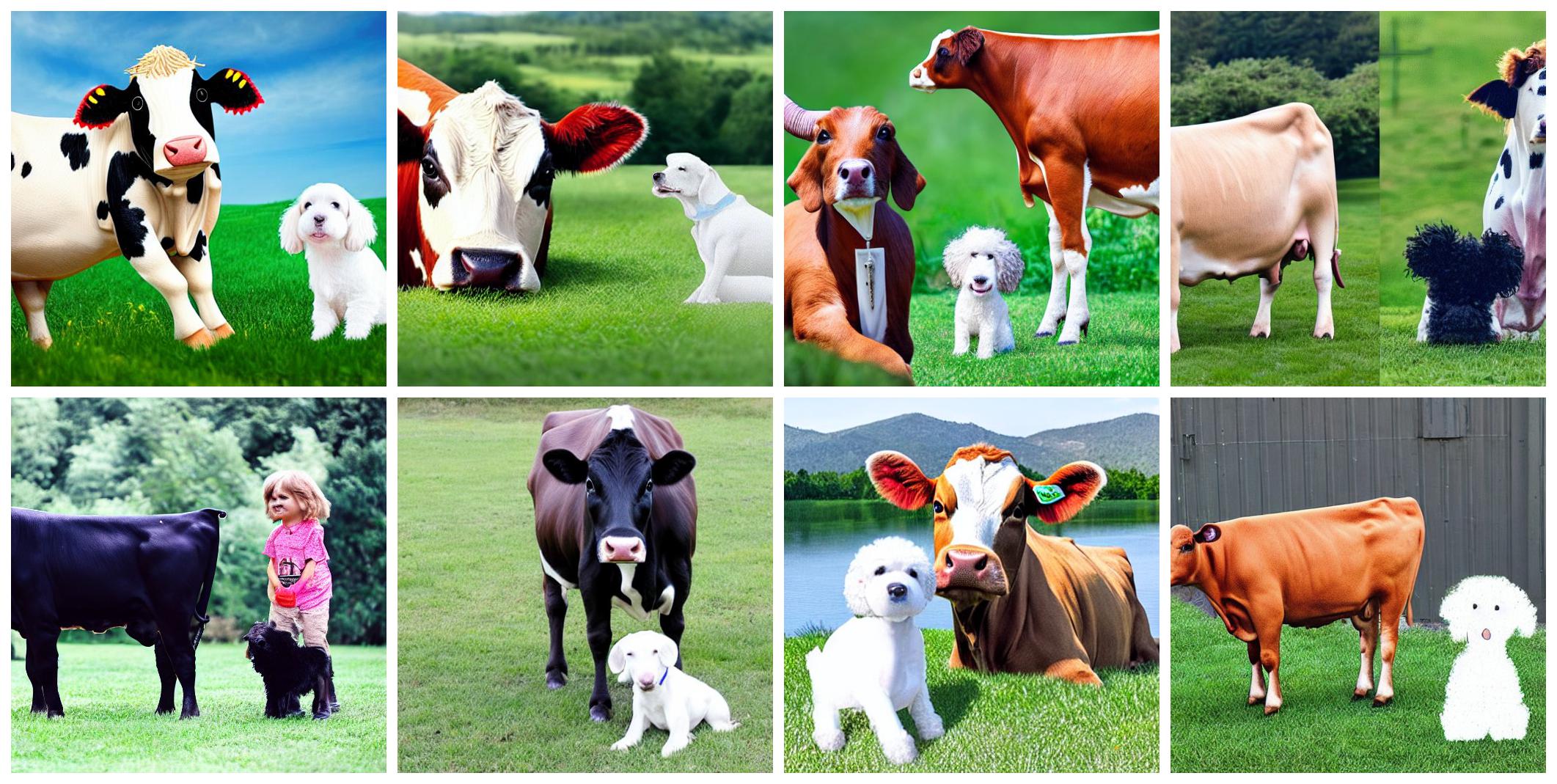} 
        \\\\
        {\raisebox{0.3in}{
        {\rotatebox{90}{\begin{tabular}{c} ``A white \textcolor{blue}{horse}\\ and a green \textcolor{blue}{bird}''
        \end{tabular}}}}}
         &
        \includegraphics[width=0.42\textwidth]{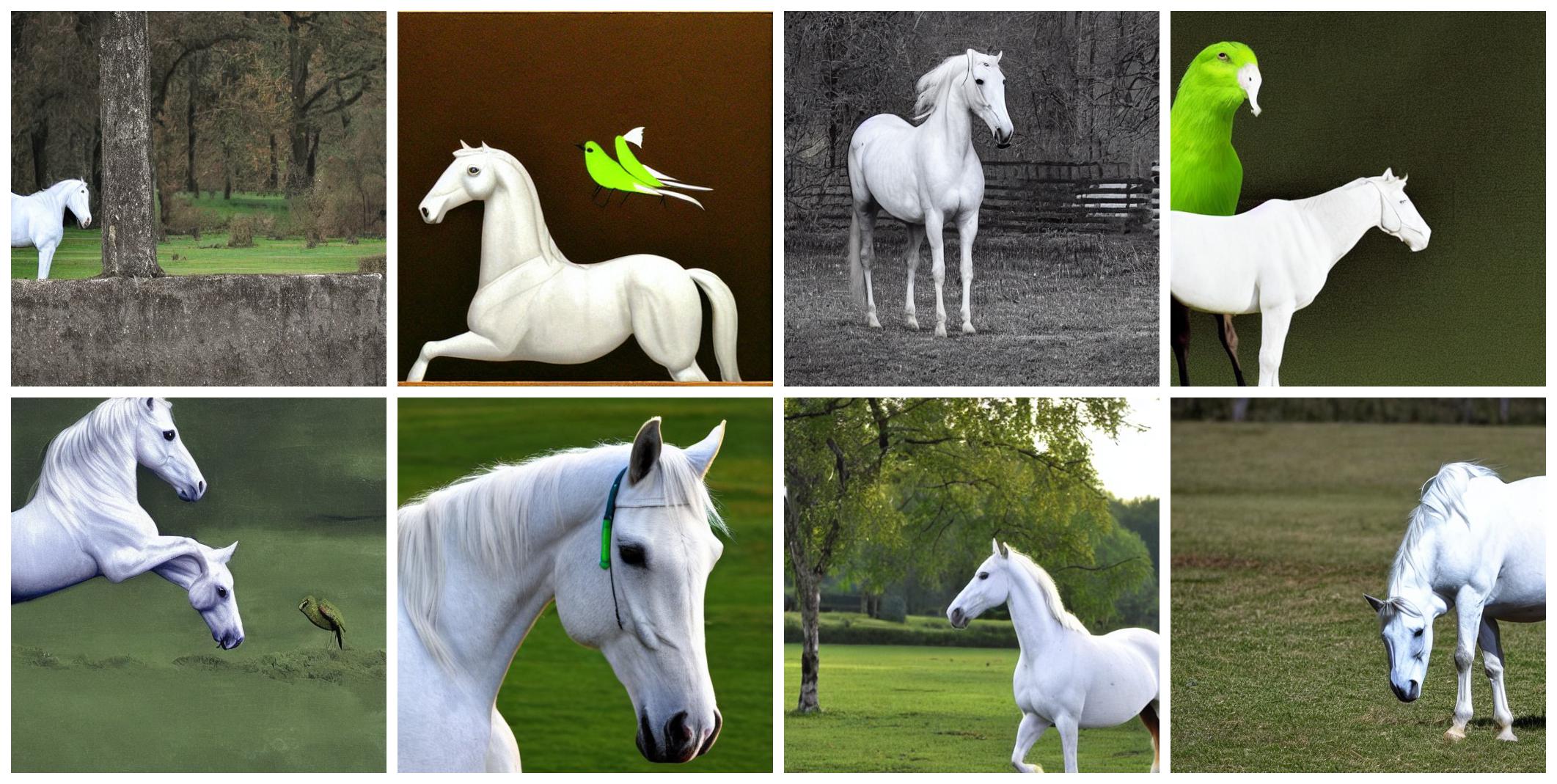} &
        \includegraphics[width=0.42\textwidth]{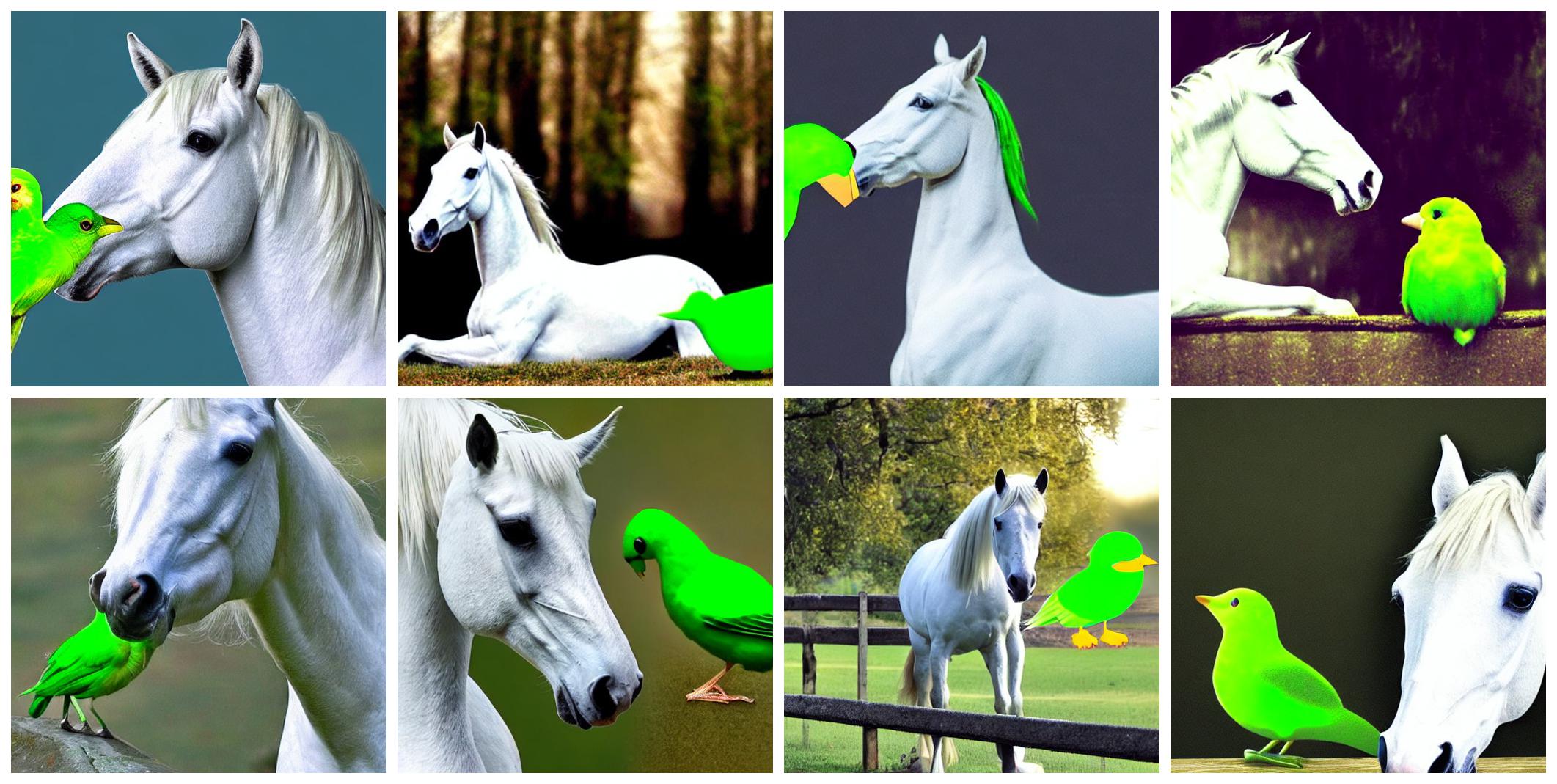} 
        \\\\

        {\raisebox{0.3in}{
        {\rotatebox{90}{\begin{tabular}{c} ``A \textcolor{blue}{turtle} and\\ an orange \textcolor{blue}{balloon}''
        \end{tabular}}}}}
         &
        \includegraphics[width=0.42\textwidth]{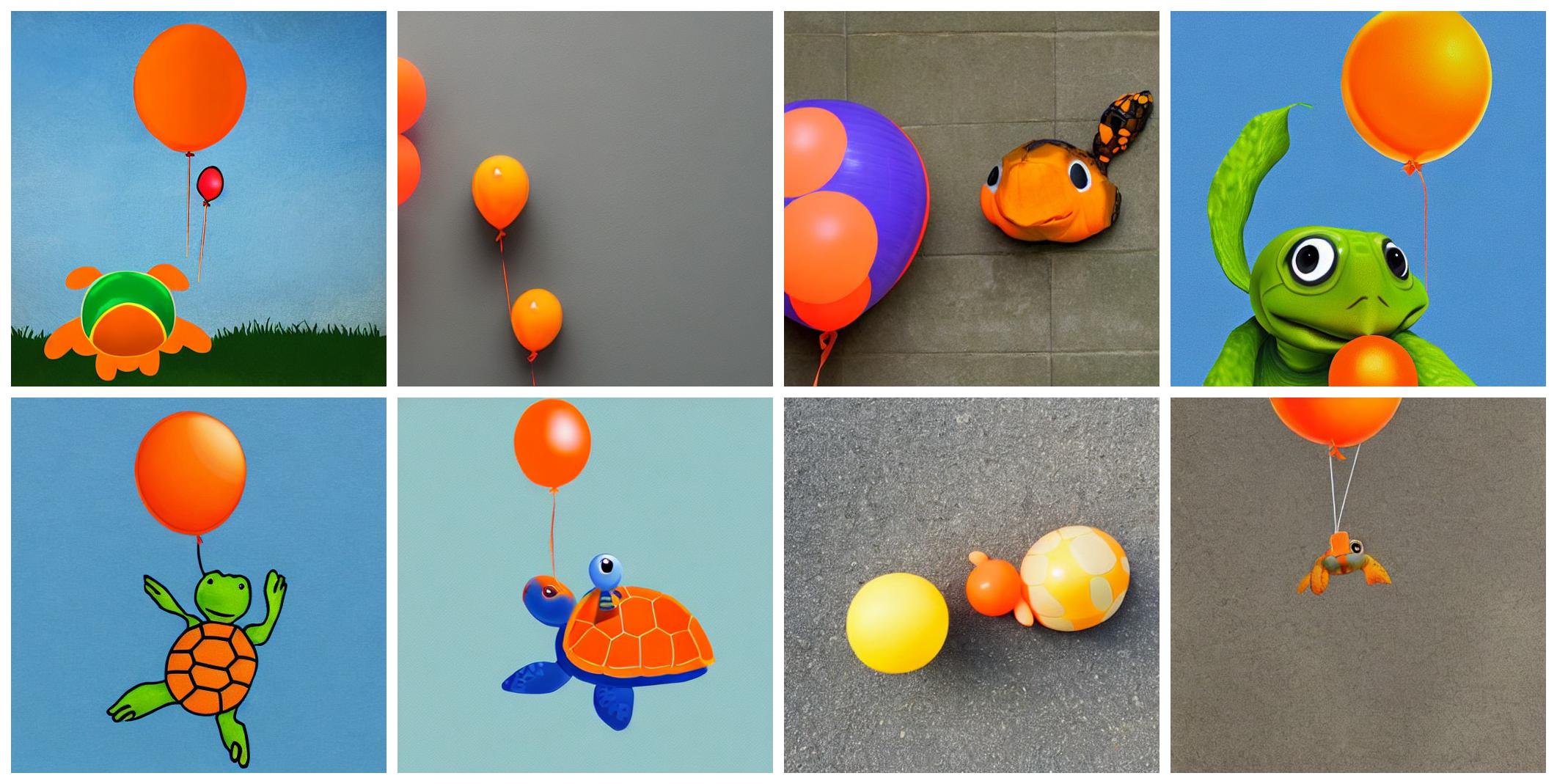} &
        \includegraphics[width=0.42\textwidth]{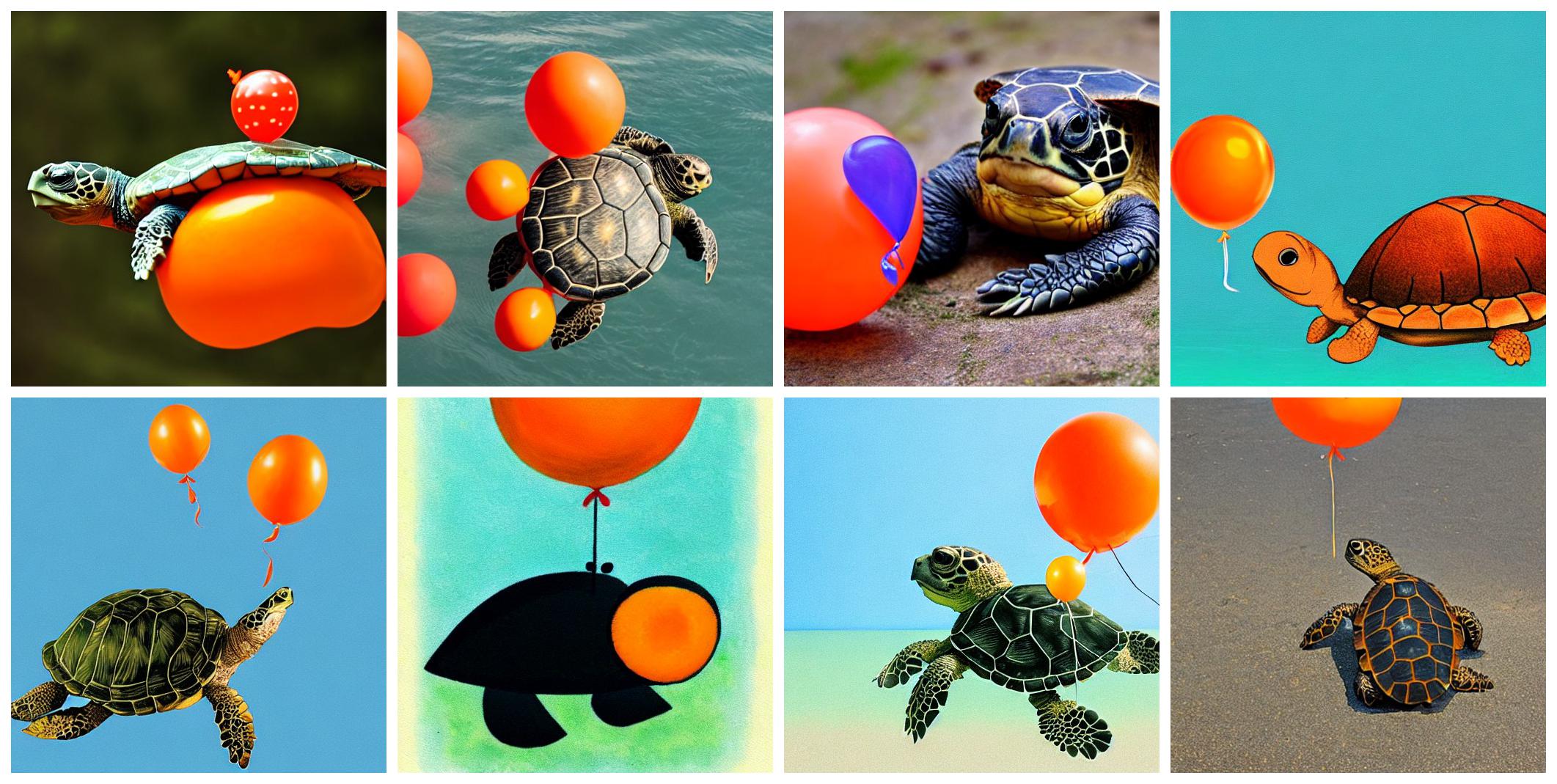} 
        \\\\
        {\raisebox{0.3in}{
        {\rotatebox{90}{\begin{tabular}{c} ``A brown \textcolor{blue}{suitcase}\\ and a white \textcolor{blue}{bench}''
        \end{tabular}}}}}
         &
        \includegraphics[width=0.42\textwidth]{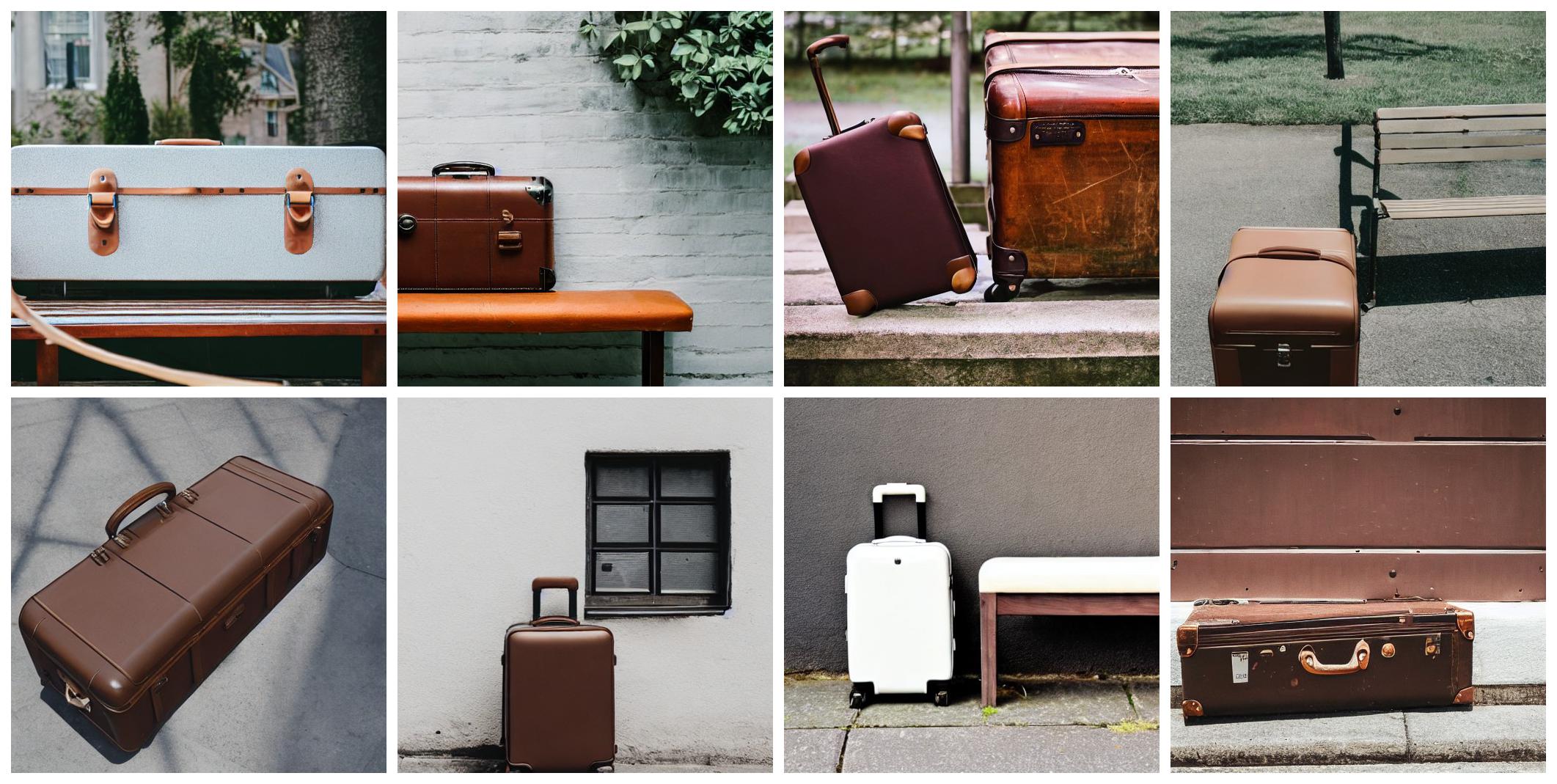} &
        \includegraphics[width=0.42\textwidth]{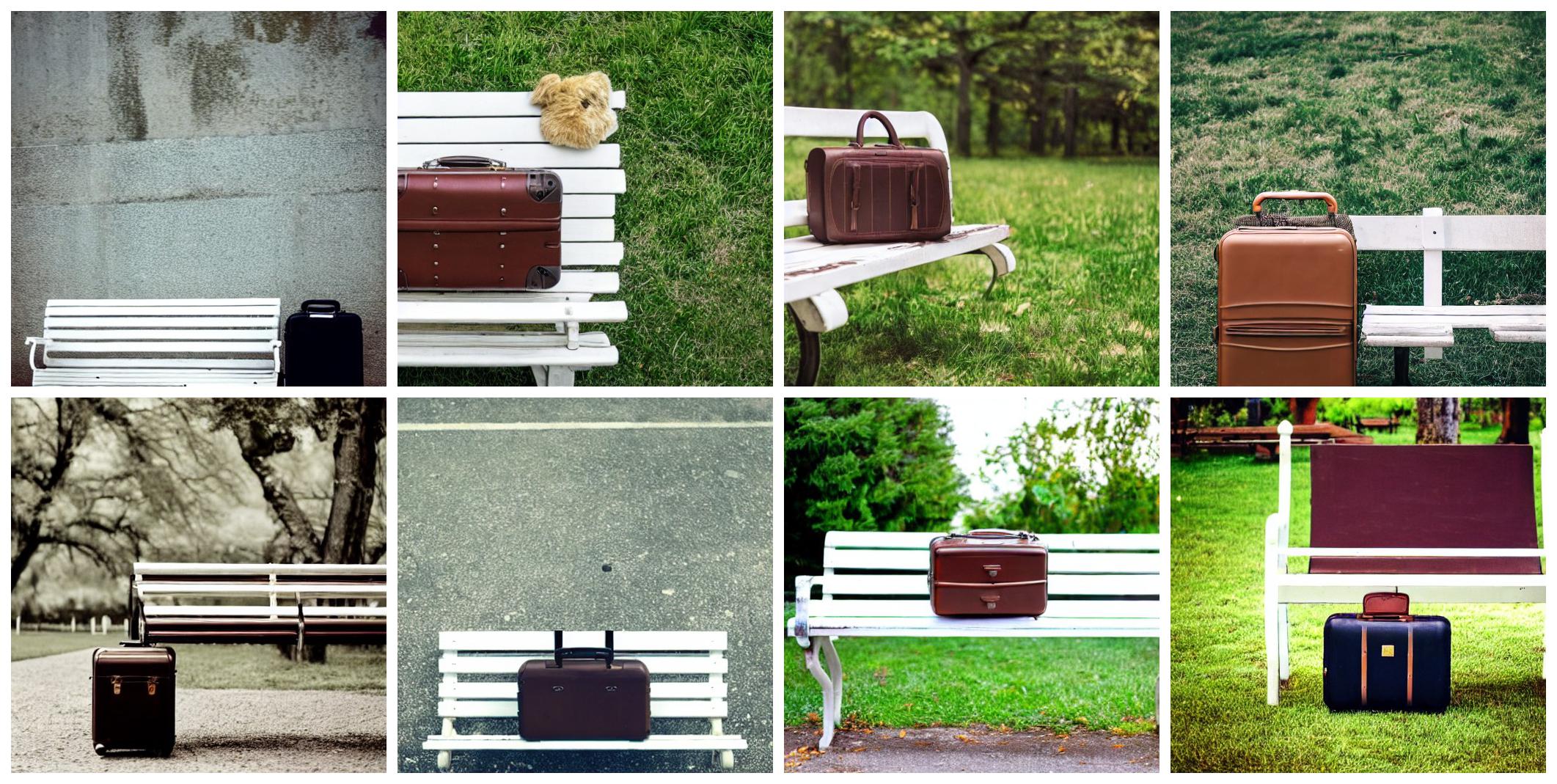} 
    \end{tabular}
    
    }
    \vspace{-0.25cm}
    \caption{Sample uncurated results achieved with Stable Diffusion~\cite{rombach2022high} with and without our Attend-and-Excite approach. For each prompt, we show eight images synthesized when optimizing over the object tokens highlighted in \textcolor{blue}{blue}. When displaying results with and without Attend-and-Excite we use the same set of random seeds without cherry-picking.}
    \label{fig:uncurated2}
\end{figure*}

In the remainder of the Appendix, we provide additional results and comparisons as follows: 
\begin{enumerate}
    \item \Cref{fig:uncurated2,fig:uncurated} present uncurated results using Stable Diffusion before and after applying Attend-and-Excite, where we show results using $8$ seeds without cherry-picking. 
    \item In~\Cref{fig:creative,fig:creative2}, we provide additional results and comparisons to Stable Diffusion on more complex prompts and styles as well as less common combinations of subjects and settings, and complex attributes.
    \item ~\Cref{fig:structured_prompts_supplementary} contains additional results for prompts from the StructureDiffusion paper.
    \item ~\Cref{fig:additional_results_supp,fig:additional_results_supp2}, present additional qualitative comparisons to all baselines.
    \item As mentioned in the results section of the main paper, the CLIP image-text similarity scores for the Composable Diffusion baseline are often high as a result of ``subject mixture'' where the generated images contain a single object that is a hybrid of all subject tokens in the input prompt. \Cref{fig:compositional_extra} presents additional results obtained using Composable Diffusion that demonstrate this phenomenon.
    \item ~\Cref{fig:additional_cls_spes} contains additional comparisons of the cross-attention maps for the subject tokens before and after applying Attend-and-Excite. As can be seen, in accordance with the results presented in the main paper, Attend-and-Excite facilitates the use of attention as an explanation.
\end{enumerate}

\begin{figure*}
    \centering
    \setlength{\tabcolsep}{0.5pt}
    \renewcommand{\arraystretch}{0.3}
    {\small

    \begin{tabular}{c c c @{\hspace{0.1cm}} c c @{\hspace{0.1cm}} c c @{\hspace{0.1cm}} c c }

        &
        \multicolumn{2}{c}{``A blue  \textcolor{blue}{dog} and a pink  \textcolor{blue}{cat}''}  &
        \multicolumn{2}{c}{\begin{tabular}{c}``A painting of a \textcolor{blue}{cat} and \\ a \textcolor{blue}{dog} in the style of Van Gogh'' \end{tabular}} &
        \multicolumn{2}{c}{\begin{tabular}{c} ``A \textcolor{blue}{swan} in the \textcolor{blue}{kitchen}'' \\\\
        \end{tabular}} &
        \multicolumn{2}{c}{``The cloud \textcolor{blue}{gate} sculpture in a  \textcolor{blue}{library}''} \\

        {\raisebox{0.3in}{
        \multirow{2}{*}{\rotatebox{90}{Stable Diffusion}}}} &
        \includegraphics[width=0.11\textwidth]{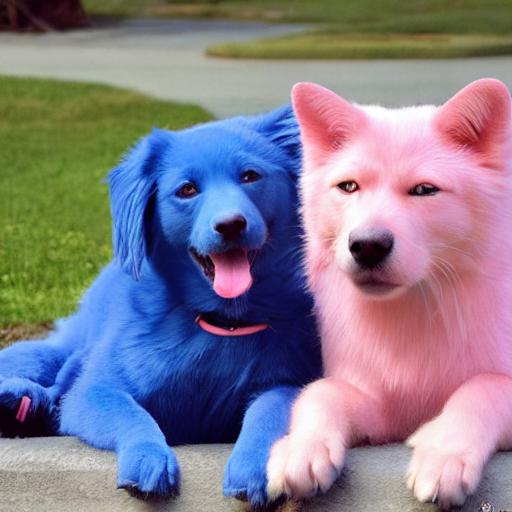} &
        \includegraphics[width=0.11\textwidth]{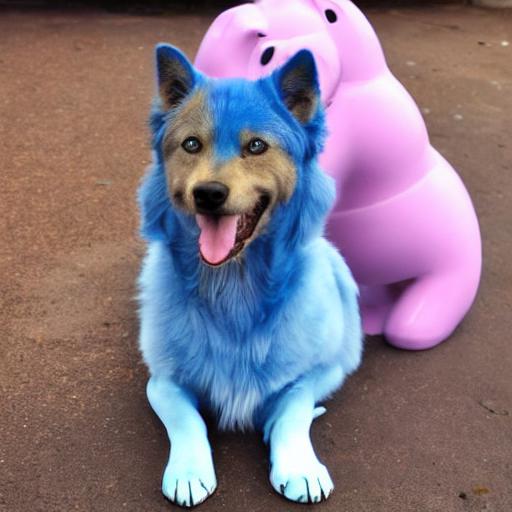} &
        \hspace{0.05cm}
        \includegraphics[width=0.11\textwidth]{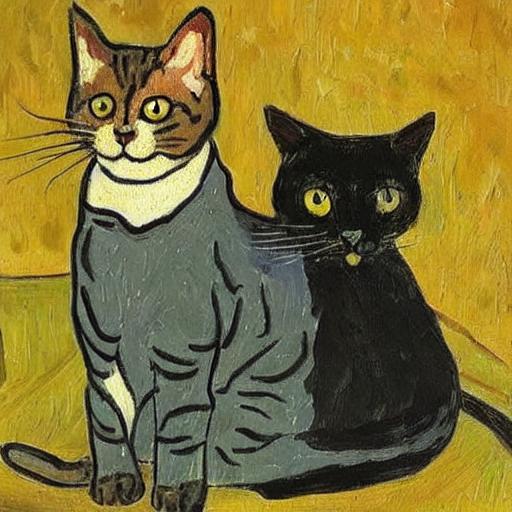} &
        \includegraphics[width=0.11\textwidth]{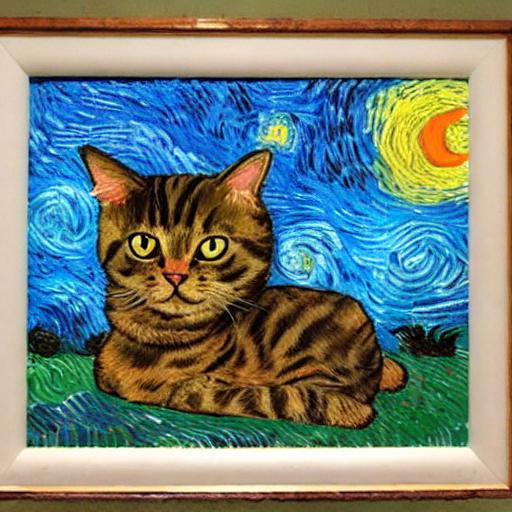} &
        \hspace{0.05cm}
        \includegraphics[width=0.11\textwidth]{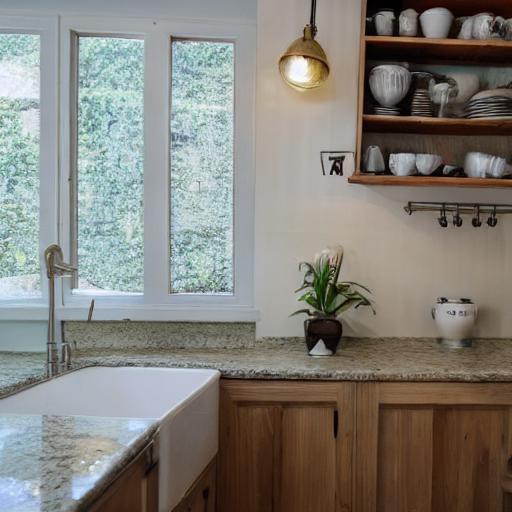} &
        \includegraphics[width=0.11\textwidth]{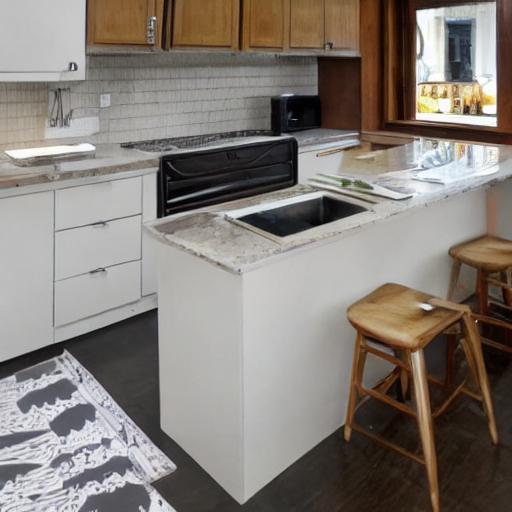} &
        \hspace{0.05cm}
        \includegraphics[width=0.11\textwidth]{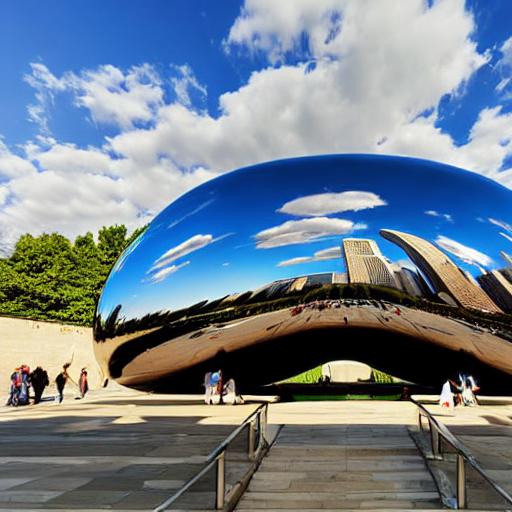} &
        \includegraphics[width=0.11\textwidth]{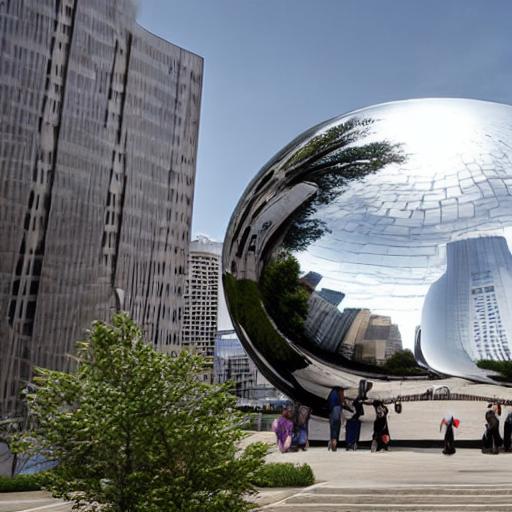} \\

        &
        \includegraphics[width=0.11\textwidth]{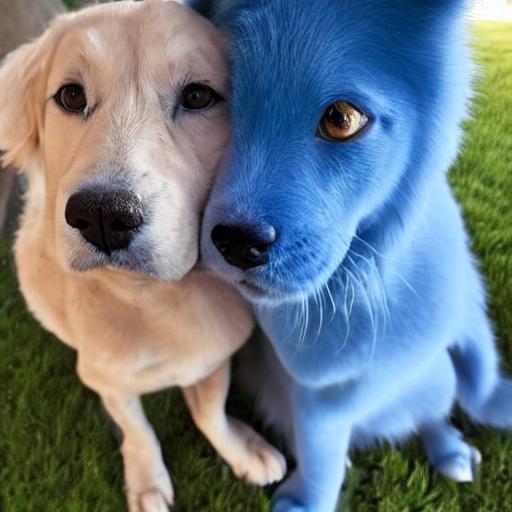} &
        \includegraphics[width=0.11\textwidth]{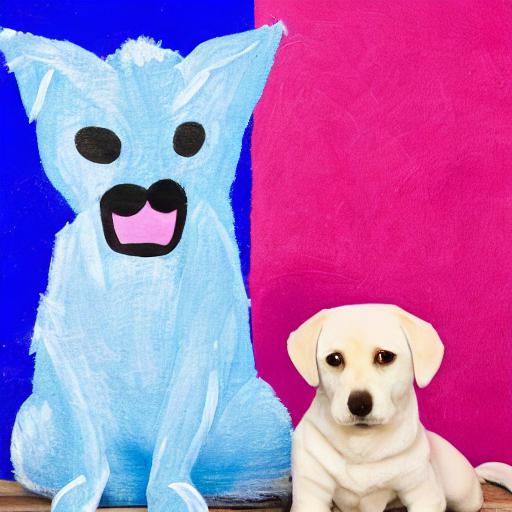} &
        \hspace{0.05cm}
        \includegraphics[width=0.11\textwidth]{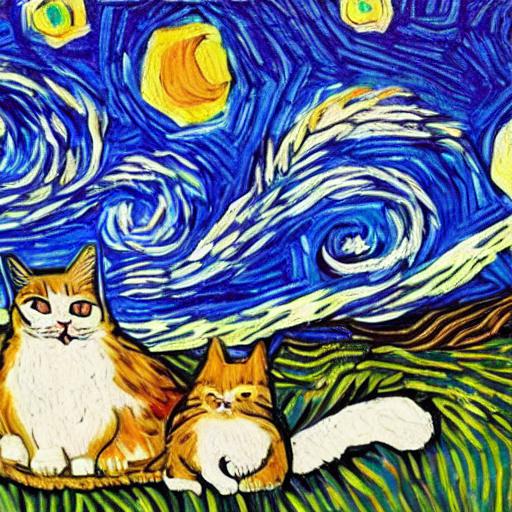} &
        \includegraphics[width=0.11\textwidth]{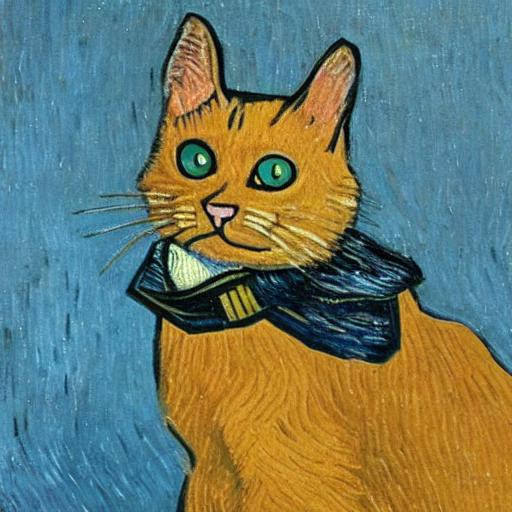} &
        \hspace{0.05cm}
        \includegraphics[width=0.11\textwidth]{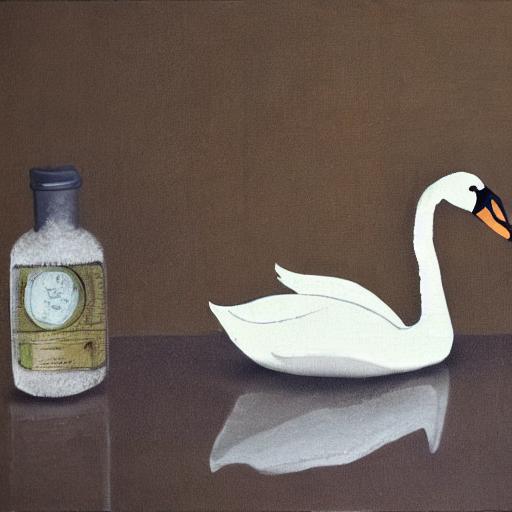} &
        \includegraphics[width=0.11\textwidth]{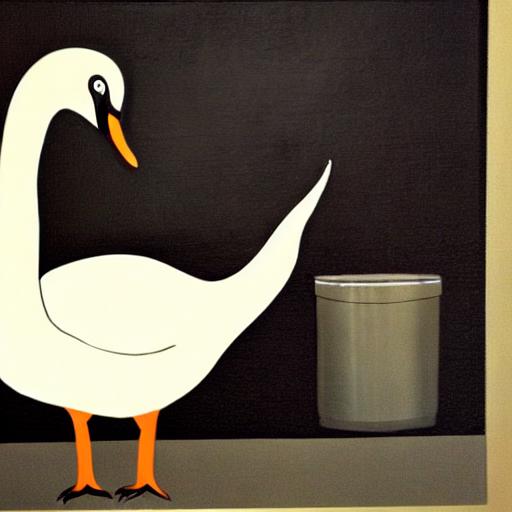} &
        \hspace{0.05cm}
        \includegraphics[width=0.11\textwidth]{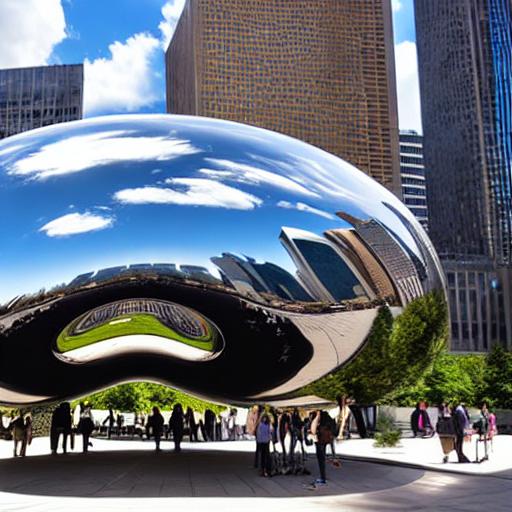} &
        \includegraphics[width=0.11\textwidth]{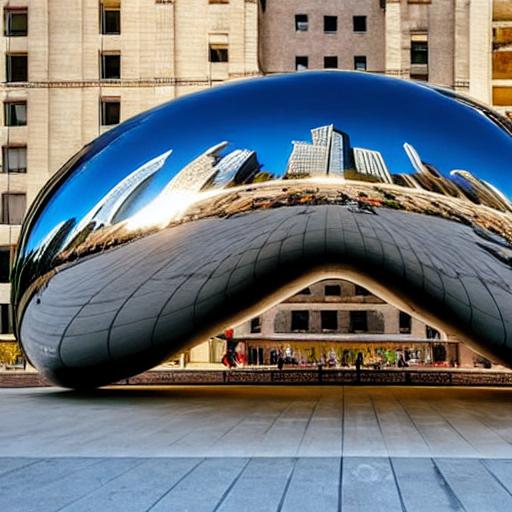} \\ \\ \\

        {\raisebox{0.425in}{
        \multirow{2}{*}{\rotatebox{90}{\begin{tabular}{c} Stable Diffusion with \\ \textcolor{blue}{Attend-and-Excite} \\ \\ \end{tabular}}}}} &
        \includegraphics[width=0.11\textwidth]{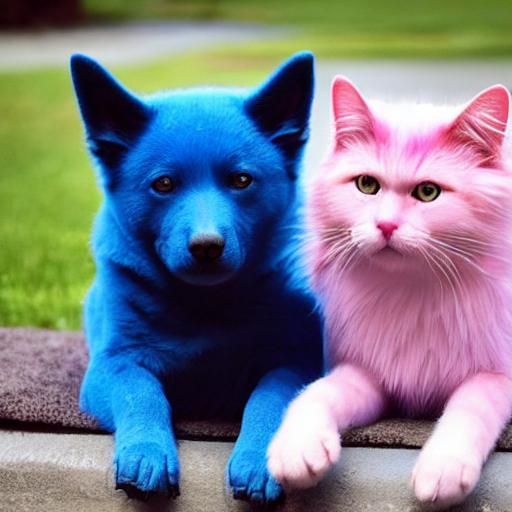} &
        \includegraphics[width=0.11\textwidth]{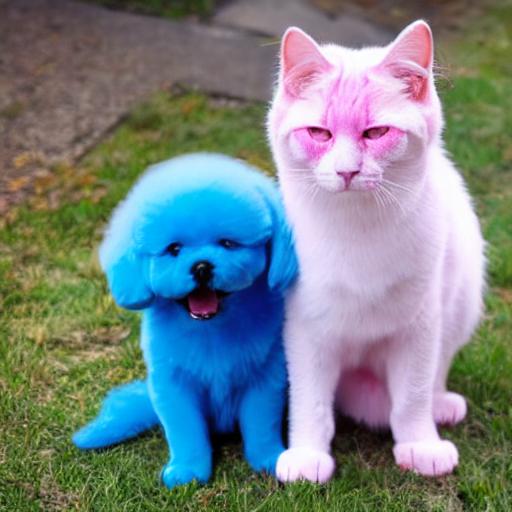} &
        \hspace{0.05cm}
        \includegraphics[width=0.11\textwidth]{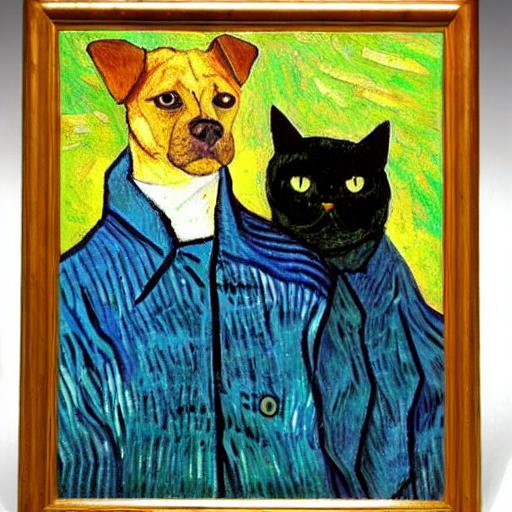} &
        \includegraphics[width=0.11\textwidth]{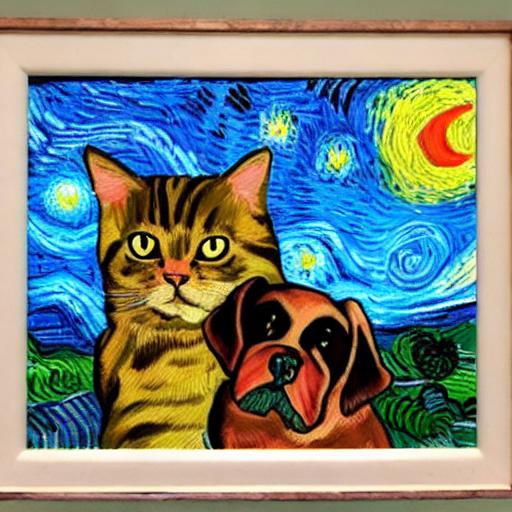} &
        \hspace{0.05cm}
        \includegraphics[width=0.11\textwidth]{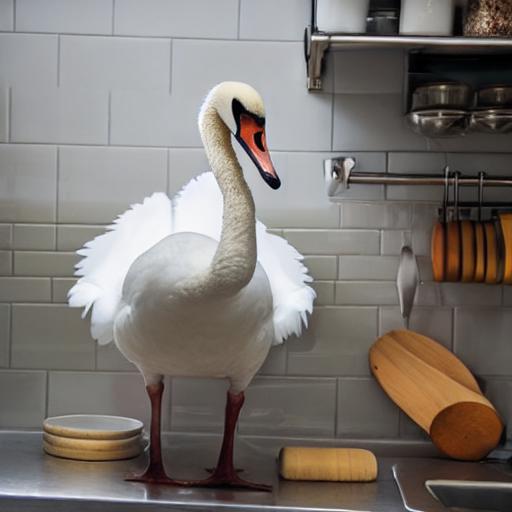} &
        \includegraphics[width=0.11\textwidth]{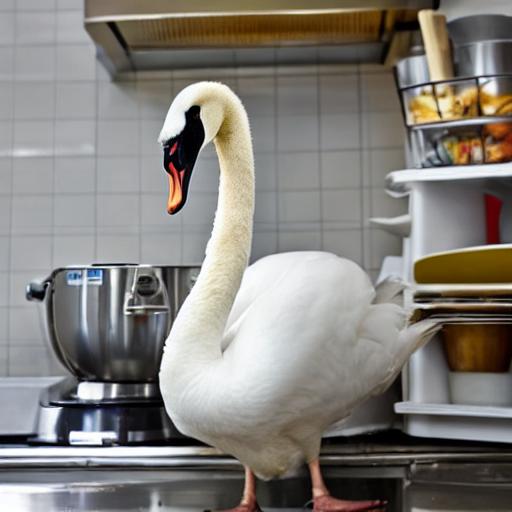} &
        \hspace{0.05cm}
        \includegraphics[width=0.11\textwidth]{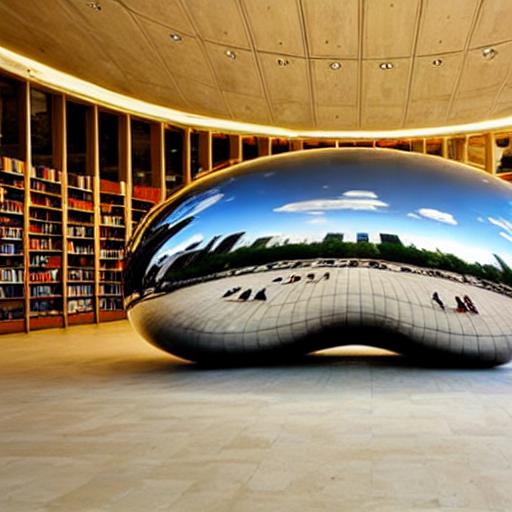} &
        \includegraphics[width=0.11\textwidth]{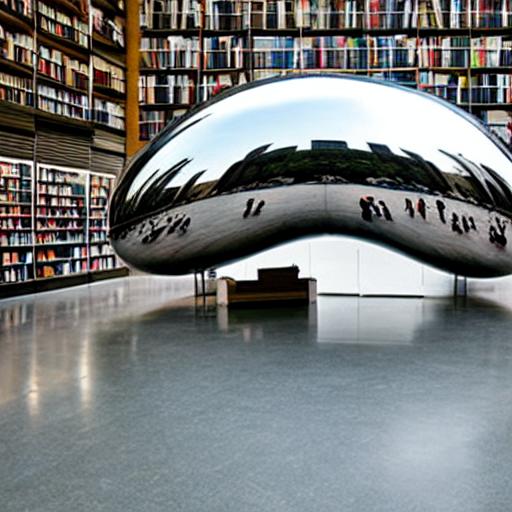} \\

        &
        \includegraphics[width=0.11\textwidth]{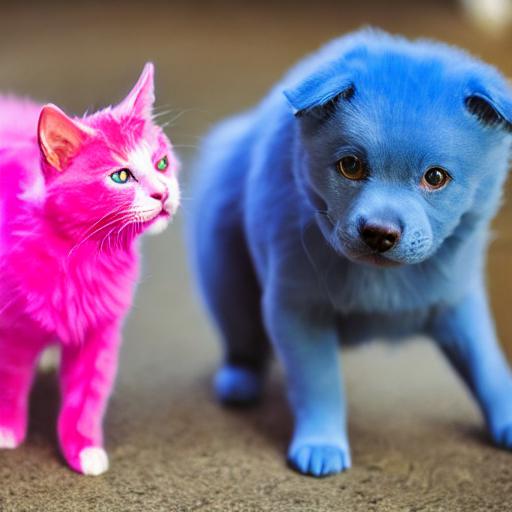} &
        \includegraphics[width=0.11\textwidth]{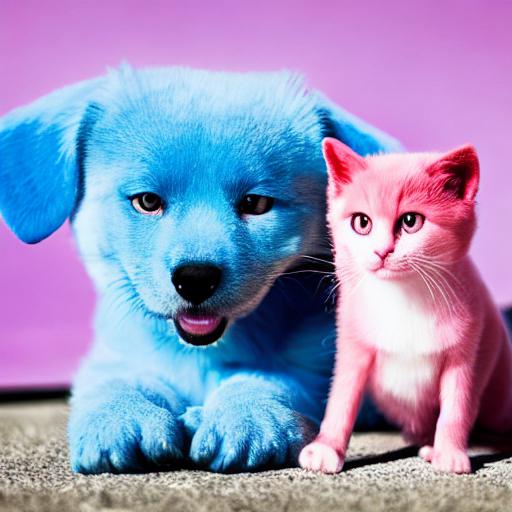} &
        \hspace{0.05cm}
        \includegraphics[width=0.11\textwidth]{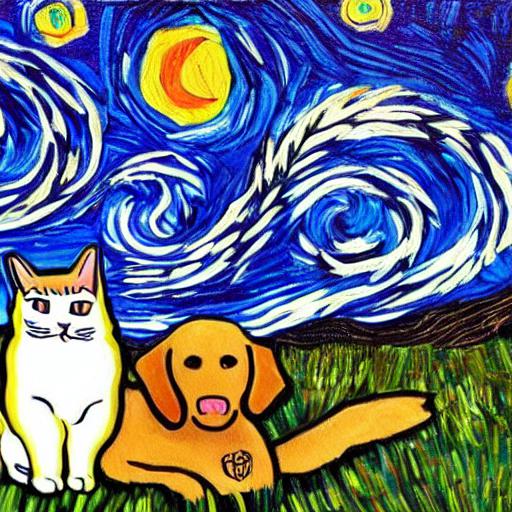} &
        \includegraphics[width=0.11\textwidth]{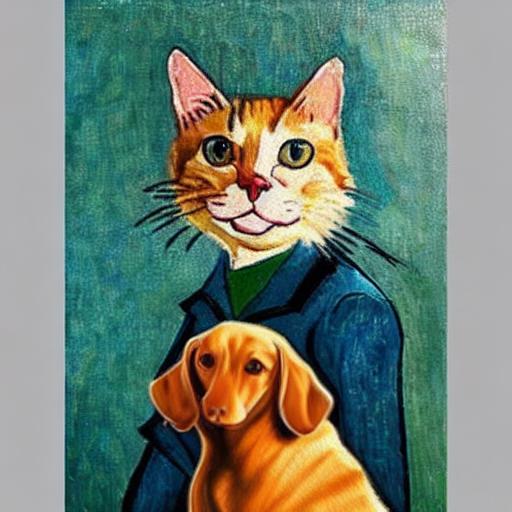} &
        \hspace{0.05cm}
        \includegraphics[width=0.11\textwidth]{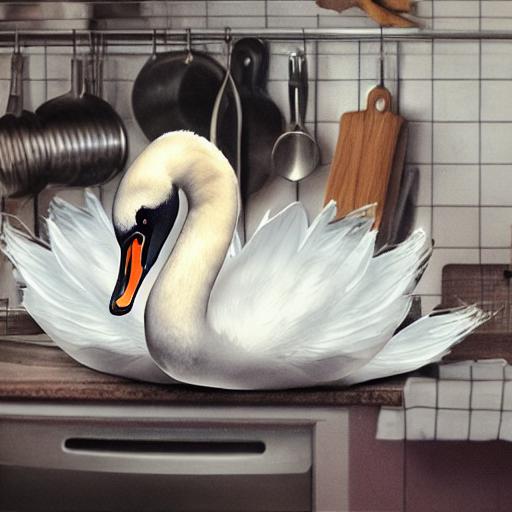} &
        \includegraphics[width=0.11\textwidth]{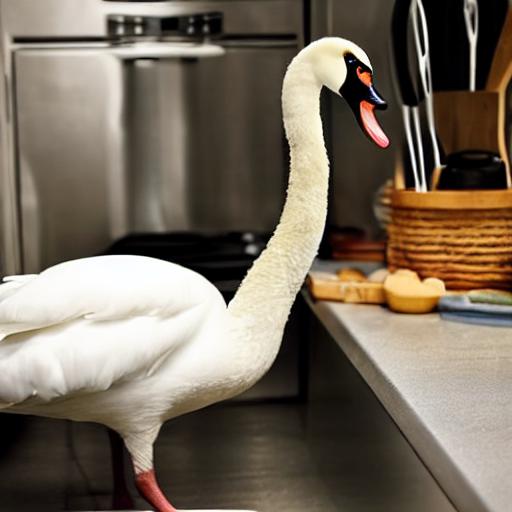} &
        \hspace{0.05cm}
        \includegraphics[width=0.11\textwidth]{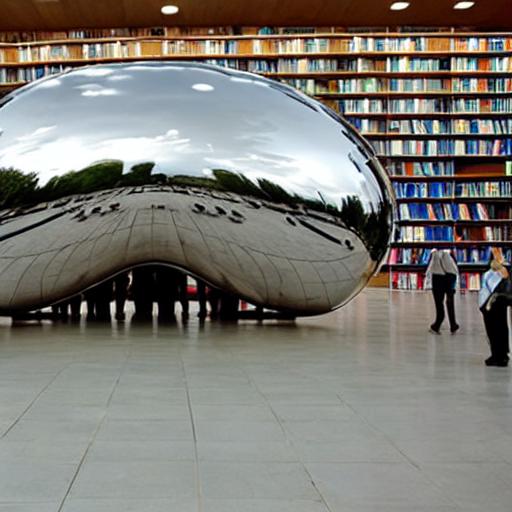} &
        \includegraphics[width=0.11\textwidth]{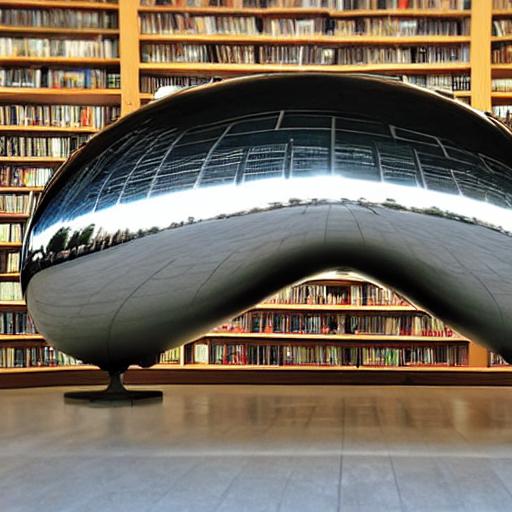} \\ \\ \\

    \end{tabular}

    \begin{tabular}{c c c @{\hspace{0.1cm}} c c @{\hspace{0.1cm}} c c @{\hspace{0.1cm}} c c }

        \\ \\

        &
        \multicolumn{2}{c}{\begin{tabular}{c} ``A green \textcolor{blue}{apple} and a red \textcolor{blue}{apple}''\end{tabular}} &
        \multicolumn{2}{c}{\begin{tabular}{c}``A \textcolor{blue}{bird} perched on \\ top of a vintage \textcolor{blue}{camera}, \\ surrounded by beautiful \textcolor{blue}{flowers}'' \end{tabular}} &
        \multicolumn{2}{c}{\begin{tabular}{c} ``An animation cartoon \\ of a \textcolor{blue}{lion} with a \textcolor{blue}{crown}'' \\\\
        \end{tabular}} &
        \multicolumn{2}{c}{``A blue  \textcolor{blue}{rabbit} and an orange  \textcolor{blue}{dog}''} \\

        {\raisebox{0.3in}{
        \multirow{2}{*}{\rotatebox{90}{Stable Diffusion}}}} &
        \includegraphics[width=0.11\textwidth]{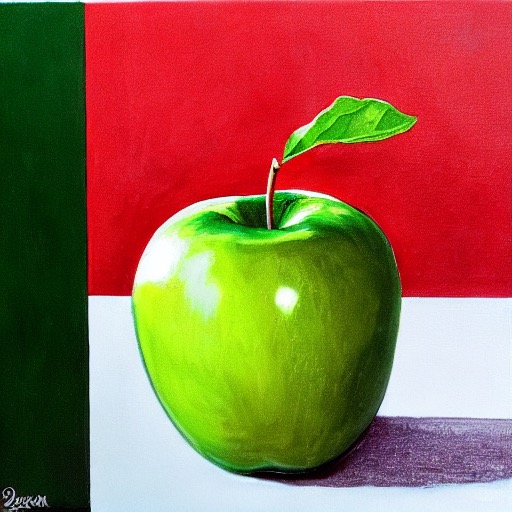} &
        \includegraphics[width=0.11\textwidth]{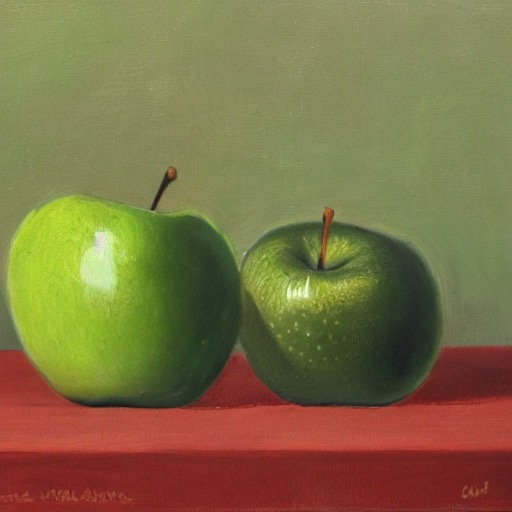} &
        \hspace{0.05cm}
        \includegraphics[width=0.11\textwidth]{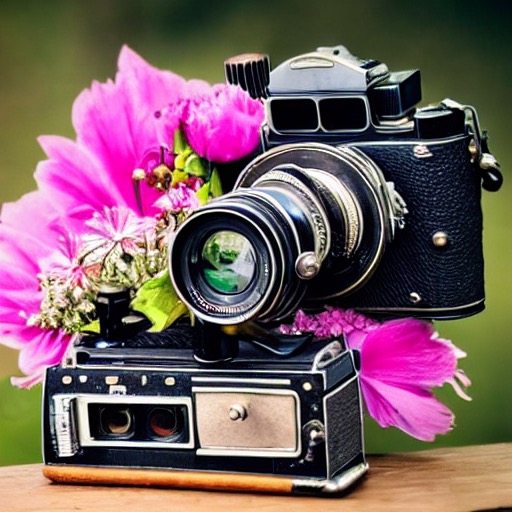} &
        \includegraphics[width=0.11\textwidth]{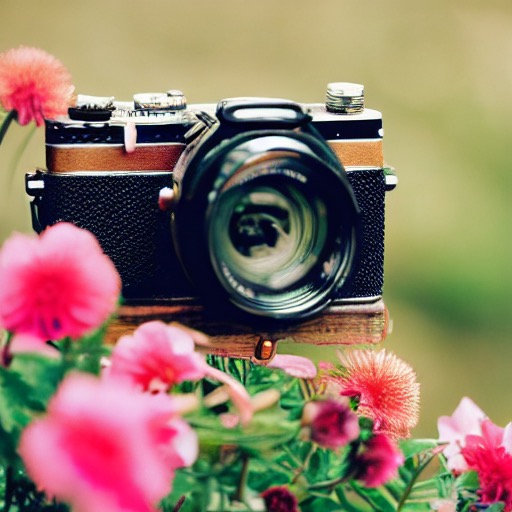} &
        \hspace{0.05cm}
        \includegraphics[width=0.11\textwidth]{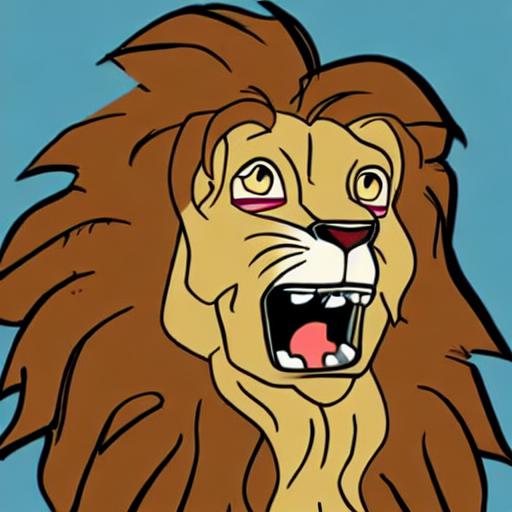} &
        \includegraphics[width=0.11\textwidth]{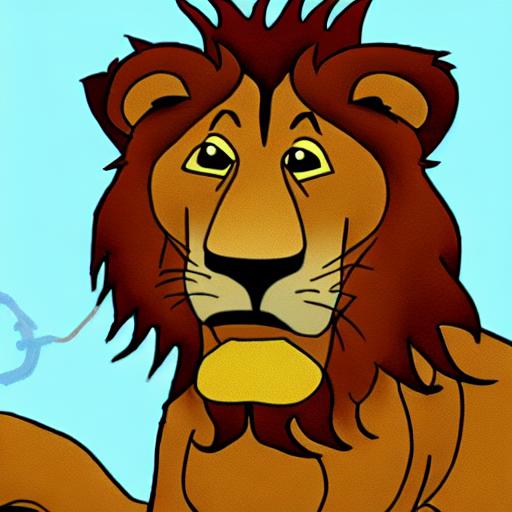} &
        \hspace{0.05cm}
        \includegraphics[width=0.11\textwidth]{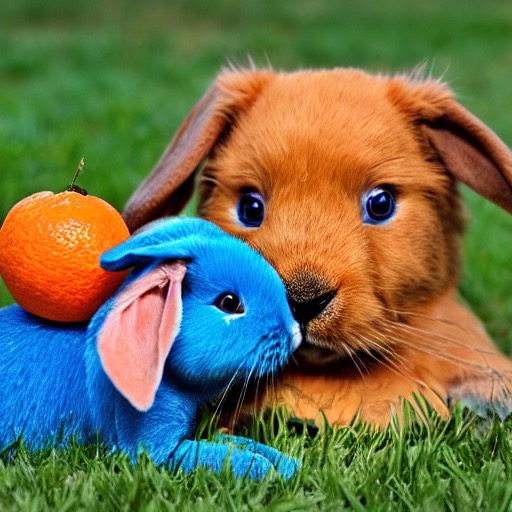} &
        \includegraphics[width=0.11\textwidth]{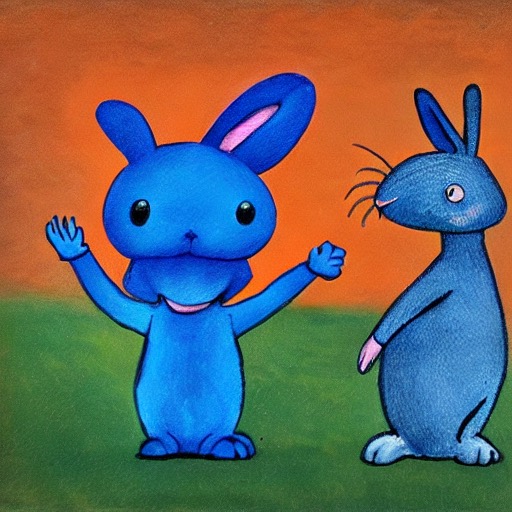} \\

        &
        \includegraphics[width=0.11\textwidth]{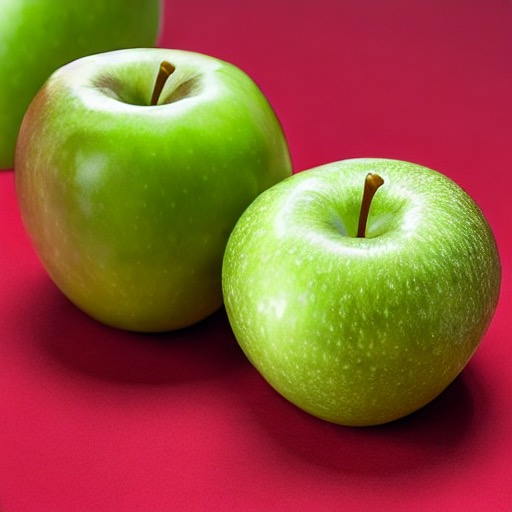} &
        \includegraphics[width=0.11\textwidth]{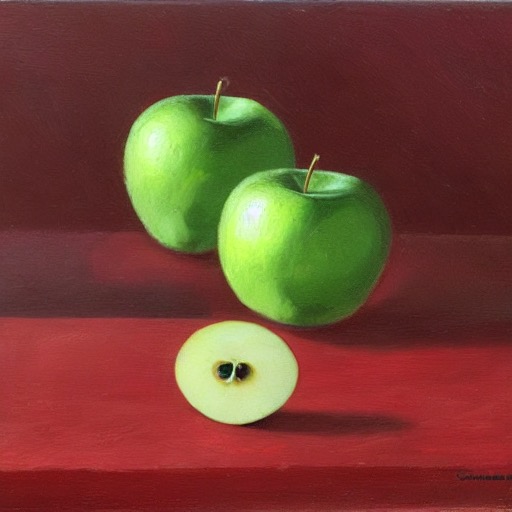} &
        \hspace{0.05cm}
        \includegraphics[width=0.11\textwidth]{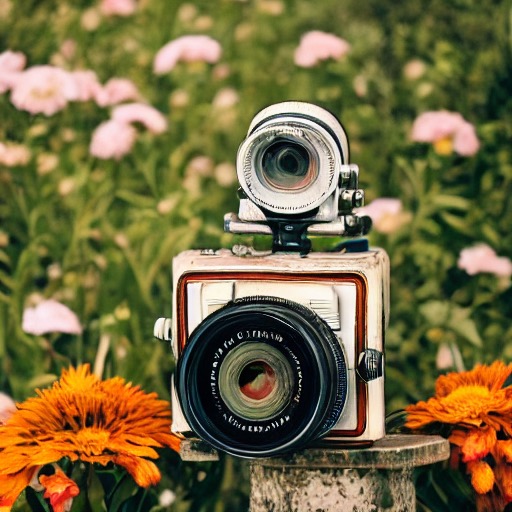} &
        \includegraphics[width=0.11\textwidth]{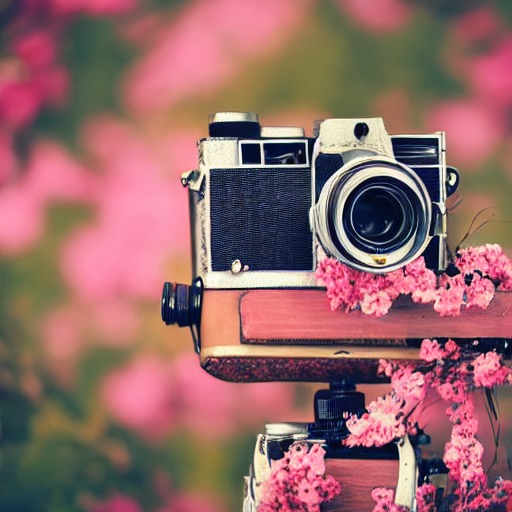} &
        \hspace{0.05cm}
        \includegraphics[width=0.11\textwidth]{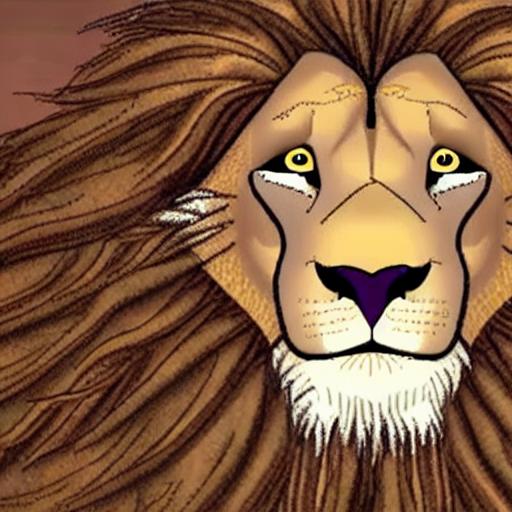} &
        \includegraphics[width=0.11\textwidth]{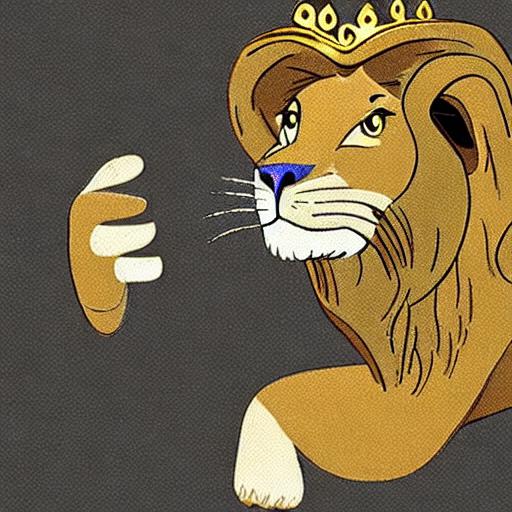} &
        \hspace{0.05cm}
        \includegraphics[width=0.11\textwidth]{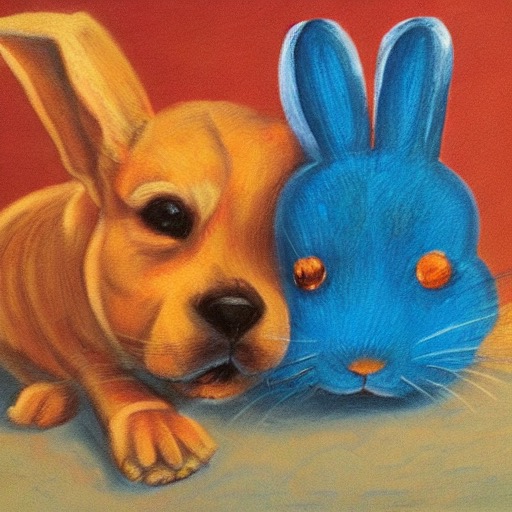} &
        \includegraphics[width=0.11\textwidth]{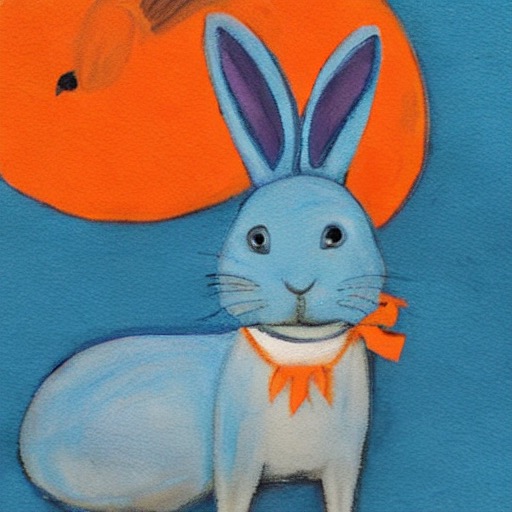} \\ \\ \\

        {\raisebox{0.425in}{
        \multirow{2}{*}{\rotatebox{90}{\begin{tabular}{c} Stable Diffusion with \\ \textcolor{blue}{Attend-and-Excite} \\ \\ \end{tabular}}}}} &
        \includegraphics[width=0.11\textwidth]{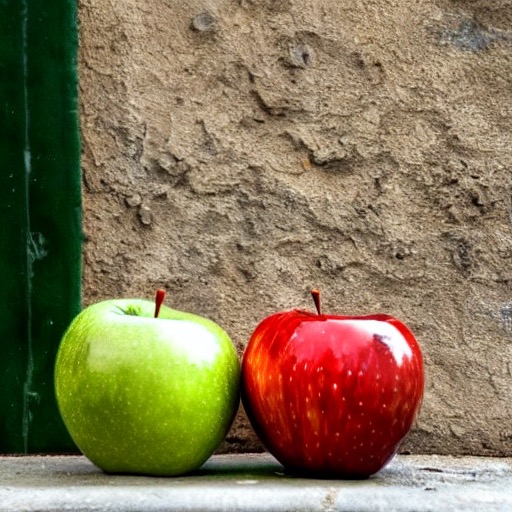} &
        \includegraphics[width=0.11\textwidth]{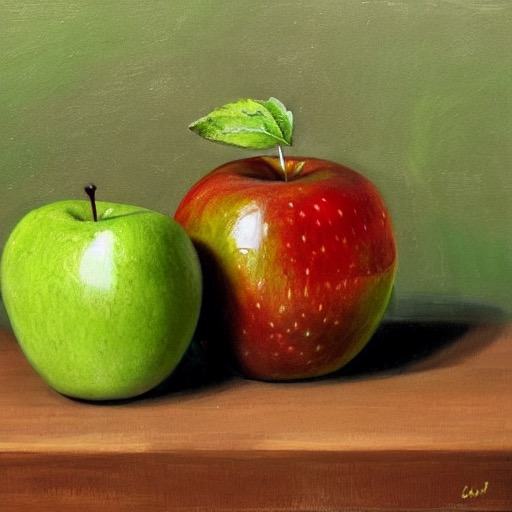} &
        \hspace{0.05cm}
        \includegraphics[width=0.11\textwidth]{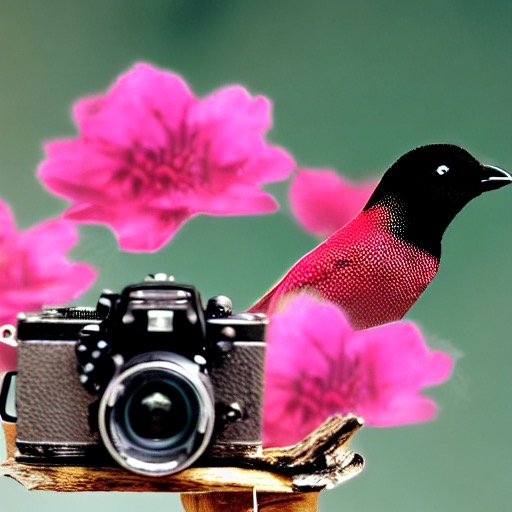} &
        \includegraphics[width=0.11\textwidth]{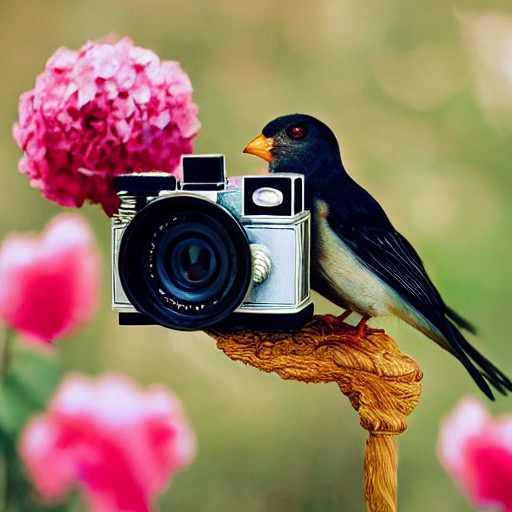} &
        \hspace{0.05cm}
        \includegraphics[width=0.11\textwidth]{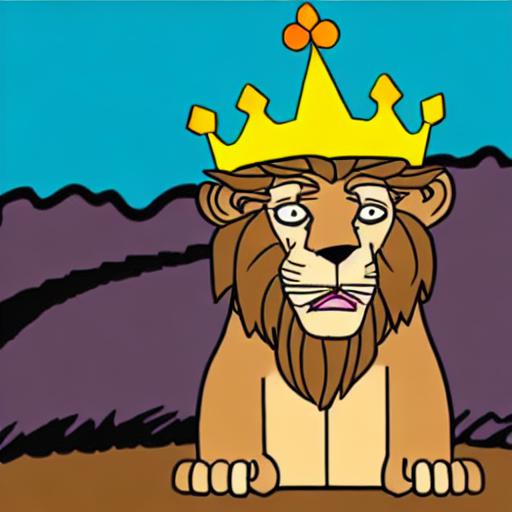} &
        \includegraphics[width=0.11\textwidth]{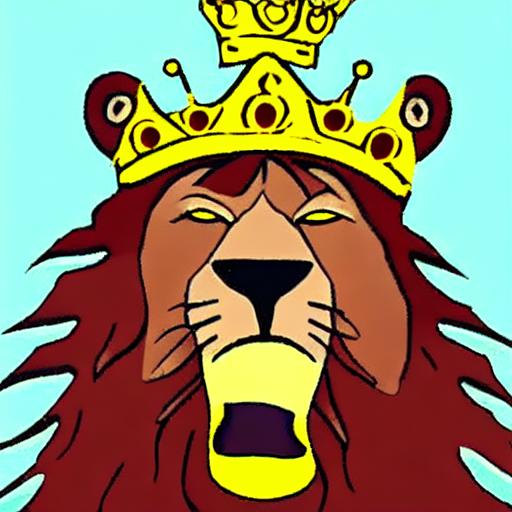} &
        \hspace{0.05cm}
        \includegraphics[width=0.11\textwidth]{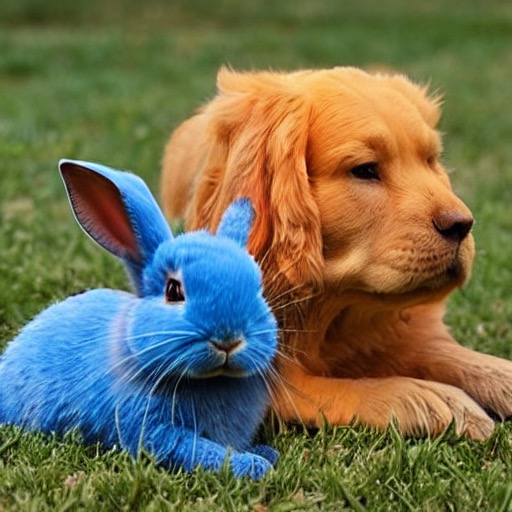} &
        \includegraphics[width=0.11\textwidth]{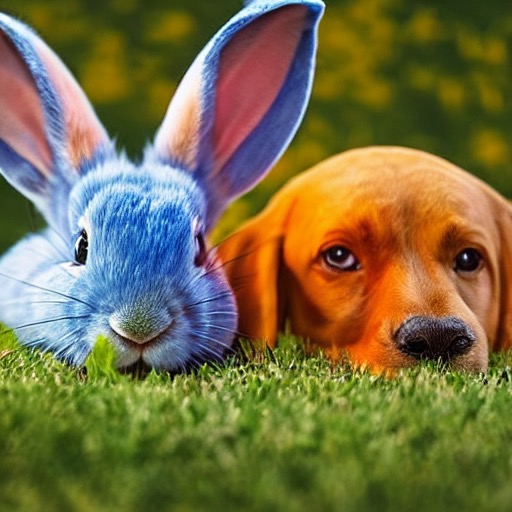} \\

        &
        \includegraphics[width=0.11\textwidth]{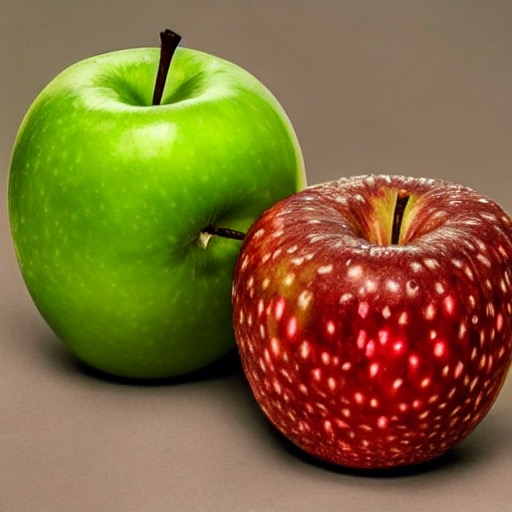} &
        \includegraphics[width=0.11\textwidth]{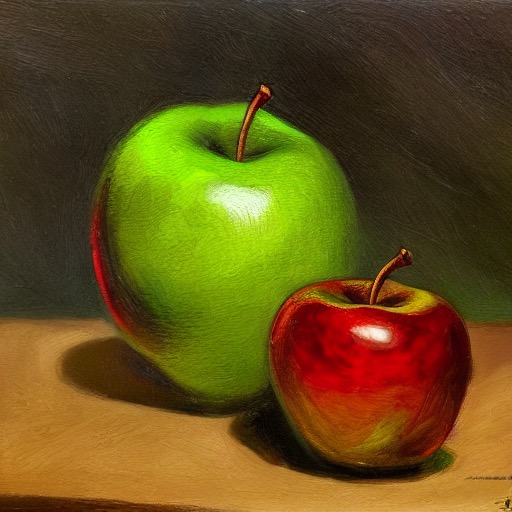} &
        \hspace{0.05cm}
        \includegraphics[width=0.11\textwidth]{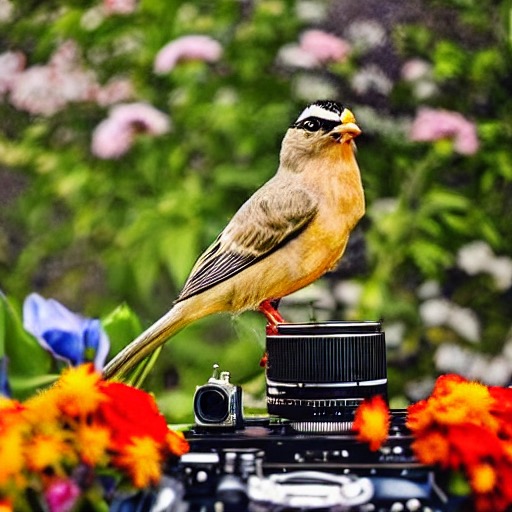} &
        \includegraphics[width=0.11\textwidth]{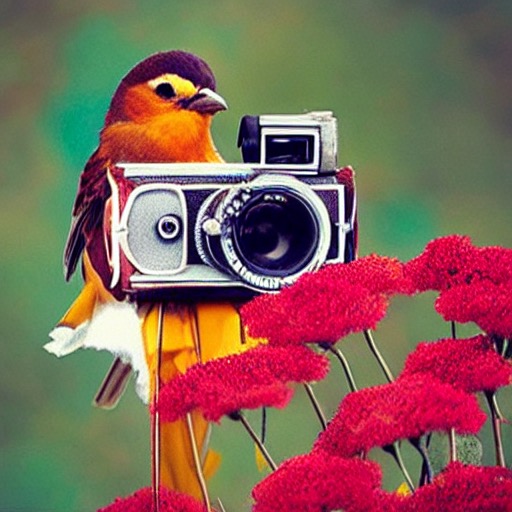} &
        \hspace{0.05cm}
        \includegraphics[width=0.11\textwidth]{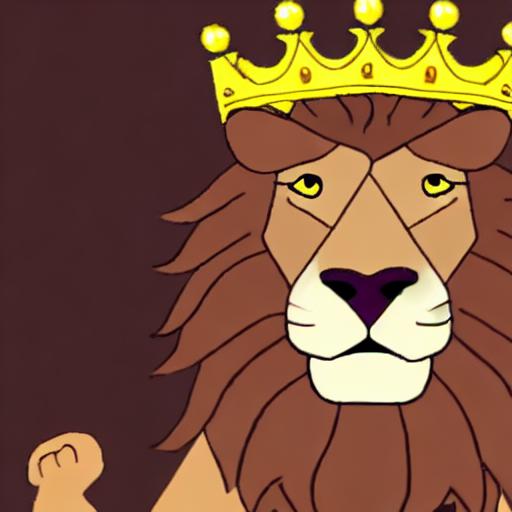} &
        \includegraphics[width=0.11\textwidth]{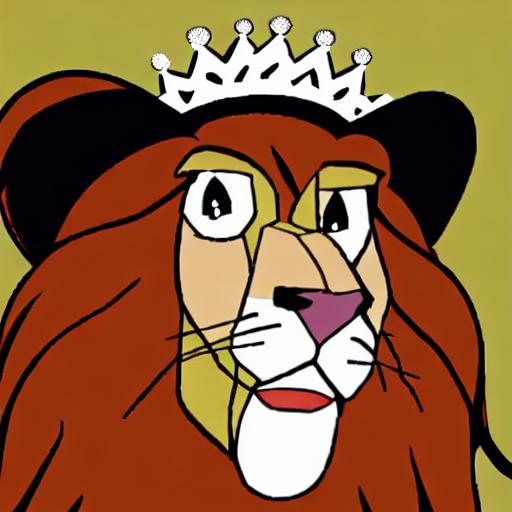} &
        \hspace{0.05cm}
        \includegraphics[width=0.11\textwidth]{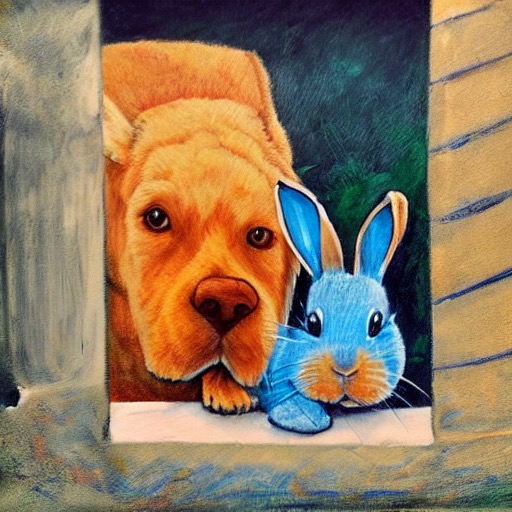} &
        \includegraphics[width=0.11\textwidth]{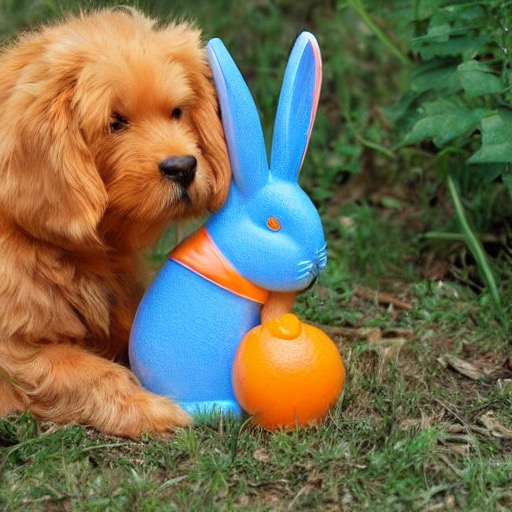} \\ \\ \\

    \end{tabular}

    }
    \vspace{-0.25cm}
    \caption{Additional comparisons with Stable Diffusion. For each prompt, we show four generated images where we use the same set of seeds for both approaches.
    The subject tokens optimized by Attend-and-Excite are highlighted in \textcolor{blue}{blue}.
    }
    \label{fig:creative}
\end{figure*}

\begin{figure*}
    \centering
    \setlength{\tabcolsep}{0.5pt}
    \renewcommand{\arraystretch}{0.3}
    {\small

    \begin{tabular}{c c c @{\hspace{0.1cm}} c c @{\hspace{0.1cm}} c c @{\hspace{0.1cm}} c c }

        & 
        \multicolumn{2}{c}{``A \textcolor{blue}{dog} walking on the \textcolor{blue}{city} street''}  &
        \multicolumn{2}{c}{\begin{tabular}{c}``A green \textcolor{blue}{dog} standing \\ over a vintage \textcolor{blue}{suitcase}'' \end{tabular}} &
        \multicolumn{2}{c}{\begin{tabular}{c} ``An orange \textcolor{blue}{cat} \\ curled up next to a warm \\ \textcolor{blue}{fireplace} with a vintage \textcolor{blue}{book}'' \\
        \end{tabular}} &
        \multicolumn{2}{c}{\begin{tabular}{c}``A nice picture of \textcolor{blue}{apples},  \textcolor{blue}{straw-} \\ \textcolor{blue}{berries}, black \textcolor{blue}{currants} and \textcolor{blue}{juice} rich \\ in vitamin c''\end{tabular}} \\

        {\raisebox{0.3in}{
        \multirow{2}{*}{\rotatebox{90}{Stable Diffusion}}}} &
        \includegraphics[width=0.11\textwidth]{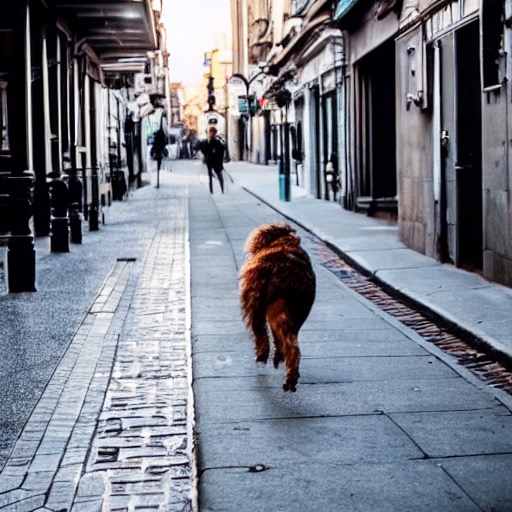} &
        \includegraphics[width=0.11\textwidth]{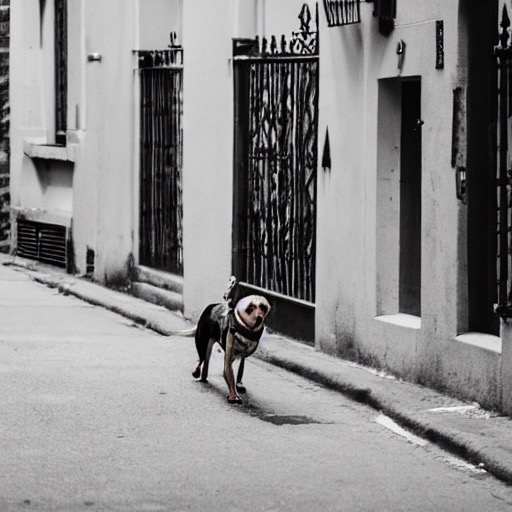} &
        \hspace{0.05cm}
        \includegraphics[width=0.11\textwidth]{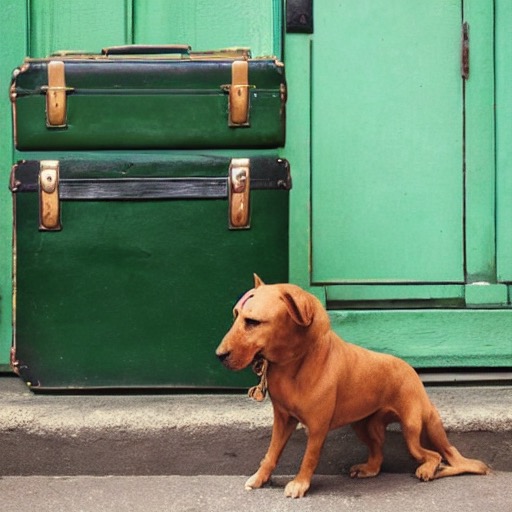} &
        \includegraphics[width=0.11\textwidth]{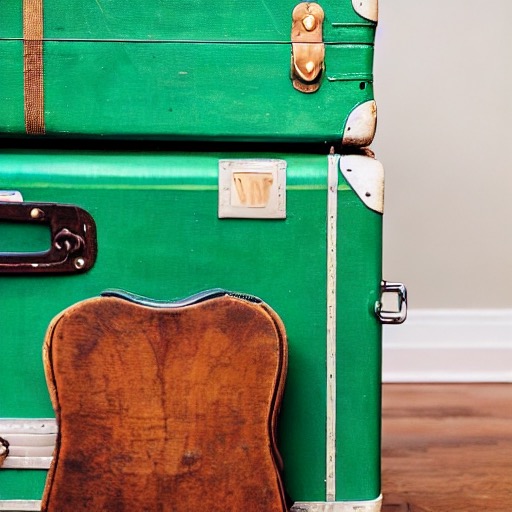} &
        \hspace{0.05cm}
        \includegraphics[width=0.11\textwidth]{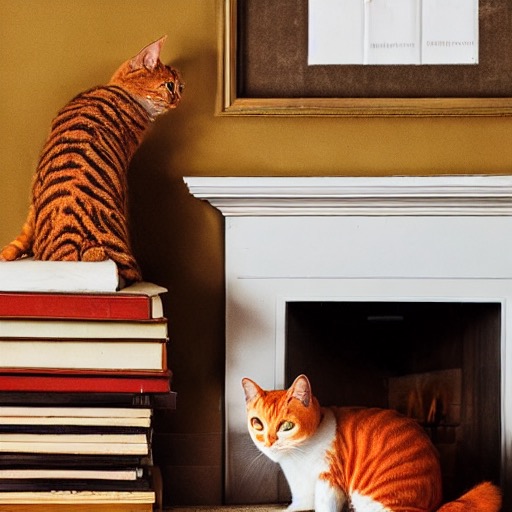} &
        \includegraphics[width=0.11\textwidth]{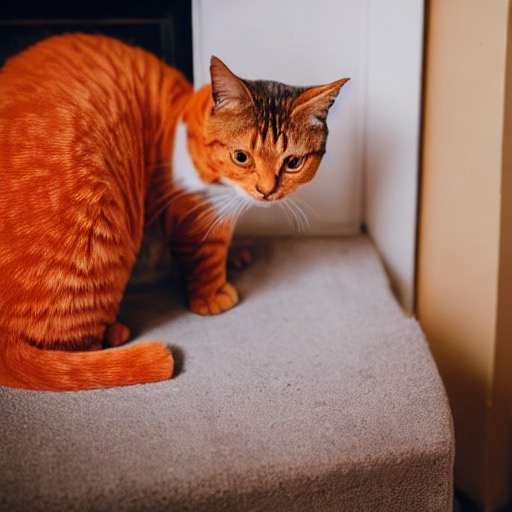} &
        \hspace{0.05cm}
        \includegraphics[width=0.11\textwidth]{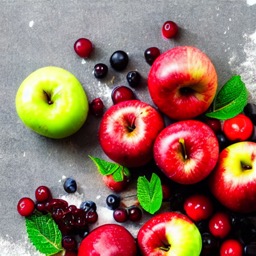} & 
        \includegraphics[width=0.11\textwidth]{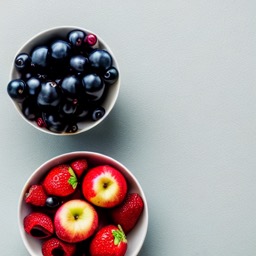} \\  

        & 
        \includegraphics[width=0.11\textwidth]{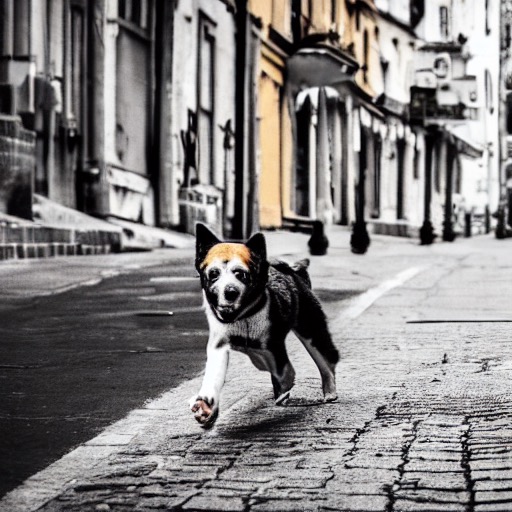} &
        \includegraphics[width=0.11\textwidth]{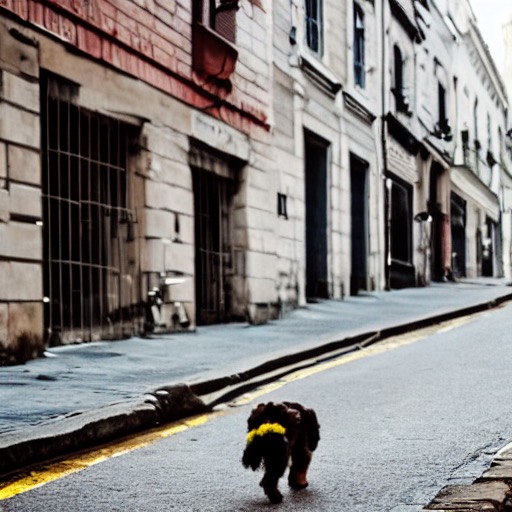} &
        \hspace{0.05cm}
        \includegraphics[width=0.11\textwidth]{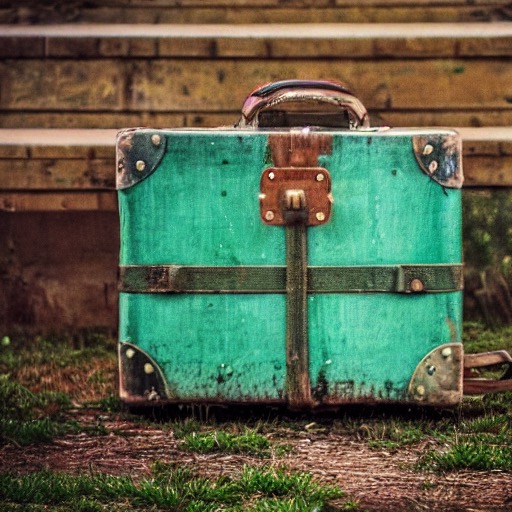} &
        \includegraphics[width=0.11\textwidth]{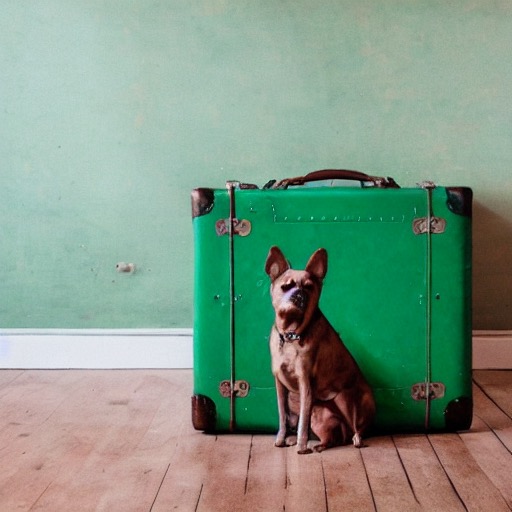} &
        \hspace{0.05cm}
        \includegraphics[width=0.11\textwidth]{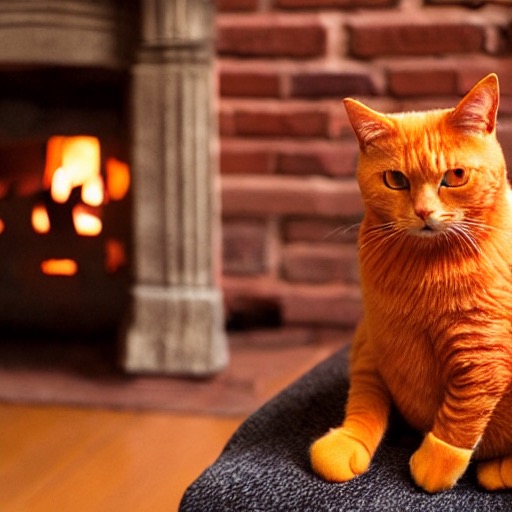} &
        \includegraphics[width=0.11\textwidth]{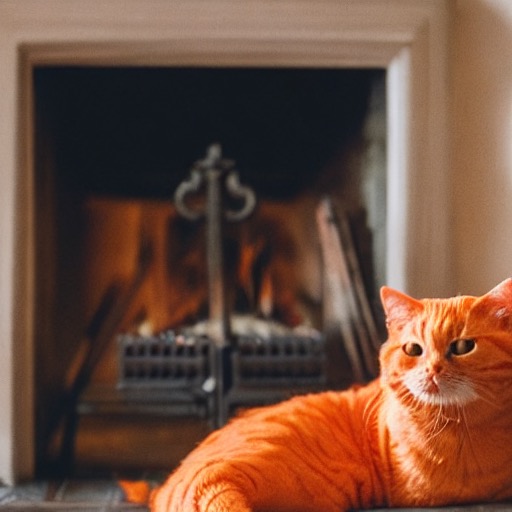} &
        \hspace{0.05cm}
        \includegraphics[width=0.11\textwidth]{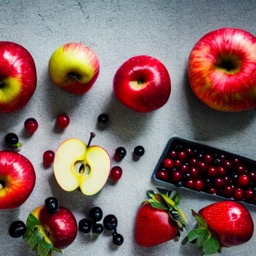} & 
        \includegraphics[width=0.11\textwidth]{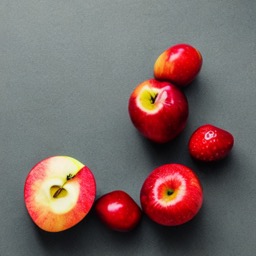} \\ \\ \\
       
        {\raisebox{0.425in}{
        \multirow{2}{*}{\rotatebox{90}{\begin{tabular}{c} Stable Diffusion with \\ \textcolor{blue}{Attend-and-Excite} \\ \\ \end{tabular}}}}} &
        \includegraphics[width=0.11\textwidth]{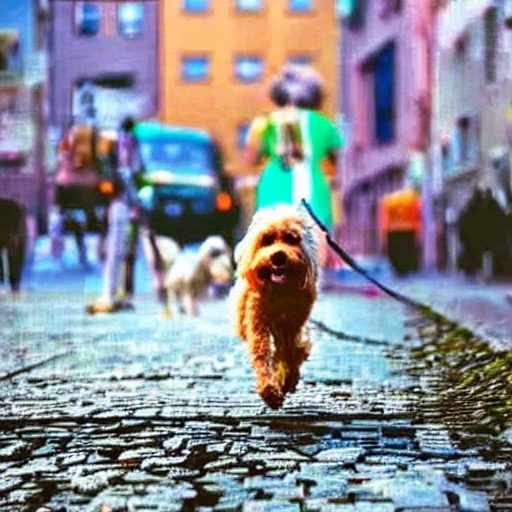} &
        \includegraphics[width=0.11\textwidth]{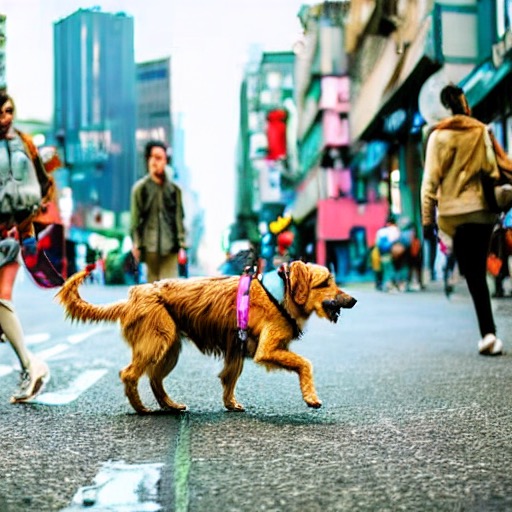} &
        \hspace{0.05cm}
        \includegraphics[width=0.11\textwidth]{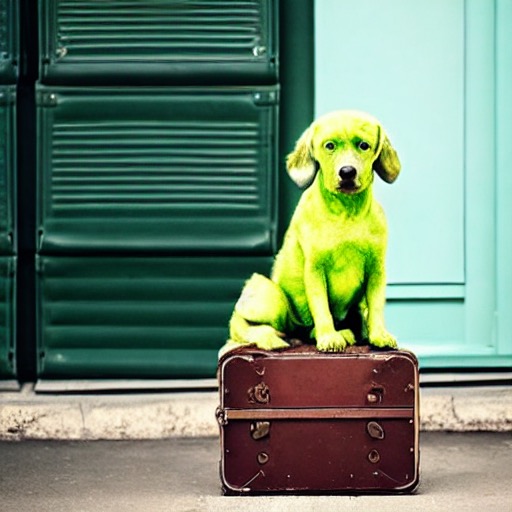} &
        \includegraphics[width=0.11\textwidth]{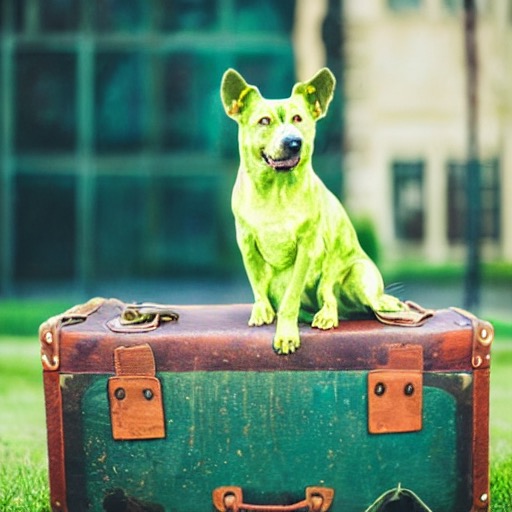} &
        \hspace{0.05cm}
        \includegraphics[width=0.11\textwidth]{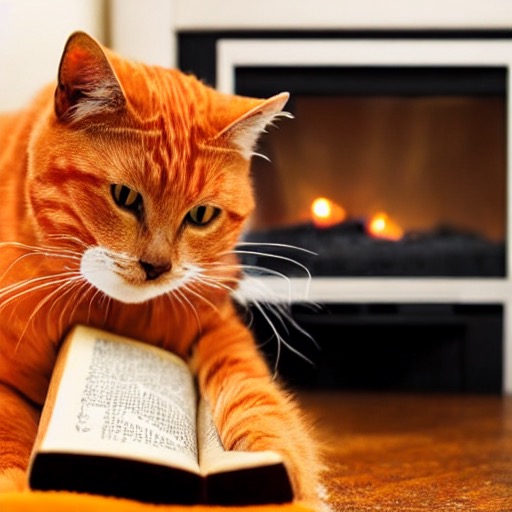} &
        \includegraphics[width=0.11\textwidth]{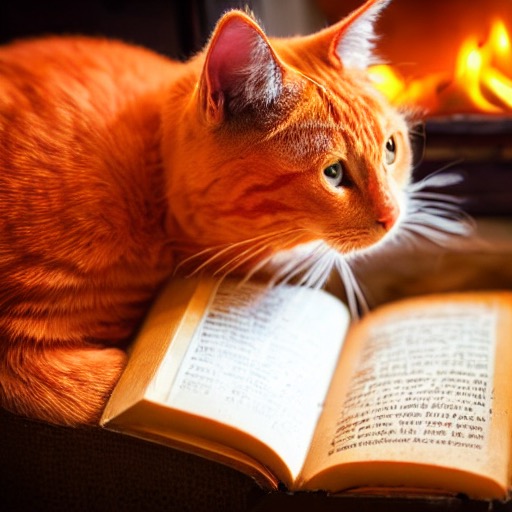} &
        \hspace{0.05cm}
        \includegraphics[width=0.11\textwidth]{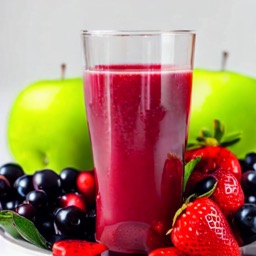} &
        \includegraphics[width=0.11\textwidth]{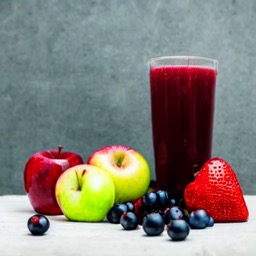} \\
        
        &
        \includegraphics[width=0.11\textwidth]{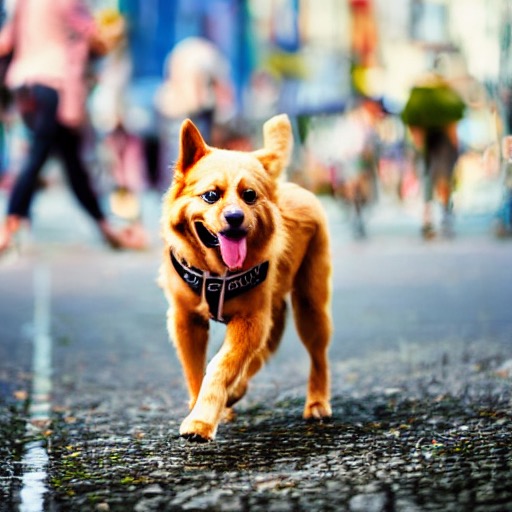} &
        \includegraphics[width=0.11\textwidth]{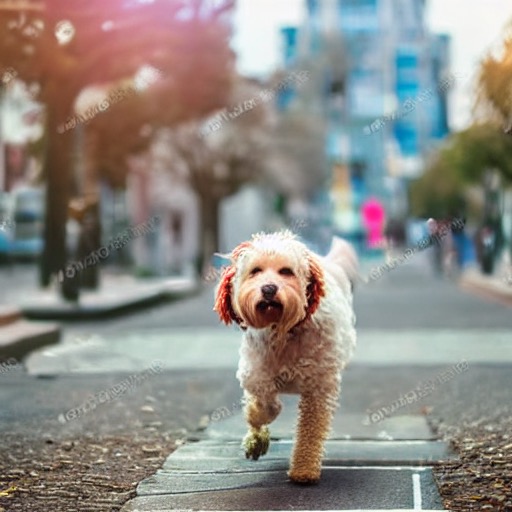} &
        \hspace{0.05cm}
        \includegraphics[width=0.11\textwidth]{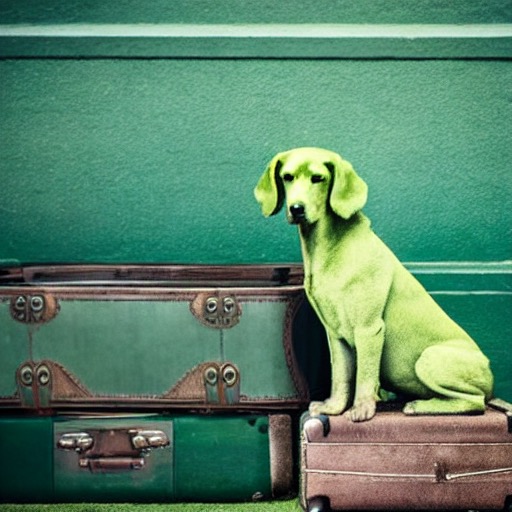} &
        \includegraphics[width=0.11\textwidth]{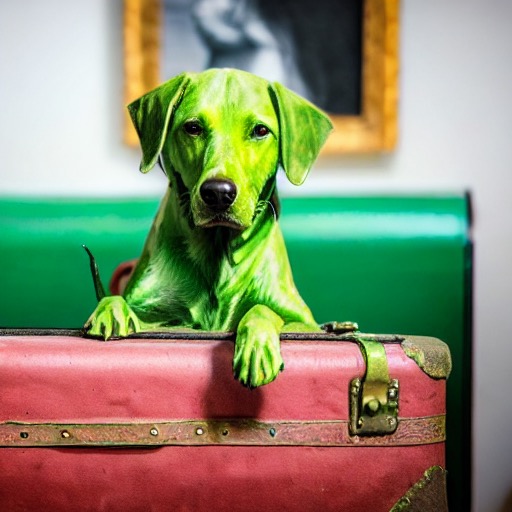} &
        \hspace{0.05cm}
        \includegraphics[width=0.11\textwidth]{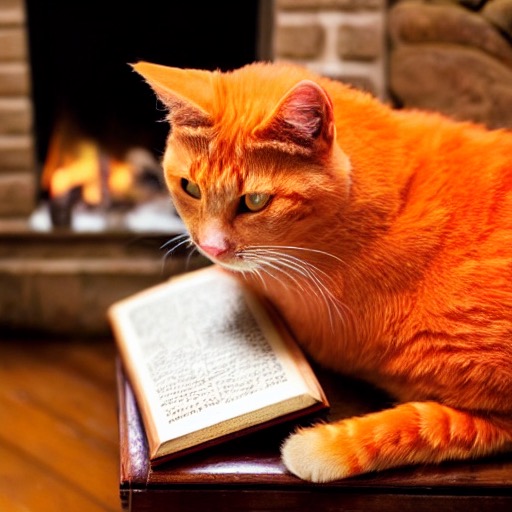} &
        \includegraphics[width=0.11\textwidth]{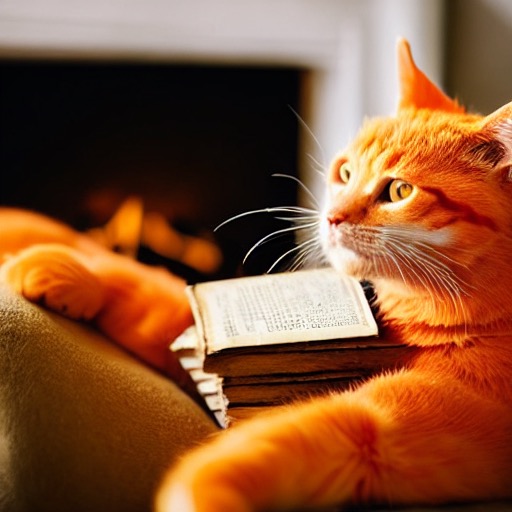} &
        \hspace{0.05cm}
        \includegraphics[width=0.11\textwidth]{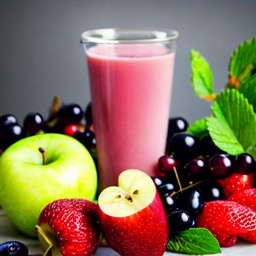} & 
        \includegraphics[width=0.11\textwidth]{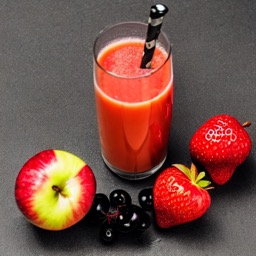} \\ \\ \\

    \end{tabular}
    
    }
    \vspace{-0.5cm}
    \caption{Additional comparisons with Stable Diffusion. For each prompt, we show four generated images where we use the same set of seeds for both approaches.
    The subject tokens optimized by Attend-and-Excite are highlighted in \textcolor{blue}{blue}.
    }
    \label{fig:creative2}
    \vspace{0.1cm}
\end{figure*}

\begin{figure*}
    \centering
    \setlength{\tabcolsep}{0.5pt}
    \renewcommand{\arraystretch}{0.3}
    \begin{tabular}{c c c @{\hspace{0.1cm}} c c c }

        \multicolumn{3}{c}{StructureDiffusion} &
        \multicolumn{3}{c}{\begin{tabular}{c} Stable Diffusion with \textcolor{blue}{Attend-and-Excite} \end{tabular}} \\ \\ \\

        \multicolumn{6}{c}{\begin{tabular}{c} ``A wooden \textcolor{blue}{building} with a stone \textcolor{blue}{bench} placed near it'' \end{tabular}} \\
        \includegraphics[width=0.115\textwidth]{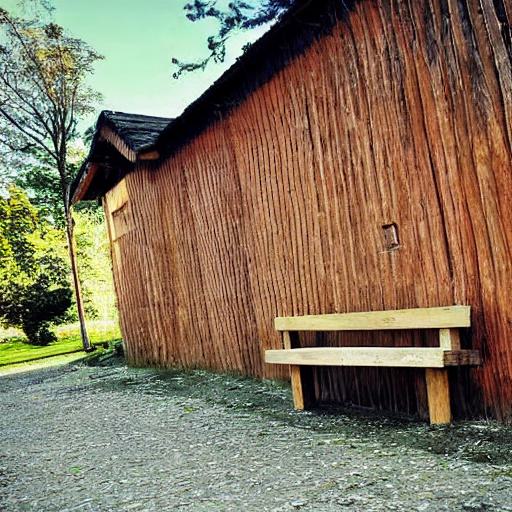} &
        \includegraphics[width=0.115\textwidth]{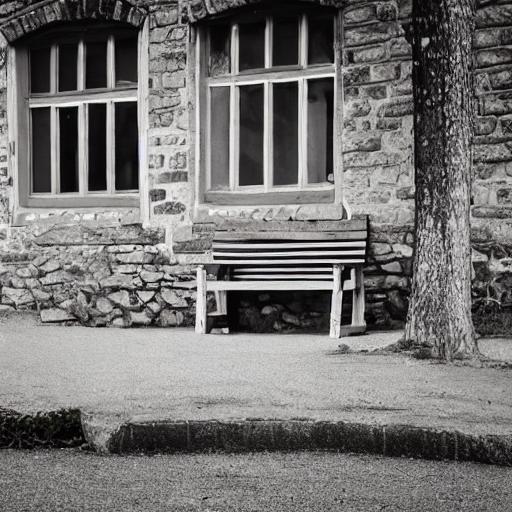} &
        \includegraphics[width=0.115\textwidth]{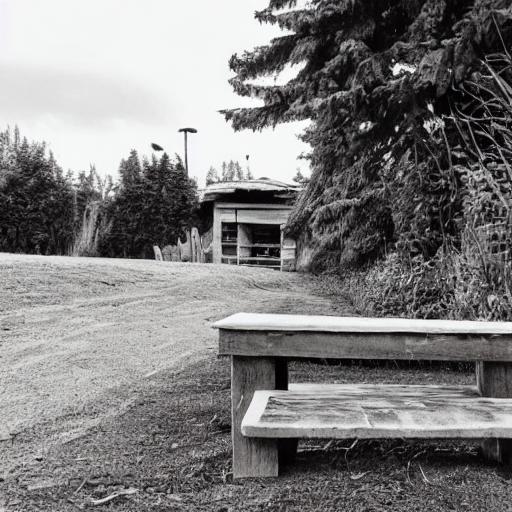} &
        \hspace{0.05cm}
        \includegraphics[width=0.115\textwidth]{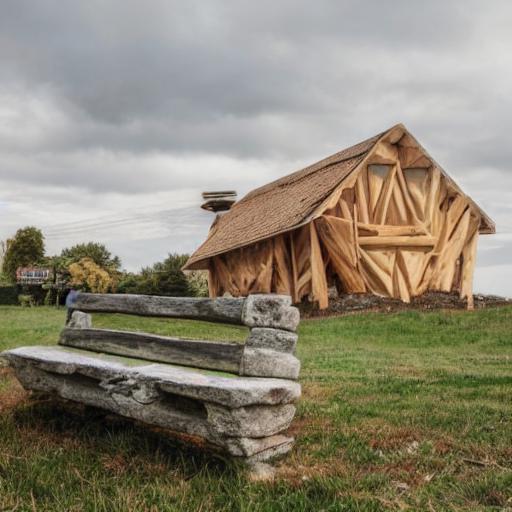} &
        \includegraphics[width=0.115\textwidth]{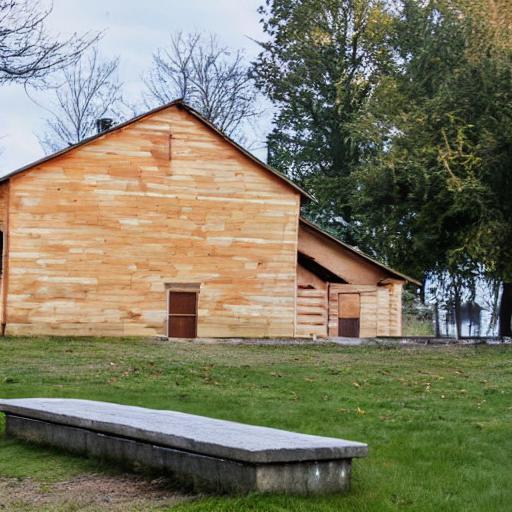} &
        \includegraphics[width=0.115\textwidth]{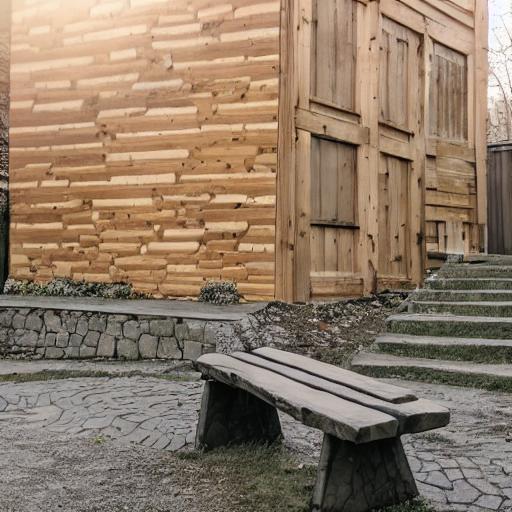} \\ \\

        \multicolumn{6}{c}{\begin{tabular}{c} ``A red \textcolor{blue}{cat} sits on a \textcolor{blue}{rug} with a black \textcolor{blue}{chord}'' \end{tabular}} \\
        \includegraphics[width=0.115\textwidth]{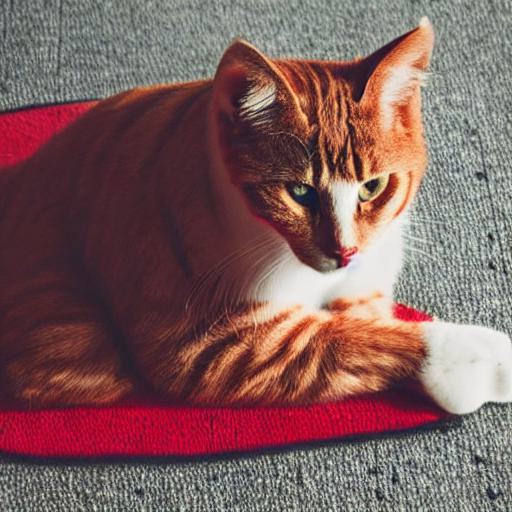} &
        \includegraphics[width=0.115\textwidth]{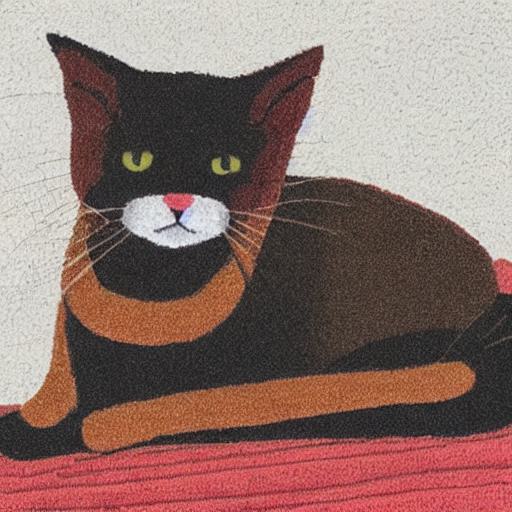} &
        \includegraphics[width=0.115\textwidth]{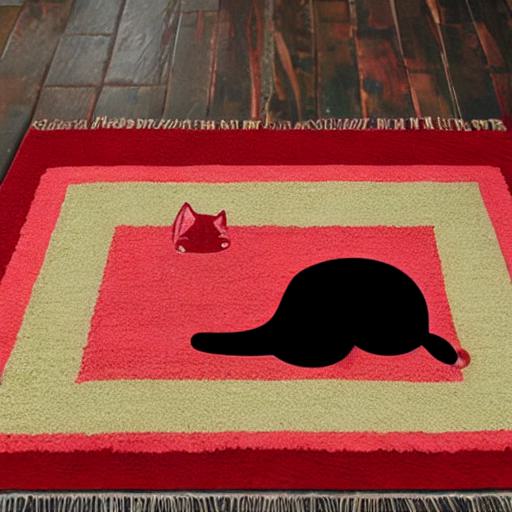} &
        \hspace{0.05cm}
        \includegraphics[width=0.115\textwidth]{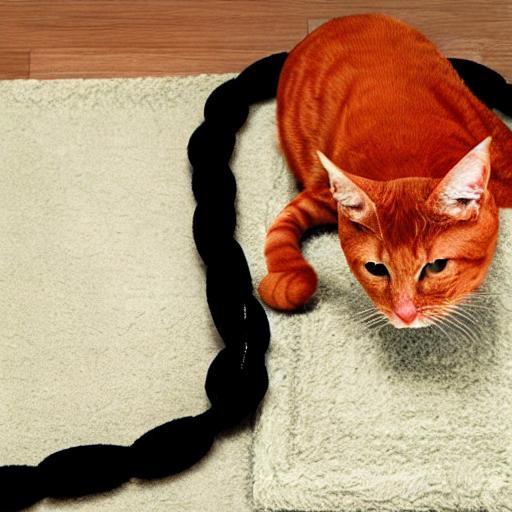} &
        \includegraphics[width=0.115\textwidth]{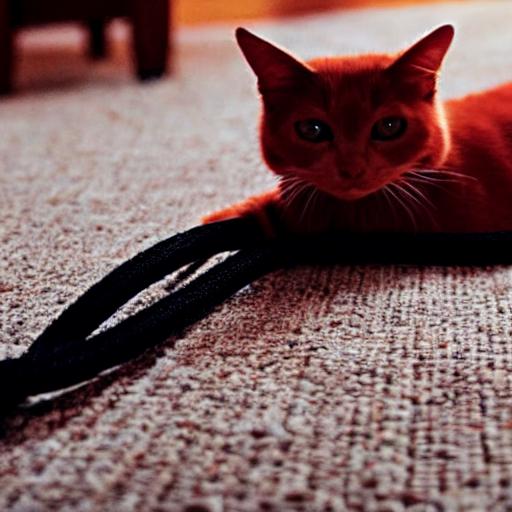} &
        \includegraphics[width=0.115\textwidth]{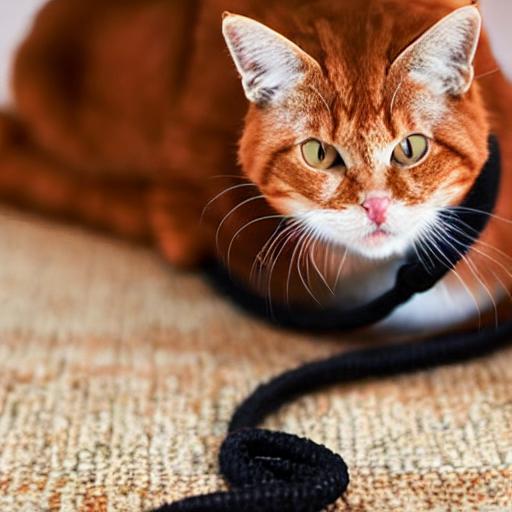} \\ \\

        \multicolumn{6}{c}{\begin{tabular}{c} ``A green \textcolor{blue}{sphere} and a red \textcolor{blue}{cube}'' \end{tabular}} \\
        \includegraphics[width=0.11\textwidth]{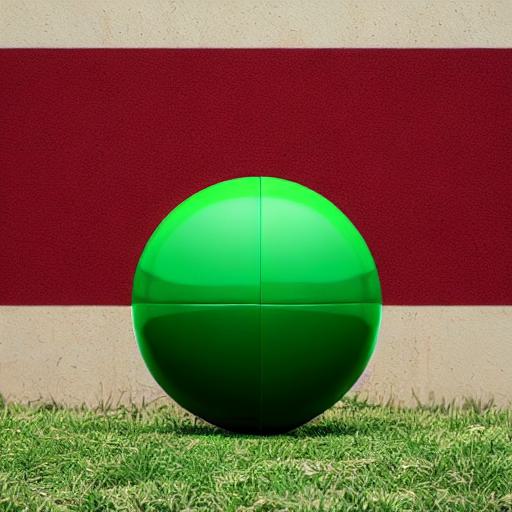} &
        \includegraphics[width=0.11\textwidth]{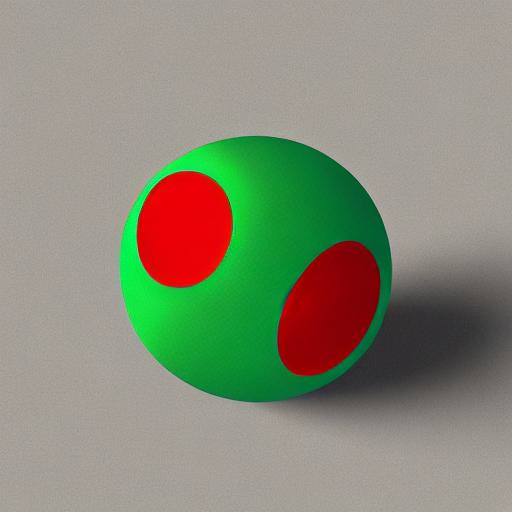} &
        \includegraphics[width=0.11\textwidth]{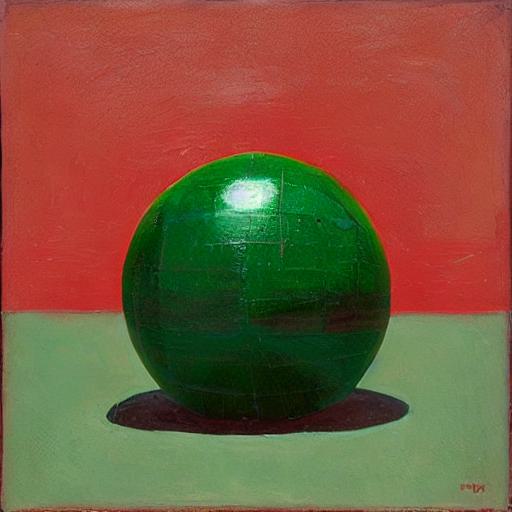} &
        \hspace{0.05cm}
        \includegraphics[width=0.11\textwidth]{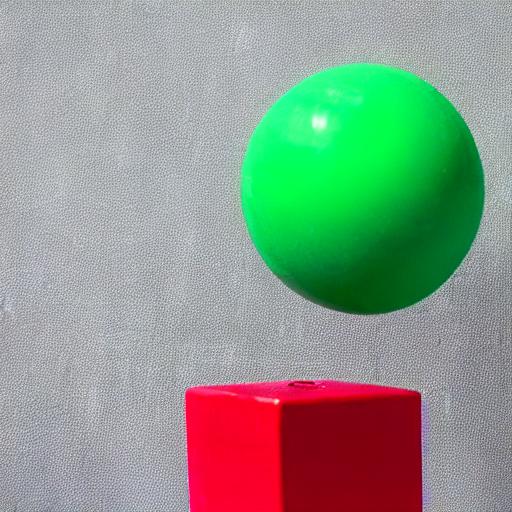} &
        \includegraphics[width=0.11\textwidth]{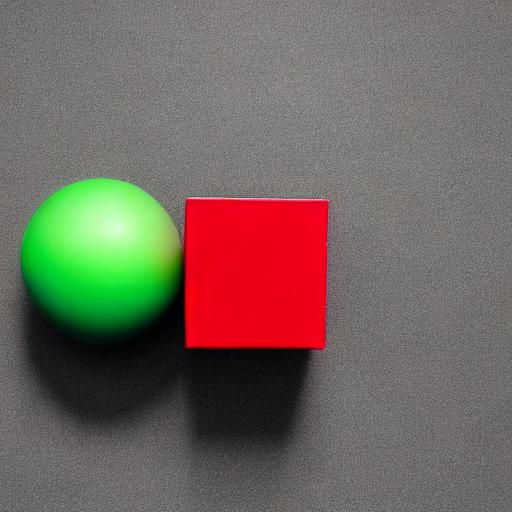} &
        \includegraphics[width=0.11\textwidth]{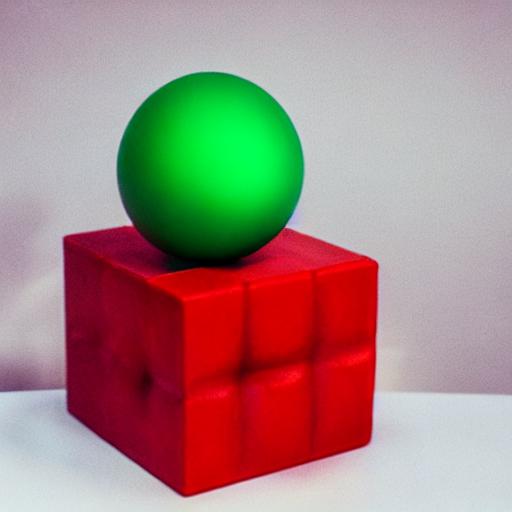} \\ \\

        \multicolumn{6}{c}{\begin{tabular}{c} ``A large \textcolor{blue}{pizza} on a white \textcolor{blue}{plate} sitting on a blue \textcolor{blue}{table}'' \end{tabular}} \\
        \includegraphics[width=0.11\textwidth]{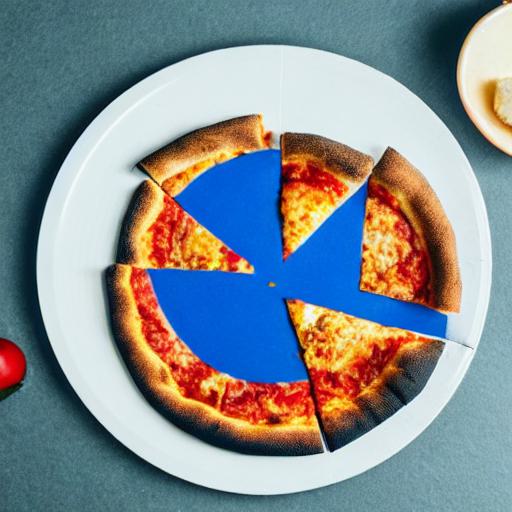} &
        \includegraphics[width=0.11\textwidth]{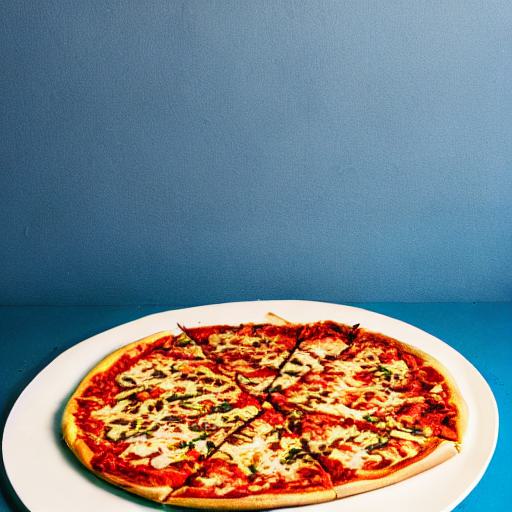} &
        \includegraphics[width=0.11\textwidth]{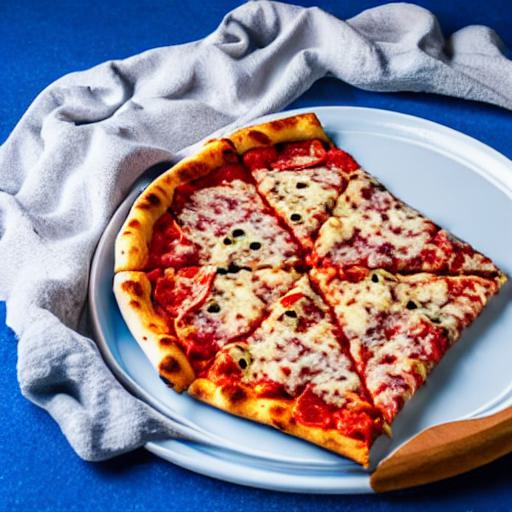} &
        \hspace{0.05cm}
        \includegraphics[width=0.11\textwidth]{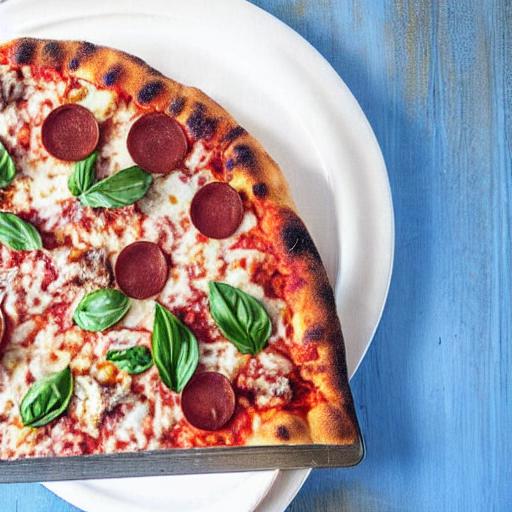} &
        \includegraphics[width=0.11\textwidth]{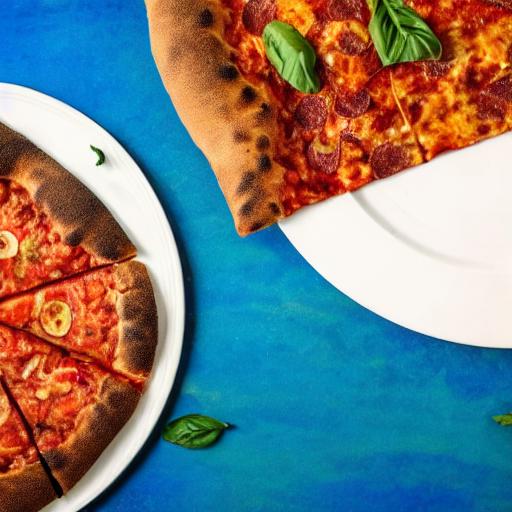} &
        \includegraphics[width=0.11\textwidth]{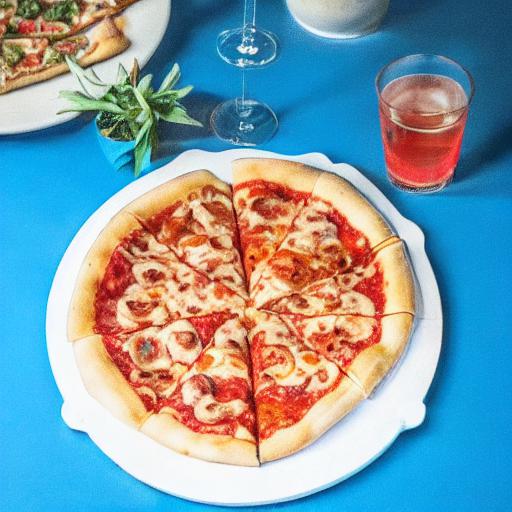} \\ \\

    \end{tabular}
    
    \vspace{-0.25cm}
    \caption{Additional comparisons with prompts appearing in Feng~\etal~\shortcite{Feng2022Training} (StructureDiffusion). For each prompt, we show results using the same set of seeds.}
    \label{fig:structured_prompts_supplementary}
\end{figure*}

\begin{figure*}
    \centering
    \setlength{\tabcolsep}{0.5pt}
    \renewcommand{\arraystretch}{0.3}
    {\small
    \begin{tabular}{c c c @{\hspace{0.1cm}} c c @{\hspace{0.1cm}} c c @{\hspace{0.1cm}} c c }

        & 
        \multicolumn{2}{c}{``A brown \textcolor{blue}{dog} and a grey \textcolor{blue}{mouse}''} &
        \multicolumn{2}{c}{``A \textcolor{blue}{bird} and a green \textcolor{blue}{car}''} &
        \multicolumn{2}{c}{``A \textcolor{blue}{mouse} with \textcolor{blue}{glasses}''} &
        \multicolumn{2}{c}{\begin{tabular}{c} ``A red \textcolor{blue}{cake} and \\ \\ a bouquet of yellow \textcolor{blue}{tulips}'' \\\\ \end{tabular}} \\

        {\raisebox{0.3in}{
        \multirow{2}{*}{\rotatebox{90}{Stable Diffusion}}}} &
        \includegraphics[width=0.11\textwidth]{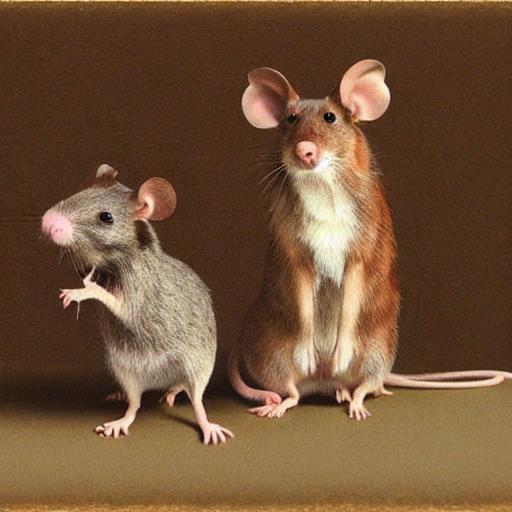} &
        \includegraphics[width=0.11\textwidth]{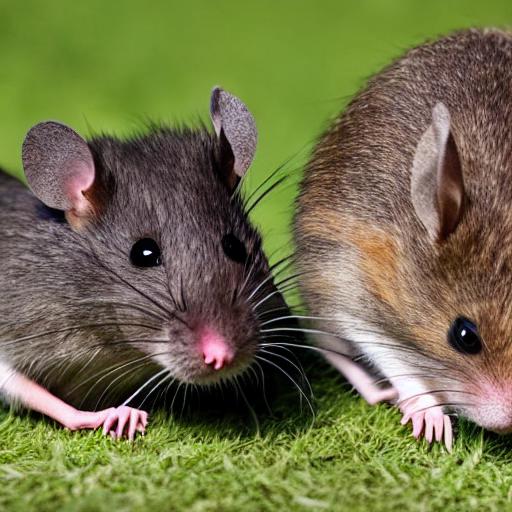} &
        \hspace{0.05cm}
        \includegraphics[width=0.11\textwidth]{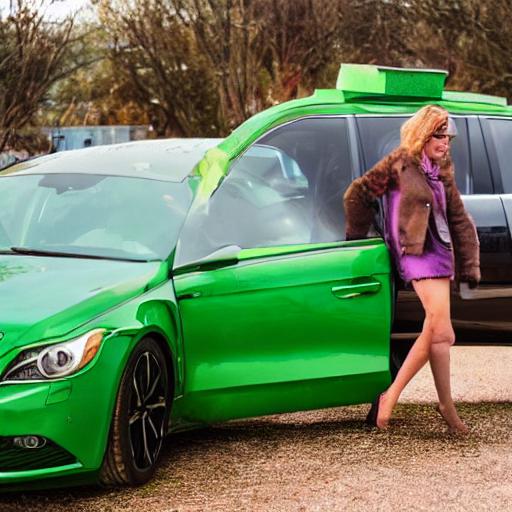} &
        \includegraphics[width=0.11\textwidth]{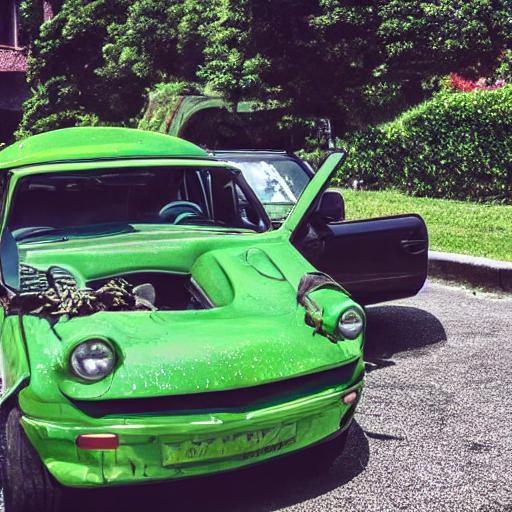} &
        \hspace{0.05cm}
        \includegraphics[width=0.11\textwidth]{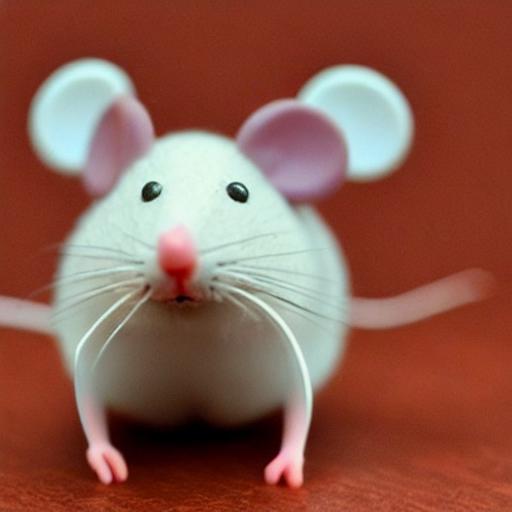} &
        \includegraphics[width=0.11\textwidth]{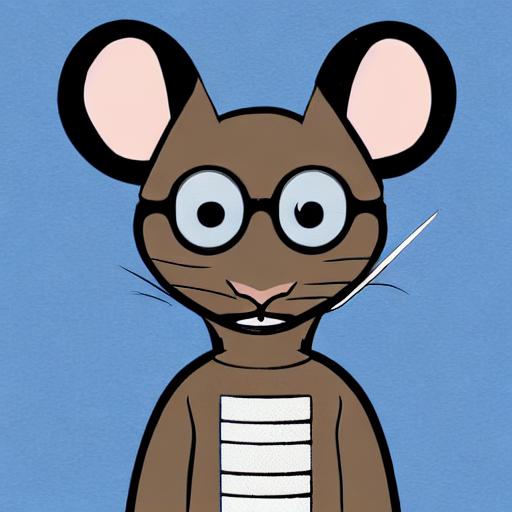} &
        \hspace{0.05cm}
        \includegraphics[width=0.11\textwidth]{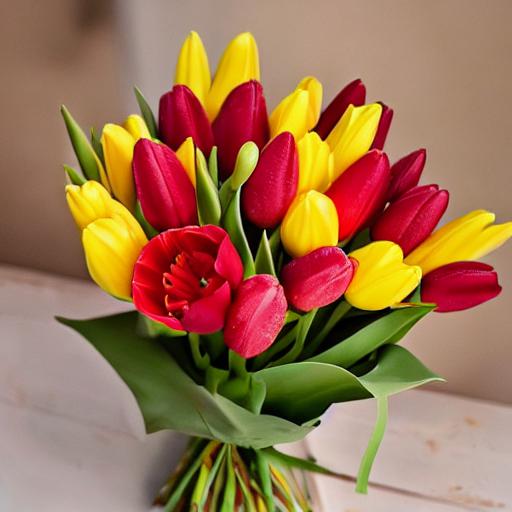} &
        \includegraphics[width=0.11\textwidth]{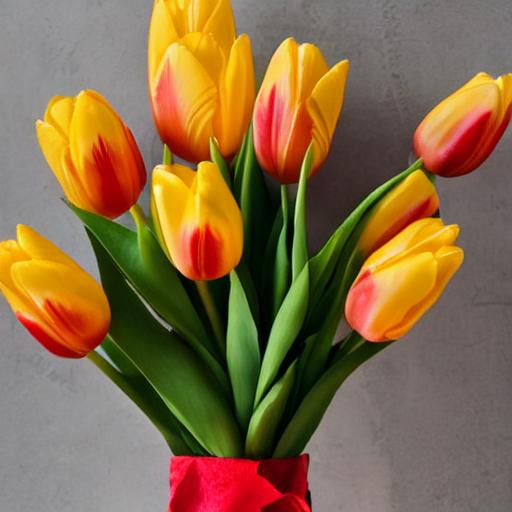} \\

        & 
        \includegraphics[width=0.11\textwidth]{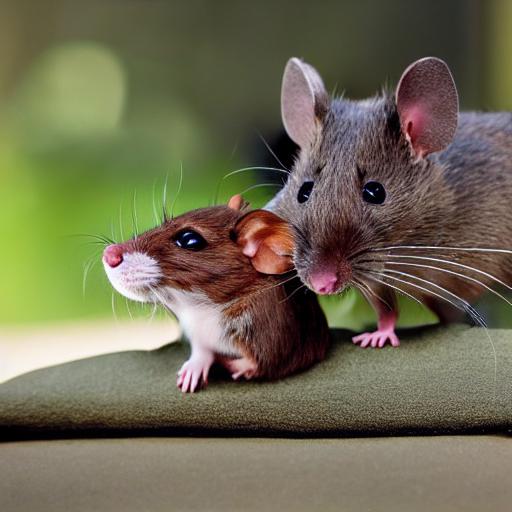} &
        \includegraphics[width=0.11\textwidth]{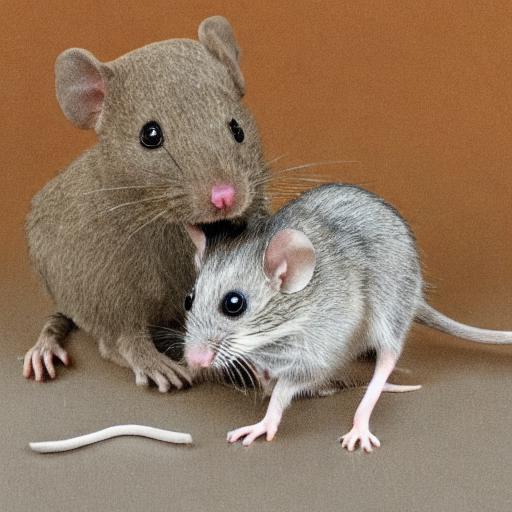} &
        \hspace{0.05cm}
        \includegraphics[width=0.11\textwidth]{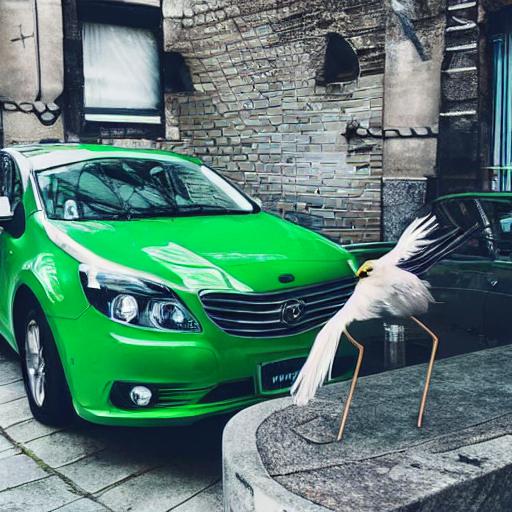} &
        \includegraphics[width=0.11\textwidth]{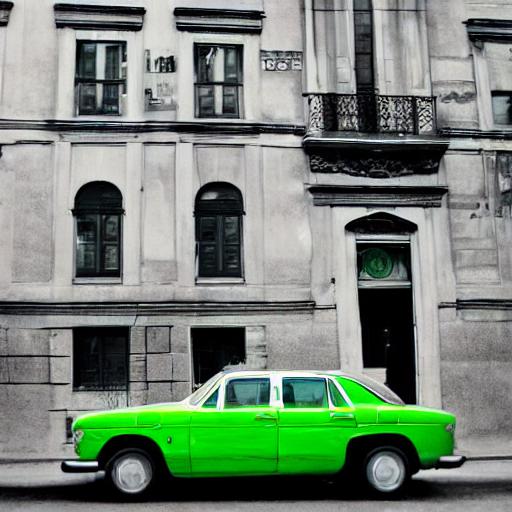} &
        \hspace{0.05cm}
        \includegraphics[width=0.11\textwidth]{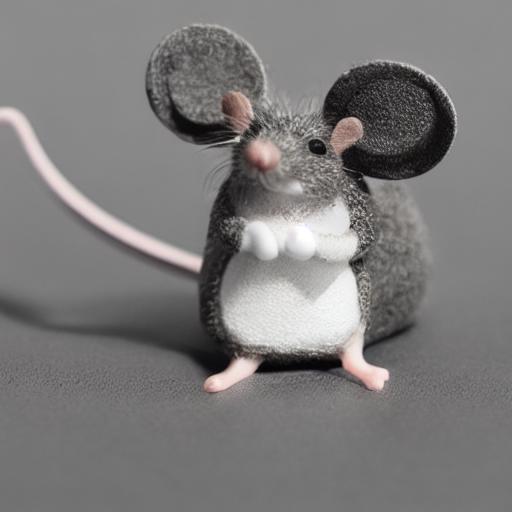} &
        \includegraphics[width=0.11\textwidth]{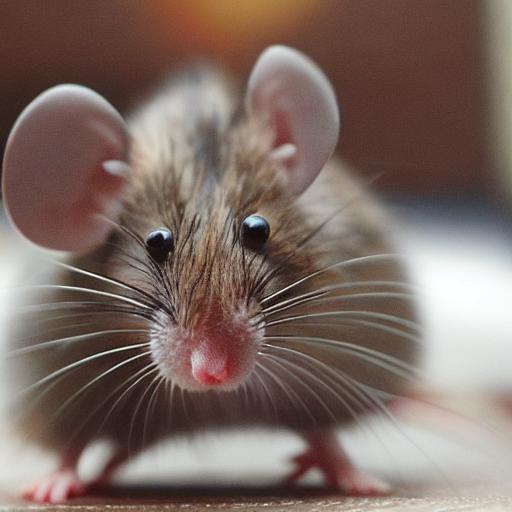} &
        \hspace{0.05cm}
        \includegraphics[width=0.11\textwidth]{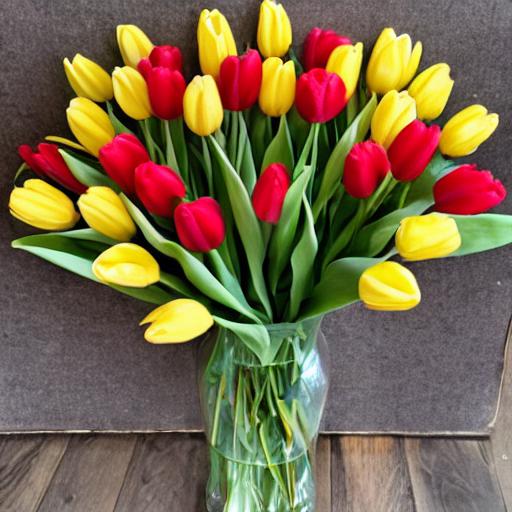} &
        \includegraphics[width=0.11\textwidth]{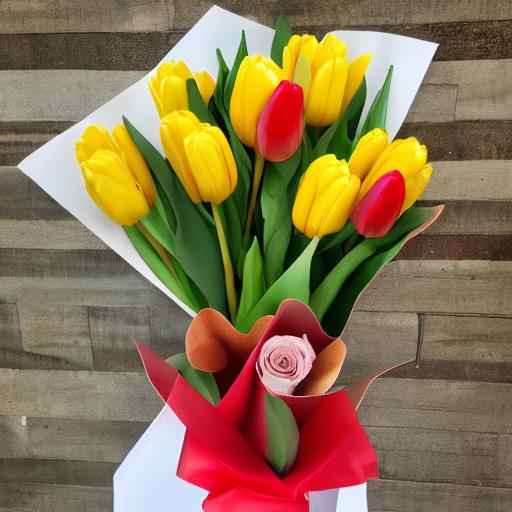}  \\ \\

        {\raisebox{0.35in}{
        \multirow{2}{*}{\rotatebox{90}{Composable Diffusion}}}} &
        \includegraphics[width=0.11\textwidth]{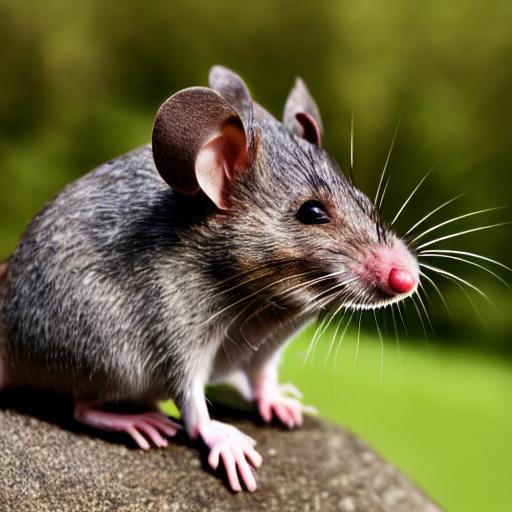} &
        \includegraphics[width=0.11\textwidth]{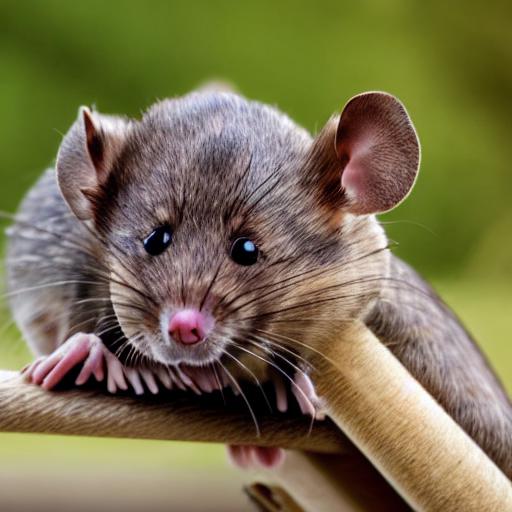} &
        \hspace{0.05cm}
        \includegraphics[width=0.11\textwidth]{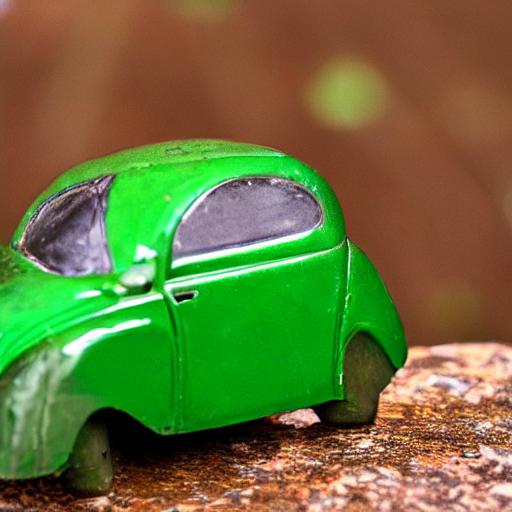} &
        \includegraphics[width=0.11\textwidth]{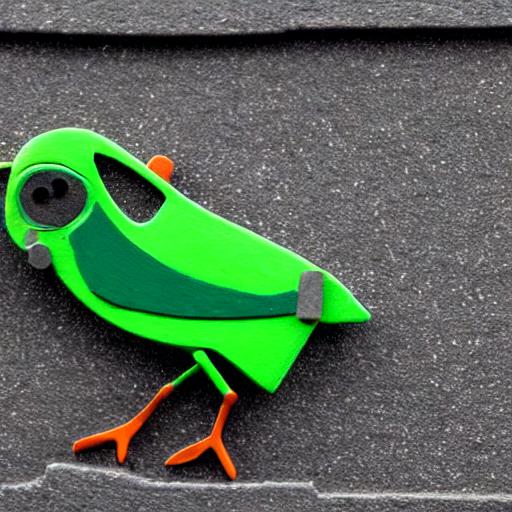} &
        \hspace{0.05cm}
        \includegraphics[width=0.11\textwidth]{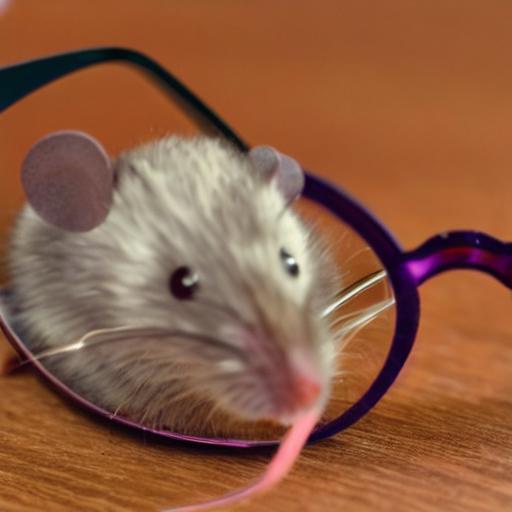} &
        \includegraphics[width=0.11\textwidth]{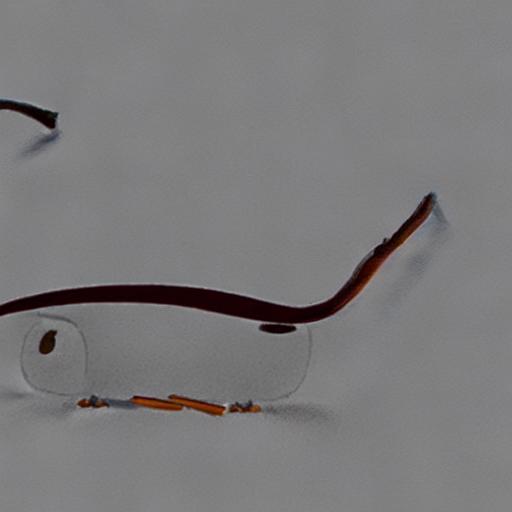} &
        \hspace{0.05cm}
        \includegraphics[width=0.11\textwidth]{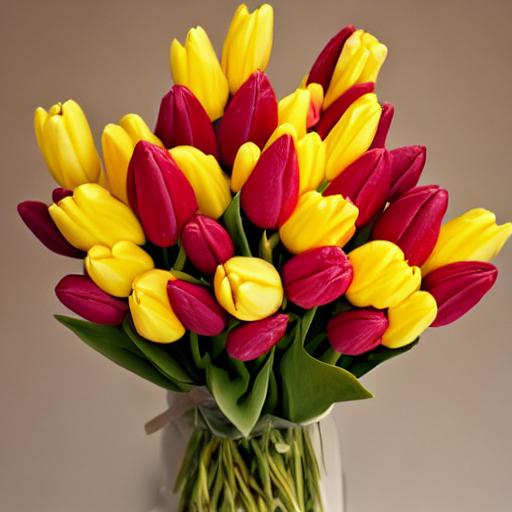} &
        \includegraphics[width=0.11\textwidth]{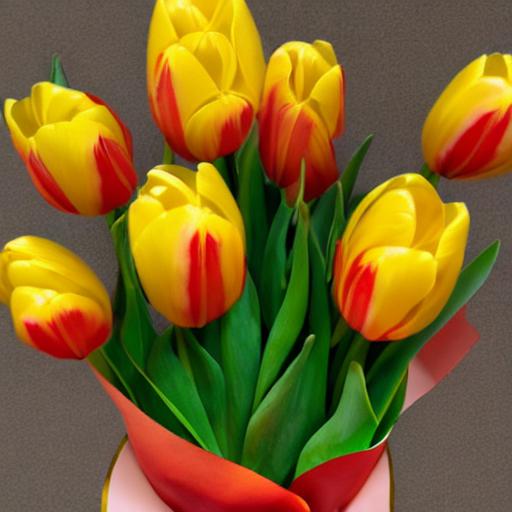} \\

        & 
        \includegraphics[width=0.11\textwidth]{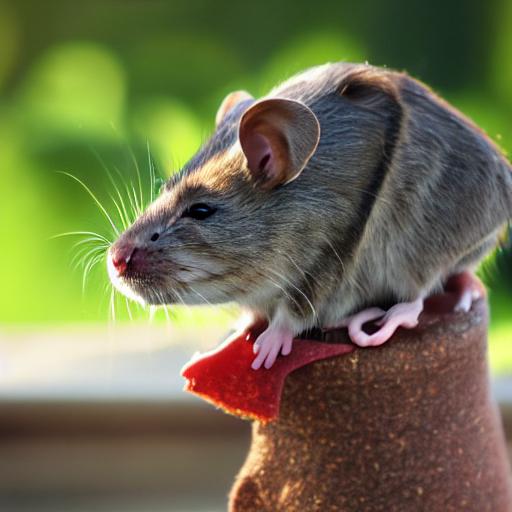} &
        \includegraphics[width=0.11\textwidth]{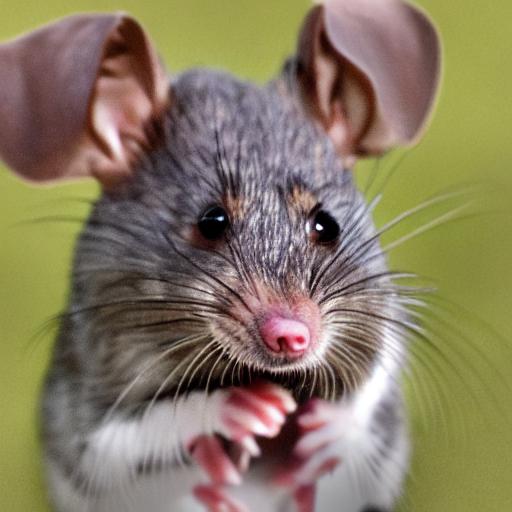} &
        \hspace{0.05cm}
        \includegraphics[width=0.11\textwidth]{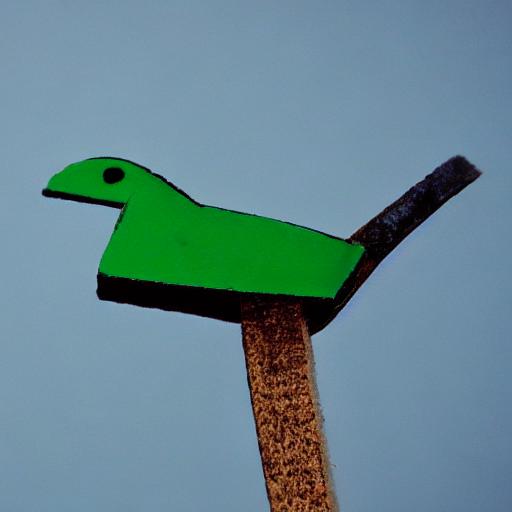} &
        \includegraphics[width=0.11\textwidth]{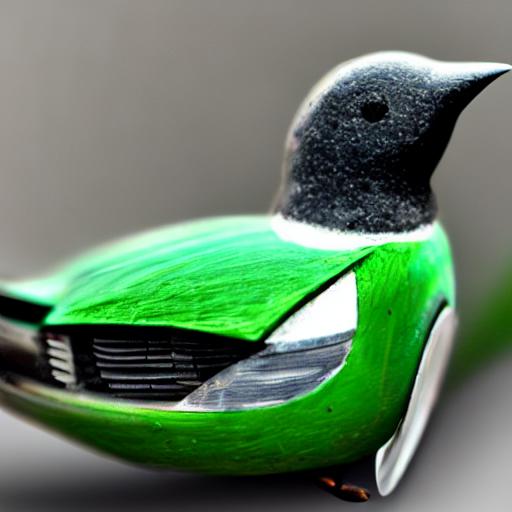} &
        \hspace{0.05cm}
        \includegraphics[width=0.11\textwidth]{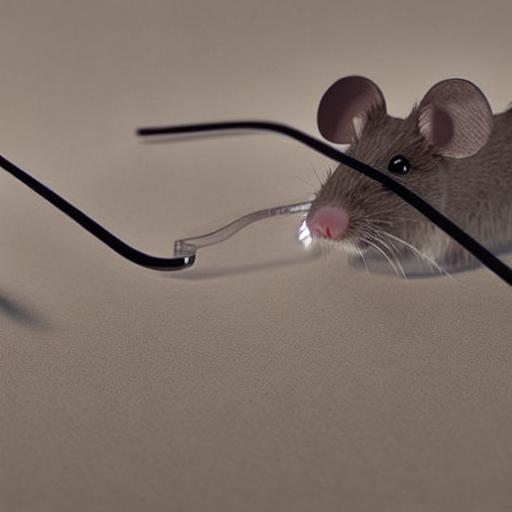} &
        \includegraphics[width=0.11\textwidth]{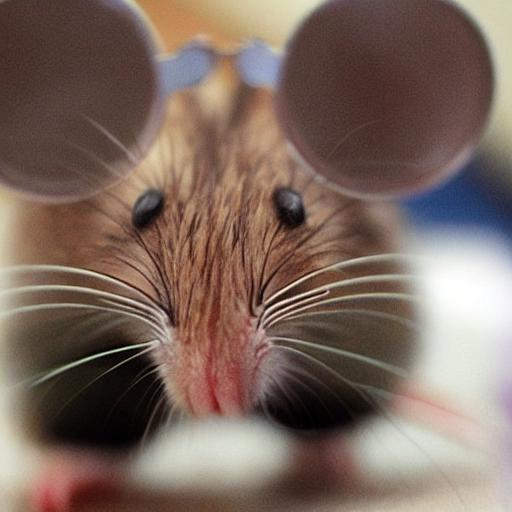} &
        \hspace{0.05cm}
        \includegraphics[width=0.11\textwidth]{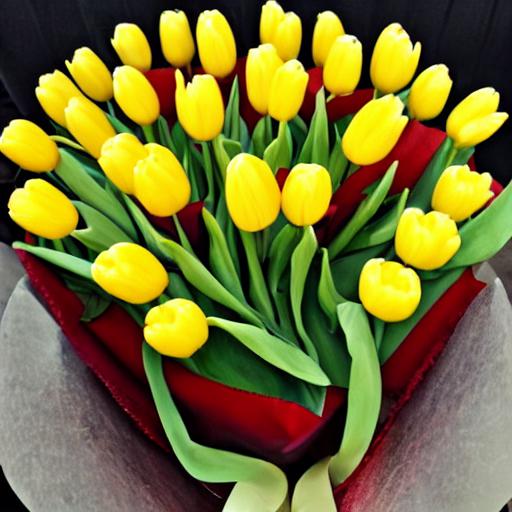} &
        \includegraphics[width=0.11\textwidth]{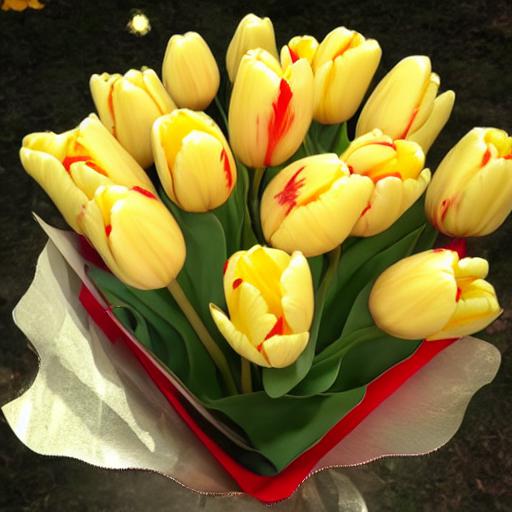}  \\ \\
        
        {\raisebox{0.35in}{
        \multirow{2}{*}{\rotatebox{90}{StructureDiffusion}}}} &
        \includegraphics[width=0.11\textwidth]{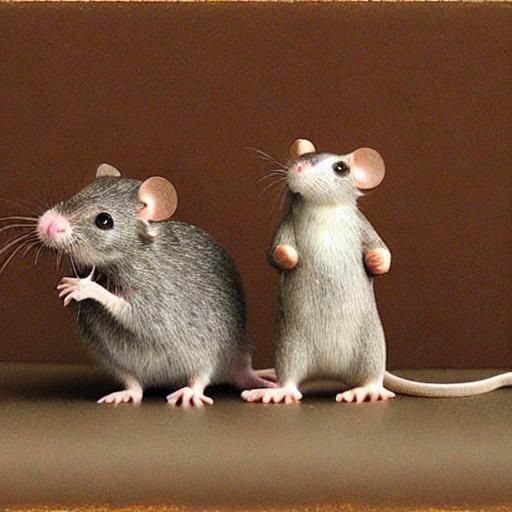} &
        \includegraphics[width=0.11\textwidth]{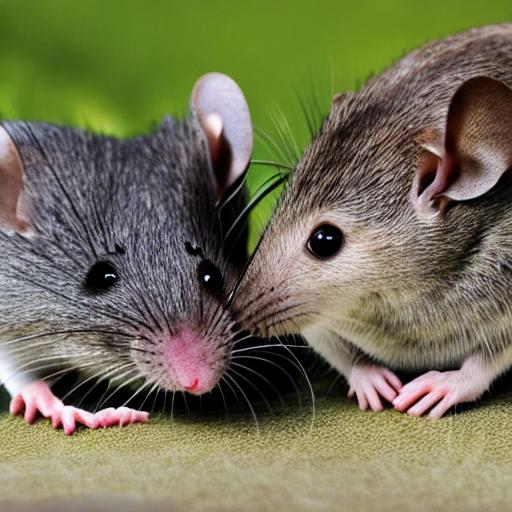} &
        \hspace{0.05cm}
        \includegraphics[width=0.11\textwidth]{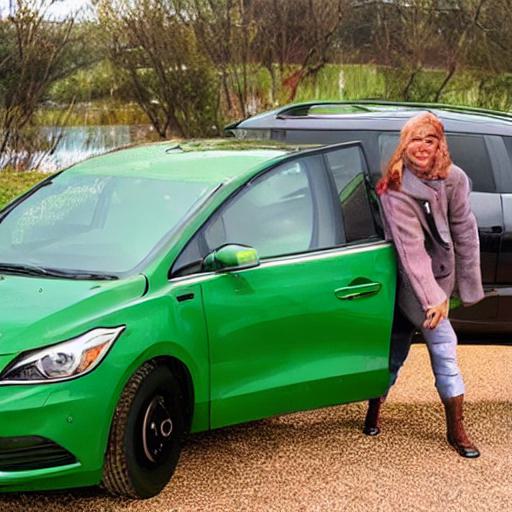} &
        \includegraphics[width=0.11\textwidth]{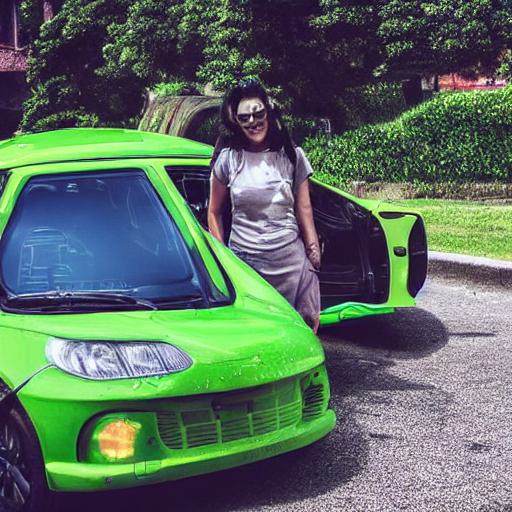} &
        \hspace{0.05cm}
        \includegraphics[width=0.11\textwidth]{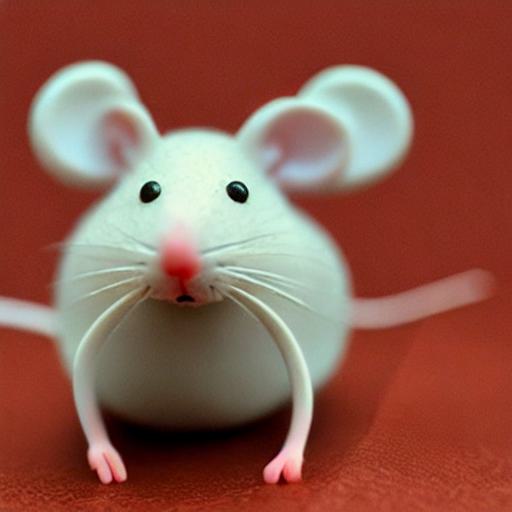} &
        \includegraphics[width=0.11\textwidth]{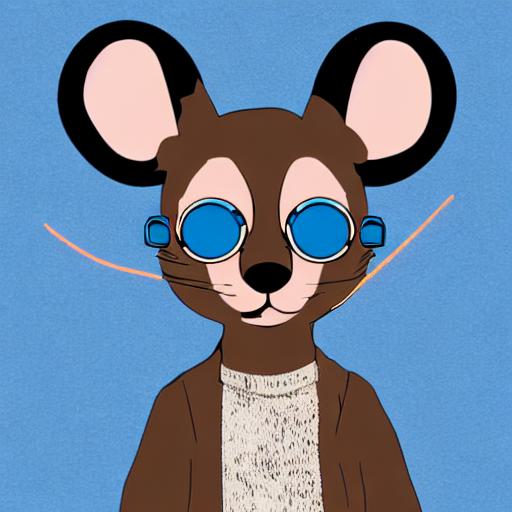} &
        \hspace{0.05cm}
        \includegraphics[width=0.11\textwidth]{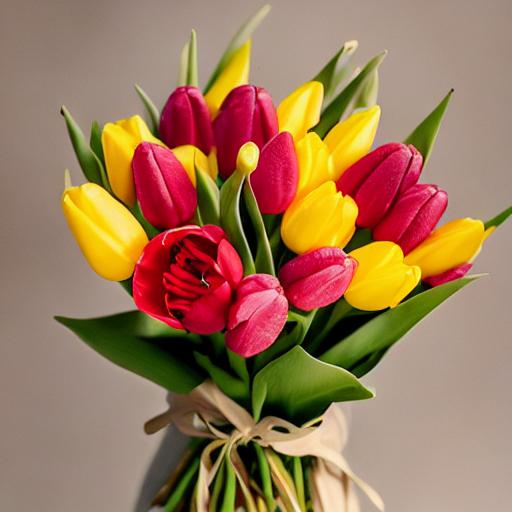} &
        \includegraphics[width=0.11\textwidth]{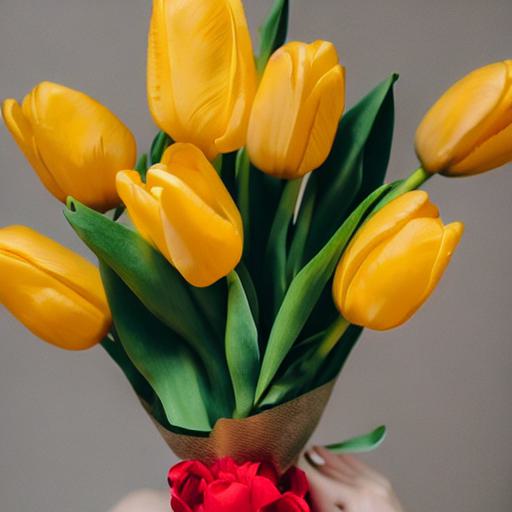} \\

        & 
        \includegraphics[width=0.11\textwidth]{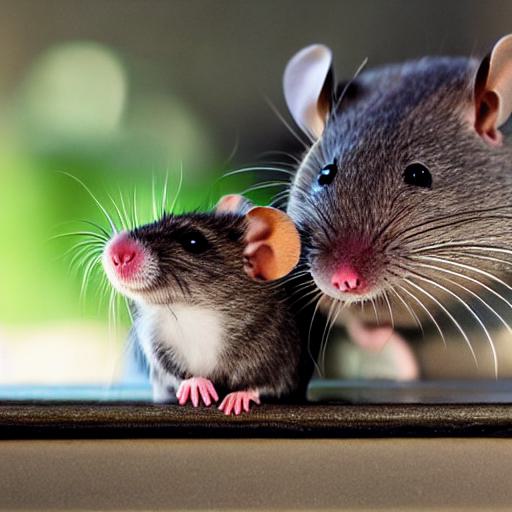} &
        \includegraphics[width=0.11\textwidth]{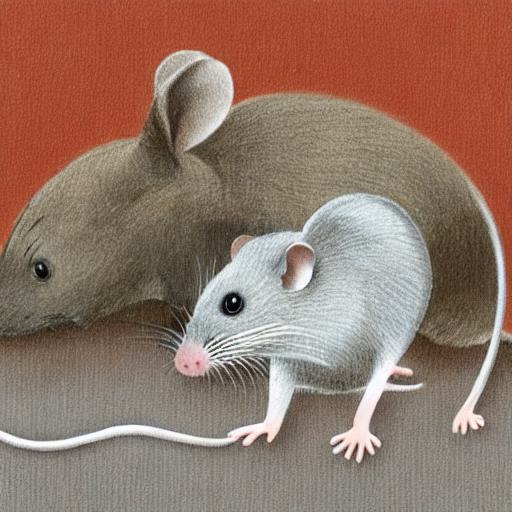} &
        \hspace{0.05cm}
        \includegraphics[width=0.11\textwidth]{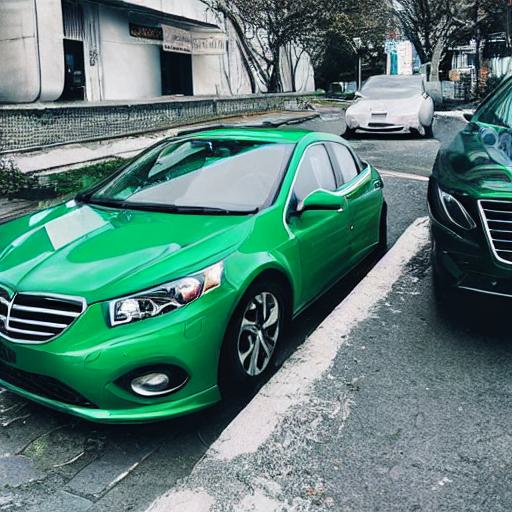} &
        \includegraphics[width=0.11\textwidth]{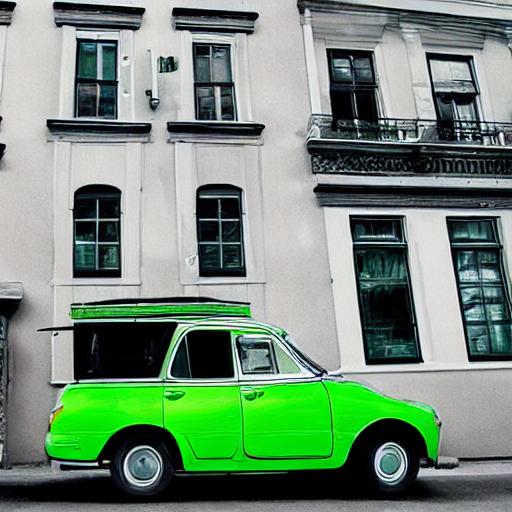} &
        \hspace{0.05cm}
        \includegraphics[width=0.11\textwidth]{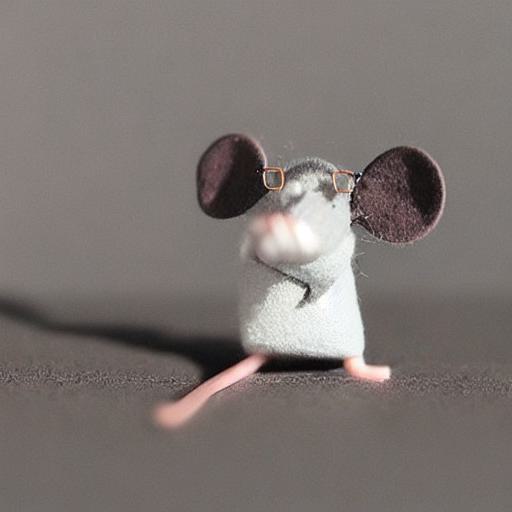} &
        \includegraphics[width=0.11\textwidth]{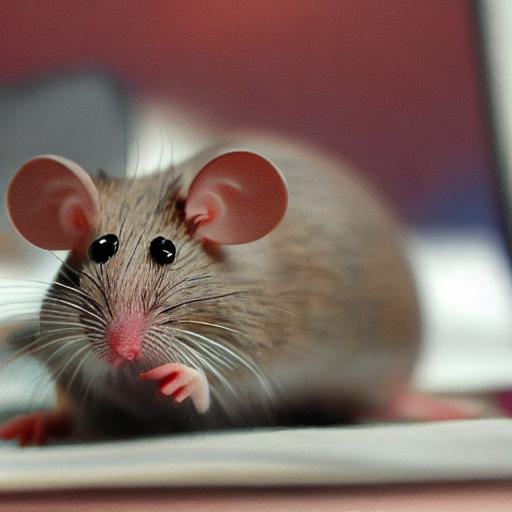} &
        \hspace{0.05cm}
        \includegraphics[width=0.11\textwidth]{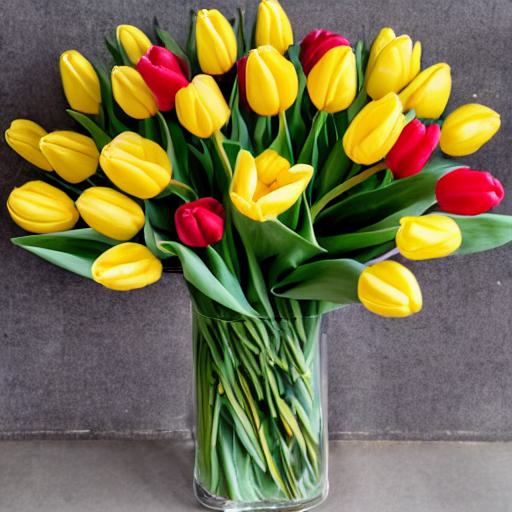} &
        \includegraphics[width=0.11\textwidth]{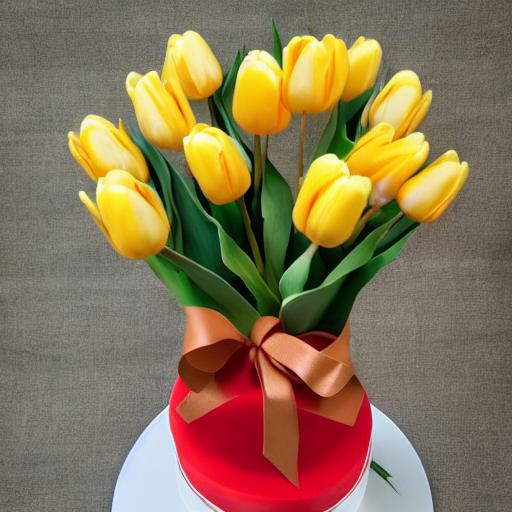}  \\ \\
        
        {\raisebox{0.425in}{
        \multirow{2}{*}{\rotatebox{90}{\begin{tabular}{c} Stable Diffusion with \\ \textcolor{blue}{Attend-and-Excite} \\ \\ \end{tabular}}}}} &
        \includegraphics[width=0.11\textwidth]{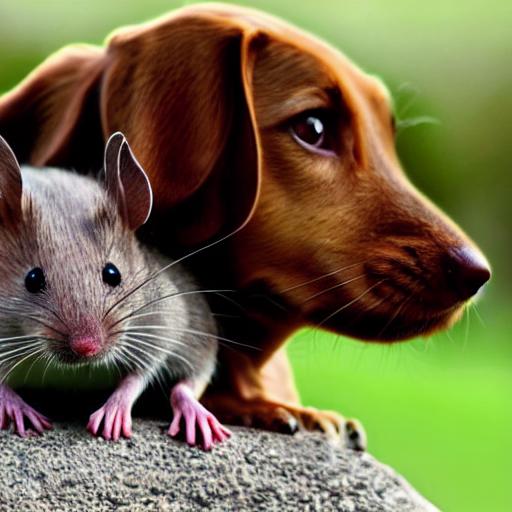} &
        \includegraphics[width=0.11\textwidth]{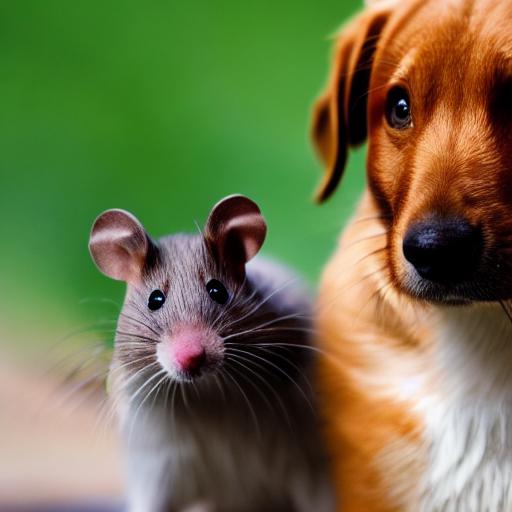} &
        \hspace{0.05cm}
        \includegraphics[width=0.11\textwidth]{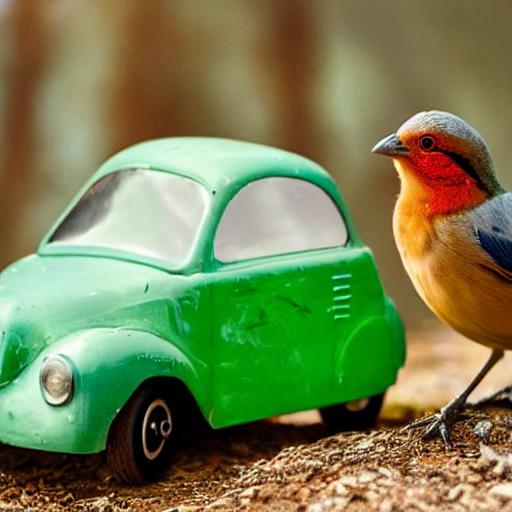} &
        \includegraphics[width=0.11\textwidth]{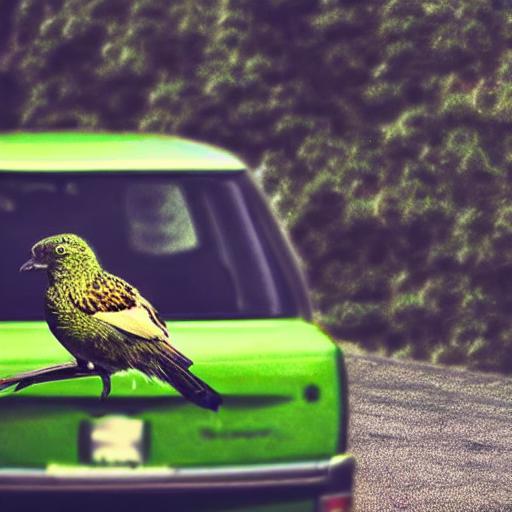} &
        \hspace{0.05cm}
        \includegraphics[width=0.11\textwidth]{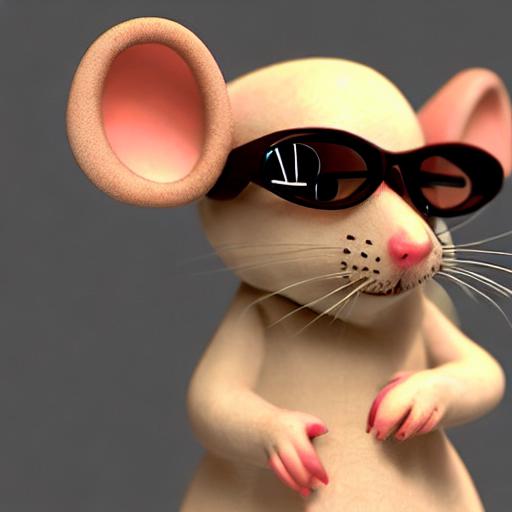} &
        \includegraphics[width=0.11\textwidth]{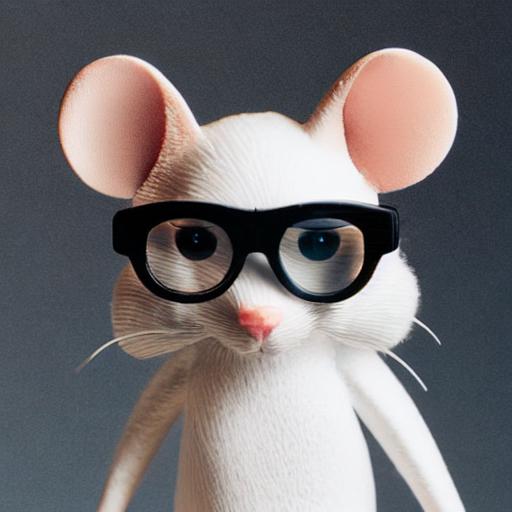} &
        \hspace{0.05cm}
        \includegraphics[width=0.11\textwidth]{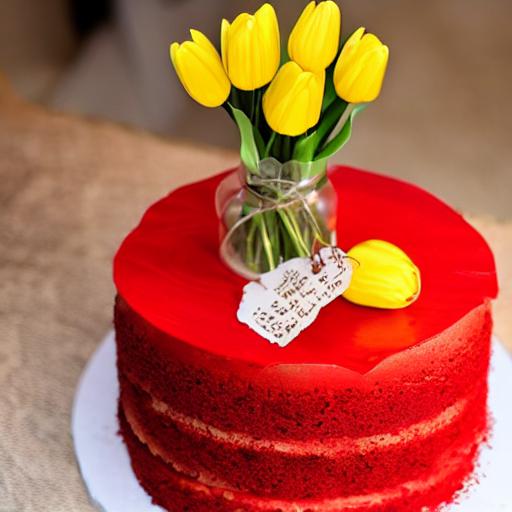} &
        \includegraphics[width=0.11\textwidth]{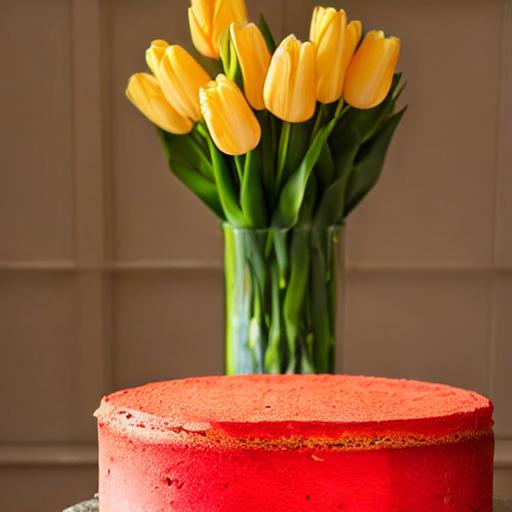} \\

        & 
        \includegraphics[width=0.11\textwidth]{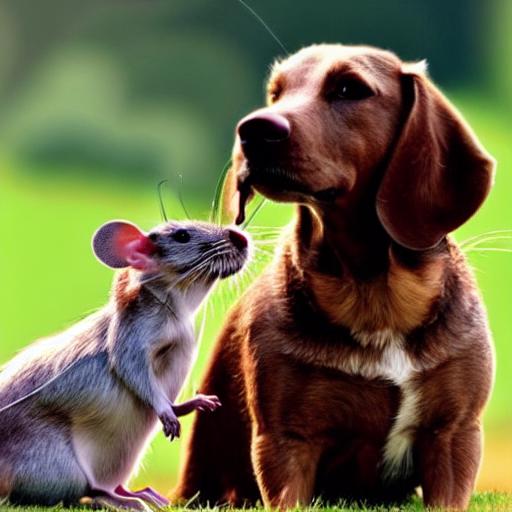} &
        \includegraphics[width=0.11\textwidth]{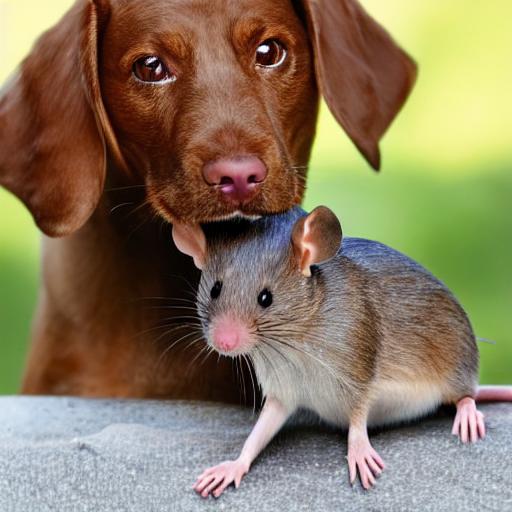} &
        \hspace{0.05cm}
        \includegraphics[width=0.11\textwidth]{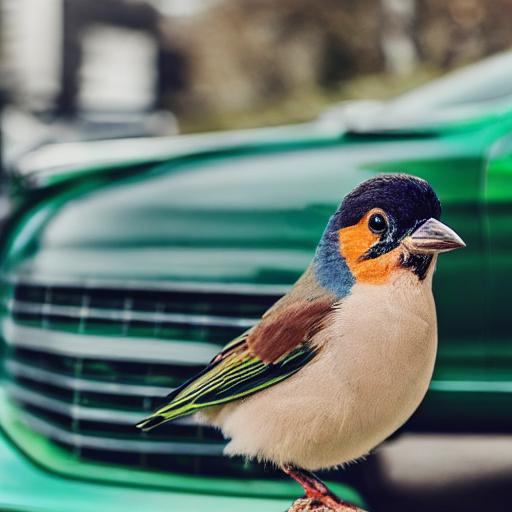} &
        \includegraphics[width=0.11\textwidth]{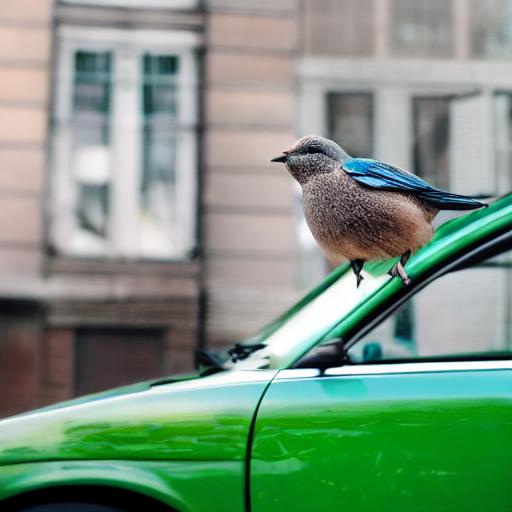} &
        \hspace{0.05cm}
        \includegraphics[width=0.11\textwidth]{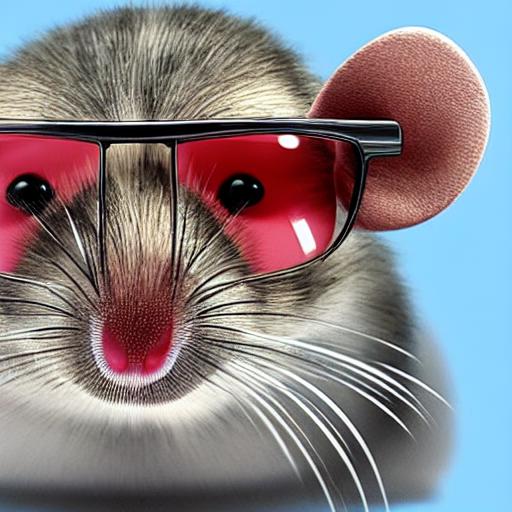} &
        \includegraphics[width=0.11\textwidth]{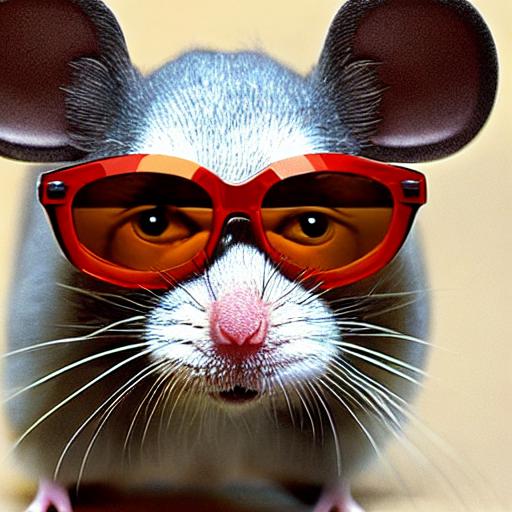} &
        \hspace{0.05cm}
        \includegraphics[width=0.11\textwidth]{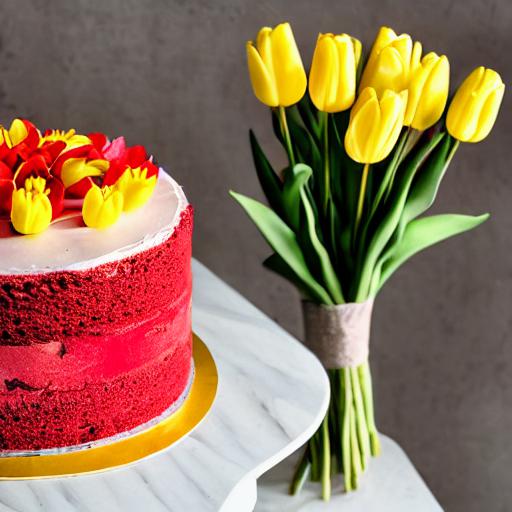} &
        \includegraphics[width=0.11\textwidth]{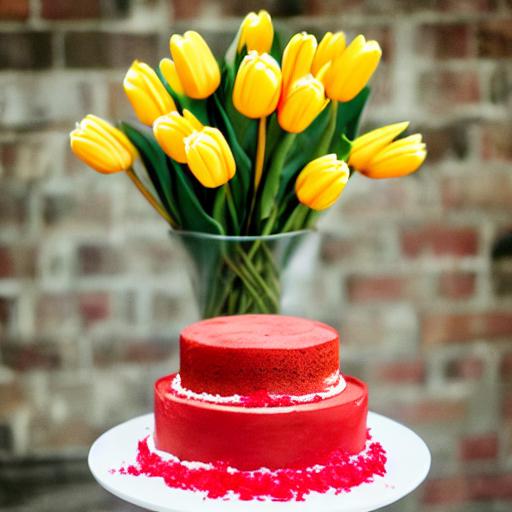}
        
    \end{tabular}
    
    }
    \vspace{-0.35cm}
    \caption{Additional Qualitative Comparison. For each prompt, we show four images generated by each of the four considered methods where we use the same set of seeds across all approaches. The subject tokens optimized by Attend-and-Excite are highlighted in \textcolor{blue}{blue}.}
    \label{fig:additional_results}

\end{figure*}
\begin{figure*}
    \centering
    \setlength{\tabcolsep}{0.5pt}
    \renewcommand{\arraystretch}{0.3}
    {\small
    \begin{tabular}{c c c @{\hspace{0.1cm}} c c @{\hspace{0.1cm}} c c @{\hspace{0.1cm}} c c }

        &
        \multicolumn{2}{c}{``A \textcolor{blue}{horse} and a \textcolor{blue}{monkey}''} &
        \multicolumn{2}{c}{``A \textcolor{blue}{bird} and a \textcolor{blue}{bear}''} &
        \multicolumn{2}{c}{``A \textcolor{blue}{rabbit} with a \textcolor{blue}{crown}''}
        &
        \multicolumn{2}{c}{\begin{tabular}{c} ``A purple \textcolor{blue}{chair} \\ \\ and a red \textcolor{blue}{bow}'' \\\\
        \end{tabular}} \\

        {\raisebox{0.3in}{
        \multirow{2}{*}{\rotatebox{90}{Stable Diffusion}}}} &
        \includegraphics[width=0.11\textwidth]{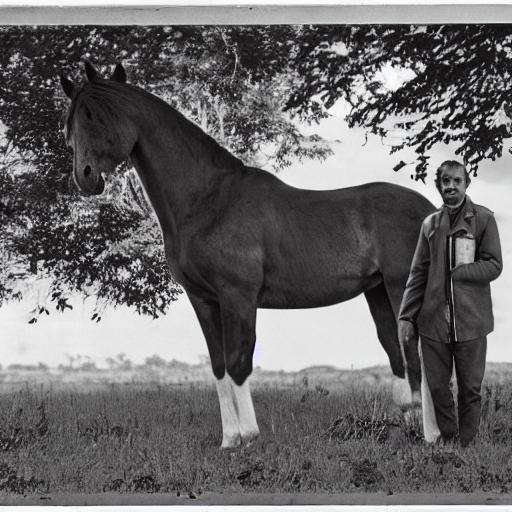} &
        \includegraphics[width=0.11\textwidth]{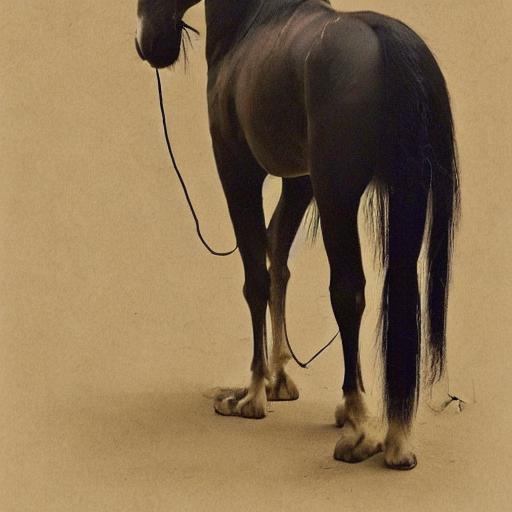} &
        \hspace{0.05cm}
        \includegraphics[width=0.11\textwidth]{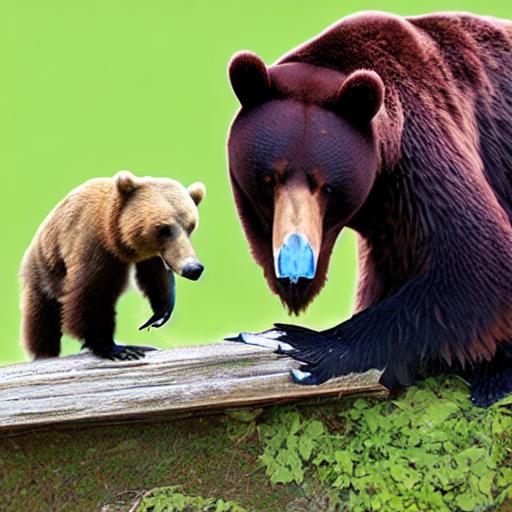} &
        \includegraphics[width=0.11\textwidth]{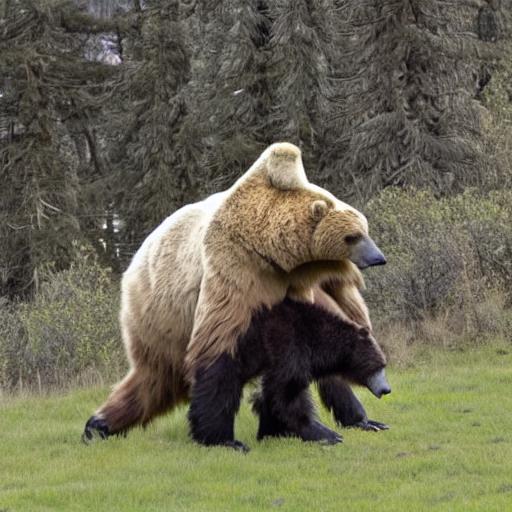} &
        \hspace{0.05cm}
        \includegraphics[width=0.11\textwidth]{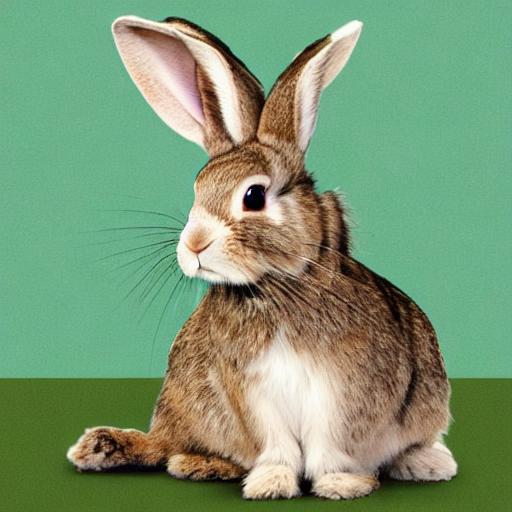} &
        \includegraphics[width=0.11\textwidth]{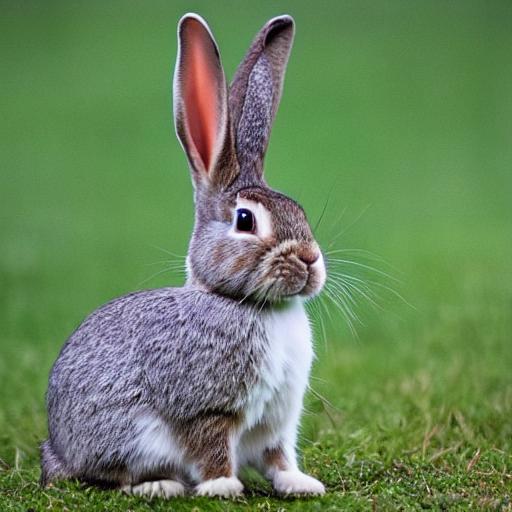} &
        \hspace{0.05cm}
        \includegraphics[width=0.11\textwidth]{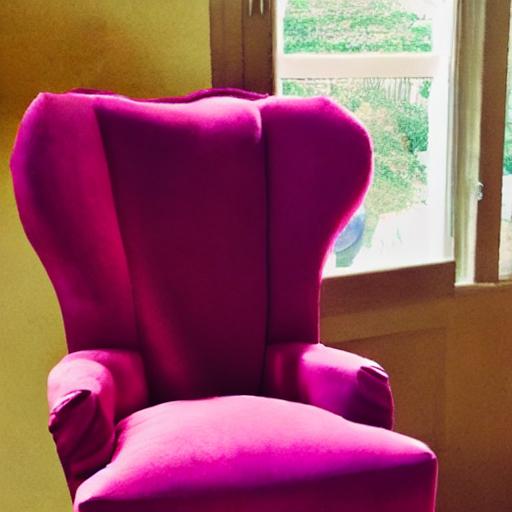} &
        \includegraphics[width=0.11\textwidth]{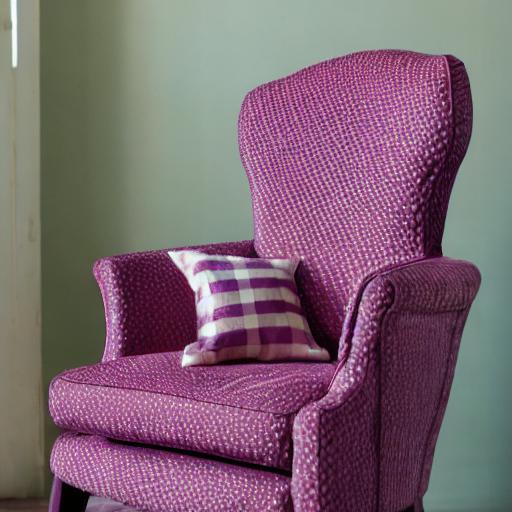} \\

        &
        \includegraphics[width=0.11\textwidth]{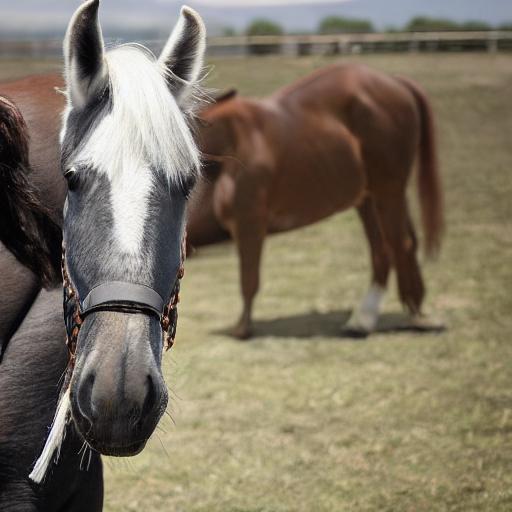} &
        \includegraphics[width=0.11\textwidth]{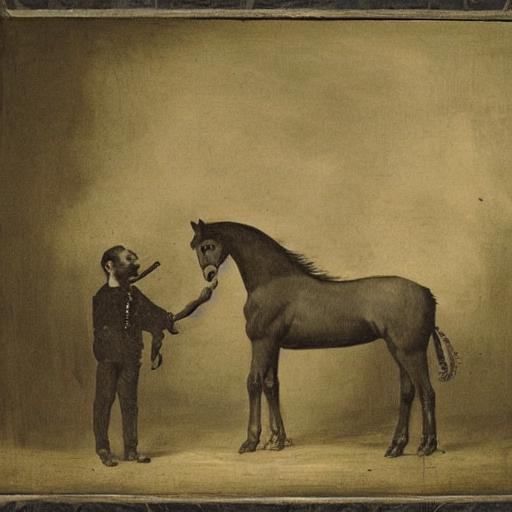} &
        \hspace{0.05cm}
        \includegraphics[width=0.11\textwidth]{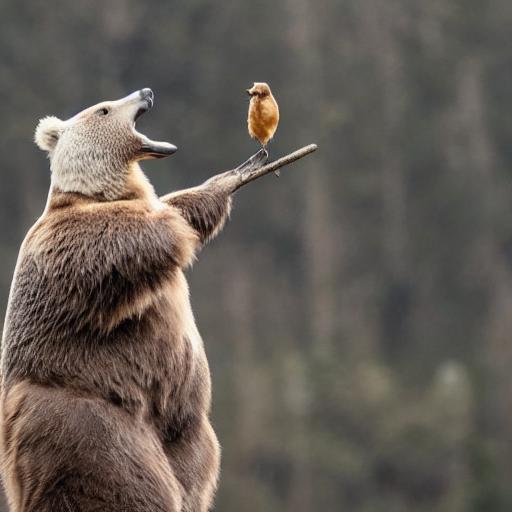} &
        \includegraphics[width=0.11\textwidth]{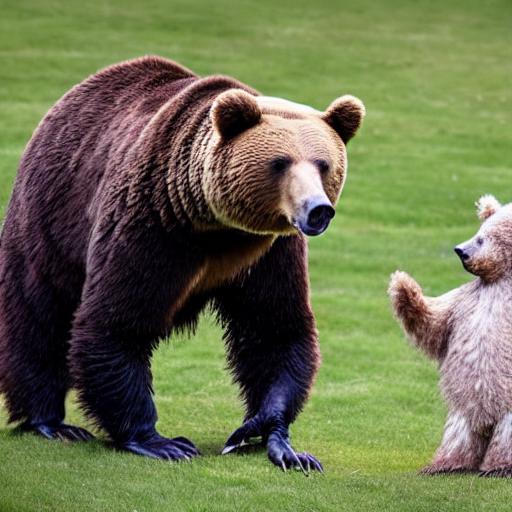} &
        \hspace{0.05cm}
        \includegraphics[width=0.11\textwidth]{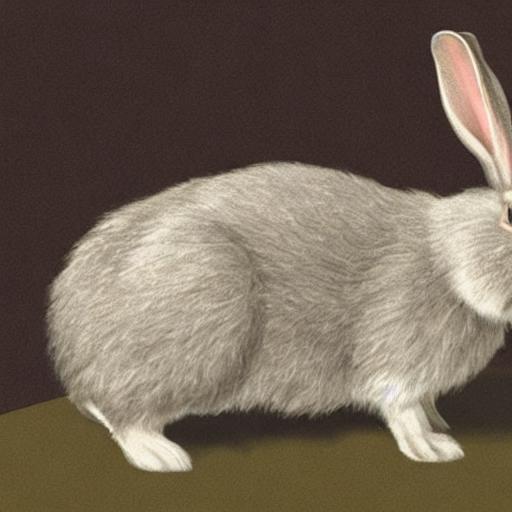} &
        \includegraphics[width=0.11\textwidth]{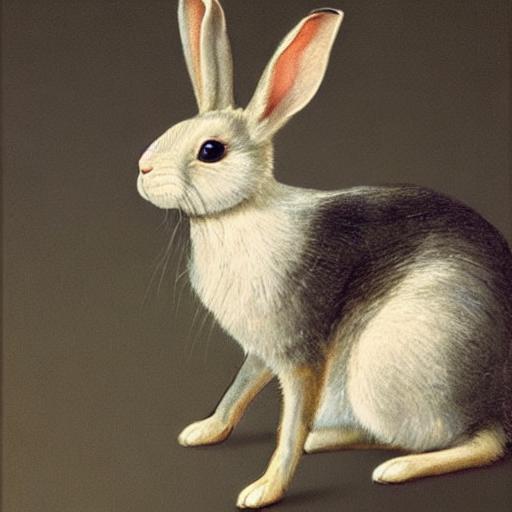} &
        \hspace{0.05cm}
        \includegraphics[width=0.11\textwidth]{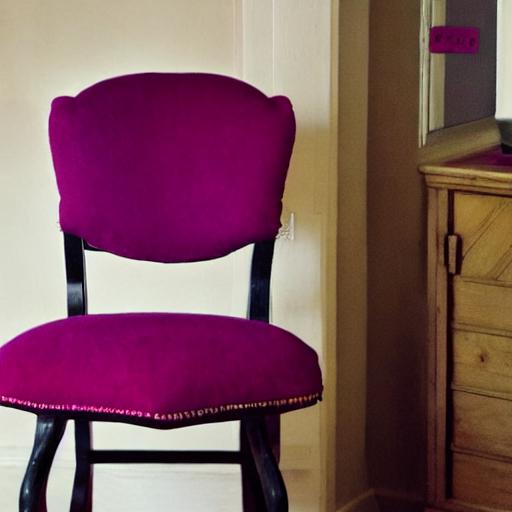} &
        \includegraphics[width=0.11\textwidth]{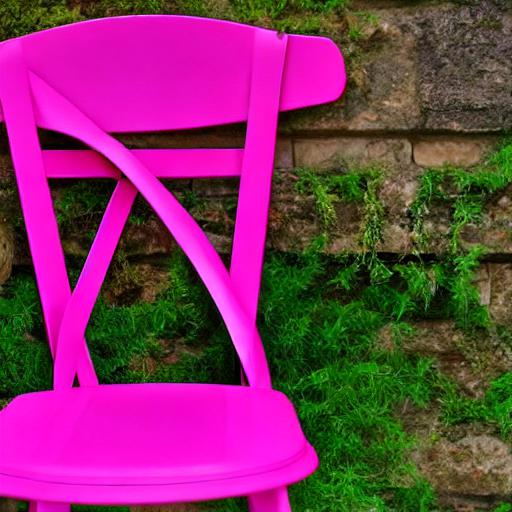}  \\ \\ \\

        {\raisebox{0.3in}{
        \multirow{2}{*}{\rotatebox{90}{Composable Diffusion}}}} &
        \includegraphics[width=0.11\textwidth]{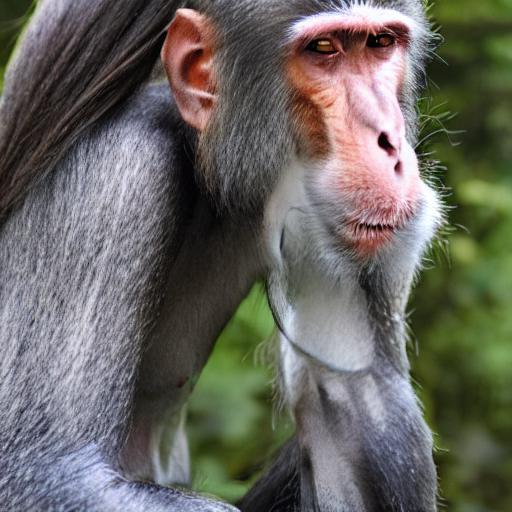} &
        \includegraphics[width=0.11\textwidth]{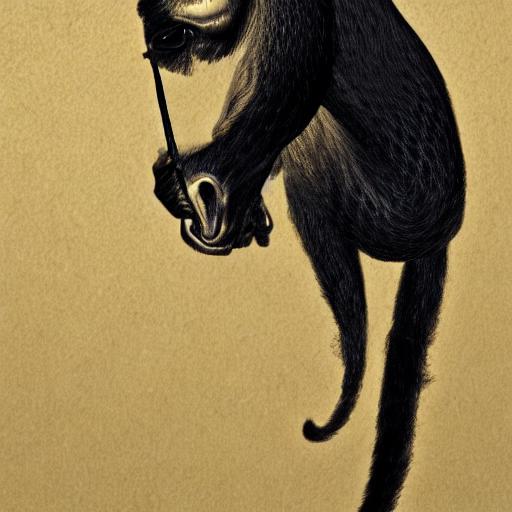} &
        \hspace{0.05cm}
        \includegraphics[width=0.11\textwidth]{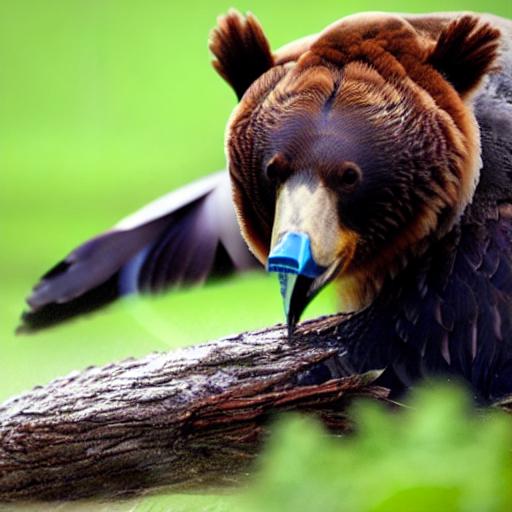} &
        \includegraphics[width=0.11\textwidth]{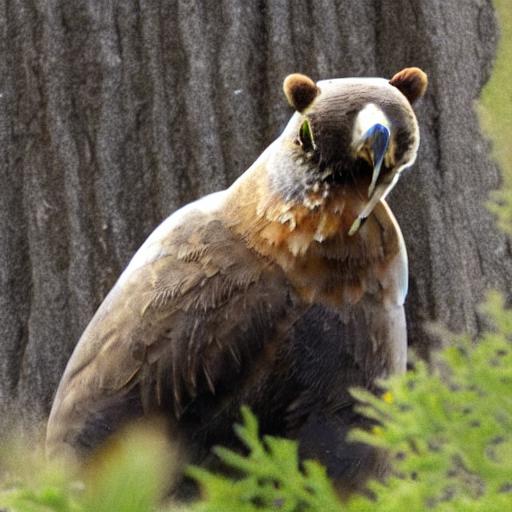} &
        \hspace{0.05cm}
        \includegraphics[width=0.11\textwidth]{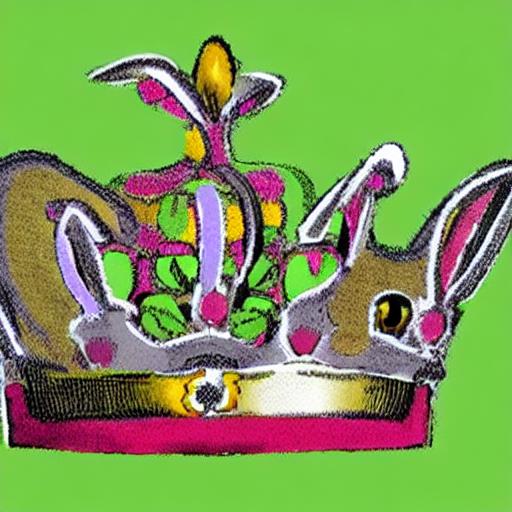} &
        \includegraphics[width=0.11\textwidth]{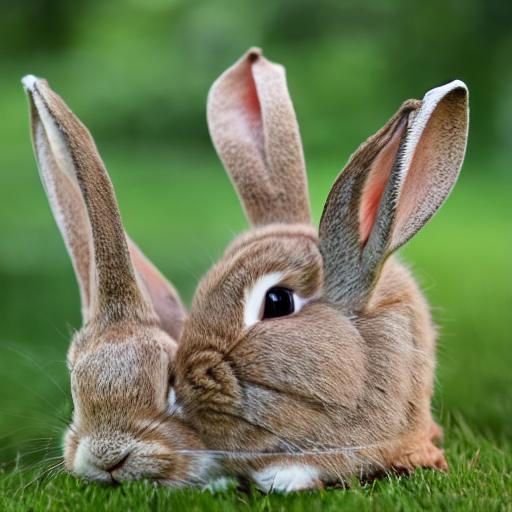} &
        \hspace{0.05cm}
        \includegraphics[width=0.11\textwidth]{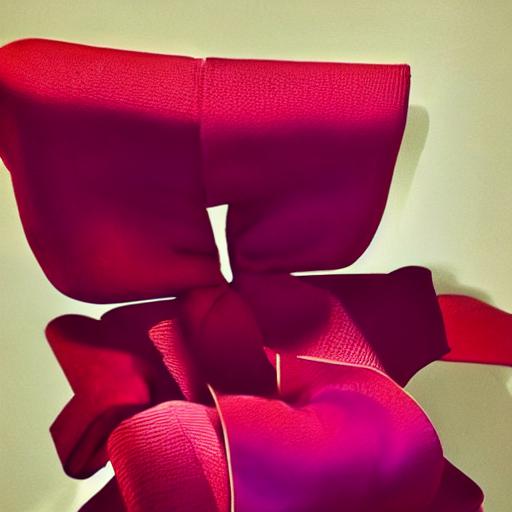} &
        \includegraphics[width=0.11\textwidth]{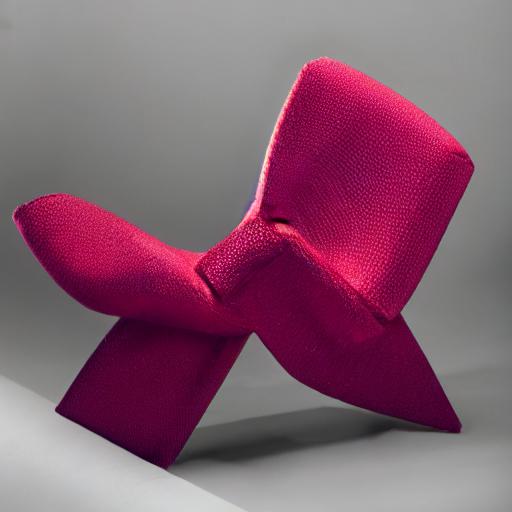} \\

        &
        \includegraphics[width=0.11\textwidth]{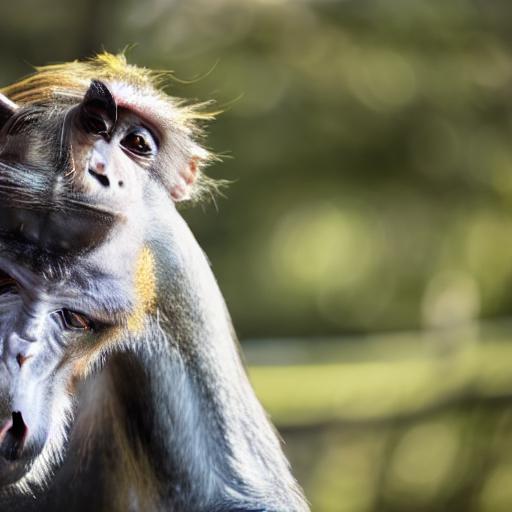} &
        \includegraphics[width=0.11\textwidth]{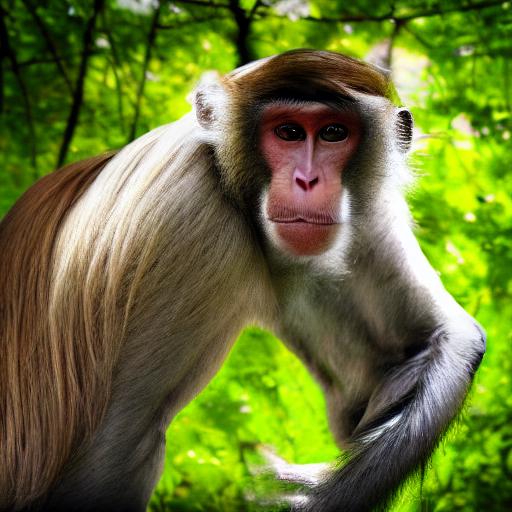} &
        \hspace{0.05cm}
        \includegraphics[width=0.11\textwidth]{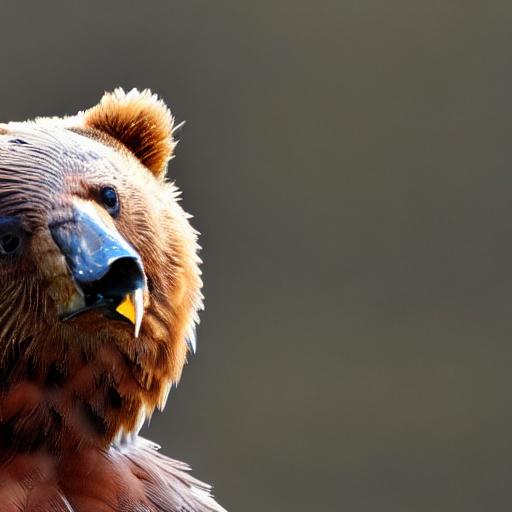} &
        \includegraphics[width=0.11\textwidth]{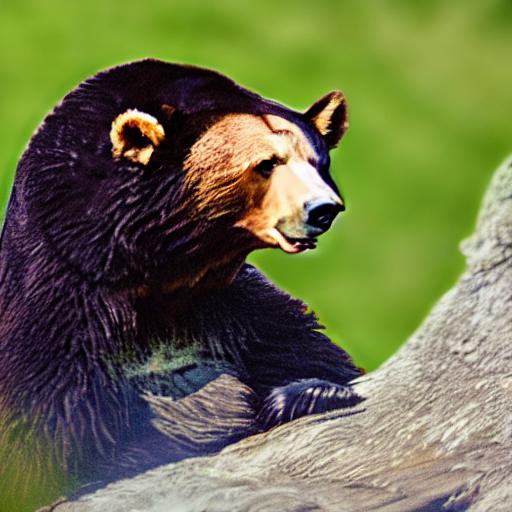} &
        \hspace{0.05cm}
        \includegraphics[width=0.11\textwidth]{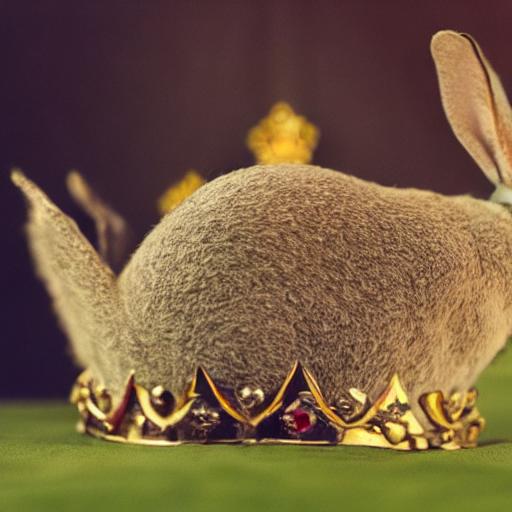} &
        \includegraphics[width=0.11\textwidth]{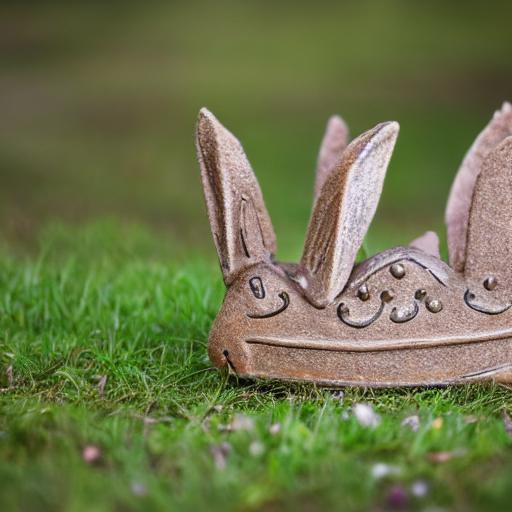} &
        \hspace{0.05cm}
        \includegraphics[width=0.11\textwidth]{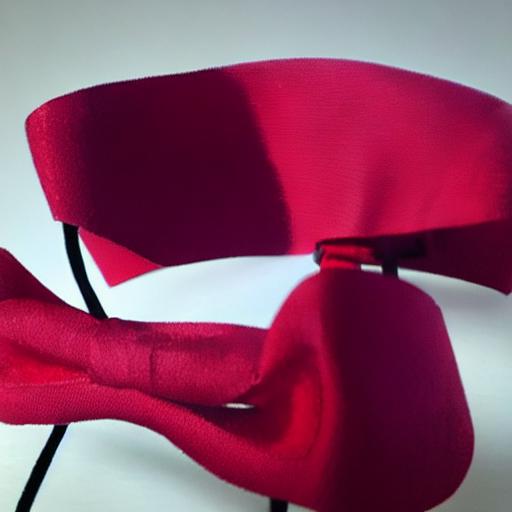} &
        \includegraphics[width=0.11\textwidth]{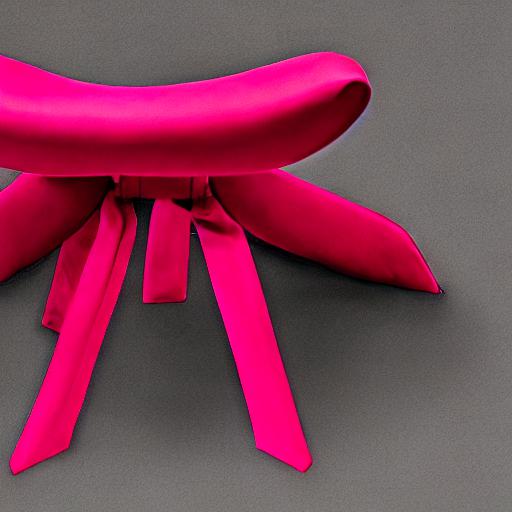}  \\ \\ \\

        {\raisebox{0.3in}{
        \multirow{2}{*}{\rotatebox{90}{Structure Diffusion}}}} &
        \includegraphics[width=0.11\textwidth]{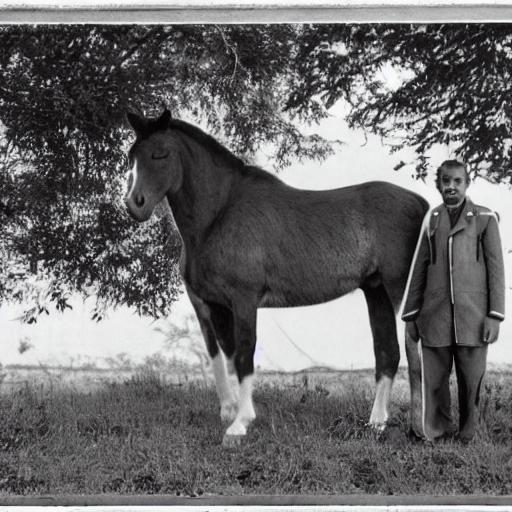} &
        \includegraphics[width=0.11\textwidth]{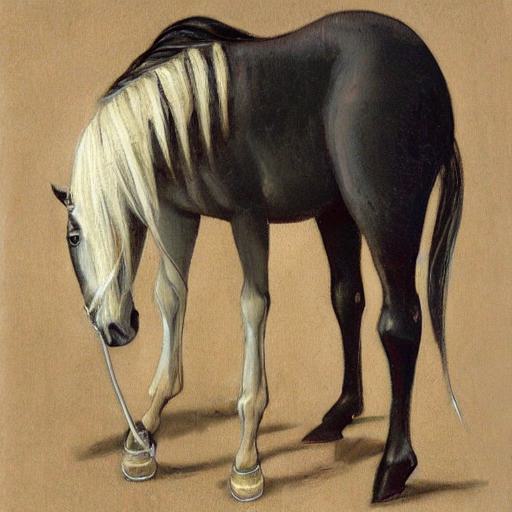} &
        \hspace{0.05cm}
        \includegraphics[width=0.11\textwidth]{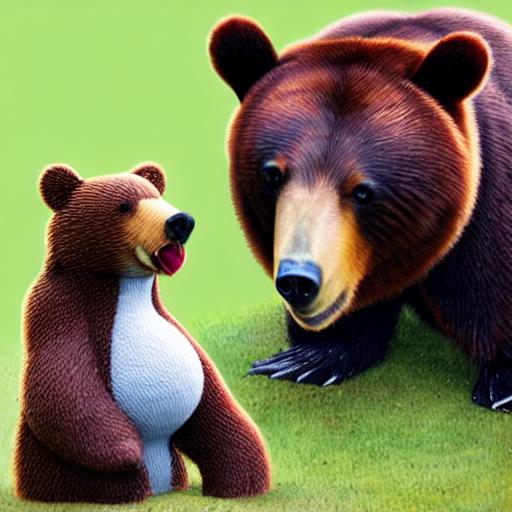} &
        \includegraphics[width=0.11\textwidth]{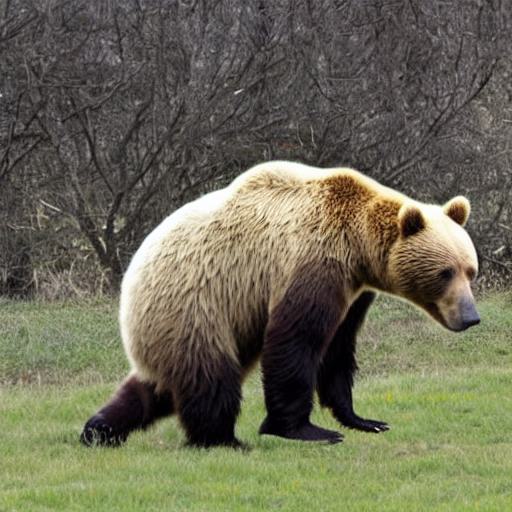} &
        \hspace{0.05cm}
        \includegraphics[width=0.11\textwidth]{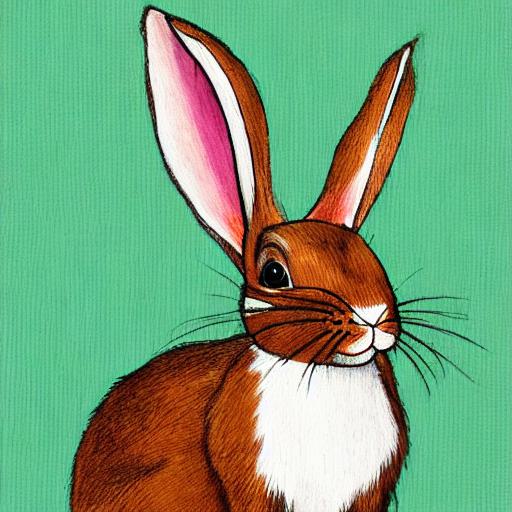} &
        \includegraphics[width=0.11\textwidth]{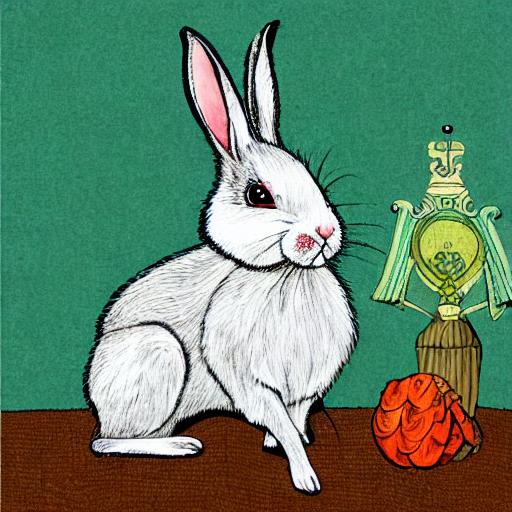} &
        \hspace{0.05cm}
        \includegraphics[width=0.11\textwidth]{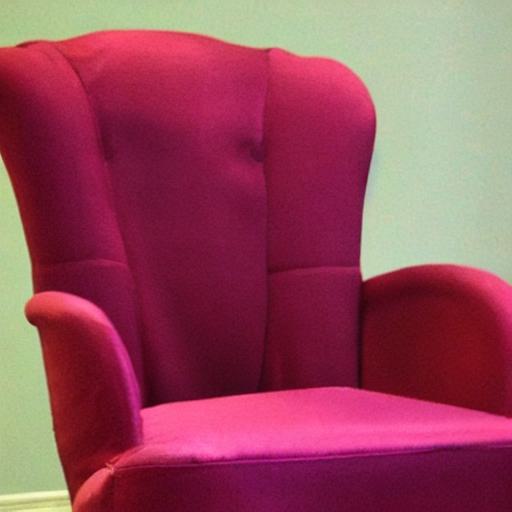} &
        \includegraphics[width=0.11\textwidth]{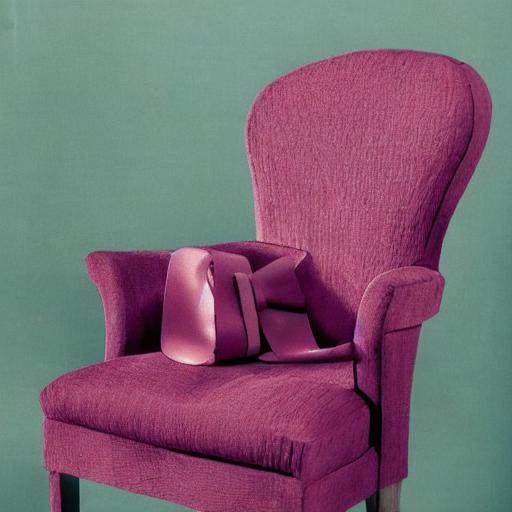} \\

        &
        \includegraphics[width=0.11\textwidth]{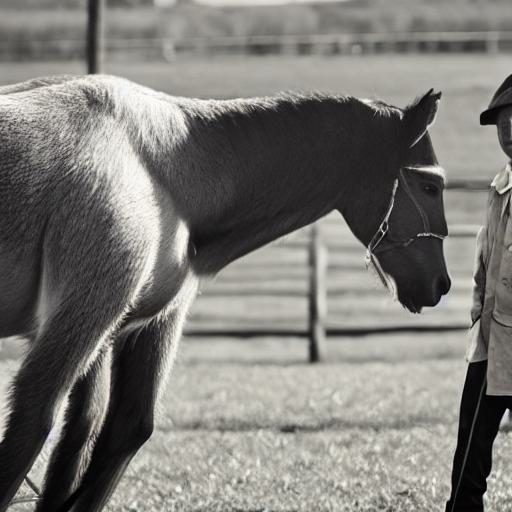} &
        \includegraphics[width=0.11\textwidth]{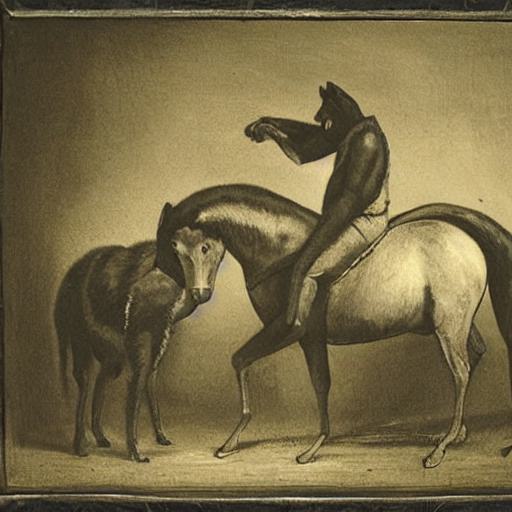} &
        \hspace{0.05cm}
        \includegraphics[width=0.11\textwidth]{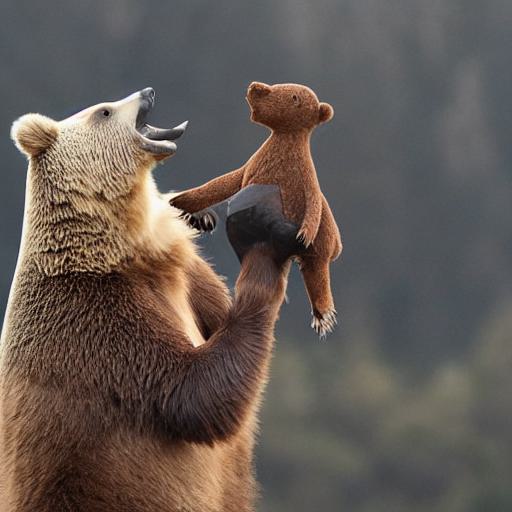} &
        \includegraphics[width=0.11\textwidth]{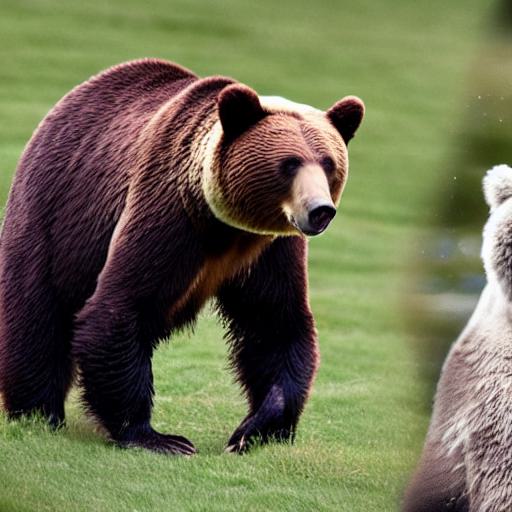} &
        \hspace{0.05cm}
        \includegraphics[width=0.11\textwidth]{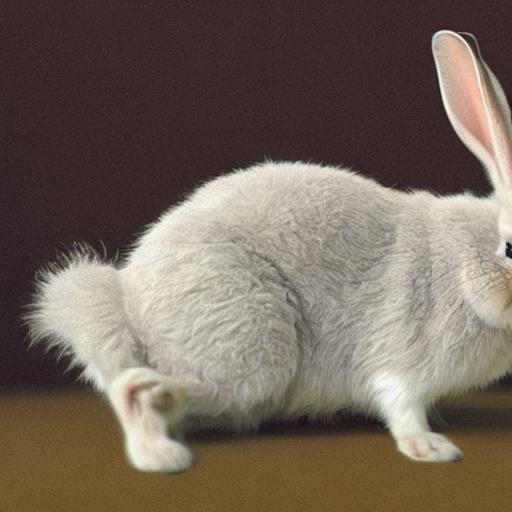} &
        \includegraphics[width=0.11\textwidth]{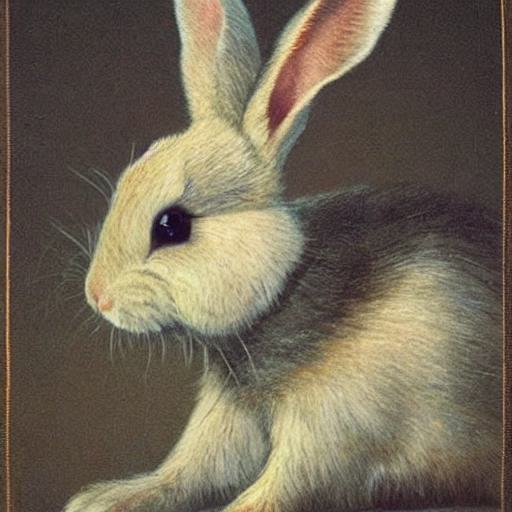} &
        \hspace{0.05cm}
        \includegraphics[width=0.11\textwidth]{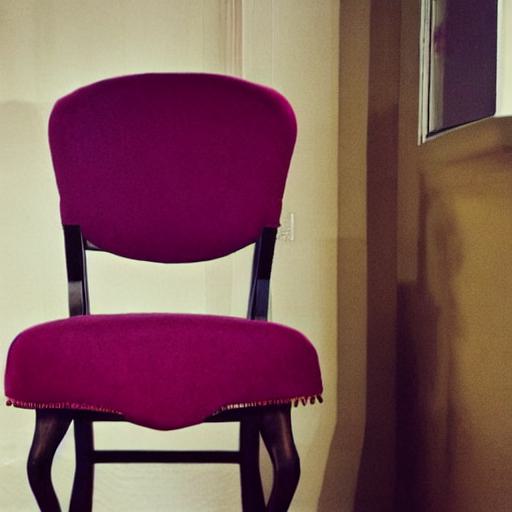} &
        \includegraphics[width=0.11\textwidth]{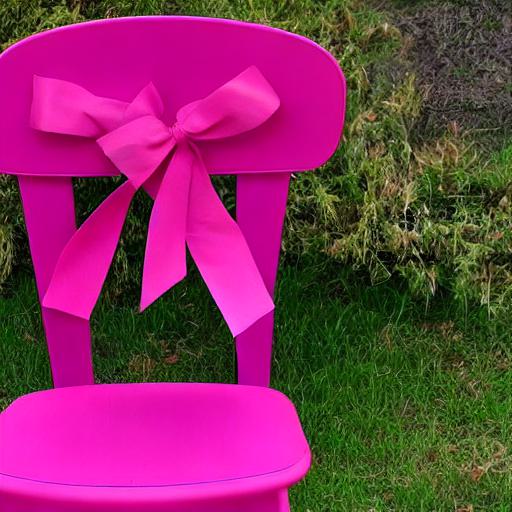}  \\ \\ \\

        {\raisebox{0.425in}{
        \multirow{2}{*}{\rotatebox{90}{\begin{tabular}{c} Stable Diffusion with \\ \textcolor{blue}{Attend-and-Excite} \\ \\ \end{tabular}}}}} &
        \includegraphics[width=0.11\textwidth]{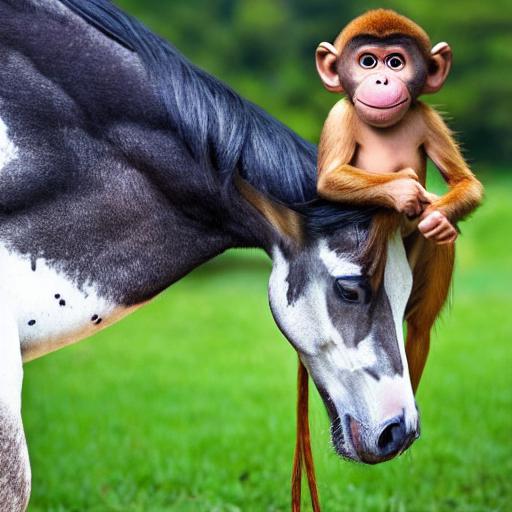} &
        \includegraphics[width=0.11\textwidth]{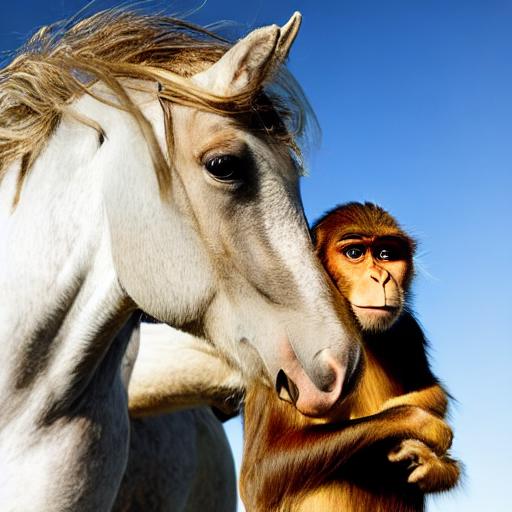} &
        \hspace{0.05cm}
        \includegraphics[width=0.11\textwidth]{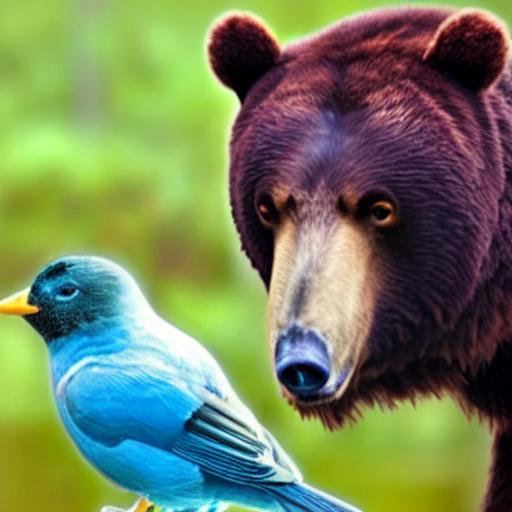} &
        \includegraphics[width=0.11\textwidth]{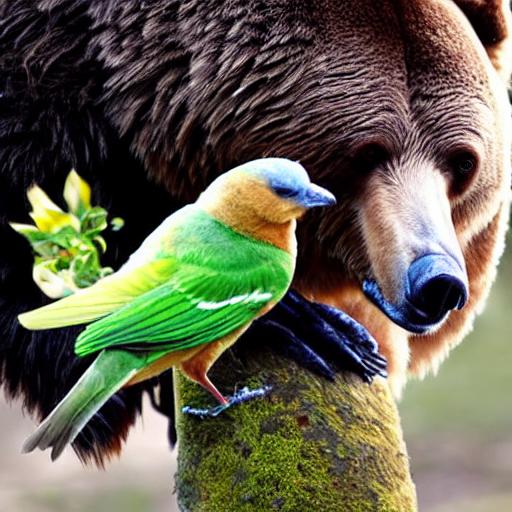} &
        \hspace{0.05cm}
         \includegraphics[width=0.11\textwidth]{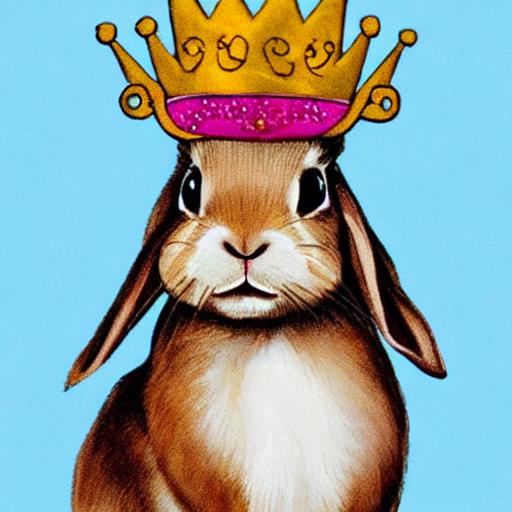} &
        \includegraphics[width=0.11\textwidth]{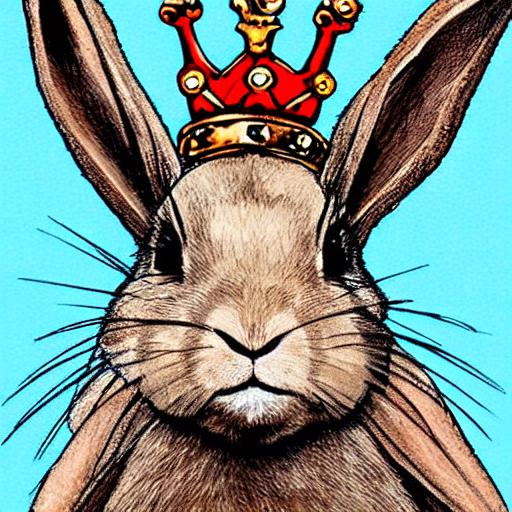} &
        \hspace{0.05cm}
        \includegraphics[width=0.11\textwidth]{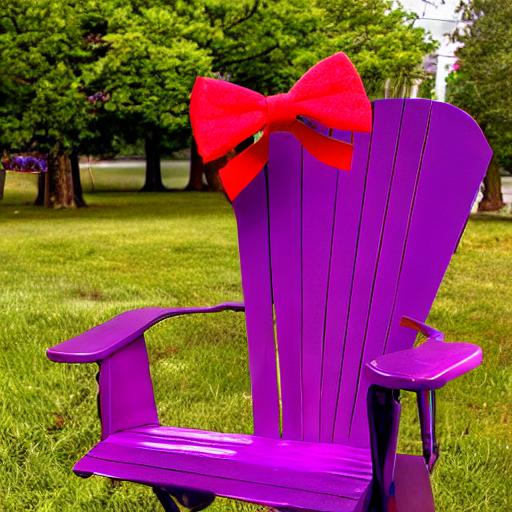} &
        \includegraphics[width=0.11\textwidth]{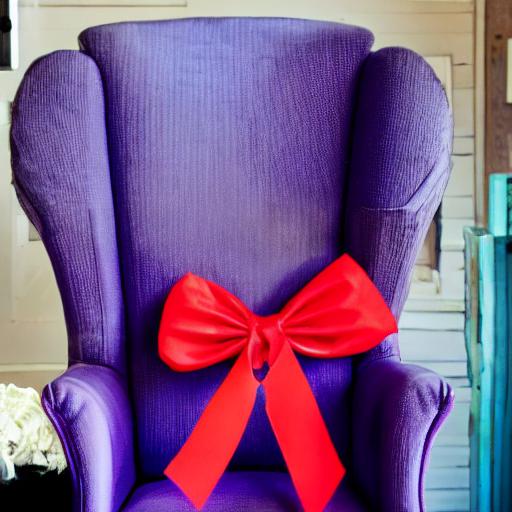}  \\

        &
        \includegraphics[width=0.11\textwidth]{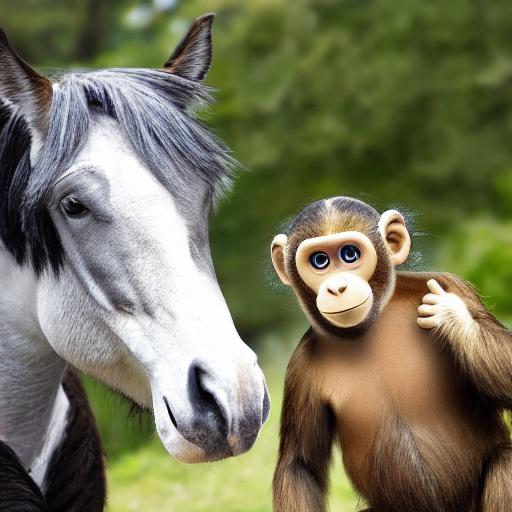} &
        \includegraphics[width=0.11\textwidth]{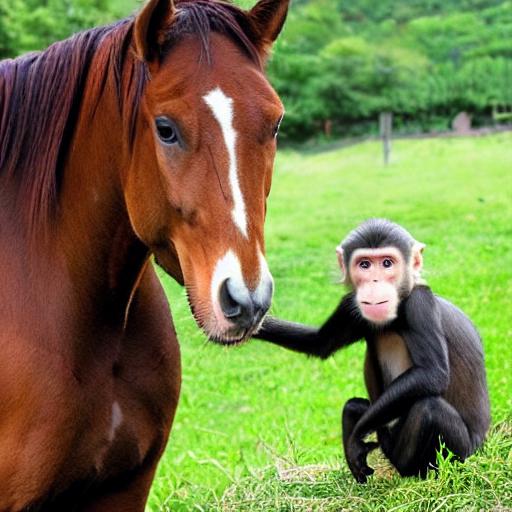} &
        \hspace{0.05cm}
        \includegraphics[width=0.11\textwidth]{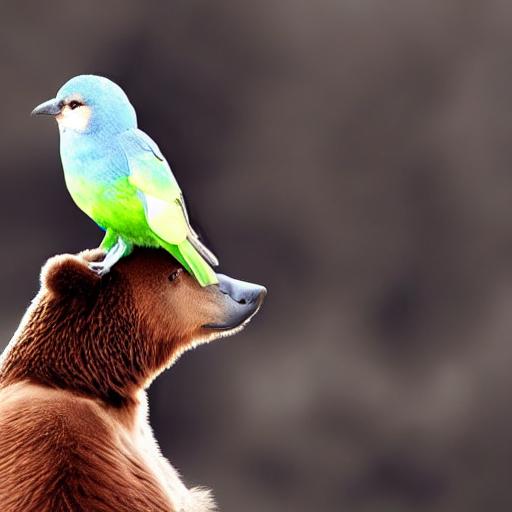} &
        \includegraphics[width=0.11\textwidth]{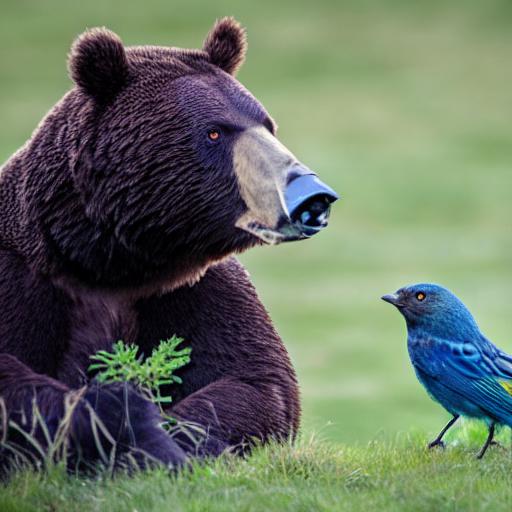} &
        \hspace{0.05cm}
        \includegraphics[width=0.11\textwidth]{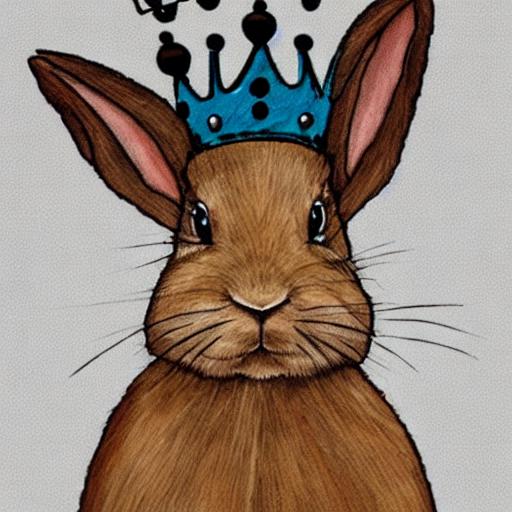} &
        \includegraphics[width=0.11\textwidth]{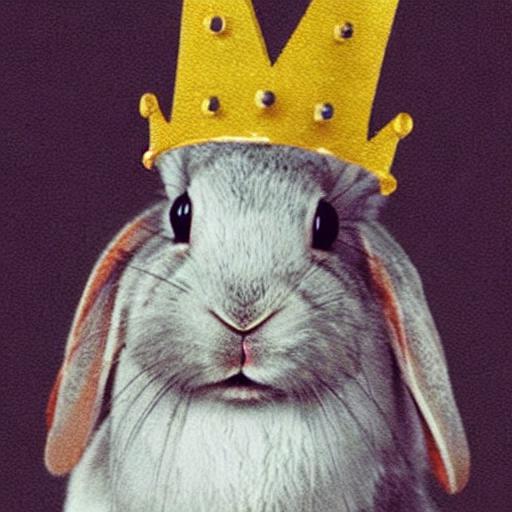} &
        \hspace{0.05cm}
        \includegraphics[width=0.11\textwidth]{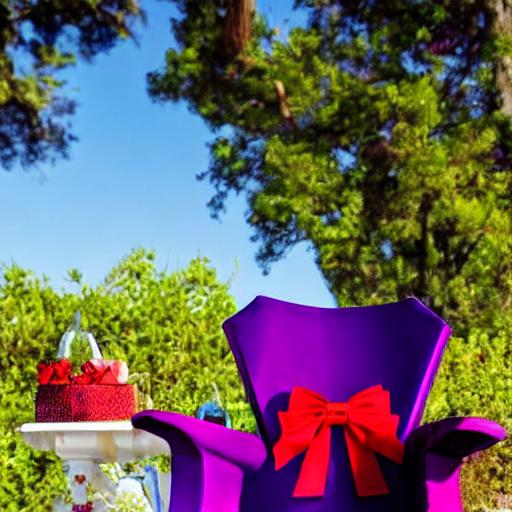} &
        \includegraphics[width=0.11\textwidth]{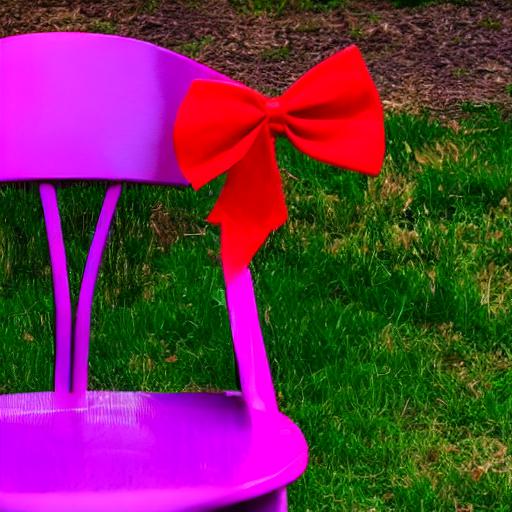}

    \end{tabular}

    }
    \vspace{-0.25cm}
    \caption{Qualitative Comparison. For each prompt, we show four images generated by each of the four considered methods where we use the same set of seeds across all approaches.
    The subject tokens optimized by Attend-and-Excite are highlighted in \textcolor{blue}{blue}.
    }
    \label{fig:additional_results_supp}
\end{figure*}

\begin{figure*}
    \centering
    \setlength{\tabcolsep}{0.5pt}
    \renewcommand{\arraystretch}{0.3}
    {\small
    \begin{tabular}{c c c @{\hspace{0.1cm}} c c @{\hspace{0.1cm}} c c @{\hspace{0.1cm}} c c }

        &
        \multicolumn{2}{c}{``A \textcolor{blue}{cat} and a \textcolor{blue}{frog}''} &
        \multicolumn{2}{c}{``A \textcolor{blue}{lion} with \textcolor{blue}{glasses}''} &
        \multicolumn{2}{c}{``A \textcolor{blue}{mouse} and a red \textcolor{blue}{car}''}
        &
        \multicolumn{2}{c}{\begin{tabular}{c} ``A green \textcolor{blue}{backpack} \\ \\ and a brown \textcolor{blue}{suitcase}'' \\\\
        \end{tabular}} \\

        {\raisebox{0.3in}{
        \multirow{2}{*}{\rotatebox{90}{Stable Diffusion}}}} &
        \includegraphics[width=0.11\textwidth]{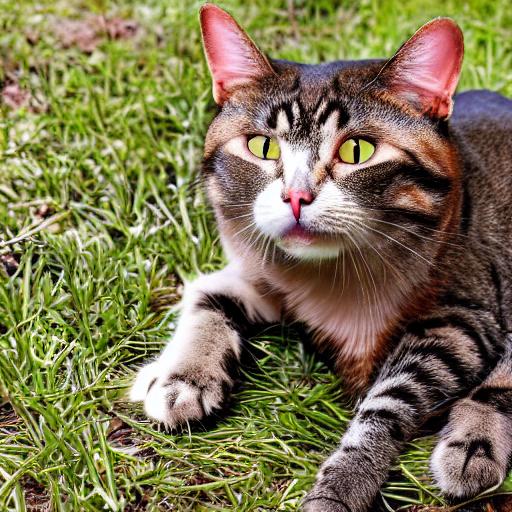} &
        \includegraphics[width=0.11\textwidth]{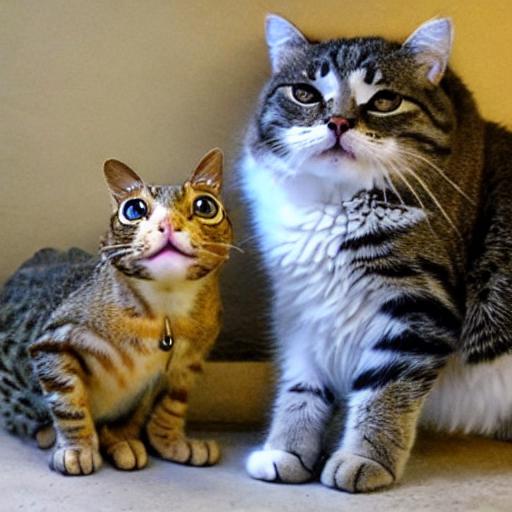} &
        \hspace{0.05cm}
        \includegraphics[width=0.11\textwidth]{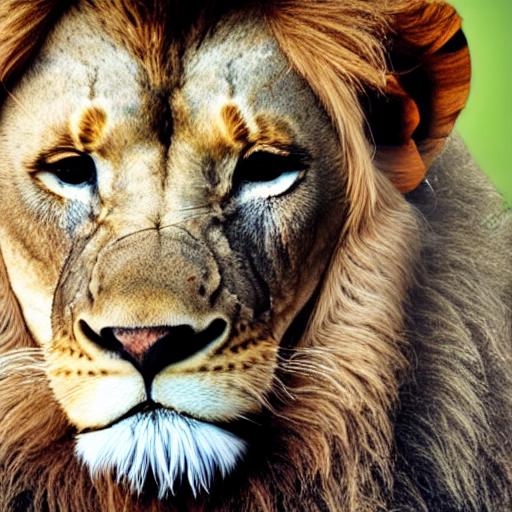} &
        \includegraphics[width=0.11\textwidth]{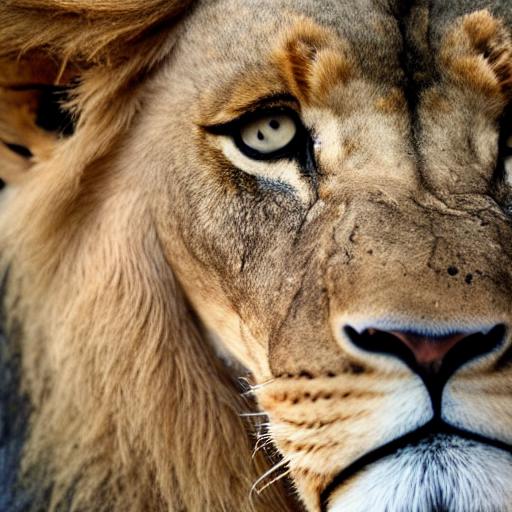} &
        \hspace{0.05cm}
        \includegraphics[width=0.11\textwidth]{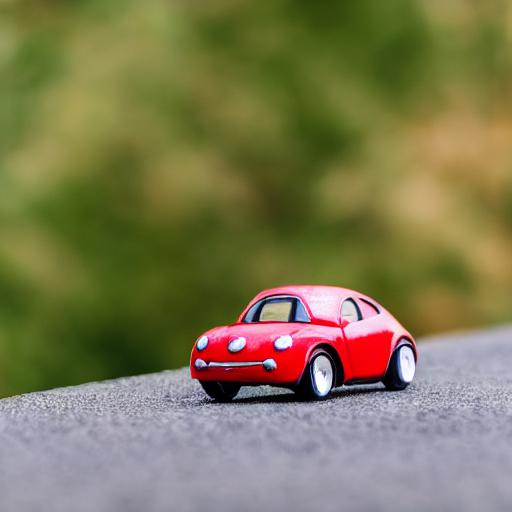} &
        \includegraphics[width=0.11\textwidth]{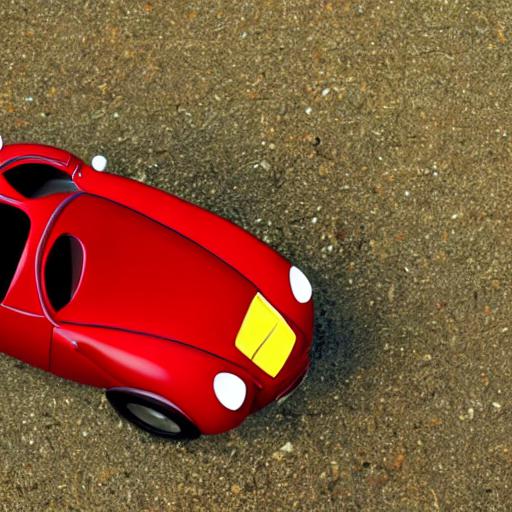} &
        \hspace{0.05cm}
        \includegraphics[width=0.11\textwidth]{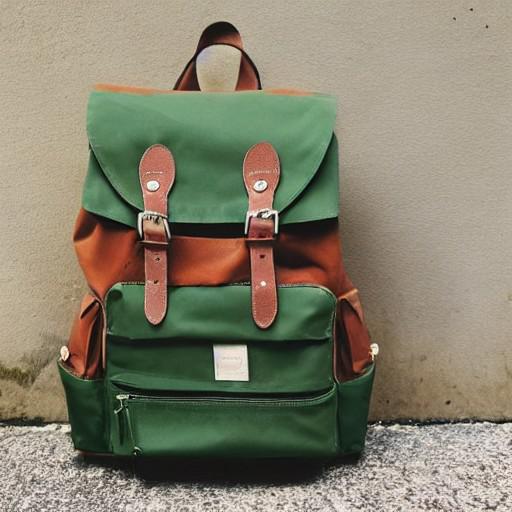} &
        \includegraphics[width=0.11\textwidth]{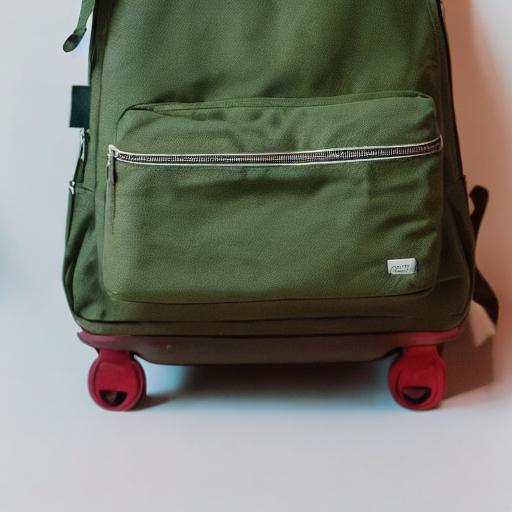} \\

        &
        \includegraphics[width=0.11\textwidth]{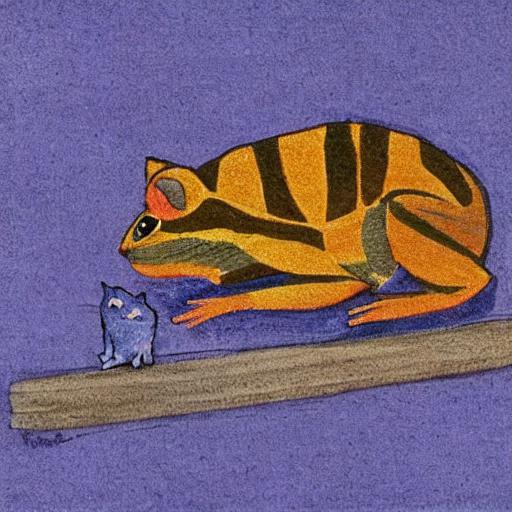} &
        \includegraphics[width=0.11\textwidth]{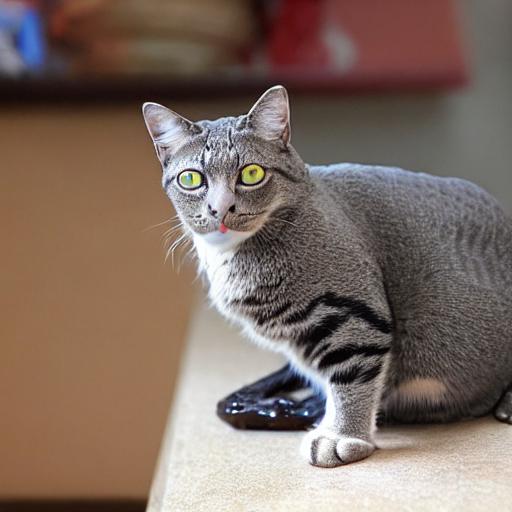} &
        \hspace{0.05cm}
        \includegraphics[width=0.11\textwidth]{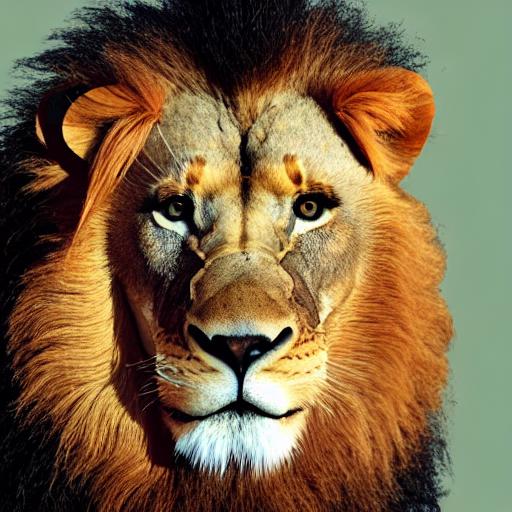} &
        \includegraphics[width=0.11\textwidth]{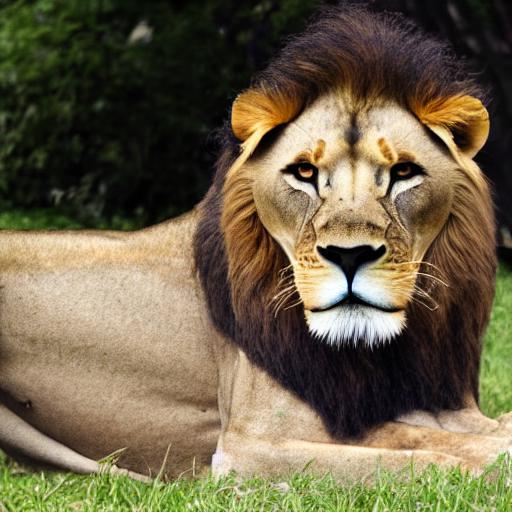} &
        \hspace{0.05cm}
        \includegraphics[width=0.11\textwidth]{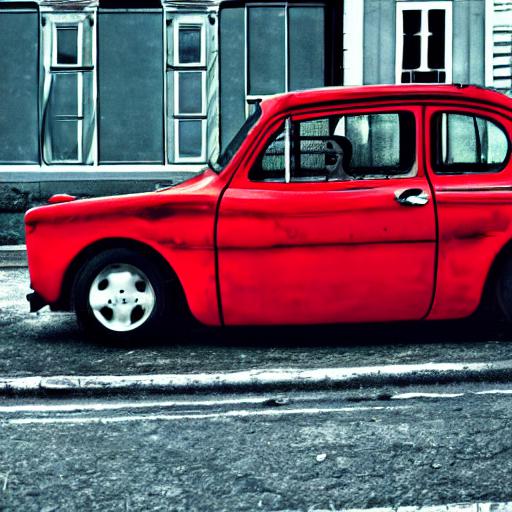} &
        \includegraphics[width=0.11\textwidth]{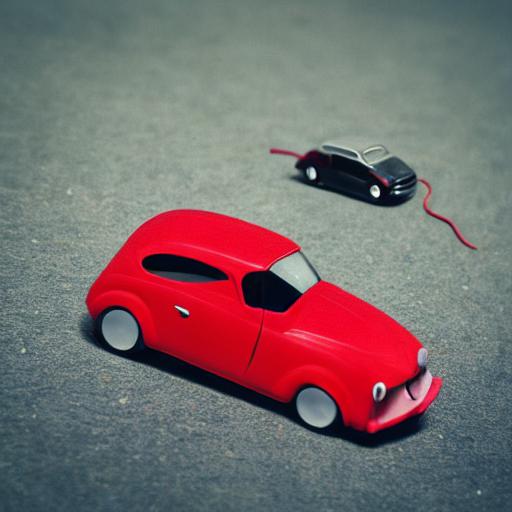} &
        \hspace{0.05cm}
        \includegraphics[width=0.11\textwidth]{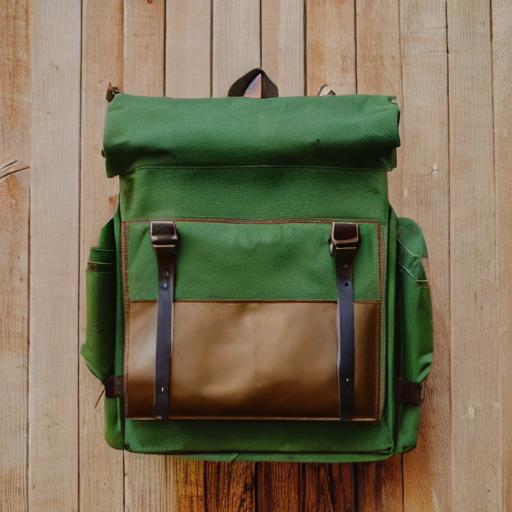} &
        \includegraphics[width=0.11\textwidth]{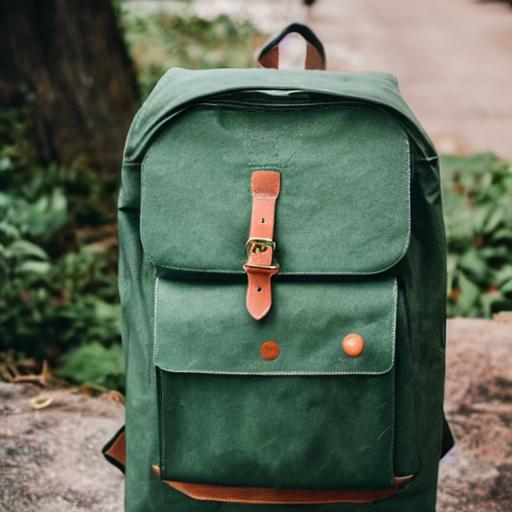}  \\ \\ \\

        {\raisebox{0.3in}{
        \multirow{2}{*}{\rotatebox{90}{Composable Diffusion}}}} &
        \includegraphics[width=0.11\textwidth]{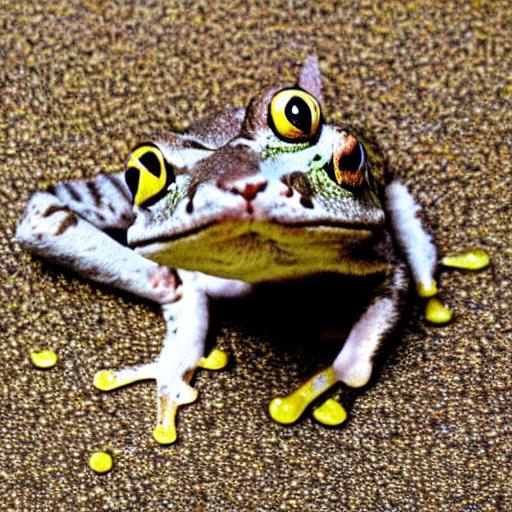} &
        \includegraphics[width=0.11\textwidth]{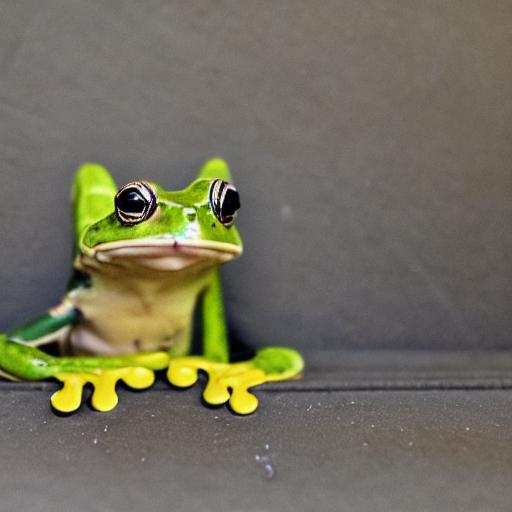} &
        \hspace{0.05cm}
        \includegraphics[width=0.11\textwidth]{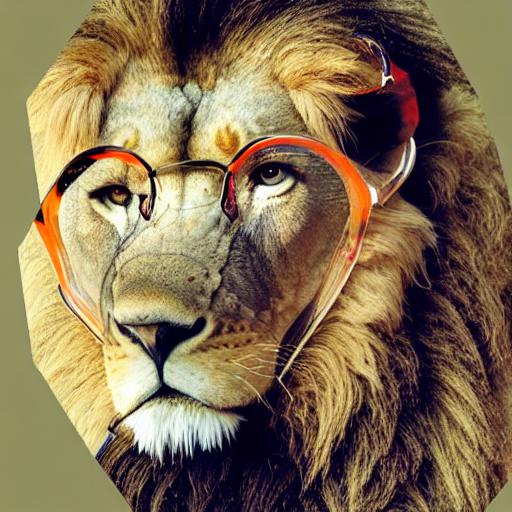} &
        \includegraphics[width=0.11\textwidth]{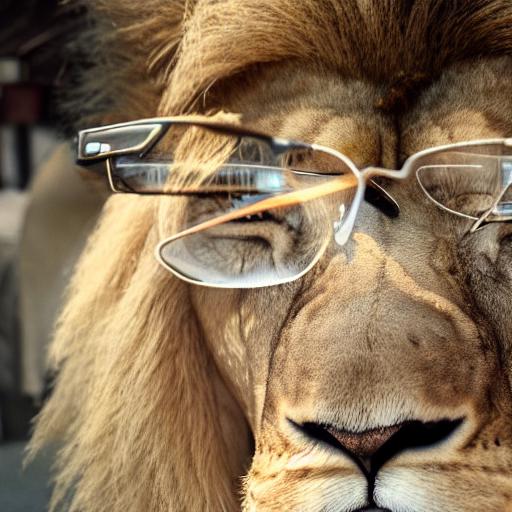} &
        \hspace{0.05cm}
        \includegraphics[width=0.11\textwidth]{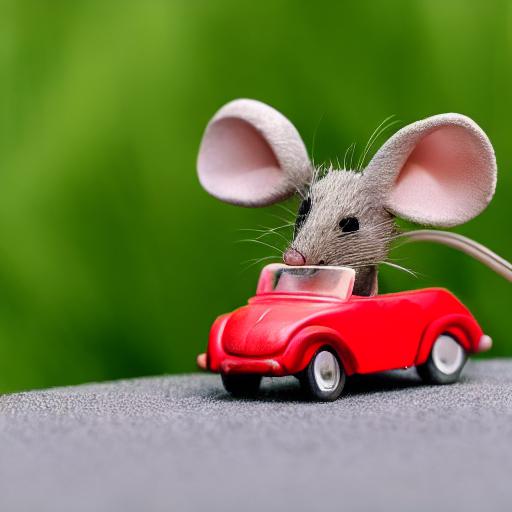} &
        \includegraphics[width=0.11\textwidth]{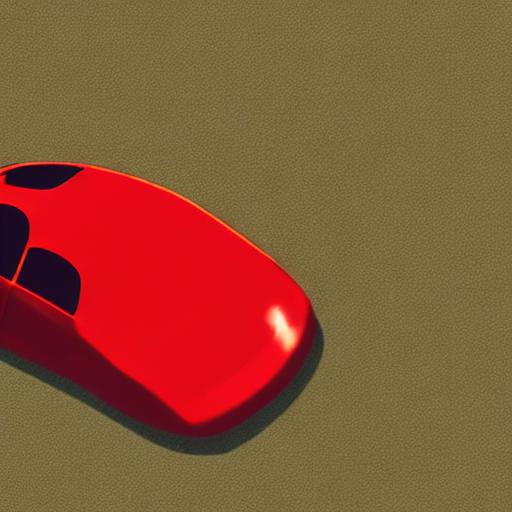} &
        \hspace{0.05cm}
        \includegraphics[width=0.11\textwidth]{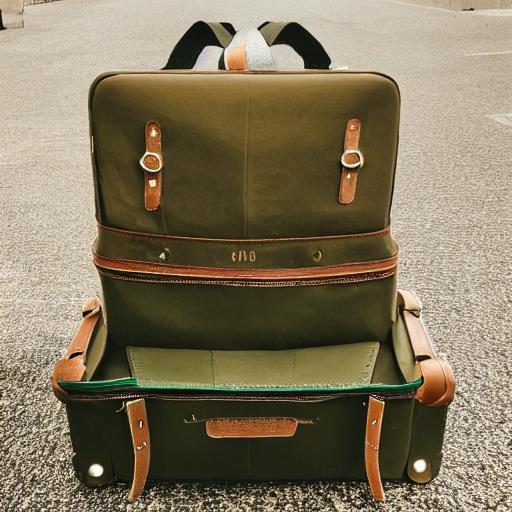} &
        \includegraphics[width=0.11\textwidth]{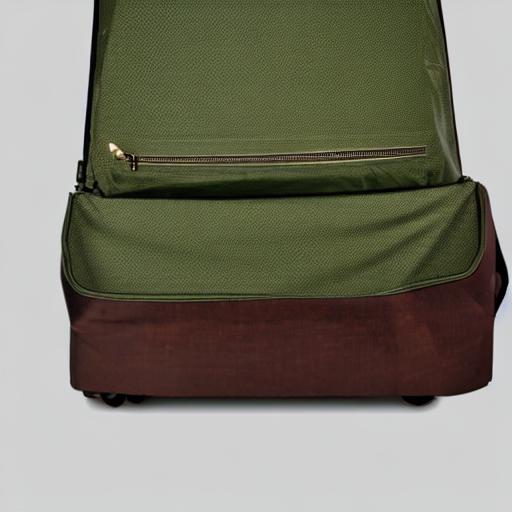} \\

        &
        \includegraphics[width=0.11\textwidth]{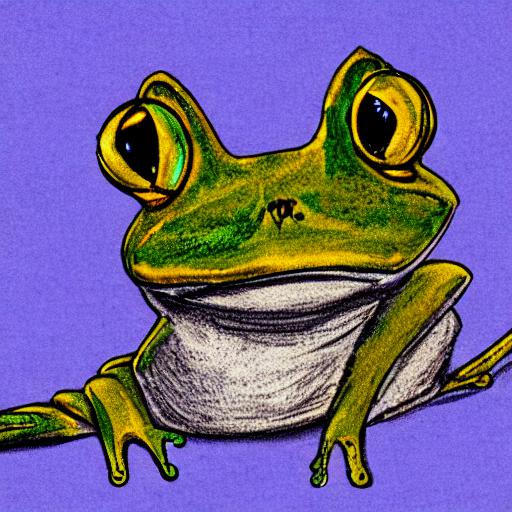} &
        \includegraphics[width=0.11\textwidth]{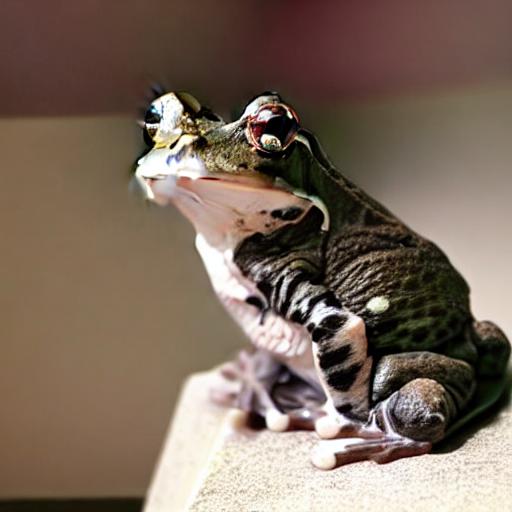} &
        \hspace{0.05cm}
        \includegraphics[width=0.11\textwidth]{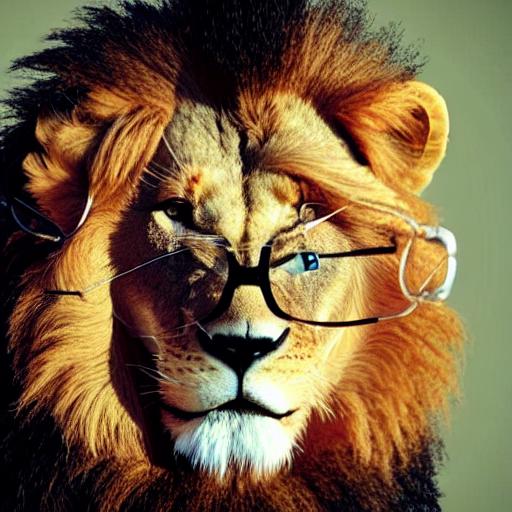} &
        \includegraphics[width=0.11\textwidth]{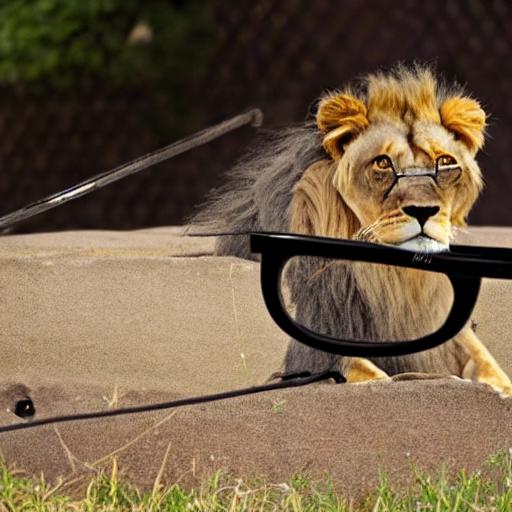} &
        \hspace{0.05cm}
        \includegraphics[width=0.11\textwidth]{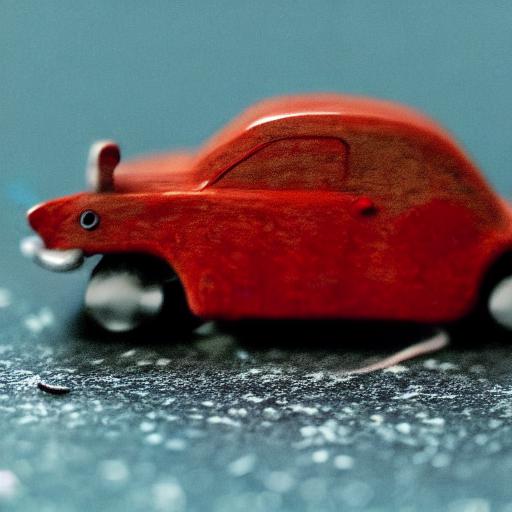} &
        \includegraphics[width=0.11\textwidth]{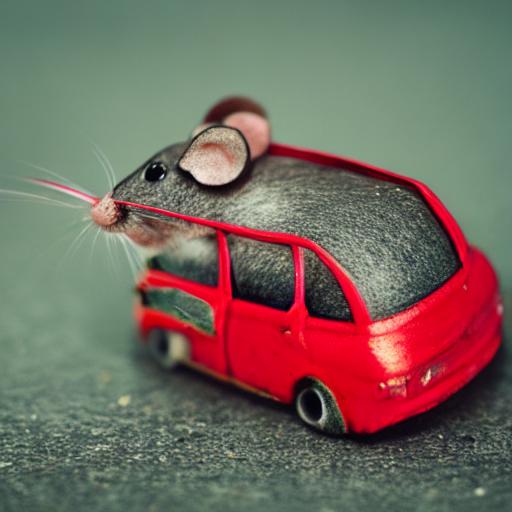} &
        \hspace{0.05cm}
        \includegraphics[width=0.11\textwidth]{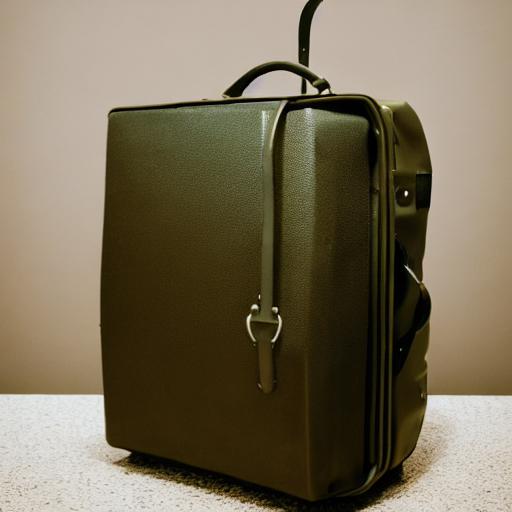} &
        \includegraphics[width=0.11\textwidth]{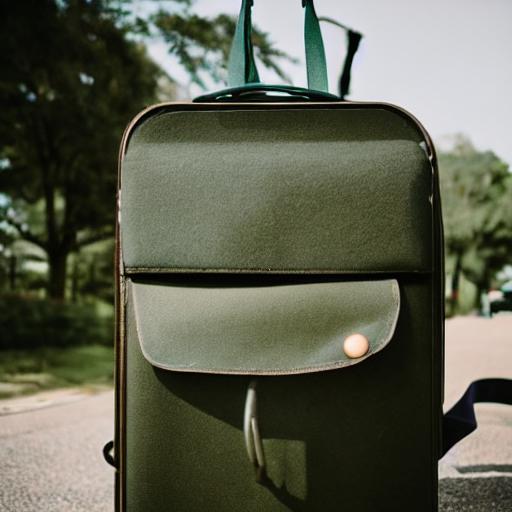}  \\ \\ \\

        {\raisebox{0.3in}{
        \multirow{2}{*}{\rotatebox{90}{StructureDiffusion}}}} &
        \includegraphics[width=0.11\textwidth]{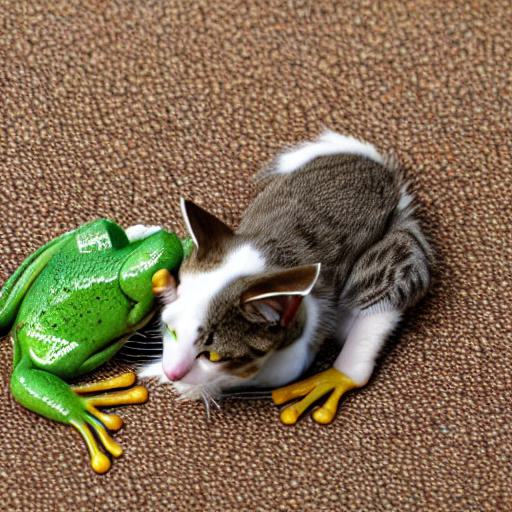} &
        \includegraphics[width=0.11\textwidth]{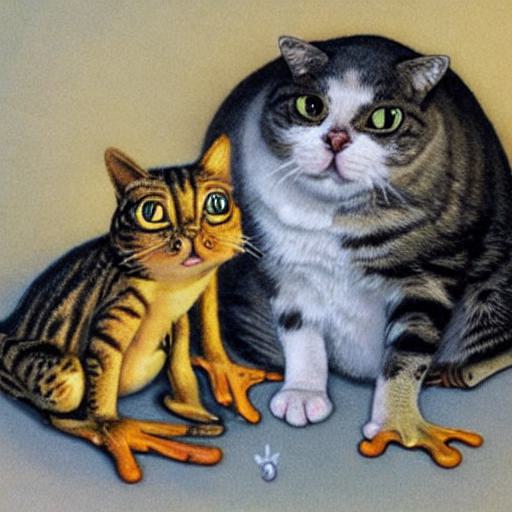} &
        \hspace{0.05cm}
        \includegraphics[width=0.11\textwidth]{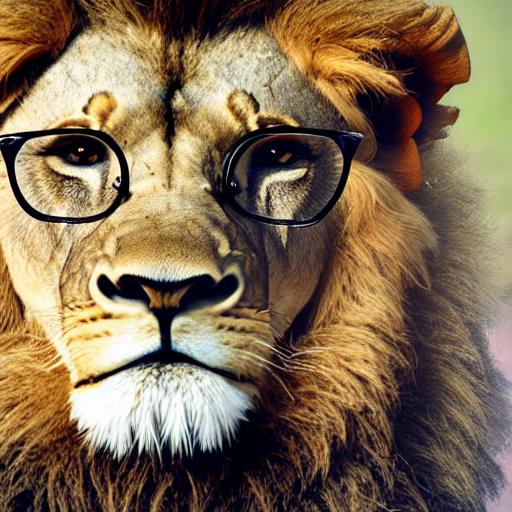} &
        \includegraphics[width=0.11\textwidth]{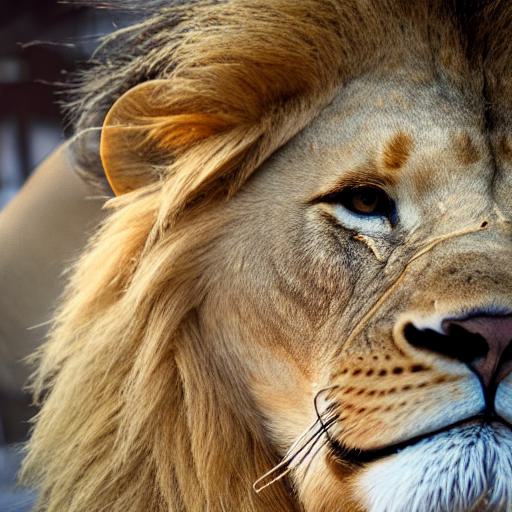} &
        \hspace{0.05cm}
        \includegraphics[width=0.11\textwidth]{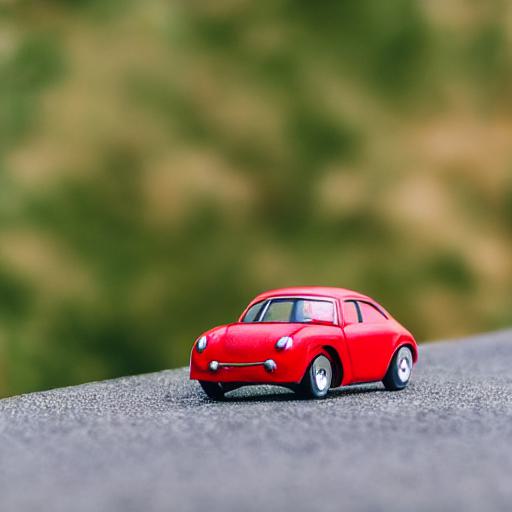} &
        \includegraphics[width=0.11\textwidth]{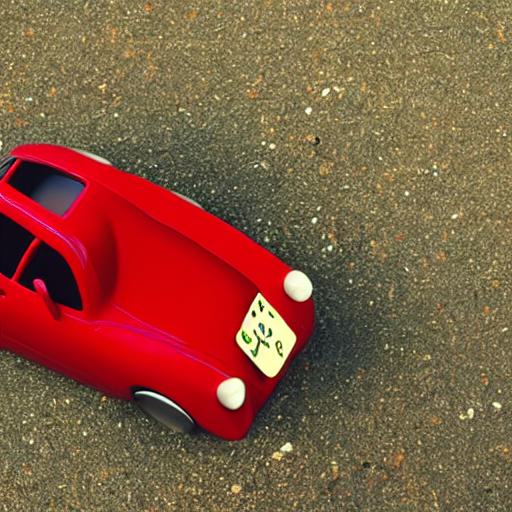} &
        \hspace{0.05cm}
        \includegraphics[width=0.11\textwidth]{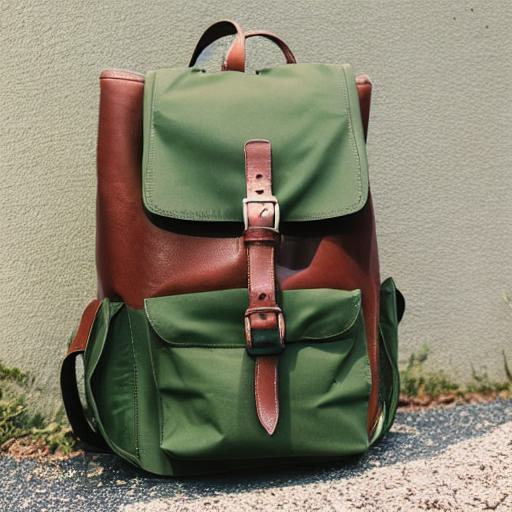} &
        \includegraphics[width=0.11\textwidth]{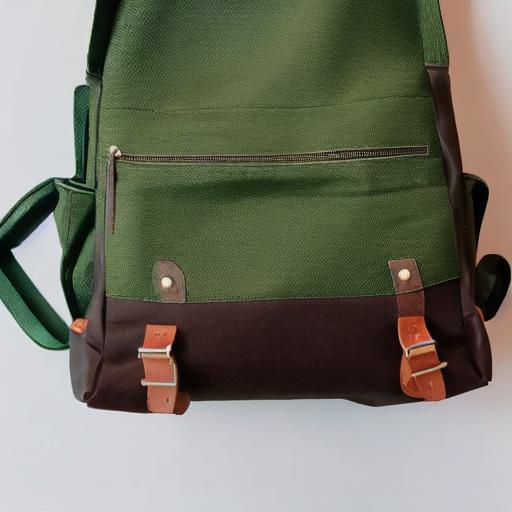} \\

        &
        \includegraphics[width=0.11\textwidth]{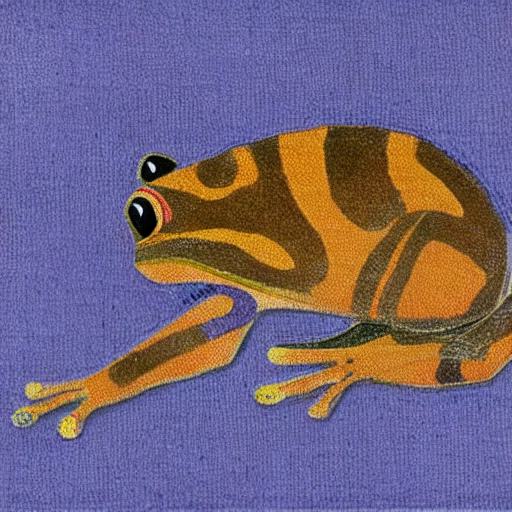} &
        \includegraphics[width=0.11\textwidth]{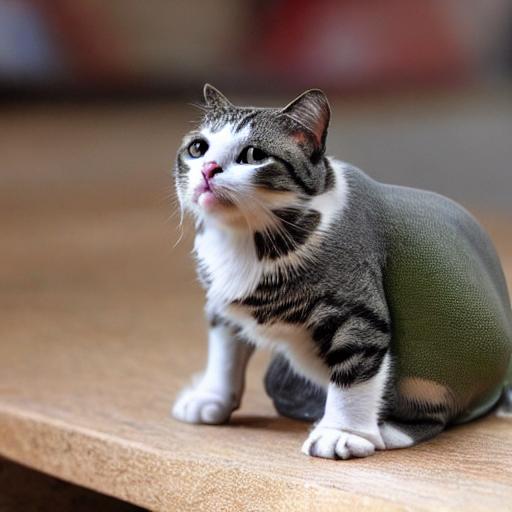} &
        \hspace{0.05cm}
        \includegraphics[width=0.11\textwidth]{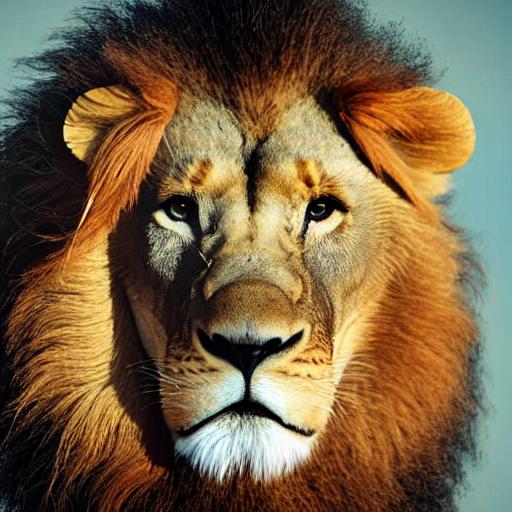} &
        \includegraphics[width=0.11\textwidth]{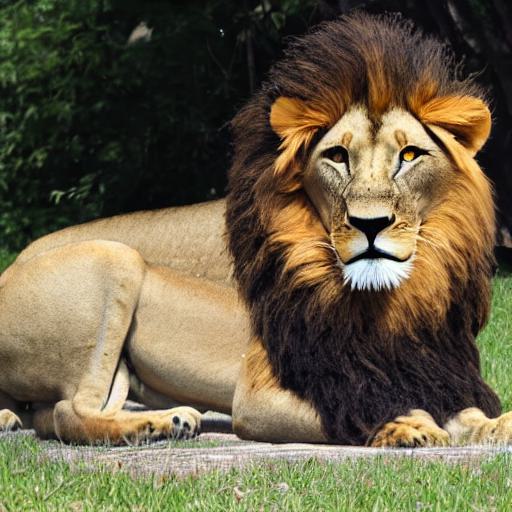} &
        \hspace{0.05cm}
        \includegraphics[width=0.11\textwidth]{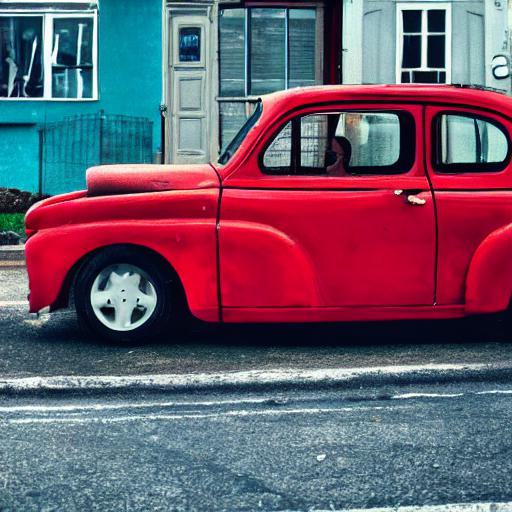} &
        \includegraphics[width=0.11\textwidth]{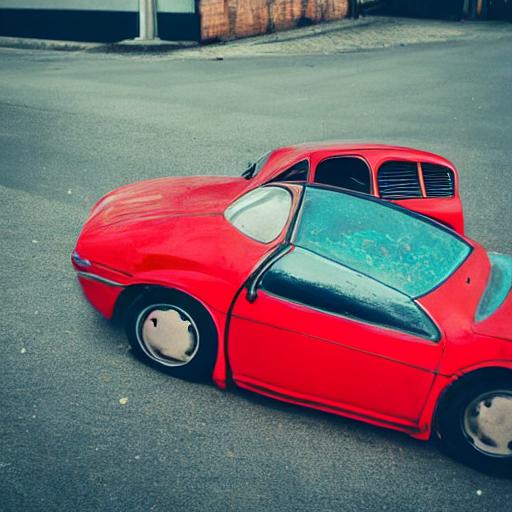} &
        \hspace{0.05cm}
        \includegraphics[width=0.11\textwidth]{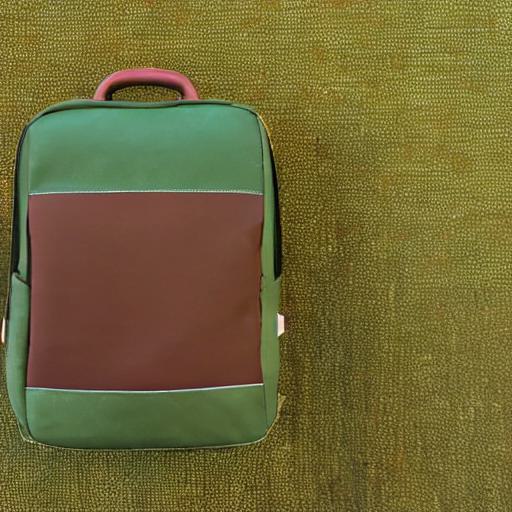} &
        \includegraphics[width=0.11\textwidth]{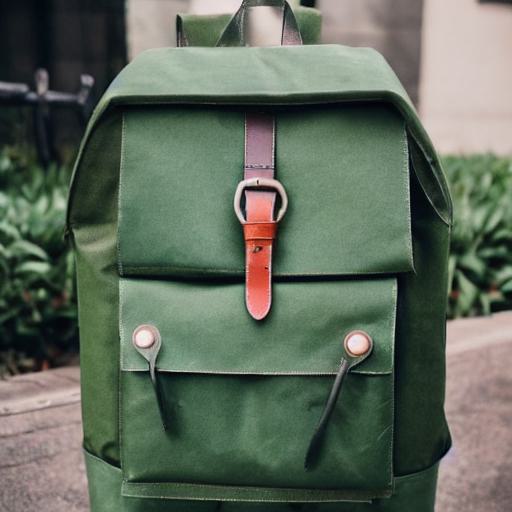}  \\ \\ \\

        {\raisebox{0.475in}{
        \multirow{2}{*}{\rotatebox{90}{\begin{tabular}{c} Stable Diffusion with \\ \textcolor{blue}{Attend-and-Excite} \\ \\ \end{tabular}}}}} &
        \includegraphics[width=0.11\textwidth]{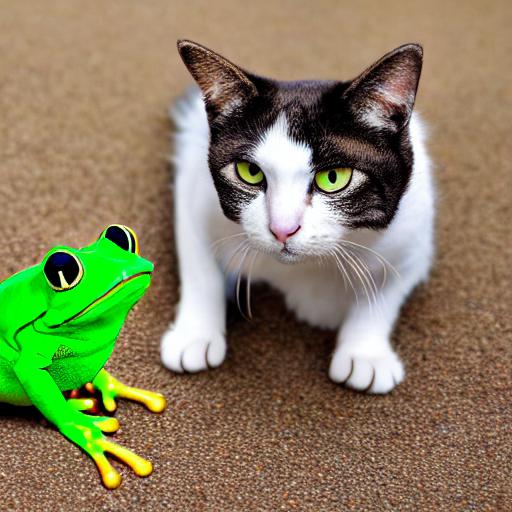} &
        \includegraphics[width=0.11\textwidth]{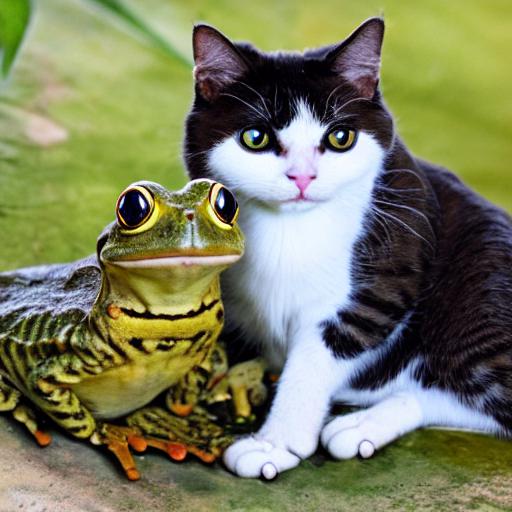} &
        \hspace{0.05cm}
         \includegraphics[width=0.11\textwidth]{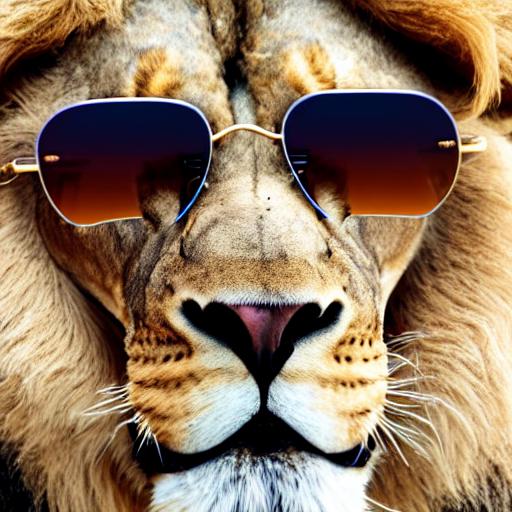} &
        \includegraphics[width=0.11\textwidth]{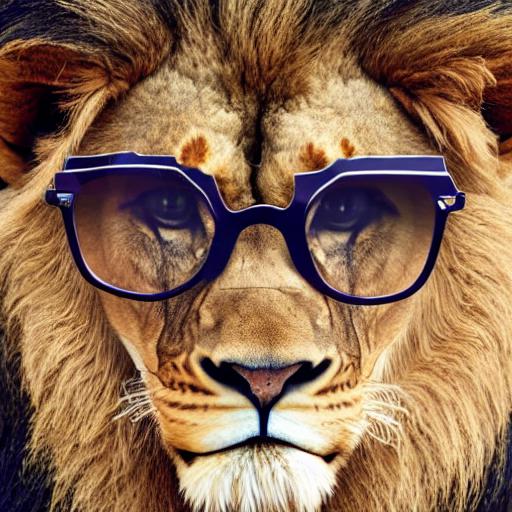} &
        \hspace{0.05cm}
        \includegraphics[width=0.11\textwidth]{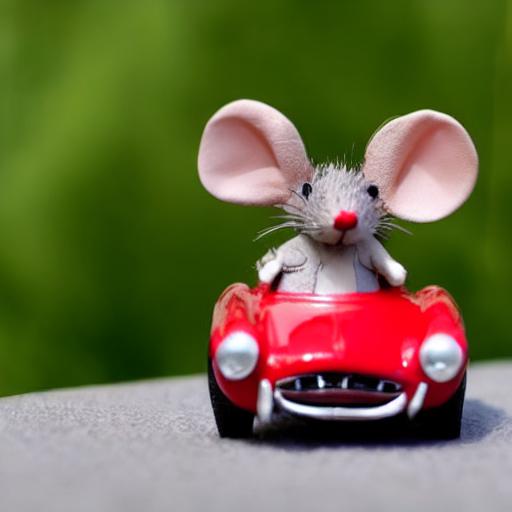} &
        \includegraphics[width=0.11\textwidth]{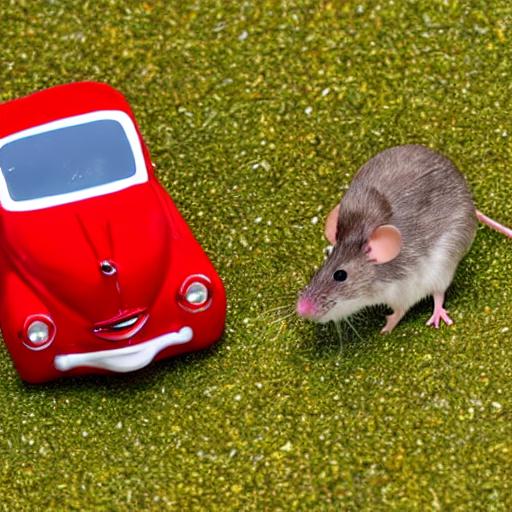} &
        \hspace{0.05cm}
        \includegraphics[width=0.11\textwidth]{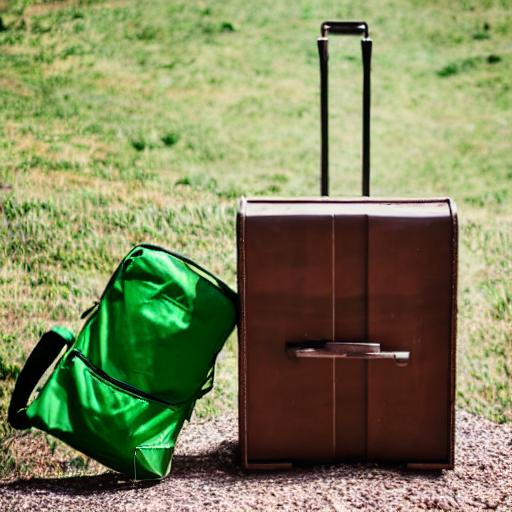} &
        \includegraphics[width=0.11\textwidth]{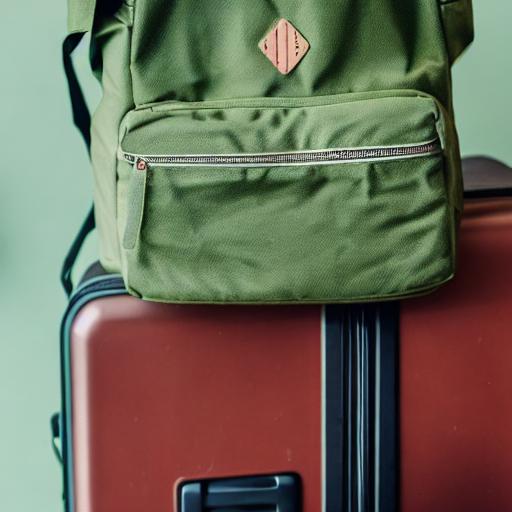}  \\

        &
        \includegraphics[width=0.11\textwidth]{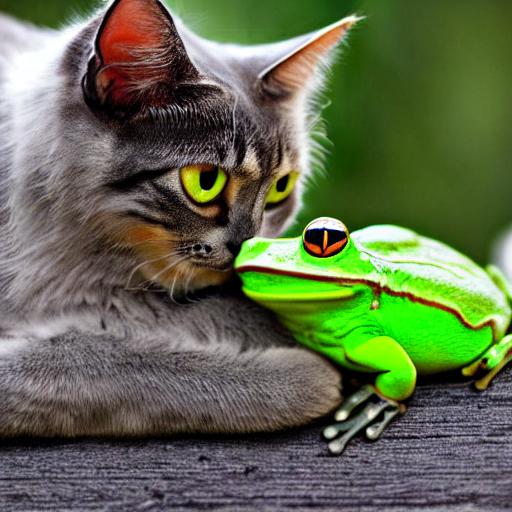} &
        \includegraphics[width=0.11\textwidth]{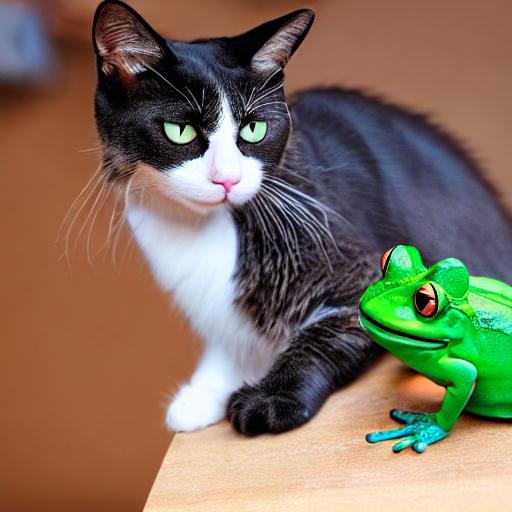} &
        \hspace{0.05cm}
        \includegraphics[width=0.11\textwidth]{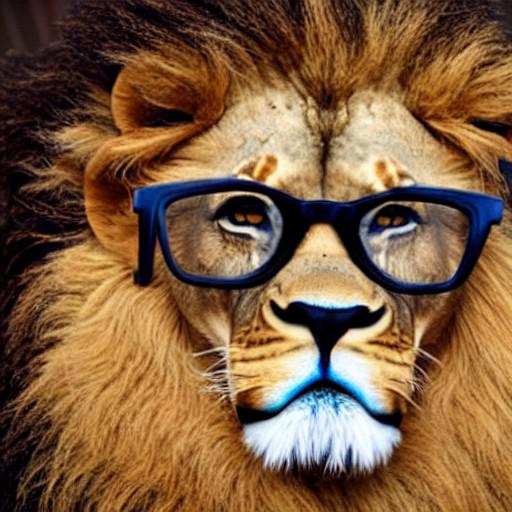} &
        \includegraphics[width=0.11\textwidth]{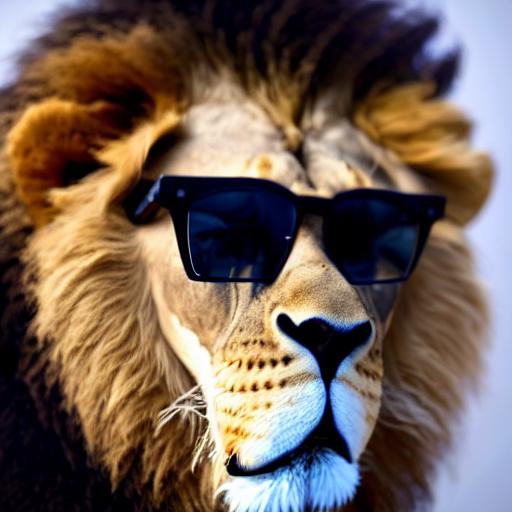} &
        \hspace{0.05cm}
        \includegraphics[width=0.11\textwidth]{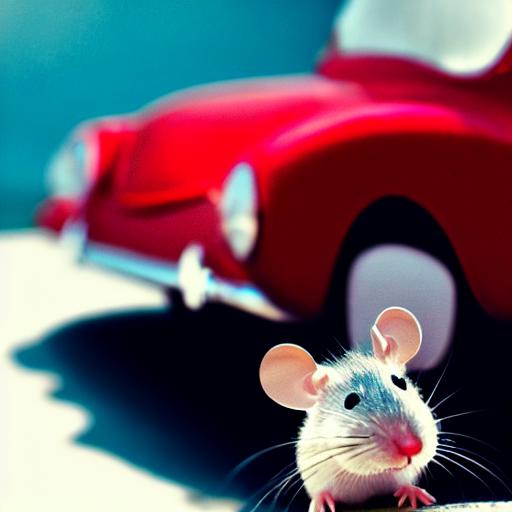} &
        \includegraphics[width=0.11\textwidth]{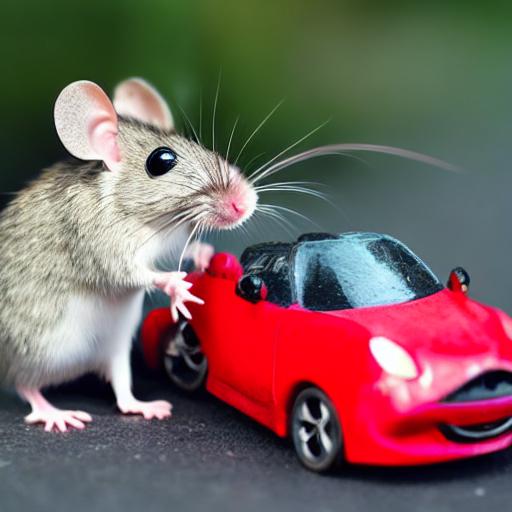} &
        \hspace{0.05cm}
        \includegraphics[width=0.11\textwidth]{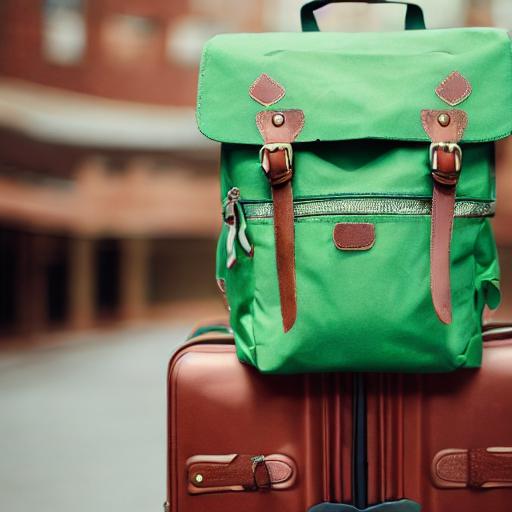} &
        \includegraphics[width=0.11\textwidth]{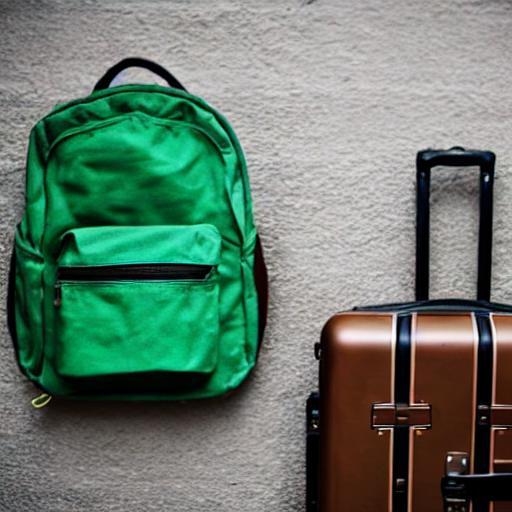}

    \end{tabular}

    }
    \vspace{-0.25cm}
    \caption{Qualitative Comparison. For each prompt, we show four images generated by each of the four considered methods where we use the same set of seeds across all approaches.
    The subject tokens optimized by Attend-and-Excite are highlighted in \textcolor{blue}{blue}.}
    \label{fig:additional_results_supp2}
\end{figure*}
\begin{figure*}
    \centering
    \setlength{\tabcolsep}{0.5pt}
    \renewcommand{\arraystretch}{0.3}
    {\small
    \begin{tabular}{c c c @{\hspace{0.1cm}} c c @{\hspace{0.1cm}} c c @{\hspace{0.1cm}} c c }

        & 
        \multicolumn{2}{c}{``A \textcolor{blue}{cat} and a \textcolor{blue}{dog}''} &
        \multicolumn{2}{c}{``A \textcolor{blue}{lion} and a \textcolor{blue}{rabbit}''} \\

        {\raisebox{0.275in}{
        \rotatebox{90}{\begin{tabular}{c} Composable \\ Diffusion \\ \\ \end{tabular}}}} &
        \includegraphics[width=0.17\textwidth]{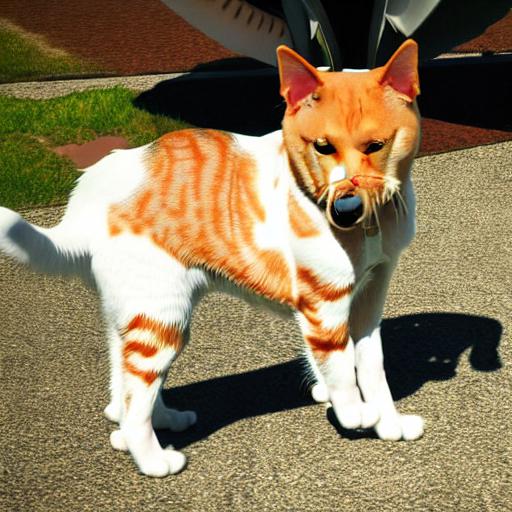} &
        \includegraphics[width=0.17\textwidth]{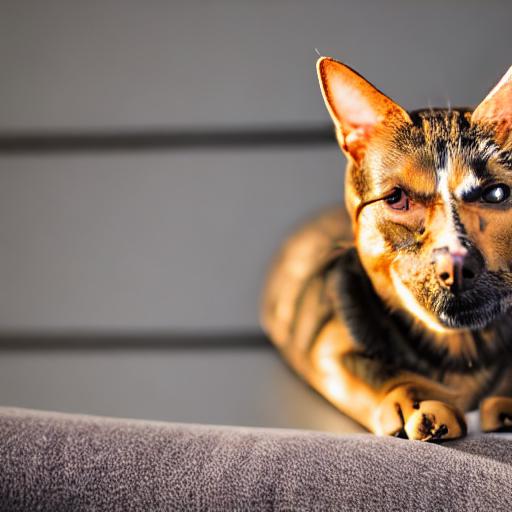} &
        \hspace{0.05cm}
        \includegraphics[width=0.17\textwidth]{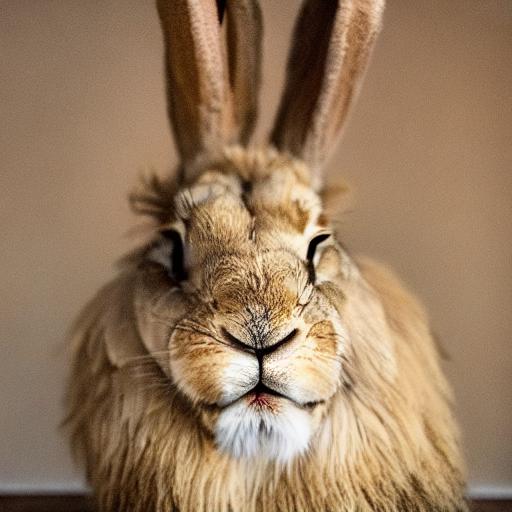} &
        \includegraphics[width=0.17\textwidth]{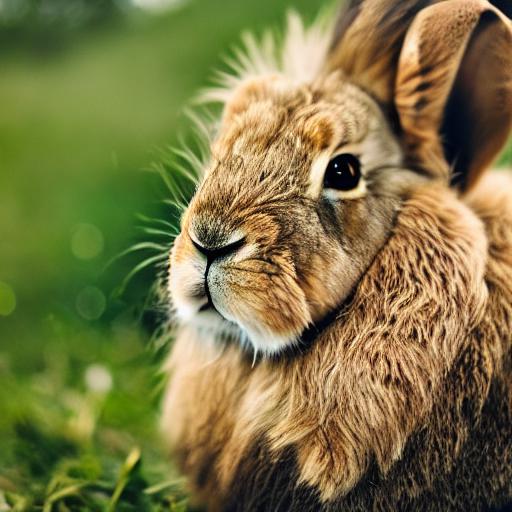} \\

        {\raisebox{0.225in}{
        \rotatebox{90}{\begin{tabular}{c} Attend-\&-Excite \\ \\ \end{tabular}}}} &
        \includegraphics[width=0.17\textwidth]{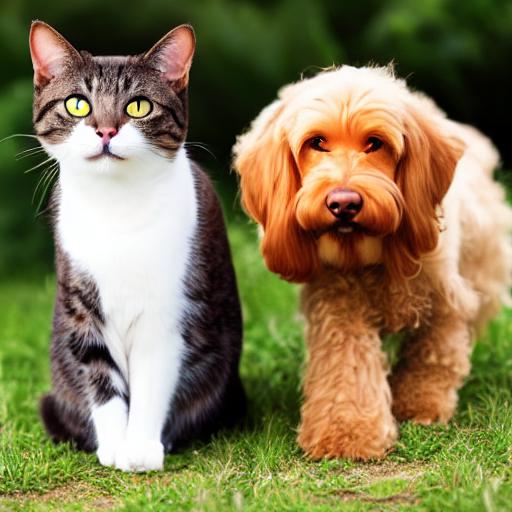} &
        \includegraphics[width=0.17\textwidth]{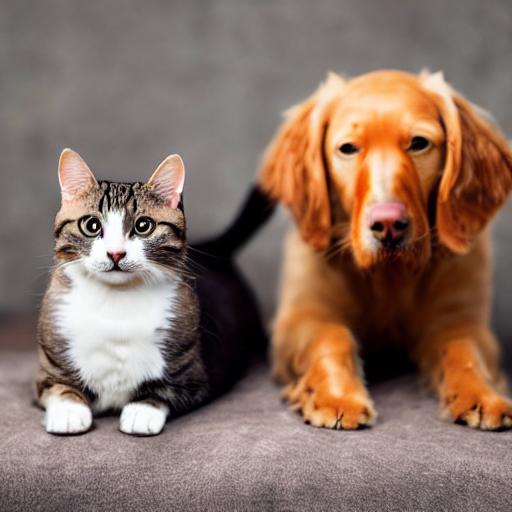} &
        \hspace{0.05cm}
        \includegraphics[width=0.17\textwidth]{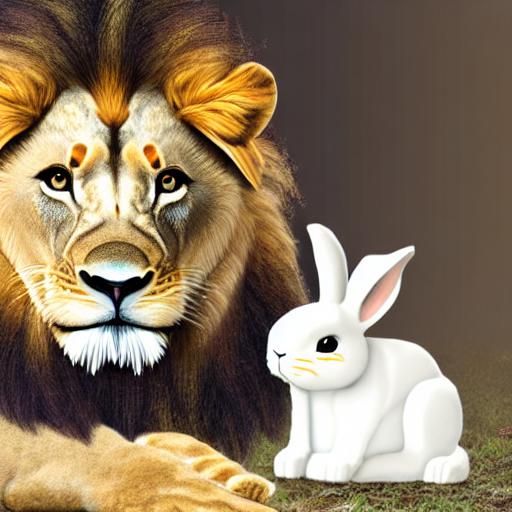} &
        \includegraphics[width=0.17\textwidth]{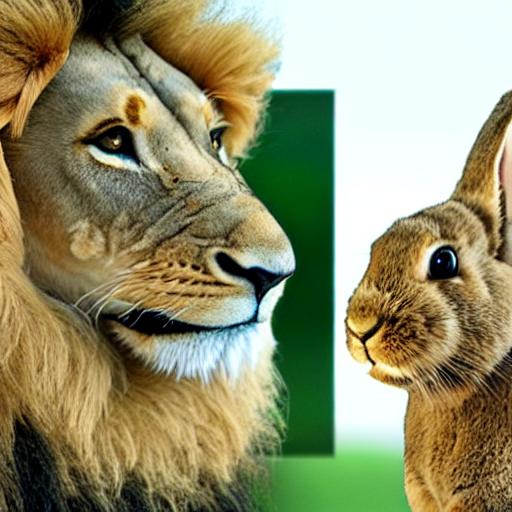} \\ \\ \\

        & 
        \multicolumn{2}{c}{\begin{tabular}{c}``An orange \textcolor{blue}{suitcase} \\ and a brown \textcolor{blue}{bench}''\end{tabular}} &
        \multicolumn{2}{c}{\begin{tabular}{c}``A purple \textcolor{blue}{balloon} \\ and a white \textcolor{blue}{clock}''\end{tabular}} \\

        {\raisebox{0.275in}{
        \rotatebox{90}{\begin{tabular}{c} Composable \\ Diffusion \\ \\ \end{tabular}}}} &
        \includegraphics[width=0.17\textwidth]{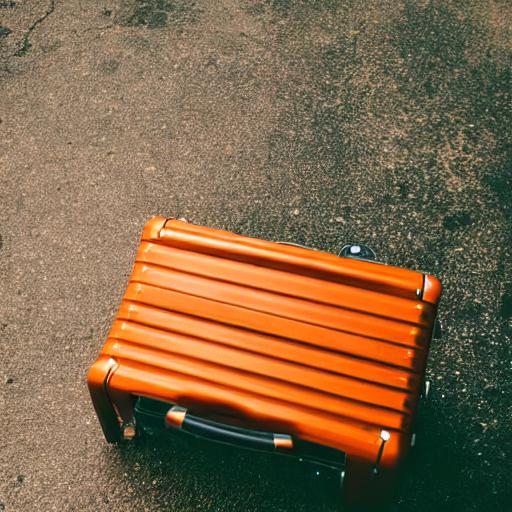} &
        \includegraphics[width=0.17\textwidth]{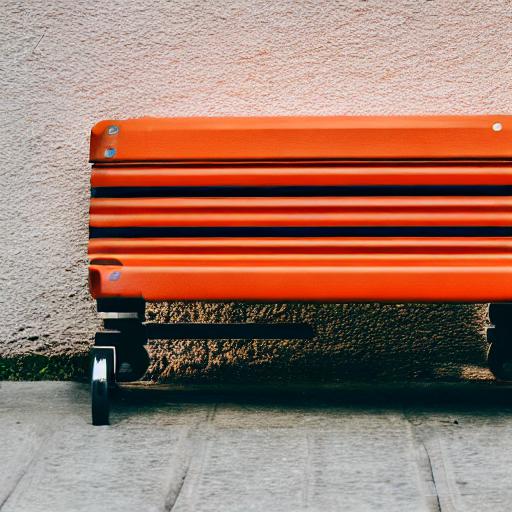} &
        \hspace{0.05cm}
        \includegraphics[width=0.17\textwidth]{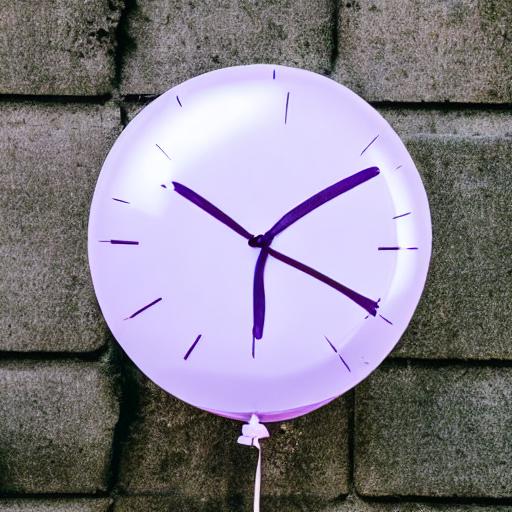} &
        \includegraphics[width=0.17\textwidth]{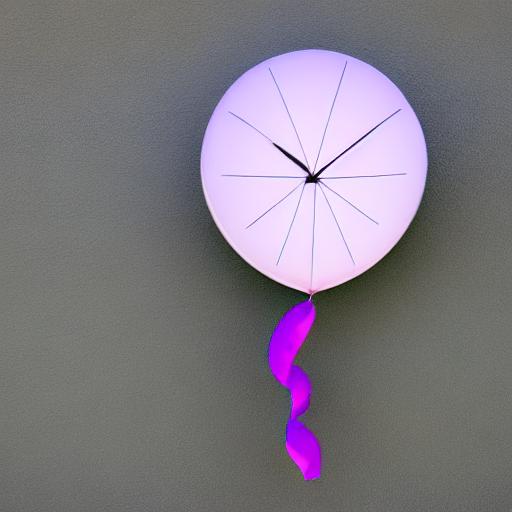} \\

        {\raisebox{0.225in}{
        \rotatebox{90}{\begin{tabular}{c} Attend-\&-Excite \\ \\ \end{tabular}}}} &
        \includegraphics[width=0.17\textwidth]{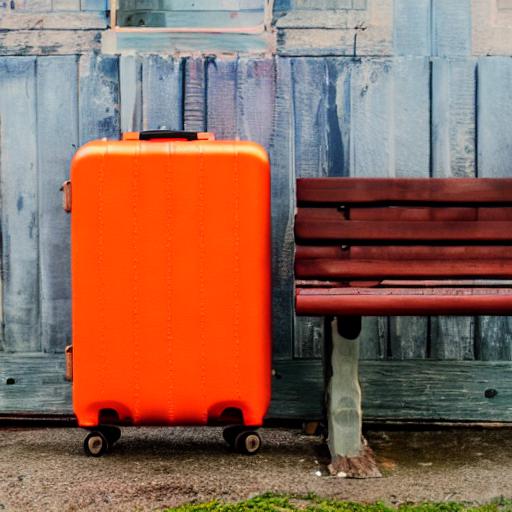} &
        \includegraphics[width=0.17\textwidth]{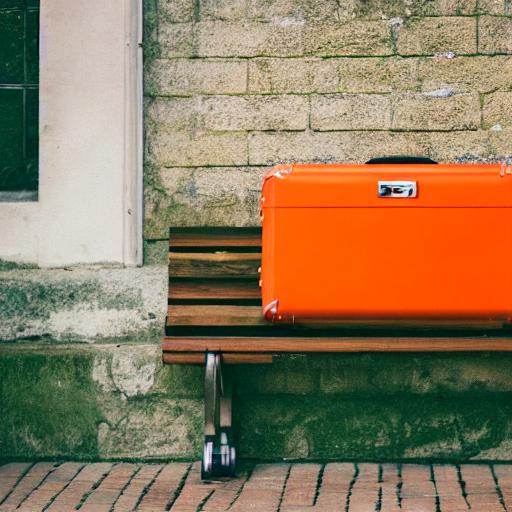} &
        \hspace{0.05cm}
        \includegraphics[width=0.17\textwidth]{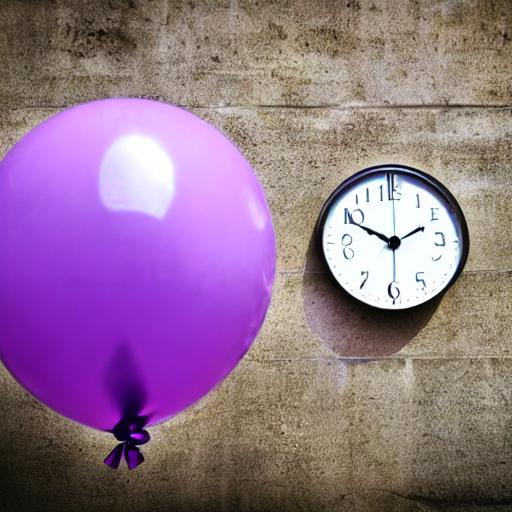} &
        \includegraphics[width=0.17\textwidth]{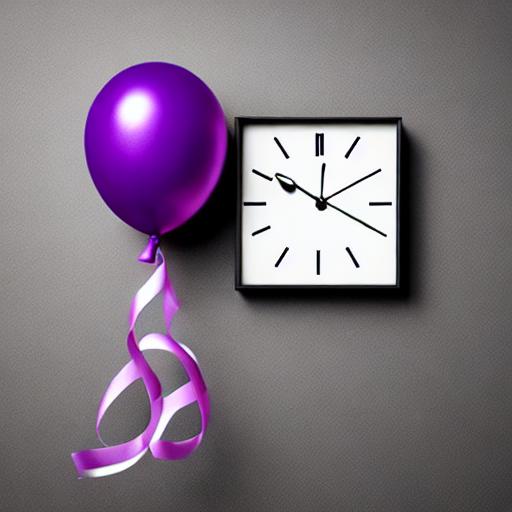} \\ \\ \\

        & 
        \multicolumn{2}{c}{\begin{tabular}{c}``A \textcolor{blue}{rabbit} \\ and a blue \textcolor{blue}{bowl}''\end{tabular}} &
        \multicolumn{2}{c}{\begin{tabular}{c}``A \textcolor{blue}{frog} \\ and a pink \textcolor{blue}{bench}''\end{tabular}} \\

        {\raisebox{0.275in}{
        \rotatebox{90}{\begin{tabular}{c} Composable \\ Diffusion \\ \\ \end{tabular}}}} &
        \includegraphics[width=0.17\textwidth]{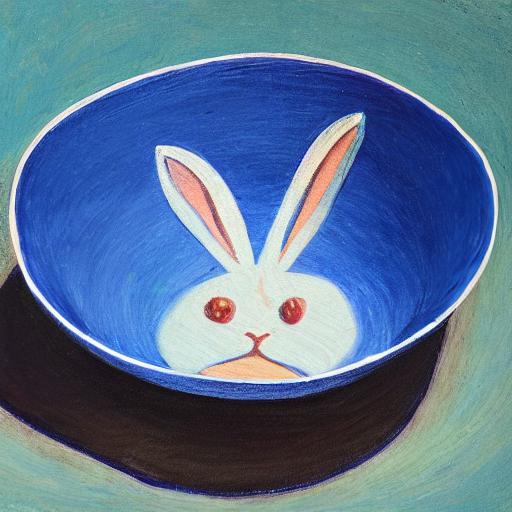} &
        \includegraphics[width=0.17\textwidth]{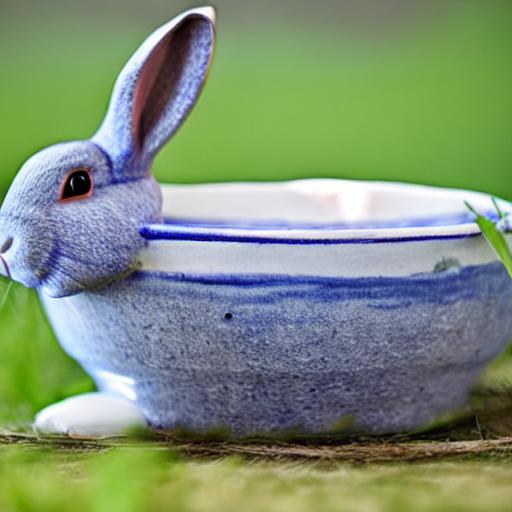} &
        \hspace{0.05cm}
        \includegraphics[width=0.17\textwidth]{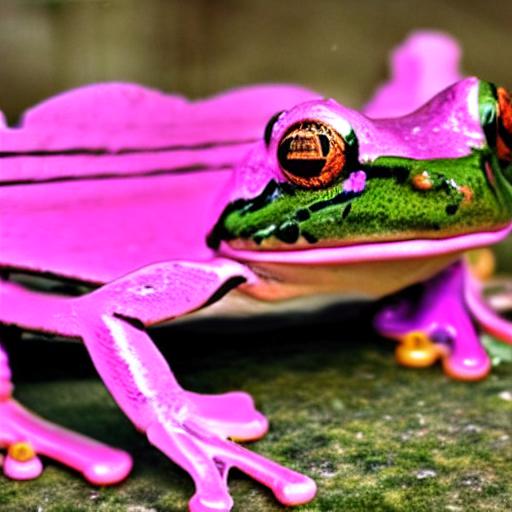} &
        \includegraphics[width=0.17\textwidth]{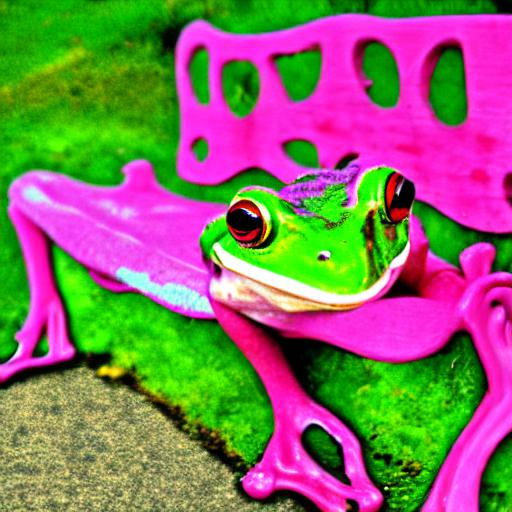} \\

        {\raisebox{0.225in}{
        \rotatebox{90}{\begin{tabular}{c} Attend-\&-Excite \\ \\ \end{tabular}}}} &
        \includegraphics[width=0.17\textwidth]{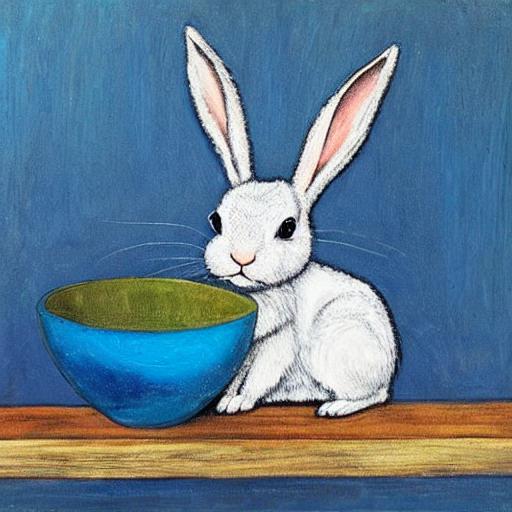} &
        \includegraphics[width=0.17\textwidth]{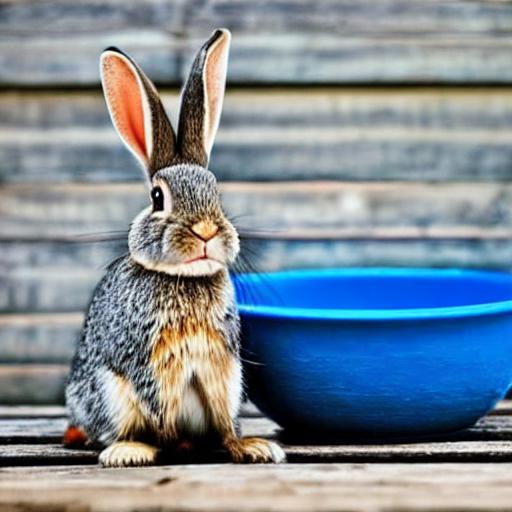} &
        \hspace{0.05cm}
        \includegraphics[width=0.17\textwidth]{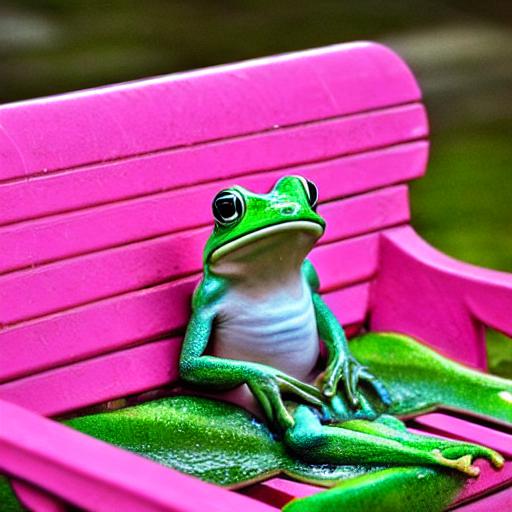} &
        \includegraphics[width=0.17\textwidth]{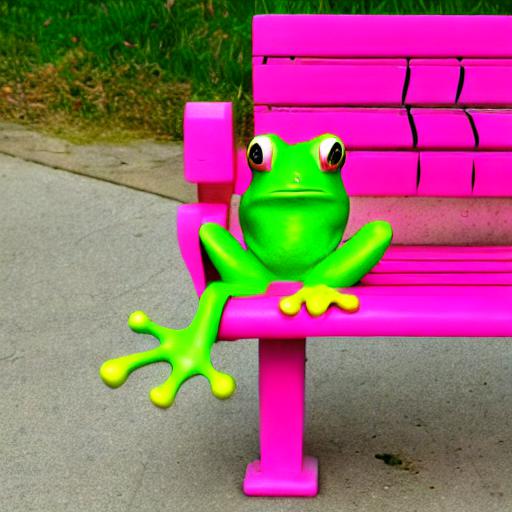} \\

    \end{tabular}
    
    }
    \vspace{-0.25cm}
    \caption{Qualitative comparison to Composable Diffusion. Composable Diffusion often results in ``hybrid'' objects which mix the subjects in the input prompt. 
    }
    \label{fig:compositional_extra}
\end{figure*}
\begin{figure*}
    \begin{tabular}{c @{\hspace{0.3cm}} c}

    \\ \\ \\ \\ \\ \\
    
    \includegraphics[width=0.42\linewidth]{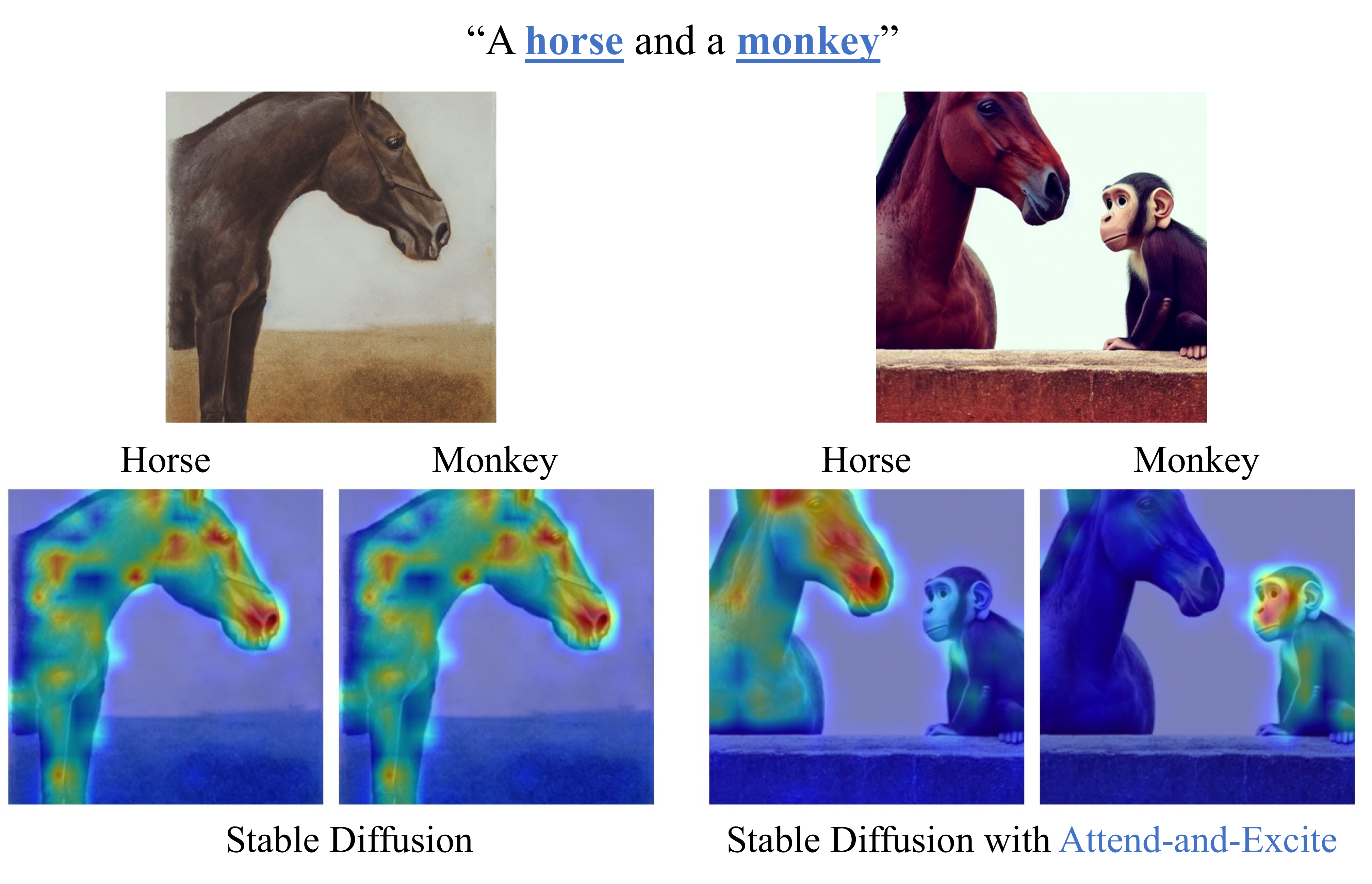} &
    \hspace{0.15cm}
    \includegraphics[width=0.42\linewidth]{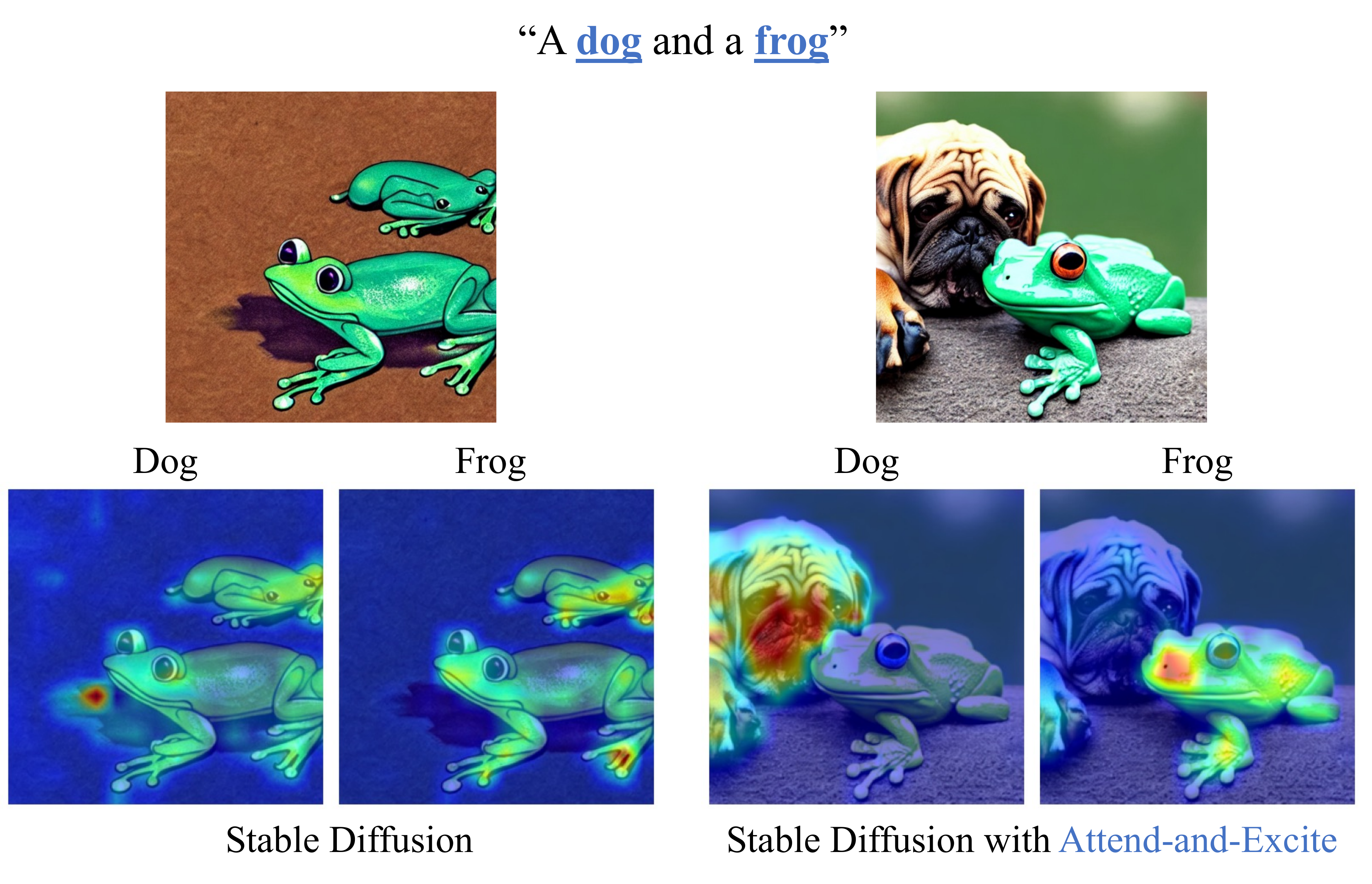} \\ \\
    
    \includegraphics[width=0.42\linewidth]{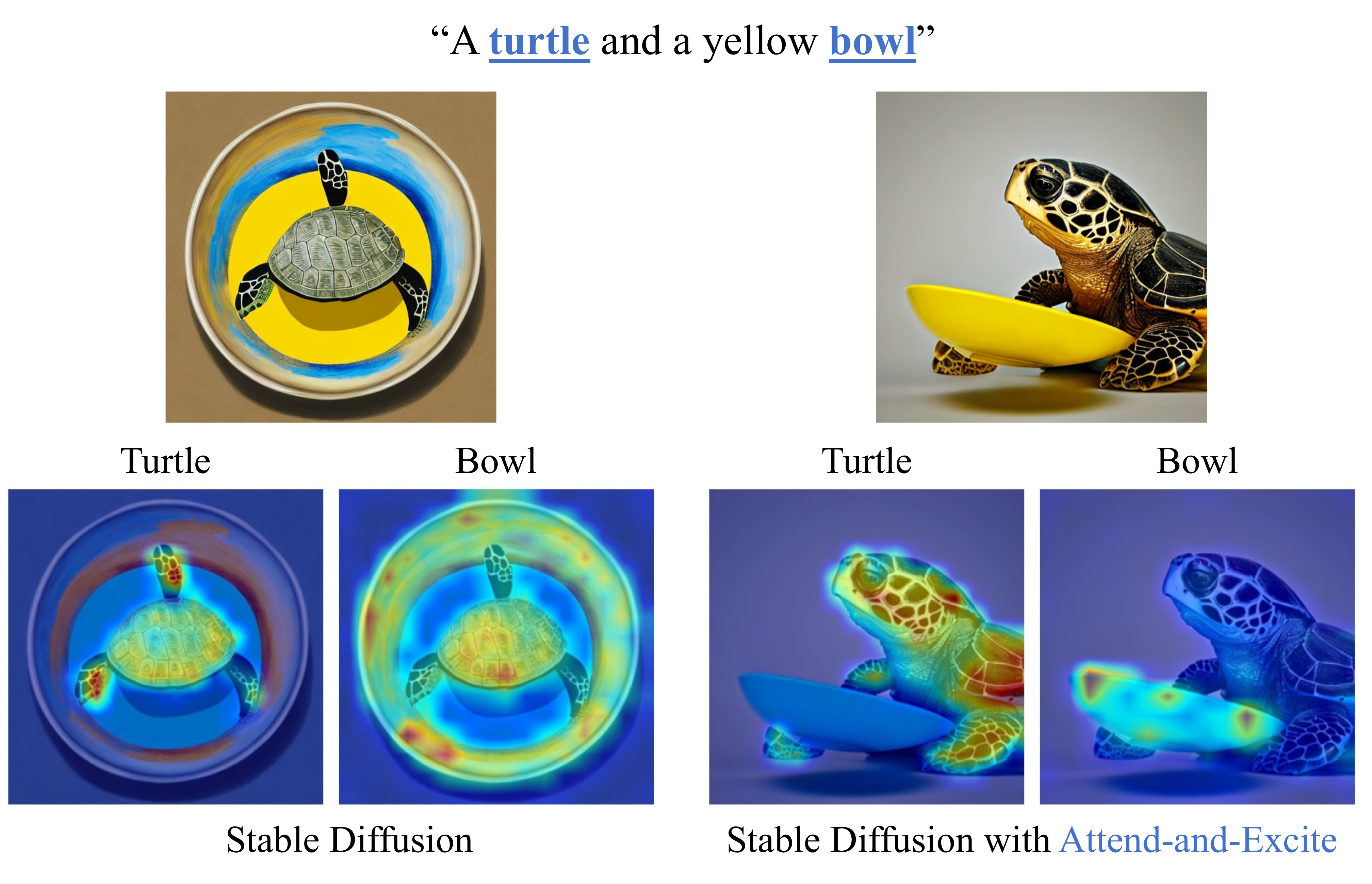} &
    \hspace{0.15cm}
    \includegraphics[width=0.42\linewidth]{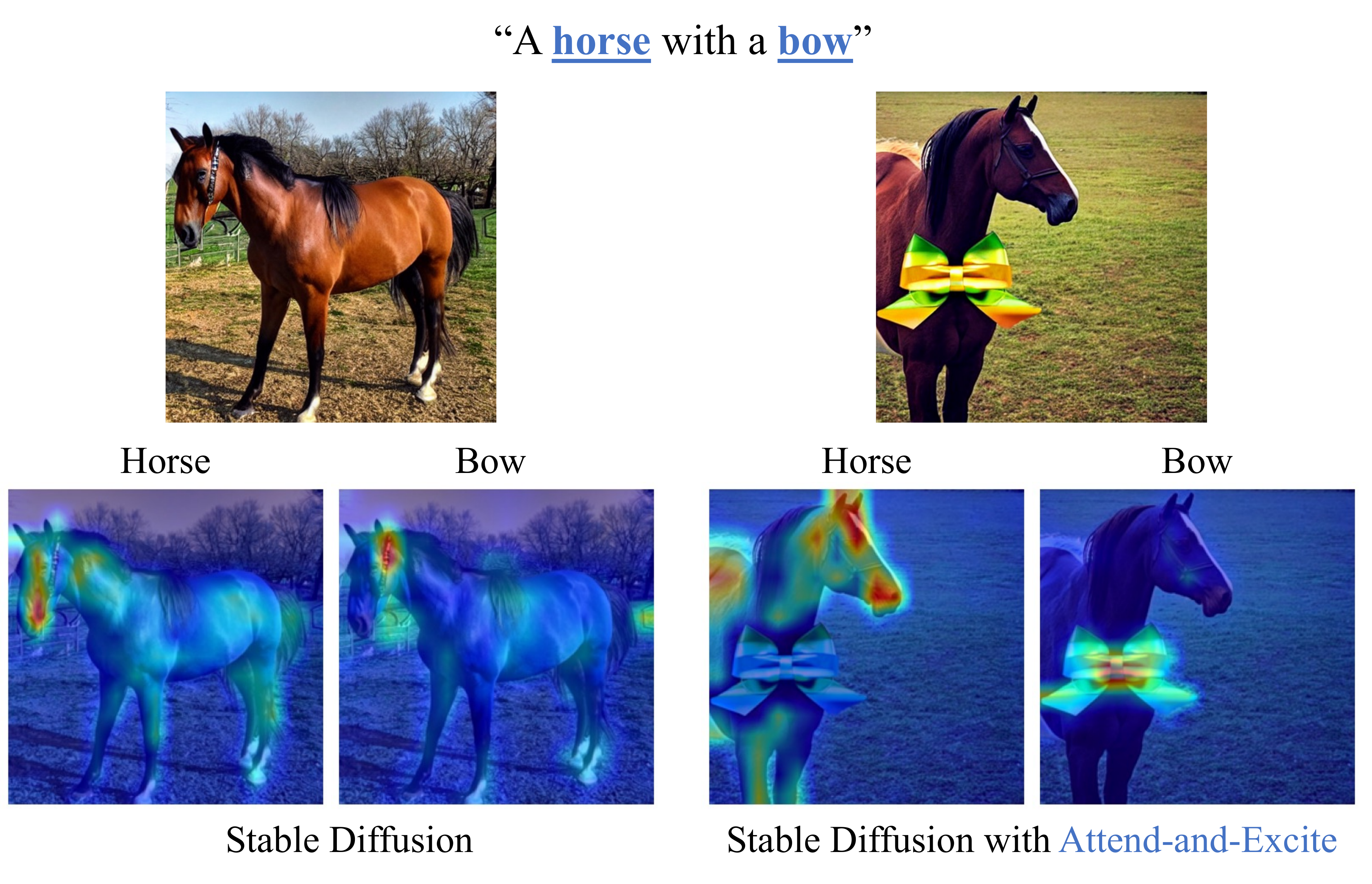} \\ \\

    \includegraphics[width=0.42\linewidth]{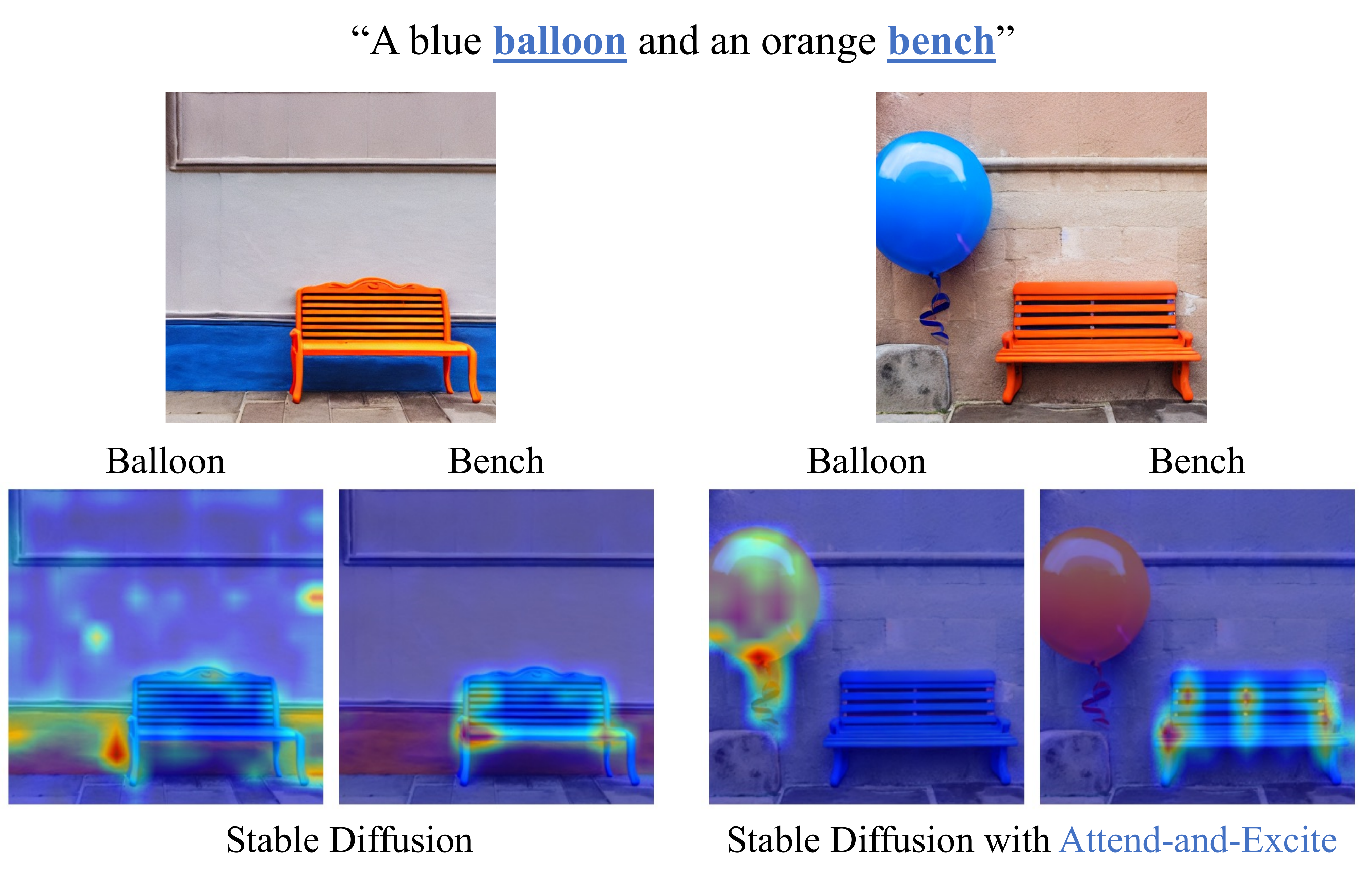} &
    \hspace{0.15cm}
    \includegraphics[width=0.42\linewidth]{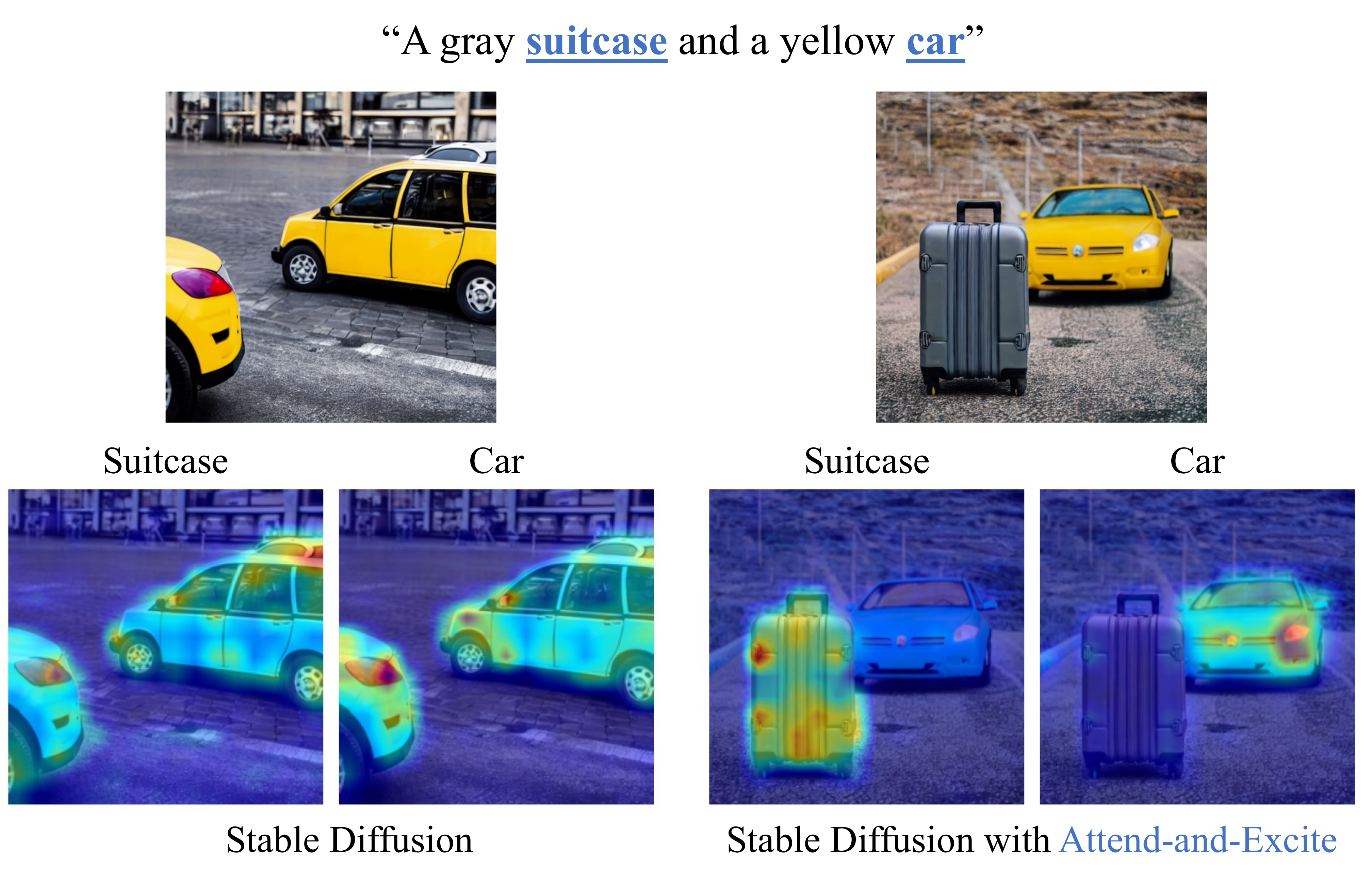} \\ \\

    \end{tabular}
    \vspace{-0.4cm}
    \caption{Visualization of the cross-attention maps per object before and after applying Attend and Excite on Stable Diffusion.}
    \label{fig:additional_cls_spes}
\end{figure*}

\end{document}